%% file: arxiv.tex
\documentclass[10pt,twocolumn,letterpaper]{article}

\usepackage[pagenumbers]{cvpr}
\providecommand{\linenumbers}{}
\providecommand{\nolinenumbers}{}

\input{preamble}
\input{numbers/supp_bench}

\input{numbers/supp_protocol}
\input{numbers/supp_training}
\input{numbers/supp_interp}

\input{numbers/supp_t4}
\usepackage{listings}
\usepackage{placeins}
\AtEndPreamble{
    \crefname{appendix}{Sec.}{Secs.}
    \Crefname{appendix}{Section}{Sections}
    \crefname{subappendix}{Sec.}{Secs.}
    \Crefname{subappendix}{Section}{Sections}
}

\definecolor{linkblue}{rgb}{0.21,0.49,0.74}
\usepackage[pagebackref,breaklinks,colorlinks,allcolors=linkblue]{hyperref}
\hypersetup{
    pdftitle={You Only Look Multiple Times: YoloMT, a Real-Time Streaming Model on a T4},
    pdfauthor={Nessim Ben Abbes, Duc Han Le, Sabri Mtibaa, Van-Tam Nguyen}
}

\title{You Only Look Multiple Times:\\YoloMT, a Real-Time Streaming Model on a T4}

\author{Nessim Ben Abbes$^{1,2}$\qquad Duc Han Le$^{2}$\qquad Sabri Mtibaa$^{1}$\qquad Van-Tam Nguyen$^{2}$\\
$^{1}$Protectline\qquad $^{2}$T\'el\'ecom Paris, Institut Polytechnique de Paris\\
{\tt\footnotesize \{nessim.benabbes, sabri.mtibaa\}@protectline.fr\qquad \{duc-han.le, van-tam.nguyen\}@telecom-paris.fr}}

\newcounter{mainpages}

\begin{document}
\addtocontents{toc}{\protect\setcounter{tocdepth}{-1}}
\twocolumn[{%
\renewcommand\twocolumn[1][]{#1}%
\maketitle
\input{floats/fig_lookback}
}]
\input{sec/0_abstract}
\input{sec/1_intro}

\input{sec/2_related}

\input{sec/3_method}

\input{sec/4_results}
\input{sec/5_discussion}
\input{sec/6_llm}
{
    \small
    \bibliographystyle{ieeenat_fullname}
    \bibliography{references}
}

\clearpage
\setcounter{mainpages}{\value{page}}
\let\mainmaketitlesupplementary\maketitlesupplementary
\renewcommand{\maketitlesupplementary}{\setcounter{page}{\value{mainpages}}\mainmaketitlesupplementary}
\setcounter{figure}{0}
\setcounter{table}{0}
\setcounter{equation}{0}
\renewcommand{\thefigure}{S\arabic{figure}}
\renewcommand{\thetable}{S\arabic{table}}
\renewcommand{\theequation}{S\arabic{equation}}
\renewcommand{\theHfigure}{S\arabic{figure}}
\renewcommand{\theHtable}{S\arabic{table}}
\renewcommand{\theHequation}{S\arabic{equation}}
\renewcommand{\dblfloatpagefraction}{0.75}
\renewcommand{\floatpagefraction}{0.85}
\addtocontents{toc}{\protect\setcounter{tocdepth}{2}}
\input{tables/supp_prompts}
\input{sec/X_suppl}

\end{document}

%% file: preamble.tex
\usepackage{multirow}
\usepackage{microtype}
\usepackage{tikz}
\usepackage{capt-of}
\usetikzlibrary{arrows.meta,positioning,calc,fit,backgrounds,decorations.pathreplacing}

\newcommand{\model}{YoloMT\xspace}

\usepackage{xr-hyper}
\makeatletter
\newcommand*{\addFileDependency}[1]{\typeout{(#1)}\@addtofilelist{#1}\IfFileExists{#1}{}{\typeout{No file #1.}}}
\makeatother
\newif\ifcrossdocument

\makeatletter
\def\cref@getref@kit#1#2{%
  \expandafter\let\expandafter#2\csname r@#1@cref\endcsname%
  \expandafter\expandafter\expandafter\def%
    \expandafter\expandafter\expandafter#2%
    \expandafter\expandafter\expandafter{%
      \expandafter\@firstoftwo#2}}%
\def\cref@firstgroup#1#2\cref@firstgroup@end{#1}
\AtBeginDocument{\ifcrossdocument
  \ifx\cref@getref\cref@getref@kit\else
    \PackageError{preamble}{cleveref's \string\cref@getref\space is not 0.21.4's}{Check the cross-document patch.}%
  \fi
  \def\cref@getref#1#2{%
    \expandafter\let\expandafter#2\csname r@#1@cref\endcsname%
    \expandafter\expandafter\expandafter\def%
      \expandafter\expandafter\expandafter#2%
      \expandafter\expandafter\expandafter{%
        \expandafter\cref@firstgroup#2\cref@firstgroup@end}}%
  \fi}
\makeatother

\xspaceaddexceptions{\% ] \,}

\input{numbers/repo}

\input{numbers/repo_k}

\input{numbers/repo_deploy}

\input{numbers/repo_ablate}

\input{numbers/tenx}

\input{numbers/ksweep}

\input{numbers/final}

\input{numbers/extra}

\input{numbers/method}

\input{numbers/results}

\input{numbers/figures}

\input{numbers/intro}

%% file: numbers/repo.tex
\newcommand{\pilotWindows}{1,103\xspace}

\newcommand{\shareLoraKsixteen}{58.0\xspace}
\newcommand{\shareBigKsixteen}{74.0\xspace}
\newcommand{\shareTeacherKsixteen}{122.3\xspace}

\newcommand{\mainOverSFTKzero}{+5.2\xspace}
\newcommand{\mainOverSFTKzeroLo}{+3.8\xspace}
\newcommand{\mainOverSFTKzeroHi}{+6.5\xspace}
\newcommand{\mainOverSFTKsixteen}{+6.6\xspace}
\newcommand{\mainOverSFTKsixteenLo}{+5.1\xspace}
\newcommand{\mainOverSFTKsixteenHi}{+8.0\xspace}

\newcommand{\fullOverLoraKzero}{+3.4\xspace}

\newcommand{\fullOverLoraKsixteen}{+7.3\xspace}

\newcommand{\distMaxAbs}{0.8\xspace}

\newcommand{\recallMainKsixteen}{0.502\xspace}
\newcommand{\precisionMainKsixteen}{0.580\xspace}

\newcommand{\recallRLKsixteen}{0.453\xspace}
\newcommand{\precisionRLKsixteen}{0.741\xspace}

\newcommand{\recallSFTKsixteen}{0.487\xspace}
\newcommand{\precisionSFTKsixteen}{0.502\xspace}

\newcommand{\trtVersion}{10.16.1.11\xspace}

\newcommand{\promptBase}{79\xspace}
\newcommand{\blockTokens}{18\xspace}

\newcommand{\maxChunk}{115\xspace}
\newcommand{\decodeSteps}{4\xspace}

\newcommand{\depthMin}{74.1\xspace}
\newcommand{\depthMax}{74.7\xspace}
\newcommand{\depthControl}{74.8\xspace}

\newcommand{\latentBest}{72.9\xspace}

\newcommand{\latentBase}{73.5\xspace}
\newcommand{\twoSecRLOne}{85.9\xspace}
\newcommand{\twoSecRLFive}{89.4\xspace}

\newcommand{\feedbackRepeatMin}{86\xspace}
\newcommand{\feedbackRepeatMax}{88\xspace}
\newcommand{\freshRepeatMax}{8\xspace}

\newcommand{\pilotVideos}{42\xspace}
\newcommand{\tVideos}{56\xspace}
\newcommand{\ovVideos}{466\xspace}
\newcommand{\tWindows}{1,518\xspace}
\newcommand{\ovWindows}{1,008\xspace}

\newcommand{\oLfmSmallKzero}{58.3\xspace}
\newcommand{\oLfmSmallKsixteen}{56.7\xspace}
\newcommand{\oLfmBigKzero}{66.1\xspace}
\newcommand{\oLfmBigKsixteen}{49.9\xspace}
\newcommand{\oSmolSmallKzero}{50.0\xspace}
\newcommand{\oSmolSmallKsixteen}{52.8\xspace}
\newcommand{\oSmolBigKzero}{59.8\xspace}
\newcommand{\oSmolBigKsixteen}{66.9\xspace}
\newcommand{\oInternOneKzero}{67.8\xspace}
\newcommand{\oInternOneKsixteen}{71.6\xspace}
\newcommand{\oInternTwoKzero}{67.9\xspace}
\newcommand{\oInternTwoKsixteen}{66.9\xspace}
\newcommand{\oQwenTwoKzero}{72.1\xspace}

\newcommand{\oQwenTwoKsixteen}{83.4\xspace}
\newcommand{\oQwenFourKzero}{74.3\xspace}

\newcommand{\oQwenFourKsixteen}{86.1\xspace}
\newcommand{\oQwenEightKzero}{73.4\xspace}

\newcommand{\oQwenEightKsixteen}{87.3\xspace}

\newcommand{\tQwenTwoKsixteen}{75.0\xspace}

\newcommand{\tQwenFourKsixteen}{86.8\xspace}

\newcommand{\tQwenEightKsixteen}{86.8\xspace}

\newcommand{\oQwenSevenTwoFiveKzero}{67.3\xspace}

\newcommand{\oQwenSevenTwoFiveKsixteen}{56.0\xspace}

\newcommand{\oLiveCcKzero}{49.3\xspace}

\newcommand{\oLiveCcKsixteen}{49.2\xspace}

\newcommand{\ovoQuestions}{837\xspace}

\newcommand{\mvbQuestions}{3,800\xspace}

\newcommand{\eFourMoreSecondsBig}{+10.2\xspace}
\newcommand{\eFourMoreSecondsBigLo}{+8.5\xspace}
\newcommand{\eFourMoreSecondsBigHi}{+11.9\xspace}
\newcommand{\eFourMoreSecondsSmall}{+5.8\xspace}
\newcommand{\eFourMoreSecondsSmallLo}{+4.1\xspace}
\newcommand{\eFourMoreSecondsSmallHi}{+7.4\xspace}

\newcommand{\tKlMinusRLKsixteen}{$-$2.3\xspace}
\newcommand{\tKlMinusRLKsixteenLo}{$-$3.6\xspace}
\newcommand{\tKlMinusRLKsixteenHi}{$-$0.9\xspace}

\newcommand{\genCloserRL}{86.2\xspace}

\newcommand{\genCloserKl}{2.7\xspace}

%% file: numbers/repo_k.tex
\newcommand{\nbQuestionsSb}{2,498\xspace}

\newcommand{\nbQuestionsVm}{900\xspace}

%% file: numbers/repo_deploy.tex
\newcommand{\qOQIntKzero}{67.9\xspace}

\newcommand{\qOQIntKsixteen}{72.0\xspace}

\newcommand{\qOQQwenKzero}{72.4\xspace}

\newcommand{\qOQQwenKsixteen}{80.8\xspace}

\newcommand{\qOQIntKzeroReward}{63.6\xspace}

\newcommand{\qOQIntKsixteenReward}{66.9\xspace}

\newcommand{\qOQQwenKzeroReward}{70.3\xspace}

\newcommand{\qOQQwenKsixteenReward}{79.2\xspace}

%% file: numbers/repo_ablate.tex
\newcommand{\ablBaseOverOnlyKKzero}{+2.7\xspace}
\newcommand{\ablBaseOverOnlyKKzeroLo}{+1.3\xspace}
\newcommand{\ablBaseOverOnlyKKzeroHi}{+4.1\xspace}
\newcommand{\ablBaseOverRandKzero}{+0.2\xspace}
\newcommand{\ablBaseOverRandKzeroLo}{$-$1.1\xspace}
\newcommand{\ablBaseOverRandKzeroHi}{+1.4\xspace}

\newcommand{\ablBaseOverOnlyKKsixteen}{$-$2.4\xspace}
\newcommand{\ablBaseOverOnlyKKsixteenLo}{$-$3.8\xspace}
\newcommand{\ablBaseOverOnlyKKsixteenHi}{$-$1.0\xspace}
\newcommand{\ablBaseOverRandKsixteen}{$-$0.2\xspace}
\newcommand{\ablBaseOverRandKsixteenLo}{$-$1.7\xspace}
\newcommand{\ablBaseOverRandKsixteenHi}{+1.2\xspace}

\newcommand{\tbLfmKsixteen}{46.6\xspace}

\newcommand{\tbSmolKsixteen}{43.5\xspace}

\newcommand{\tbQjLfmKsixteen}{40.1\xspace}

\newcommand{\tbQjSmolKsixteen}{38.0\xspace}

%% file: numbers/tenx.tex
\newcommand{\tenOverEsixMean}{+0.7\xspace}
\newcommand{\tenOverEsixMeanLo}{$-$0.3\xspace}
\newcommand{\tenOverEsixMeanHi}{+1.8\xspace}

\newcommand{\esixOverSftTenMean}{+1.1\xspace}
\newcommand{\esixOverSftTenMeanLo}{0.0\xspace}
\newcommand{\esixOverSftTenMeanHi}{+2.2\xspace}

\newcommand{\ovSftOverSftMean}{+1.6\xspace}
\newcommand{\ovSftOverSftMeanLo}{+0.4\xspace}
\newcommand{\ovSftOverSftMeanHi}{+2.7\xspace}

\newcommand{\ovEqOverSftMean}{$-$0.4\xspace}
\newcommand{\ovEqOverSftMeanLo}{$-$1.5\xspace}
\newcommand{\ovEqOverSftMeanHi}{+0.8\xspace}

\newcommand{\tenOKzero}{83.8\xspace}
\newcommand{\tenOKsixteen}{92.6\xspace}

\newcommand{\tenOverEsixOKsixteen}{+9.4\xspace}
\newcommand{\tenOverEsixOKsixteenLo}{+8.2\xspace}
\newcommand{\tenOverEsixOKsixteenHi}{+10.6\xspace}
\newcommand{\tenOverOvSftOKzero}{+4.3\xspace}
\newcommand{\tenOverOvSftOKzeroLo}{+3.3\xspace}
\newcommand{\tenOverOvSftOKzeroHi}{+5.3\xspace}
\newcommand{\tenOverOvSftOKsixteen}{+4.6\xspace}
\newcommand{\tenOverOvSftOKsixteenLo}{+3.6\xspace}
\newcommand{\tenOverOvSftOKsixteenHi}{+5.6\xspace}

\newcommand{\tenTKsixteen}{71.8\xspace}

\newcommand{\tenOverEsixTKsixteen}{$-$0.2\xspace}
\newcommand{\tenOverEsixTKsixteenLo}{$-$1.4\xspace}
\newcommand{\tenOverEsixTKsixteenHi}{+1.0\xspace}
\newcommand{\tenOverOvSftTKzero}{+4.8\xspace}
\newcommand{\tenOverOvSftTKzeroLo}{+3.7\xspace}
\newcommand{\tenOverOvSftTKzeroHi}{+5.9\xspace}
\newcommand{\tenOverOvSftTKsixteen}{+5.3\xspace}
\newcommand{\tenOverOvSftTKsixteenLo}{+4.0\xspace}
\newcommand{\tenOverOvSftTKsixteenHi}{+6.5\xspace}

\newcommand{\jqTenOKzero}{84.0\xspace}
\newcommand{\jqTenOKsixteen}{91.7\xspace}

\newcommand{\jqTenOverOvSftOKzero}{+4.6\xspace}
\newcommand{\jqTenOverOvSftOKzeroLo}{+3.6\xspace}
\newcommand{\jqTenOverOvSftOKzeroHi}{+5.7\xspace}
\newcommand{\jqTenOverOvSftOKsixteen}{+4.0\xspace}
\newcommand{\jqTenOverOvSftOKsixteenLo}{+2.9\xspace}
\newcommand{\jqTenOverOvSftOKsixteenHi}{+5.0\xspace}

\newcommand{\jqTenTKsixteen}{69.7\xspace}

\newcommand{\jqTenOverOvSftTKzero}{+3.2\xspace}
\newcommand{\jqTenOverOvSftTKzeroLo}{+2.0\xspace}
\newcommand{\jqTenOverOvSftTKzeroHi}{+4.4\xspace}
\newcommand{\jqTenOverOvSftTKsixteen}{+3.9\xspace}
\newcommand{\jqTenOverOvSftTKsixteenLo}{+2.7\xspace}
\newcommand{\jqTenOverOvSftTKsixteenHi}{+5.2\xspace}

\newcommand{\jqOLfmSmallKzero}{54.0\xspace}
\newcommand{\jqOLfmSmallKsixteen}{52.2\xspace}
\newcommand{\jqOSmolSmallKzero}{46.3\xspace}
\newcommand{\jqOSmolSmallKsixteen}{48.0\xspace}
\newcommand{\jqOInternOneKzero}{62.6\xspace}
\newcommand{\jqOInternOneKsixteen}{66.9\xspace}
\newcommand{\jqOQwenTwoKzero}{70.4\xspace}
\newcommand{\jqOQwenTwoKsixteen}{82.4\xspace}
\newcommand{\jqTQwenTwoKsixteen}{72.3\xspace}
\newcommand{\jqOLfmBigKzero}{63.0\xspace}
\newcommand{\jqOLfmBigKsixteen}{45.0\xspace}
\newcommand{\jqOSmolBigKzero}{56.4\xspace}
\newcommand{\jqOSmolBigKsixteen}{65.3\xspace}
\newcommand{\jqOInternTwoKzero}{64.1\xspace}
\newcommand{\jqOInternTwoKsixteen}{62.7\xspace}
\newcommand{\jqOQwenFourKzero}{71.5\xspace}
\newcommand{\jqOQwenFourKsixteen}{84.5\xspace}
\newcommand{\jqTQwenFourKsixteen}{82.1\xspace}
\newcommand{\jqOQwenSevenTwoFiveKzero}{61.8\xspace}
\newcommand{\jqOQwenSevenTwoFiveKsixteen}{53.2\xspace}
\newcommand{\jqOLiveCcKzero}{46.5\xspace}
\newcommand{\jqOLiveCcKsixteen}{45.5\xspace}
\newcommand{\jqOQwenEightKzero}{71.0\xspace}
\newcommand{\jqOQwenEightKsixteen}{85.2\xspace}
\newcommand{\jqTQwenEightKsixteen}{81.3\xspace}
\newcommand{\sftTenHours}{14.5\xspace}
\newcommand{\rlTenHours}{12.7\xspace}

%% file: numbers/ksweep.tex
\newcommand{\devTenKzero}{52.8\xspace}

\newcommand{\devTenKtwo}{57.5\xspace}

\newcommand{\devTenKfour}{60.3\xspace}

\newcommand{\devTenKsixteen}{72.2\xspace}

\newcommand{\devOvSftKzero}{48.8\xspace}

\newcommand{\devOvSftKtwo}{53.2\xspace}

\newcommand{\devOvSftKfour}{56.3\xspace}

\newcommand{\devOvSftKsixteen}{67.4\xspace}

\newcommand{\devEsixKzero}{52.5\xspace}

\newcommand{\devEsixKtwo}{58.2\xspace}

\newcommand{\devEsixKfour}{61.1\xspace}

\newcommand{\devEsixKsixteen}{71.8\xspace}

\newcommand{\devSftKzero}{47.8\xspace}

\newcommand{\devSftKtwo}{52.3\xspace}

\newcommand{\devSftKfour}{54.8\xspace}

\newcommand{\devSftKsixteen}{64.9\xspace}

\newcommand{\devTenRecallKsixteen}{0.505\xspace}
\newcommand{\devTenPrecisionKsixteen}{0.580\xspace}
\newcommand{\devTenOverEsixKzero}{+0.3\xspace}
\newcommand{\devTenOverEsixKzeroLo}{$-$1.0\xspace}
\newcommand{\devTenOverEsixKzeroHi}{+1.6\xspace}
\newcommand{\devOvSftOverSftKzero}{+1.0\xspace}
\newcommand{\devOvSftOverSftKzeroLo}{$-$0.3\xspace}
\newcommand{\devOvSftOverSftKzeroHi}{+2.3\xspace}
\newcommand{\devTenOverOvSftKzero}{+4.0\xspace}
\newcommand{\devTenOverOvSftKzeroLo}{+2.7\xspace}
\newcommand{\devTenOverOvSftKzeroHi}{+5.2\xspace}
\newcommand{\devEsixOverSftKzero}{+4.6\xspace}
\newcommand{\devEsixOverSftKzeroLo}{+3.4\xspace}
\newcommand{\devEsixOverSftKzeroHi}{+6.0\xspace}
\newcommand{\devTenOverEsixKsixteen}{+0.4\xspace}
\newcommand{\devTenOverEsixKsixteenLo}{$-$1.0\xspace}
\newcommand{\devTenOverEsixKsixteenHi}{+1.8\xspace}
\newcommand{\devOvSftOverSftKsixteen}{+2.5\xspace}
\newcommand{\devOvSftOverSftKsixteenLo}{+1.0\xspace}
\newcommand{\devOvSftOverSftKsixteenHi}{+4.0\xspace}
\newcommand{\devTenOverOvSftKsixteen}{+4.8\xspace}
\newcommand{\devTenOverOvSftKsixteenLo}{+3.5\xspace}
\newcommand{\devTenOverOvSftKsixteenHi}{+6.1\xspace}
\newcommand{\devEsixOverSftKsixteen}{+6.9\xspace}
\newcommand{\devEsixOverSftKsixteenLo}{+5.4\xspace}
\newcommand{\devEsixOverSftKsixteenHi}{+8.3\xspace}

\newcommand{\devQTenOverOvSftKzero}{+4.3\xspace}
\newcommand{\devQTenOverOvSftKzeroLo}{+3.1\xspace}
\newcommand{\devQTenOverOvSftKzeroHi}{+5.5\xspace}

\newcommand{\devQTenOverOvSftKsixteen}{+4.3\xspace}
\newcommand{\devQTenOverOvSftKsixteenLo}{+3.0\xspace}
\newcommand{\devQTenOverOvSftKsixteenHi}{+5.7\xspace}

\newcommand{\devQTenGain}{+19.4\xspace}
\newcommand{\devQTenGainLo}{+17.6\xspace}
\newcommand{\devQTenGainHi}{+21.2\xspace}

%% file: numbers/final.tex
\newcommand{\oneAccOursOvo}{45.9\xspace}
\newcommand{\oneAccOursMvb}{39.5\xspace}
\newcommand{\oneAccOursSb}{50.2\xspace}
\newcommand{\oneAccOursVm}{46.2\xspace}
\newcommand{\oneAccOursMean}{45.5\xspace}
\newcommand{\oneWordsOurs}{40\xspace}

\newcommand{\oneAccLfmSmallOvo}{38.6\xspace}
\newcommand{\oneAccLfmSmallMvb}{35.5\xspace}
\newcommand{\oneAccLfmSmallSb}{45.6\xspace}
\newcommand{\oneAccLfmSmallVm}{42.0\xspace}
\newcommand{\oneAccLfmSmallMean}{40.4\xspace}
\newcommand{\oneWordsLfmSmall}{18\xspace}

\newcommand{\oneAccSmolSmallOvo}{40.3\xspace}
\newcommand{\oneAccSmolSmallMvb}{36.1\xspace}
\newcommand{\oneAccSmolSmallSb}{46.0\xspace}
\newcommand{\oneAccSmolSmallVm}{39.8\xspace}
\newcommand{\oneAccSmolSmallMean}{40.5\xspace}
\newcommand{\oneWordsSmolSmall}{22\xspace}

\newcommand{\oneAccMobileOvo}{35.4\xspace}
\newcommand{\oneAccMobileMvb}{33.8\xspace}
\newcommand{\oneAccMobileSb}{43.4\xspace}
\newcommand{\oneAccMobileVm}{36.3\xspace}
\newcommand{\oneAccMobileMean}{37.2\xspace}
\newcommand{\oneWordsMobile}{11\xspace}

\newcommand{\oneAccInternOneOvo}{46.4\xspace}
\newcommand{\oneAccInternOneMvb}{43.8\xspace}
\newcommand{\oneAccInternOneSb}{52.4\xspace}
\newcommand{\oneAccInternOneVm}{51.3\xspace}
\newcommand{\oneAccInternOneMean}{48.5\xspace}
\newcommand{\oneWordsInternOne}{39\xspace}

\newcommand{\oneAccInternOneIntOvo}{49.0\xspace}
\newcommand{\oneAccInternOneIntMvb}{43.8\xspace}
\newcommand{\oneAccInternOneIntSb}{52.9\xspace}
\newcommand{\oneAccInternOneIntVm}{50.9\xspace}
\newcommand{\oneAccInternOneIntMean}{49.1\xspace}
\newcommand{\oneWordsInternOneInt}{39\xspace}

\newcommand{\oneAccLfmBigOvo}{38.9\xspace}
\newcommand{\oneAccLfmBigMvb}{37.9\xspace}
\newcommand{\oneAccLfmBigSb}{47.2\xspace}
\newcommand{\oneAccLfmBigVm}{43.7\xspace}
\newcommand{\oneAccLfmBigMean}{42.0\xspace}
\newcommand{\oneWordsLfmBig}{44\xspace}

\newcommand{\oneAccQwenTwoOvo}{49.5\xspace}
\newcommand{\oneAccQwenTwoMvb}{45.5\xspace}
\newcommand{\oneAccQwenTwoSb}{52.9\xspace}
\newcommand{\oneAccQwenTwoVm}{53.1\xspace}
\newcommand{\oneAccQwenTwoMean}{50.2\xspace}
\newcommand{\oneWordsQwenTwo}{46\xspace}

\newcommand{\oneAccQwenTwoIntOvo}{47.4\xspace}
\newcommand{\oneAccQwenTwoIntMvb}{44.0\xspace}
\newcommand{\oneAccQwenTwoIntSb}{52.8\xspace}
\newcommand{\oneAccQwenTwoIntVm}{53.9\xspace}
\newcommand{\oneAccQwenTwoIntMean}{49.5\xspace}
\newcommand{\oneWordsQwenTwoInt}{43\xspace}

\newcommand{\oneAccSmolBigOvo}{43.8\xspace}
\newcommand{\oneAccSmolBigMvb}{39.1\xspace}
\newcommand{\oneAccSmolBigSb}{48.8\xspace}
\newcommand{\oneAccSmolBigVm}{43.3\xspace}
\newcommand{\oneAccSmolBigMean}{43.8\xspace}
\newcommand{\oneWordsSmolBig}{35\xspace}

\newcommand{\oneAccInternTwoOvo}{49.2\xspace}
\newcommand{\oneAccInternTwoMvb}{45.4\xspace}
\newcommand{\oneAccInternTwoSb}{53.2\xspace}
\newcommand{\oneAccInternTwoVm}{56.8\xspace}
\newcommand{\oneAccInternTwoMean}{51.2\xspace}
\newcommand{\oneWordsInternTwo}{52\xspace}

\newcommand{\oneAccQwenFourOvo}{49.6\xspace}
\newcommand{\oneAccQwenFourMvb}{45.3\xspace}
\newcommand{\oneAccQwenFourSb}{55.6\xspace}
\newcommand{\oneAccQwenFourVm}{57.0\xspace}
\newcommand{\oneAccQwenFourMean}{51.9\xspace}
\newcommand{\oneWordsQwenFour}{42\xspace}

\newcommand{\oneAccQwenSevenTwoFiveOvo}{44.7\xspace}
\newcommand{\oneAccQwenSevenTwoFiveMvb}{40.7\xspace}
\newcommand{\oneAccQwenSevenTwoFiveSb}{48.0\xspace}
\newcommand{\oneAccQwenSevenTwoFiveVm}{42.6\xspace}
\newcommand{\oneAccQwenSevenTwoFiveMean}{44.0\xspace}
\newcommand{\oneWordsQwenSevenTwoFive}{19\xspace}

\newcommand{\oneAccLiveCcOvo}{38.4\xspace}
\newcommand{\oneAccLiveCcMvb}{37.3\xspace}
\newcommand{\oneAccLiveCcSb}{46.0\xspace}
\newcommand{\oneAccLiveCcVm}{41.1\xspace}
\newcommand{\oneAccLiveCcMean}{40.7\xspace}
\newcommand{\oneWordsLiveCc}{14\xspace}

\newcommand{\oneAccQwenEightOvo}{49.9\xspace}
\newcommand{\oneAccQwenEightMvb}{46.1\xspace}
\newcommand{\oneAccQwenEightSb}{55.9\xspace}
\newcommand{\oneAccQwenEightVm}{57.1\xspace}
\newcommand{\oneAccQwenEightMean}{52.3\xspace}
\newcommand{\oneWordsQwenEight}{39\xspace}

\newcommand{\oneAccTeacherOvo}{52.7\xspace}
\newcommand{\oneAccTeacherMvb}{47.2\xspace}
\newcommand{\oneAccTeacherSb}{54.8\xspace}
\newcommand{\oneAccTeacherVm}{54.0\xspace}
\newcommand{\oneAccTeacherMean}{52.2\xspace}
\newcommand{\oneWordsTeacher}{43\xspace}

\newcommand{\oneAccFloorOvo}{27.8\xspace}
\newcommand{\oneAccFloorMvb}{27.6\xspace}
\newcommand{\oneAccFloorSb}{33.5\xspace}
\newcommand{\oneAccFloorVm}{28.7\xspace}
\newcommand{\oneAccFloorMean}{29.4\xspace}

\newcommand{\oneOverLfmSmallMean}{+5.0\xspace}
\newcommand{\oneOverLfmSmallMeanLo}{+3.7\xspace}
\newcommand{\oneOverLfmSmallMeanHi}{+6.4\xspace}

\newcommand{\oneOverSmolSmallMean}{+4.9\xspace}
\newcommand{\oneOverSmolSmallMeanLo}{+3.6\xspace}
\newcommand{\oneOverSmolSmallMeanHi}{+6.2\xspace}

\newcommand{\oneOverMobileMean}{+8.2\xspace}
\newcommand{\oneOverMobileMeanLo}{+6.8\xspace}
\newcommand{\oneOverMobileMeanHi}{+9.6\xspace}

\newcommand{\oneLeadInternOneMean}{+3.0\xspace}

\newcommand{\oneOverLfmBigMean}{+3.5\xspace}
\newcommand{\oneOverLfmBigMeanLo}{+2.1\xspace}
\newcommand{\oneOverLfmBigMeanHi}{+4.9\xspace}

\newcommand{\oneLeadQwenTwoMean}{+4.8\xspace}

\newcommand{\oneOverSmolBigMean}{+1.7\xspace}
\newcommand{\oneOverSmolBigMeanLo}{+0.4\xspace}
\newcommand{\oneOverSmolBigMeanHi}{+3.0\xspace}

\newcommand{\oneLeadInternTwoMean}{+5.7\xspace}

\newcommand{\oneLeadQwenEightMean}{+6.8\xspace}

\newcommand{\oneCutOverLfmSmallMean}{+1.3\xspace}
\newcommand{\oneCutOverLfmSmallMeanLo}{+0.1\xspace}
\newcommand{\oneCutOverLfmSmallMeanHi}{+2.5\xspace}

\newcommand{\oneCutOverSmolSmallMean}{+1.3\xspace}
\newcommand{\oneCutOverSmolSmallMeanLo}{0.0\xspace}
\newcommand{\oneCutOverSmolSmallMeanHi}{+2.6\xspace}

\newcommand{\oneCutOverMobileMean}{+1.8\xspace}
\newcommand{\oneCutOverMobileMeanLo}{+0.6\xspace}
\newcommand{\oneCutOverMobileMeanHi}{+3.1\xspace}

\newcommand{\nbAccFastVlmOvo}{43.7\xspace}
\newcommand{\nbAccFastVlmMvb}{37.4\xspace}
\newcommand{\nbAccFastVlmSb}{47.4\xspace}
\newcommand{\nbAccFastVlmVm}{40.4\xspace}
\newcommand{\nbAccFastVlmMean}{42.3\xspace}
\newcommand{\nbWordsFastVlm}{42\xspace}

\newcommand{\nbOverFastVlmMean}{+3.3\xspace}
\newcommand{\nbOverFastVlmMeanLo}{+2.0\xspace}
\newcommand{\nbOverFastVlmMeanHi}{+4.6\xspace}

\newcommand{\nbAccStreamingVlmOvo}{43.5\xspace}
\newcommand{\nbAccStreamingVlmMvb}{44.4\xspace}
\newcommand{\nbAccStreamingVlmSb}{50.8\xspace}
\newcommand{\nbAccStreamingVlmVm}{49.7\xspace}
\newcommand{\nbAccStreamingVlmMean}{47.1\xspace}
\newcommand{\nbWordsStreamingVlm}{67\xspace}

\newcommand{\nbLeadStreamingVlmMean}{+1.5\xspace}
\newcommand{\nbLeadStreamingVlmMeanLo}{+0.1\xspace}
\newcommand{\nbLeadStreamingVlmMeanHi}{+3.0\xspace}
\newcommand{\nbAccTimeChatOvo}{49.2\xspace}
\newcommand{\nbAccTimeChatMvb}{42.1\xspace}
\newcommand{\nbAccTimeChatSb}{51.8\xspace}
\newcommand{\nbAccTimeChatVm}{55.9\xspace}
\newcommand{\nbAccTimeChatMean}{49.7\xspace}
\newcommand{\nbWordsTimeChat}{49\xspace}

\newcommand{\nbLeadTimeChatMean}{+4.2\xspace}
\newcommand{\nbLeadTimeChatMeanLo}{+2.8\xspace}
\newcommand{\nbLeadTimeChatMeanHi}{+5.5\xspace}
\newcommand{\nbTimeChatKept}{51\xspace}

\newcommand{\gaGainOursMean}{+2.5\xspace}
\newcommand{\gaGainOursMeanLo}{+1.2\xspace}
\newcommand{\gaGainOursMeanHi}{+3.7\xspace}
\newcommand{\gaGainLfmSmallMean}{$-$1.7\xspace}
\newcommand{\gaGainLfmSmallMeanLo}{$-$3.0\xspace}
\newcommand{\gaGainLfmSmallMeanHi}{$-$0.4\xspace}
\newcommand{\gaOverLfmSmallMean}{+4.2\xspace}
\newcommand{\gaOverLfmSmallMeanLo}{+2.5\xspace}
\newcommand{\gaOverLfmSmallMeanHi}{+5.9\xspace}

\newcommand{\gaOverSmolSmallMean}{+1.0\xspace}
\newcommand{\gaOverSmolSmallMeanLo}{$-$0.7\xspace}
\newcommand{\gaOverSmolSmallMeanHi}{+2.6\xspace}

\newcommand{\gaOverMobileMean}{+1.1\xspace}
\newcommand{\gaOverMobileMeanLo}{$-$0.6\xspace}
\newcommand{\gaOverMobileMeanHi}{+2.9\xspace}

\newcommand{\gaOverInternOneMean}{$-$0.2\xspace}
\newcommand{\gaOverInternOneMeanLo}{$-$1.9\xspace}
\newcommand{\gaOverInternOneMeanHi}{+1.5\xspace}

\newcommand{\gaOverLfmBigMean}{+6.3\xspace}
\newcommand{\gaOverLfmBigMeanLo}{+4.5\xspace}
\newcommand{\gaOverLfmBigMeanHi}{+8.1\xspace}

\newcommand{\gaOverQwenTwoMean}{$-$0.7\xspace}
\newcommand{\gaOverQwenTwoMeanLo}{$-$2.3\xspace}
\newcommand{\gaOverQwenTwoMeanHi}{+1.0\xspace}

\newcommand{\gaOverSmolBigMean}{+1.0\xspace}
\newcommand{\gaOverSmolBigMeanLo}{$-$0.7\xspace}
\newcommand{\gaOverSmolBigMeanHi}{+2.7\xspace}

\newcommand{\gaOverInternTwoMean}{$-$1.5\xspace}
\newcommand{\gaOverInternTwoMeanLo}{$-$3.2\xspace}
\newcommand{\gaOverInternTwoMeanHi}{+0.2\xspace}

\newcommand{\gaOverQwenFourMean}{$-$0.9\xspace}
\newcommand{\gaOverQwenFourMeanLo}{$-$2.7\xspace}
\newcommand{\gaOverQwenFourMeanHi}{+0.8\xspace}

\newcommand{\gaOverQwenSevenTwoFiveMean}{+5.2\xspace}
\newcommand{\gaOverQwenSevenTwoFiveMeanLo}{+3.4\xspace}
\newcommand{\gaOverQwenSevenTwoFiveMeanHi}{+7.0\xspace}

\newcommand{\gaOverLiveCcMean}{+3.3\xspace}
\newcommand{\gaOverLiveCcMeanLo}{+1.6\xspace}
\newcommand{\gaOverLiveCcMeanHi}{+5.1\xspace}

\newcommand{\gaOverQwenEightMean}{$-$1.6\xspace}
\newcommand{\gaOverQwenEightMeanLo}{$-$3.3\xspace}
\newcommand{\gaOverQwenEightMeanHi}{0.0\xspace}

\newcommand{\gaOverTeacherMean}{$-$4.3\xspace}
\newcommand{\gaOverTeacherMeanLo}{$-$6.0\xspace}
\newcommand{\gaOverTeacherMeanHi}{$-$2.6\xspace}

\newcommand{\gaOverFastVlmMean}{+3.0\xspace}
\newcommand{\gaOverFastVlmMeanLo}{+1.3\xspace}
\newcommand{\gaOverFastVlmMeanHi}{+4.7\xspace}

\newcommand{\gaOverStreamingVlmMean}{$-$3.5\xspace}
\newcommand{\gaOverStreamingVlmMeanLo}{$-$5.3\xspace}
\newcommand{\gaOverStreamingVlmMeanHi}{$-$1.7\xspace}

\newcommand{\gaOverTimeChatMean}{$-$0.8\xspace}
\newcommand{\gaOverTimeChatMeanLo}{$-$2.5\xspace}
\newcommand{\gaOverTimeChatMeanHi}{+0.9\xspace}

\newcommand{\cyPlainKzero}{52.7\xspace}
\newcommand{\cyPlainKsixteen}{72.2\xspace}
\newcommand{\cyRepeat}{52.2\xspace}
\newcommand{\cyForeign}{18.7\xspace}
\newcommand{\cyShuffled}{71.0\xspace}
\newcommand{\cyConsensusKsixteen}{73.5\xspace}
\newcommand{\cyConsensusKzero}{53.7\xspace}

\newcommand{\cyRepeatOverZero}{$-$0.6\xspace}
\newcommand{\cyRepeatOverZeroLo}{$-$1.8\xspace}
\newcommand{\cyRepeatOverZeroHi}{+0.7\xspace}
\newcommand{\cyRepeatDrop}{20.0\xspace}
\newcommand{\cyRepeatDropLo}{+18.2\xspace}
\newcommand{\cyRepeatDropHi}{+21.8\xspace}

\newcommand{\cyShuffledCost}{1.2\xspace}
\newcommand{\cyShuffledCostLo}{$-$0.1\xspace}
\newcommand{\cyShuffledCostHi}{+2.5\xspace}
\newcommand{\cyConsensusGapKsixteen}{+1.3\xspace}
\newcommand{\cyConsensusGapKsixteenLo}{+0.3\xspace}
\newcommand{\cyConsensusGapKsixteenHi}{+2.3\xspace}
\newcommand{\cyConsensusGapKzero}{+1.0\xspace}
\newcommand{\cyConsensusGapKzeroLo}{$-$0.1\xspace}
\newcommand{\cyConsensusGapKzeroHi}{+2.0\xspace}
\newcommand{\cyRepeatVsReal}{$-$20.0\xspace}
\newcommand{\cyRepeatVsRealLo}{$-$21.8\xspace}
\newcommand{\cyRepeatVsRealHi}{$-$18.2\xspace}
\newcommand{\cyForeignVsReal}{$-$53.5\xspace}
\newcommand{\cyForeignVsRealLo}{$-$55.5\xspace}
\newcommand{\cyForeignVsRealHi}{$-$51.5\xspace}
\newcommand{\cyShuffledVsReal}{$-$1.2\xspace}
\newcommand{\cyShuffledVsRealLo}{$-$2.5\xspace}
\newcommand{\cyShuffledVsRealHi}{+0.1\xspace}
\newcommand{\cyMbrOfGain}{7\xspace}

\newcommand{\tsInternOneKsixteen}{68.4\xspace}

\newcommand{\tsInternTwoKsixteen}{59.3\xspace}

\newcommand{\tsLfmBigKsixteen}{49.0\xspace}

\newcommand{\tsSmolBigKsixteen}{54.4\xspace}

\newcommand{\tsQwenSevenTwoFiveKsixteen}{44.1\xspace}

\newcommand{\tsLiveCcKsixteen}{46.5\xspace}

\newcommand{\jqTsInternOneKsixteen}{62.3\xspace}

\newcommand{\jqTsInternTwoKsixteen}{54.8\xspace}

\newcommand{\jqTsLfmBigKsixteen}{41.9\xspace}

\newcommand{\jqTsSmolBigKsixteen}{50.5\xspace}

\newcommand{\jqTsQwenSevenTwoFiveKsixteen}{38.3\xspace}

\newcommand{\jqTsLiveCcKsixteen}{41.4\xspace}

\newcommand{\fcTMobileKsixteen}{37.9\xspace}

\newcommand{\fcTInternOneIntKsixteen}{68.2\xspace}

\newcommand{\fcTQwenTwoIntKsixteen}{73.2\xspace}

\newcommand{\fcOMobileKzero}{39.2\xspace}
\newcommand{\fcOMobileKsixteen}{35.2\xspace}

\newcommand{\fcQTMobileKsixteen}{29.7\xspace}

\newcommand{\fcQTInternOneIntKsixteen}{62.8\xspace}

\newcommand{\fcQTQwenTwoIntKsixteen}{70.5\xspace}

\newcommand{\fcQOMobileKzero}{35.4\xspace}
\newcommand{\fcQOMobileKsixteen}{31.0\xspace}

\newcommand{\lengthCellOOursKzero}{83.8\xspace}

\newcommand{\lengthCellOOursKsixteen}{92.6\xspace}
\newcommand{\freeTOursKsixteen}{71.8\xspace}
\newcommand{\lengthCellTOursKsixteen}{71.8\xspace}

\newcommand{\lengthCellOLfmSmallKzero}{58.3\xspace}

\newcommand{\lengthCellOLfmSmallKsixteen}{56.7\xspace}
\newcommand{\freeTLfmSmallKsixteen}{46.7\xspace}
\newcommand{\lengthCellTLfmSmallKsixteen}{46.7\,(46.6)\xspace}

\newcommand{\lengthCellOSmolSmallKzero}{50.0\xspace}

\newcommand{\lengthCellOSmolSmallKsixteen}{53.4\,(52.8)\xspace}
\newcommand{\freeTSmolSmallKsixteen}{44.5\xspace}
\newcommand{\lengthCellTSmolSmallKsixteen}{44.5\,(43.5)\xspace}

\newcommand{\lengthCellOMobileKzero}{39.2\xspace}

\newcommand{\lengthCellOMobileKsixteen}{35.2\xspace}

\newcommand{\lengthCellTMobileKsixteen}{37.9\xspace}

\newcommand{\lengthCellOInternOneKzero}{67.8\xspace}

\newcommand{\lengthCellOInternOneKsixteen}{74.7\,(71.6)\xspace}

\newcommand{\lengthCellTInternOneKsixteen}{70.2\,(68.4)\xspace}

\newcommand{\lengthCellOInternOneIntKzero}{67.9\xspace}

\newcommand{\lengthCellOInternOneIntKsixteen}{74.9\,(72.0)\xspace}

\newcommand{\lengthCellTInternOneIntKsixteen}{70.3\,(68.2)\xspace}

\newcommand{\lengthCellOLfmBigKzero}{66.3\,(66.1)\xspace}

\newcommand{\lengthCellOLfmBigKsixteen}{51.7\,(49.9)\xspace}

\newcommand{\lengthCellTLfmBigKsixteen}{50.2\,(49.0)\xspace}

\newcommand{\lengthCellOQwenTwoKzero}{73.0\,(72.1)\xspace}

\newcommand{\lengthCellOQwenTwoKsixteen}{84.5\,(83.4)\xspace}

\newcommand{\lengthCellTQwenTwoKsixteen}{80.7\,(75.0)\xspace}

\newcommand{\lengthCellOQwenTwoIntKzero}{72.9\,(72.4)\xspace}

\newcommand{\lengthCellOQwenTwoIntKsixteen}{82.2\,(80.8)\xspace}

\newcommand{\lengthCellTQwenTwoIntKsixteen}{77.6\,(73.2)\xspace}

\newcommand{\lengthCellOSmolBigKzero}{59.9\,(59.8)\xspace}

\newcommand{\lengthCellOSmolBigKsixteen}{68.1\,(66.9)\xspace}

\newcommand{\lengthCellTSmolBigKsixteen}{61.2\,(54.4)\xspace}

\newcommand{\lengthCellOInternTwoKzero}{71.3\,(67.9)\xspace}

\newcommand{\lengthCellOInternTwoKsixteen}{80.1\,(66.9)\xspace}
\newcommand{\freeTInternTwoKsixteen}{78.3\xspace}
\newcommand{\lengthCellTInternTwoKsixteen}{78.3\,(59.3)\xspace}

\newcommand{\lengthCellOQwenFourKzero}{74.3\xspace}

\newcommand{\lengthCellOQwenFourKsixteen}{86.1\xspace}

\newcommand{\lengthCellTQwenFourKsixteen}{86.8\xspace}

\newcommand{\lengthCellOQwenSevenTwoFiveKzero}{67.6\,(67.3)\xspace}

\newcommand{\lengthCellOQwenSevenTwoFiveKsixteen}{57.1\,(56.0)\xspace}

\newcommand{\lengthCellTQwenSevenTwoFiveKsixteen}{47.9\,(44.1)\xspace}

\newcommand{\lengthCellOLiveCcKzero}{49.3\xspace}

\newcommand{\lengthCellOLiveCcKsixteen}{51.5\,(49.2)\xspace}

\newcommand{\lengthCellTLiveCcKsixteen}{48.9\,(46.5)\xspace}

\newcommand{\lengthCellOQwenEightKzero}{73.4\xspace}

\newcommand{\lengthCellOQwenEightKsixteen}{87.3\xspace}

\newcommand{\lengthCellTQwenEightKsixteen}{86.8\xspace}

\newcommand{\nfOFastVlmKzero}{41.9\xspace}

\newcommand{\lengthCellOFastVlmKzero}{57.0\,(41.9)\xspace}
\newcommand{\nfOStreamingVlmKzero}{1.7\xspace}

\newcommand{\lengthCellOStreamingVlmKzero}{14.9\,(1.7)\xspace}
\newcommand{\nfOTimeChatKzero}{66.1\xspace}

\newcommand{\lengthCellOTimeChatKzero}{68.9\,(66.1)\xspace}
\newcommand{\nfOFastVlmKsixteen}{36.7\xspace}

\newcommand{\lengthCellOFastVlmKsixteen}{54.9\,(36.7)\xspace}
\newcommand{\nfOStreamingVlmKsixteen}{1.6\xspace}

\newcommand{\lengthCellOStreamingVlmKsixteen}{21.9\,(1.6)\xspace}
\newcommand{\nfOTimeChatKsixteen}{73.6\xspace}

\newcommand{\lengthCellOTimeChatKsixteen}{79.1\,(73.6)\xspace}
\newcommand{\nfTFastVlmKsixteen}{31.8\xspace}

\newcommand{\lengthCellTFastVlmKsixteen}{47.2\,(31.8)\xspace}
\newcommand{\nfTStreamingVlmKsixteen}{3.1\xspace}

\newcommand{\lengthCellTStreamingVlmKsixteen}{41.0\,(3.1)\xspace}
\newcommand{\nfTTimeChatKsixteen}{55.2\xspace}
\newcommand{\freeTTimeChatKsixteen}{74.3\xspace}
\newcommand{\lengthCellTTimeChatKsixteen}{74.3\,(55.2)\xspace}
\newcommand{\nfQOFastVlmKzero}{38.1\xspace}

\newcommand{\nfQOStreamingVlmKzero}{1.4\xspace}

\newcommand{\nfQOTimeChatKzero}{63.8\xspace}

\newcommand{\nfQOFastVlmKsixteen}{32.7\xspace}

\newcommand{\nfQOStreamingVlmKsixteen}{1.4\xspace}

\newcommand{\nfQOTimeChatKsixteen}{71.6\xspace}

\newcommand{\nfQTFastVlmKsixteen}{27.3\xspace}

\newcommand{\nfQTStreamingVlmKsixteen}{2.5\xspace}

\newcommand{\nfQTTimeChatKsixteen}{52.6\xspace}

\newcommand{\eyPNinetyNineMin}{12.66\xspace}
\newcommand{\eyPNinetyNineMax}{12.69\xspace}

\newcommand{\eyGB}{1.05\xspace}
\newcommand{\eyHeadRows}{16,575\xspace}
\newcommand{\eyHeadCover}{99.9\xspace}

\newcommand{\eyLmGB}{0.61\xspace}
\newcommand{\eyTokenMs}{1.9\xspace}

\newcommand{\efYoloSteps}{4\xspace}
\newcommand{\efYoloPNinetyNineMin}{12.44\xspace}
\newcommand{\efYoloPNinetyNineMax}{12.66\xspace}

\newcommand{\efYoloLagKeightSecMax}{0.43\xspace}

\newcommand{\efYoloGpuKeightMax}{0.17\xspace}

\newcommand{\efYoloLagKsixteenSecMax}{0.50\xspace}

\newcommand{\efYoloGpuKsixteenMax}{0.19\xspace}
\newcommand{\efLfmSteps}{3\xspace}
\newcommand{\efLfmPNinetyNineMin}{11.74\xspace}
\newcommand{\efLfmPNinetyNineMax}{11.90\xspace}

\newcommand{\efLfmLagKeightSecMax}{0.73\xspace}

\newcommand{\efLfmGpuKeightMax}{0.26\xspace}

\newcommand{\efLfmLagKsixteenSecMax}{0.93\xspace}

\newcommand{\efLfmGpuKsixteenMax}{0.33\xspace}
\newcommand{\efLfmTwoSteps}{2\xspace}
\newcommand{\efLfmTwoPNinetyNineMin}{11.44\xspace}
\newcommand{\efLfmTwoPNinetyNineMax}{11.84\xspace}

\newcommand{\efLfmTwoLagKeightSecMax}{0.97\xspace}

\newcommand{\efLfmTwoGpuKeightMax}{0.26\xspace}

\newcommand{\efSmolSteps}{2\xspace}
\newcommand{\efSmolPNinetyNineMin}{15.09\xspace}
\newcommand{\efSmolPNinetyNineMax}{19.19\xspace}

\newcommand{\stOneMin}{14.3\xspace}
\newcommand{\stOneMax}{15.0\xspace}
\newcommand{\stOneHostMax}{15.9\xspace}

\newcommand{\stThreeMax}{13.2\xspace}
\newcommand{\stThreeHostMax}{13.8\xspace}
\newcommand{\stFourMin}{17.3\xspace}

\newcommand{\stAlignedTwoMin}{26.7\xspace}
\newcommand{\parMobile}{0.66\xspace}

\newcommand{\parFastVlm}{0.62\xspace}

\newcommand{\parStreamingVlm}{8.29\xspace}

\newcommand{\parTimeChat}{8.29\xspace}

\newcommand{\teacherRecallKsixteen}{0.841\xspace}
\newcommand{\teacherPrecisionKsixteen}{0.887\xspace}
\newcommand{\promptPilotWindows}{100\xspace}
\newcommand{\rlRollouts}{512\xspace}
\newcommand{\rlGroupSize}{8\xspace}
\newcommand{\rlBeta}{0.04\xspace}
\newcommand{\rlMaxProb}{0.5\xspace}
\newcommand{\rlTemperature}{1\xspace}
\newcommand{\rlMaxNew}{80\xspace}
\newcommand{\rlClip}{0.2\xspace}
\newcommand{\fluencyCaptions}{20,000\xspace}
\newcommand{\fluencyBGood}{0.9339\xspace}
\newcommand{\fluencyBShuffled}{3.7250\xspace}
\newcommand{\fluencyContentThreshold}{5\xspace}
\newcommand{\fluencyBigramThreshold}{3\xspace}

\newcommand{\latSecOursKzero}{1.17\xspace}
\newcommand{\latSecOursKsixteen}{1.13\xspace}

\newcommand{\runGBOurs}{1.24\xspace}
\newcommand{\latSecLfmSmallKzero}{1.27\xspace}
\newcommand{\latSecLfmSmallKsixteen}{1.35\xspace}

\newcommand{\runGBLfmSmall}{1.43\xspace}
\newcommand{\latSecSmolSmallKzero}{2.59\xspace}
\newcommand{\latSecSmolSmallKsixteen}{3.13\xspace}

\newcommand{\runGBSmolSmall}{2.10\xspace}
\newcommand{\latSecMobileKzero}{2.24\xspace}
\newcommand{\latSecMobileKsixteen}{2.25\xspace}

\newcommand{\runGBMobile}{3.23\xspace}
\newcommand{\latSecInternOneKzero}{2.81\xspace}
\newcommand{\latSecInternOneKsixteen}{7.25\xspace}

\newcommand{\runGBInternOne}{6.25\xspace}
\newcommand{\latSecLfmBigKzero}{1.33\xspace}
\newcommand{\latSecLfmBigKsixteen}{1.74\xspace}

\newcommand{\runGBLfmBig}{3.98\xspace}
\newcommand{\latSecQwenTwoKzero}{2.84\xspace}
\newcommand{\latSecQwenTwoKsixteen}{3.12\xspace}

\newcommand{\runGBQwenTwo}{4.67\xspace}
\newcommand{\latSecSmolBigKzero}{1.65\xspace}
\newcommand{\latSecSmolBigKsixteen}{2.73\xspace}

\newcommand{\runGBSmolBig}{5.43\xspace}
\newcommand{\latSecInternTwoKzero}{2.89\xspace}
\newcommand{\latSecInternTwoKsixteen}{8.17\xspace}

\newcommand{\runGBInternTwo}{8.83\xspace}
\newcommand{\latSecQwenFourKzero}{3.86\xspace}
\newcommand{\latSecQwenFourKsixteen}{4.39\xspace}

\newcommand{\runGBQwenFour}{9.51\xspace}
\newcommand{\latPNinetyNineOursKsixteen}{1.24\xspace}
\newcommand{\latSecInternOneIntKzero}{0.38\xspace}
\newcommand{\latSecInternOneIntKsixteen}{4.87\xspace}

\newcommand{\runGBInternOneInt}{4.47\xspace}
\newcommand{\latSecQwenTwoIntKzero}{0.47\xspace}
\newcommand{\latSecQwenTwoIntKsixteen}{0.84\xspace}

\newcommand{\runGBQwenTwoInt}{5.24\xspace}

\newcommand{\tFourGB}{16.11\xspace}
\newcommand{\wgtGBQwenSevenTwoFive}{16.58\xspace}
\newcommand{\wgtGBLiveCc}{16.58\xspace}
\newcommand{\wgtGBQwenEight}{17.53\xspace}
\newcommand{\wgtGBTeacher}{27.78\xspace}
\newcommand{\wgtGBStreamingVlm}{16.58\xspace}
\newcommand{\wgtGBTimeChat}{16.58\xspace}
\newcommand{\latSecFastVlmKzero}{1.54\xspace}
\newcommand{\latSecFastVlmKsixteen}{3.30\xspace}

\newcommand{\runGBFastVlm}{5.46\xspace}

%% file: numbers/extra.tex
\newcommand{\sftTrainTen}{941,483\xspace}

\newcommand{\sftOpenvidTen}{862,267\xspace}

\newcommand{\rlTen}{39,502\xspace}

\newcommand{\sftStepsTen}{7,354\xspace}
\newcommand{\sftBestTen}{7,250\xspace}
\newcommand{\rlTripTen}{150\xspace}
\newcommand{\rlExitTen}{450\xspace}
\newcommand{\rlBestTen}{400\xspace}

%% file: numbers/method.tex
\newcommand{\poolerParams}{1.05M\xspace}
\newcommand{\lmParams}{354.5M\xspace}

\newcommand{\totalParams}{449.8M\xspace}
\newcommand{\templateTokens}{13\xspace}
\newcommand{\captionTokens}{52\xspace}

\newcommand{\gpuMsPerSecond}{480\xspace}
\newcommand{\qwenTwoLmParams}{1.72B\xspace}
\newcommand{\qwenTwoLmGB}{3.44\xspace}
\newcommand{\qwenTwoTokenMs}{10.8\xspace}
\newcommand{\qwenTwoCaptionMs}{559\xspace}

\newcommand{\sftTrain}{79,216\xspace}

\newcommand{\sftSources}{11\xspace}

\newcommand{\rlVal}{526\xspace}

\newcommand{\sftSteps}{618\xspace}
\newcommand{\sftBatch}{256\xspace}
\newcommand{\sftEpochs}{two\xspace}
\newcommand{\sftLr}{$10^{-5}$\xspace}
\newcommand{\sftVisionLr}{$2{\times}10^{-6}$\xspace}

\newcommand{\preKmax}{5\xspace}

\newcommand{\rlWindowsStep}{64\xspace}
\newcommand{\rlGroup}{eight\xspace}
\newcommand{\rlLr}{$10^{-6}$\xspace}
\newcommand{\rlVisionLr}{$2{\times}10^{-7}$\xspace}
\newcommand{\rlEvalEvery}{50\xspace}

%% file: numbers/results.tex
\newcommand{\parOurs}{0.45\xspace}

\newcommand{\parLfmSmall}{0.45\xspace}

\newcommand{\parSmolSmall}{0.51\xspace}

\newcommand{\parInternOne}{1.06\xspace}

\newcommand{\parInternTwo}{2.35\xspace}

\newcommand{\parQwenTwo}{2.13\xspace}

\newcommand{\parQwenFour}{4.44\xspace}

\newcommand{\parSmolBig}{2.25\xspace}

\newcommand{\parLfmBig}{1.60\xspace}

\newcommand{\parQwenEight}{8.77\xspace}

\newcommand{\parQwenSevenTwoFive}{8.29\xspace}

\newcommand{\parLiveCc}{8.29\xspace}

\newcommand{\parTeacher}{27.78\xspace}

\newcommand{\twoSecWindows}{771\xspace}

%% file: numbers/figures.tex
\newcommand{\examplesMedianF}{0.75\xspace}
\newcommand{\examplesMedianGain}{$+$0.05\xspace}
\newcommand{\examplesPool}{16\xspace}
\newcommand{\examplesGainPos}{11\xspace}

%% file: numbers/supp_bench.tex
\newcommand{\sbTaskNOvoOjr}{184\xspace}
\newcommand{\sbTaskFloorOvoOjr}{26.1\xspace}
\newcommand{\sbTaskOursOvoOjr}{44.6\xspace}
\newcommand{\sbTaskLfmSmallOvoOjr}{38.0\xspace}
\newcommand{\sbTaskSmolSmallOvoOjr}{43.5\xspace}
\newcommand{\sbTaskFastVlmOvoOjr}{43.5\xspace}
\newcommand{\sbTaskMobileOvoOjr}{33.2\xspace}
\newcommand{\sbTaskInternOneOvoOjr}{41.3\xspace}
\newcommand{\sbTaskInternOneIntOvoOjr}{46.2\xspace}
\newcommand{\sbTaskLfmBigOvoOjr}{32.1\xspace}
\newcommand{\sbTaskQwenTwoOvoOjr}{50.0\xspace}
\newcommand{\sbTaskQwenTwoIntOvoOjr}{45.1\xspace}
\newcommand{\sbTaskSmolBigOvoOjr}{47.8\xspace}
\newcommand{\sbTaskInternTwoOvoOjr}{48.4\xspace}
\newcommand{\sbTaskQwenFourOvoOjr}{48.4\xspace}
\newcommand{\sbTaskQwenSevenTwoFiveOvoOjr}{46.2\xspace}
\newcommand{\sbTaskLiveCcOvoOjr}{35.3\xspace}
\newcommand{\sbTaskStreamingVlmOvoOjr}{38.6\xspace}
\newcommand{\sbTaskTimeChatOvoOjr}{43.5\xspace}
\newcommand{\sbTaskQwenEightOvoOjr}{46.2\xspace}
\newcommand{\sbTaskTeacherOvoOjr}{53.3\xspace}
\newcommand{\sbTaskNOvoStu}{178\xspace}
\newcommand{\sbTaskFloorOvoStu}{28.7\xspace}
\newcommand{\sbTaskOursOvoStu}{39.3\xspace}
\newcommand{\sbTaskLfmSmallOvoStu}{32.0\xspace}
\newcommand{\sbTaskSmolSmallOvoStu}{31.5\xspace}
\newcommand{\sbTaskFastVlmOvoStu}{34.3\xspace}
\newcommand{\sbTaskMobileOvoStu}{27.0\xspace}
\newcommand{\sbTaskInternOneOvoStu}{35.4\xspace}
\newcommand{\sbTaskInternOneIntOvoStu}{35.4\xspace}
\newcommand{\sbTaskLfmBigOvoStu}{33.1\xspace}
\newcommand{\sbTaskQwenTwoOvoStu}{39.9\xspace}
\newcommand{\sbTaskQwenTwoIntOvoStu}{41.0\xspace}
\newcommand{\sbTaskSmolBigOvoStu}{33.1\xspace}
\newcommand{\sbTaskInternTwoOvoStu}{35.4\xspace}
\newcommand{\sbTaskQwenFourOvoStu}{44.9\xspace}
\newcommand{\sbTaskQwenSevenTwoFiveOvoStu}{38.2\xspace}
\newcommand{\sbTaskLiveCcOvoStu}{28.7\xspace}
\newcommand{\sbTaskStreamingVlmOvoStu}{37.6\xspace}
\newcommand{\sbTaskTimeChatOvoStu}{44.9\xspace}
\newcommand{\sbTaskQwenEightOvoStu}{41.6\xspace}
\newcommand{\sbTaskTeacherOvoStu}{50.6\xspace}
\newcommand{\sbTaskNOvoOcr}{149\xspace}
\newcommand{\sbTaskFloorOvoOcr}{28.2\xspace}
\newcommand{\sbTaskOursOvoOcr}{40.9\xspace}
\newcommand{\sbTaskLfmSmallOvoOcr}{32.9\xspace}
\newcommand{\sbTaskSmolSmallOvoOcr}{37.6\xspace}
\newcommand{\sbTaskFastVlmOvoOcr}{40.3\xspace}
\newcommand{\sbTaskMobileOvoOcr}{30.9\xspace}
\newcommand{\sbTaskInternOneOvoOcr}{47.0\xspace}
\newcommand{\sbTaskInternOneIntOvoOcr}{50.3\xspace}
\newcommand{\sbTaskLfmBigOvoOcr}{33.6\xspace}
\newcommand{\sbTaskQwenTwoOvoOcr}{47.7\xspace}
\newcommand{\sbTaskQwenTwoIntOvoOcr}{41.6\xspace}
\newcommand{\sbTaskSmolBigOvoOcr}{34.9\xspace}
\newcommand{\sbTaskInternTwoOvoOcr}{53.7\xspace}
\newcommand{\sbTaskQwenFourOvoOcr}{45.6\xspace}
\newcommand{\sbTaskQwenSevenTwoFiveOvoOcr}{36.9\xspace}
\newcommand{\sbTaskLiveCcOvoOcr}{33.6\xspace}
\newcommand{\sbTaskStreamingVlmOvoOcr}{44.3\xspace}
\newcommand{\sbTaskTimeChatOvoOcr}{52.3\xspace}
\newcommand{\sbTaskQwenEightOvoOcr}{49.0\xspace}
\newcommand{\sbTaskTeacherOvoOcr}{47.0\xspace}
\newcommand{\sbTaskNOvoAtr}{116\xspace}
\newcommand{\sbTaskFloorOvoAtr}{33.6\xspace}
\newcommand{\sbTaskOursOvoAtr}{52.6\xspace}
\newcommand{\sbTaskLfmSmallOvoAtr}{42.2\xspace}
\newcommand{\sbTaskSmolSmallOvoAtr}{36.2\xspace}
\newcommand{\sbTaskFastVlmOvoAtr}{43.1\xspace}
\newcommand{\sbTaskMobileOvoAtr}{39.7\xspace}
\newcommand{\sbTaskInternOneOvoAtr}{47.4\xspace}
\newcommand{\sbTaskInternOneIntOvoAtr}{54.3\xspace}
\newcommand{\sbTaskLfmBigOvoAtr}{48.3\xspace}
\newcommand{\sbTaskQwenTwoOvoAtr}{47.4\xspace}
\newcommand{\sbTaskQwenTwoIntOvoAtr}{44.0\xspace}
\newcommand{\sbTaskSmolBigOvoAtr}{50.9\xspace}
\newcommand{\sbTaskInternTwoOvoAtr}{46.6\xspace}
\newcommand{\sbTaskQwenFourOvoAtr}{52.6\xspace}
\newcommand{\sbTaskQwenSevenTwoFiveOvoAtr}{45.7\xspace}
\newcommand{\sbTaskLiveCcOvoAtr}{42.2\xspace}
\newcommand{\sbTaskStreamingVlmOvoAtr}{46.6\xspace}
\newcommand{\sbTaskTimeChatOvoAtr}{48.3\xspace}
\newcommand{\sbTaskQwenEightOvoAtr}{51.7\xspace}
\newcommand{\sbTaskTeacherOvoAtr}{53.4\xspace}
\newcommand{\sbTaskNOvoAcr}{109\xspace}
\newcommand{\sbTaskFloorOvoAcr}{26.6\xspace}
\newcommand{\sbTaskOursOvoAcr}{48.6\xspace}
\newcommand{\sbTaskLfmSmallOvoAcr}{39.4\xspace}
\newcommand{\sbTaskSmolSmallOvoAcr}{45.9\xspace}
\newcommand{\sbTaskFastVlmOvoAcr}{50.5\xspace}
\newcommand{\sbTaskMobileOvoAcr}{40.4\xspace}
\newcommand{\sbTaskInternOneOvoAcr}{57.8\xspace}
\newcommand{\sbTaskInternOneIntOvoAcr}{56.0\xspace}
\newcommand{\sbTaskLfmBigOvoAcr}{37.6\xspace}
\newcommand{\sbTaskQwenTwoOvoAcr}{56.0\xspace}
\newcommand{\sbTaskQwenTwoIntOvoAcr}{59.6\xspace}
\newcommand{\sbTaskSmolBigOvoAcr}{46.8\xspace}
\newcommand{\sbTaskInternTwoOvoAcr}{60.6\xspace}
\newcommand{\sbTaskQwenFourOvoAcr}{51.4\xspace}
\newcommand{\sbTaskQwenSevenTwoFiveOvoAcr}{54.1\xspace}
\newcommand{\sbTaskLiveCcOvoAcr}{41.3\xspace}
\newcommand{\sbTaskStreamingVlmOvoAcr}{49.5\xspace}
\newcommand{\sbTaskTimeChatOvoAcr}{52.3\xspace}
\newcommand{\sbTaskQwenEightOvoAcr}{54.1\xspace}
\newcommand{\sbTaskTeacherOvoAcr}{53.2\xspace}
\newcommand{\sbTaskNOvoFpd}{101\xspace}
\newcommand{\sbTaskFloorOvoFpd}{23.8\xspace}
\newcommand{\sbTaskOursOvoFpd}{56.4\xspace}
\newcommand{\sbTaskLfmSmallOvoFpd}{54.5\xspace}
\newcommand{\sbTaskSmolSmallOvoFpd}{52.5\xspace}
\newcommand{\sbTaskFastVlmOvoFpd}{59.4\xspace}
\newcommand{\sbTaskMobileOvoFpd}{50.5\xspace}
\newcommand{\sbTaskInternOneOvoFpd}{60.4\xspace}
\newcommand{\sbTaskInternOneIntOvoFpd}{62.4\xspace}
\newcommand{\sbTaskLfmBigOvoFpd}{60.4\xspace}
\newcommand{\sbTaskQwenTwoOvoFpd}{63.4\xspace}
\newcommand{\sbTaskQwenTwoIntOvoFpd}{62.4\xspace}
\newcommand{\sbTaskSmolBigOvoFpd}{57.4\xspace}
\newcommand{\sbTaskInternTwoOvoFpd}{59.4\xspace}
\newcommand{\sbTaskQwenFourOvoFpd}{60.4\xspace}
\newcommand{\sbTaskQwenSevenTwoFiveOvoFpd}{53.5\xspace}
\newcommand{\sbTaskLiveCcOvoFpd}{60.4\xspace}
\newcommand{\sbTaskStreamingVlmOvoFpd}{51.5\xspace}
\newcommand{\sbTaskTimeChatOvoFpd}{60.4\xspace}
\newcommand{\sbTaskQwenEightOvoFpd}{66.3\xspace}
\newcommand{\sbTaskTeacherOvoFpd}{62.4\xspace}

\newcommand{\sbTaskNMvbActionSequence}{200\xspace}
\newcommand{\sbTaskFloorMvbActionSequence}{29.0\xspace}
\newcommand{\sbTaskOursMvbActionSequence}{51.0\xspace}
\newcommand{\sbTaskLfmSmallMvbActionSequence}{43.0\xspace}
\newcommand{\sbTaskSmolSmallMvbActionSequence}{47.5\xspace}
\newcommand{\sbTaskFastVlmMvbActionSequence}{40.5\xspace}
\newcommand{\sbTaskMobileMvbActionSequence}{39.0\xspace}
\newcommand{\sbTaskInternOneMvbActionSequence}{59.0\xspace}
\newcommand{\sbTaskInternOneIntMvbActionSequence}{60.0\xspace}
\newcommand{\sbTaskLfmBigMvbActionSequence}{48.5\xspace}
\newcommand{\sbTaskQwenTwoMvbActionSequence}{55.5\xspace}
\newcommand{\sbTaskQwenTwoIntMvbActionSequence}{55.0\xspace}
\newcommand{\sbTaskSmolBigMvbActionSequence}{52.0\xspace}
\newcommand{\sbTaskInternTwoMvbActionSequence}{62.0\xspace}
\newcommand{\sbTaskQwenFourMvbActionSequence}{57.5\xspace}
\newcommand{\sbTaskQwenSevenTwoFiveMvbActionSequence}{47.0\xspace}
\newcommand{\sbTaskLiveCcMvbActionSequence}{51.0\xspace}
\newcommand{\sbTaskStreamingVlmMvbActionSequence}{51.0\xspace}
\newcommand{\sbTaskTimeChatMvbActionSequence}{46.0\xspace}
\newcommand{\sbTaskQwenEightMvbActionSequence}{55.5\xspace}
\newcommand{\sbTaskTeacherMvbActionSequence}{62.0\xspace}
\newcommand{\sbTaskNMvbActionPrediction}{200\xspace}
\newcommand{\sbTaskFloorMvbActionPrediction}{33.5\xspace}
\newcommand{\sbTaskOursMvbActionPrediction}{42.0\xspace}
\newcommand{\sbTaskLfmSmallMvbActionPrediction}{32.0\xspace}
\newcommand{\sbTaskSmolSmallMvbActionPrediction}{46.0\xspace}
\newcommand{\sbTaskFastVlmMvbActionPrediction}{39.0\xspace}
\newcommand{\sbTaskMobileMvbActionPrediction}{35.0\xspace}
\newcommand{\sbTaskInternOneMvbActionPrediction}{50.0\xspace}
\newcommand{\sbTaskInternOneIntMvbActionPrediction}{52.5\xspace}
\newcommand{\sbTaskLfmBigMvbActionPrediction}{38.5\xspace}
\newcommand{\sbTaskQwenTwoMvbActionPrediction}{47.5\xspace}
\newcommand{\sbTaskQwenTwoIntMvbActionPrediction}{46.5\xspace}
\newcommand{\sbTaskSmolBigMvbActionPrediction}{49.0\xspace}
\newcommand{\sbTaskInternTwoMvbActionPrediction}{52.5\xspace}
\newcommand{\sbTaskQwenFourMvbActionPrediction}{46.0\xspace}
\newcommand{\sbTaskQwenSevenTwoFiveMvbActionPrediction}{47.0\xspace}
\newcommand{\sbTaskLiveCcMvbActionPrediction}{38.0\xspace}
\newcommand{\sbTaskStreamingVlmMvbActionPrediction}{37.5\xspace}
\newcommand{\sbTaskTimeChatMvbActionPrediction}{41.0\xspace}
\newcommand{\sbTaskQwenEightMvbActionPrediction}{49.0\xspace}
\newcommand{\sbTaskTeacherMvbActionPrediction}{49.0\xspace}
\newcommand{\sbTaskNMvbObjectInteraction}{200\xspace}
\newcommand{\sbTaskFloorMvbObjectInteraction}{19.5\xspace}
\newcommand{\sbTaskOursMvbObjectInteraction}{48.5\xspace}
\newcommand{\sbTaskLfmSmallMvbObjectInteraction}{40.5\xspace}
\newcommand{\sbTaskSmolSmallMvbObjectInteraction}{46.5\xspace}
\newcommand{\sbTaskFastVlmMvbObjectInteraction}{40.0\xspace}
\newcommand{\sbTaskMobileMvbObjectInteraction}{36.0\xspace}
\newcommand{\sbTaskInternOneMvbObjectInteraction}{64.5\xspace}
\newcommand{\sbTaskInternOneIntMvbObjectInteraction}{64.5\xspace}
\newcommand{\sbTaskLfmBigMvbObjectInteraction}{45.0\xspace}
\newcommand{\sbTaskQwenTwoMvbObjectInteraction}{61.0\xspace}
\newcommand{\sbTaskQwenTwoIntMvbObjectInteraction}{61.0\xspace}
\newcommand{\sbTaskSmolBigMvbObjectInteraction}{48.0\xspace}
\newcommand{\sbTaskInternTwoMvbObjectInteraction}{63.0\xspace}
\newcommand{\sbTaskQwenFourMvbObjectInteraction}{55.5\xspace}
\newcommand{\sbTaskQwenSevenTwoFiveMvbObjectInteraction}{49.5\xspace}
\newcommand{\sbTaskLiveCcMvbObjectInteraction}{48.5\xspace}
\newcommand{\sbTaskStreamingVlmMvbObjectInteraction}{54.5\xspace}
\newcommand{\sbTaskTimeChatMvbObjectInteraction}{47.0\xspace}
\newcommand{\sbTaskQwenEightMvbObjectInteraction}{58.0\xspace}
\newcommand{\sbTaskTeacherMvbObjectInteraction}{59.5\xspace}
\newcommand{\sbTaskNMvbActionAntonym}{200\xspace}
\newcommand{\sbTaskFloorMvbActionAntonym}{0.0\xspace}
\newcommand{\sbTaskOursMvbActionAntonym}{22.0\xspace}
\newcommand{\sbTaskLfmSmallMvbActionAntonym}{15.5\xspace}
\newcommand{\sbTaskSmolSmallMvbActionAntonym}{19.5\xspace}
\newcommand{\sbTaskFastVlmMvbActionAntonym}{22.0\xspace}
\newcommand{\sbTaskMobileMvbActionAntonym}{12.5\xspace}
\newcommand{\sbTaskInternOneMvbActionAntonym}{30.0\xspace}
\newcommand{\sbTaskInternOneIntMvbActionAntonym}{29.0\xspace}
\newcommand{\sbTaskLfmBigMvbActionAntonym}{24.5\xspace}
\newcommand{\sbTaskQwenTwoMvbActionAntonym}{27.0\xspace}
\newcommand{\sbTaskQwenTwoIntMvbActionAntonym}{30.5\xspace}
\newcommand{\sbTaskSmolBigMvbActionAntonym}{23.5\xspace}
\newcommand{\sbTaskInternTwoMvbActionAntonym}{35.0\xspace}
\newcommand{\sbTaskQwenFourMvbActionAntonym}{36.0\xspace}
\newcommand{\sbTaskQwenSevenTwoFiveMvbActionAntonym}{32.0\xspace}
\newcommand{\sbTaskLiveCcMvbActionAntonym}{26.0\xspace}
\newcommand{\sbTaskStreamingVlmMvbActionAntonym}{30.0\xspace}
\newcommand{\sbTaskTimeChatMvbActionAntonym}{31.5\xspace}
\newcommand{\sbTaskQwenEightMvbActionAntonym}{36.0\xspace}
\newcommand{\sbTaskTeacherMvbActionAntonym}{36.5\xspace}
\newcommand{\sbTaskNMvbFineGrainedAction}{200\xspace}
\newcommand{\sbTaskFloorMvbFineGrainedAction}{11.5\xspace}
\newcommand{\sbTaskOursMvbFineGrainedAction}{30.0\xspace}
\newcommand{\sbTaskLfmSmallMvbFineGrainedAction}{36.0\xspace}
\newcommand{\sbTaskSmolSmallMvbFineGrainedAction}{27.5\xspace}
\newcommand{\sbTaskFastVlmMvbFineGrainedAction}{29.0\xspace}
\newcommand{\sbTaskMobileMvbFineGrainedAction}{24.5\xspace}
\newcommand{\sbTaskInternOneMvbFineGrainedAction}{35.0\xspace}
\newcommand{\sbTaskInternOneIntMvbFineGrainedAction}{35.5\xspace}
\newcommand{\sbTaskLfmBigMvbFineGrainedAction}{34.5\xspace}
\newcommand{\sbTaskQwenTwoMvbFineGrainedAction}{39.5\xspace}
\newcommand{\sbTaskQwenTwoIntMvbFineGrainedAction}{39.0\xspace}
\newcommand{\sbTaskSmolBigMvbFineGrainedAction}{33.0\xspace}
\newcommand{\sbTaskInternTwoMvbFineGrainedAction}{37.0\xspace}
\newcommand{\sbTaskQwenFourMvbFineGrainedAction}{35.5\xspace}
\newcommand{\sbTaskQwenSevenTwoFiveMvbFineGrainedAction}{35.0\xspace}
\newcommand{\sbTaskLiveCcMvbFineGrainedAction}{31.5\xspace}
\newcommand{\sbTaskStreamingVlmMvbFineGrainedAction}{37.0\xspace}
\newcommand{\sbTaskTimeChatMvbFineGrainedAction}{31.0\xspace}
\newcommand{\sbTaskQwenEightMvbFineGrainedAction}{36.5\xspace}
\newcommand{\sbTaskTeacherMvbFineGrainedAction}{36.5\xspace}
\newcommand{\sbTaskNMvbUnexpectedAction}{200\xspace}
\newcommand{\sbTaskFloorMvbUnexpectedAction}{40.0\xspace}
\newcommand{\sbTaskOursMvbUnexpectedAction}{45.5\xspace}
\newcommand{\sbTaskLfmSmallMvbUnexpectedAction}{43.0\xspace}
\newcommand{\sbTaskSmolSmallMvbUnexpectedAction}{39.5\xspace}
\newcommand{\sbTaskFastVlmMvbUnexpectedAction}{51.5\xspace}
\newcommand{\sbTaskMobileMvbUnexpectedAction}{47.0\xspace}
\newcommand{\sbTaskInternOneMvbUnexpectedAction}{50.5\xspace}
\newcommand{\sbTaskInternOneIntMvbUnexpectedAction}{51.0\xspace}
\newcommand{\sbTaskLfmBigMvbUnexpectedAction}{42.5\xspace}
\newcommand{\sbTaskQwenTwoMvbUnexpectedAction}{55.0\xspace}
\newcommand{\sbTaskQwenTwoIntMvbUnexpectedAction}{51.5\xspace}
\newcommand{\sbTaskSmolBigMvbUnexpectedAction}{45.0\xspace}
\newcommand{\sbTaskInternTwoMvbUnexpectedAction}{53.0\xspace}
\newcommand{\sbTaskQwenFourMvbUnexpectedAction}{50.5\xspace}
\newcommand{\sbTaskQwenSevenTwoFiveMvbUnexpectedAction}{47.5\xspace}
\newcommand{\sbTaskLiveCcMvbUnexpectedAction}{50.5\xspace}
\newcommand{\sbTaskStreamingVlmMvbUnexpectedAction}{51.0\xspace}
\newcommand{\sbTaskTimeChatMvbUnexpectedAction}{47.5\xspace}
\newcommand{\sbTaskQwenEightMvbUnexpectedAction}{51.0\xspace}
\newcommand{\sbTaskTeacherMvbUnexpectedAction}{52.5\xspace}
\newcommand{\sbTaskNMvbObjectExistence}{200\xspace}
\newcommand{\sbTaskFloorMvbObjectExistence}{0.0\xspace}
\newcommand{\sbTaskOursMvbObjectExistence}{48.0\xspace}
\newcommand{\sbTaskLfmSmallMvbObjectExistence}{36.5\xspace}
\newcommand{\sbTaskSmolSmallMvbObjectExistence}{26.0\xspace}
\newcommand{\sbTaskFastVlmMvbObjectExistence}{46.0\xspace}
\newcommand{\sbTaskMobileMvbObjectExistence}{28.5\xspace}
\newcommand{\sbTaskInternOneMvbObjectExistence}{41.5\xspace}
\newcommand{\sbTaskInternOneIntMvbObjectExistence}{42.5\xspace}
\newcommand{\sbTaskLfmBigMvbObjectExistence}{36.5\xspace}
\newcommand{\sbTaskQwenTwoMvbObjectExistence}{60.5\xspace}
\newcommand{\sbTaskQwenTwoIntMvbObjectExistence}{56.5\xspace}
\newcommand{\sbTaskSmolBigMvbObjectExistence}{45.5\xspace}
\newcommand{\sbTaskInternTwoMvbObjectExistence}{56.0\xspace}
\newcommand{\sbTaskQwenFourMvbObjectExistence}{61.0\xspace}
\newcommand{\sbTaskQwenSevenTwoFiveMvbObjectExistence}{43.0\xspace}
\newcommand{\sbTaskLiveCcMvbObjectExistence}{32.0\xspace}
\newcommand{\sbTaskStreamingVlmMvbObjectExistence}{60.0\xspace}
\newcommand{\sbTaskTimeChatMvbObjectExistence}{61.5\xspace}
\newcommand{\sbTaskQwenEightMvbObjectExistence}{65.0\xspace}
\newcommand{\sbTaskTeacherMvbObjectExistence}{54.5\xspace}
\newcommand{\sbTaskNMvbMovingDirection}{200\xspace}
\newcommand{\sbTaskFloorMvbMovingDirection}{34.5\xspace}
\newcommand{\sbTaskOursMvbMovingDirection}{32.0\xspace}
\newcommand{\sbTaskLfmSmallMvbMovingDirection}{34.0\xspace}
\newcommand{\sbTaskSmolSmallMvbMovingDirection}{33.0\xspace}
\newcommand{\sbTaskFastVlmMvbMovingDirection}{28.5\xspace}
\newcommand{\sbTaskMobileMvbMovingDirection}{33.5\xspace}
\newcommand{\sbTaskInternOneMvbMovingDirection}{34.0\xspace}
\newcommand{\sbTaskInternOneIntMvbMovingDirection}{33.0\xspace}
\newcommand{\sbTaskLfmBigMvbMovingDirection}{27.0\xspace}
\newcommand{\sbTaskQwenTwoMvbMovingDirection}{39.5\xspace}
\newcommand{\sbTaskQwenTwoIntMvbMovingDirection}{38.0\xspace}
\newcommand{\sbTaskSmolBigMvbMovingDirection}{32.0\xspace}
\newcommand{\sbTaskInternTwoMvbMovingDirection}{37.5\xspace}
\newcommand{\sbTaskQwenFourMvbMovingDirection}{40.0\xspace}
\newcommand{\sbTaskQwenSevenTwoFiveMvbMovingDirection}{40.5\xspace}
\newcommand{\sbTaskLiveCcMvbMovingDirection}{36.0\xspace}
\newcommand{\sbTaskStreamingVlmMvbMovingDirection}{40.0\xspace}
\newcommand{\sbTaskTimeChatMvbMovingDirection}{36.5\xspace}
\newcommand{\sbTaskQwenEightMvbMovingDirection}{42.0\xspace}
\newcommand{\sbTaskTeacherMvbMovingDirection}{44.0\xspace}
\newcommand{\sbTaskNMvbMovingCount}{200\xspace}
\newcommand{\sbTaskFloorMvbMovingCount}{39.5\xspace}
\newcommand{\sbTaskOursMvbMovingCount}{40.0\xspace}
\newcommand{\sbTaskLfmSmallMvbMovingCount}{34.5\xspace}
\newcommand{\sbTaskSmolSmallMvbMovingCount}{41.0\xspace}
\newcommand{\sbTaskFastVlmMvbMovingCount}{44.0\xspace}
\newcommand{\sbTaskMobileMvbMovingCount}{37.0\xspace}
\newcommand{\sbTaskInternOneMvbMovingCount}{43.0\xspace}
\newcommand{\sbTaskInternOneIntMvbMovingCount}{42.5\xspace}
\newcommand{\sbTaskLfmBigMvbMovingCount}{42.5\xspace}
\newcommand{\sbTaskQwenTwoMvbMovingCount}{45.5\xspace}
\newcommand{\sbTaskQwenTwoIntMvbMovingCount}{41.0\xspace}
\newcommand{\sbTaskSmolBigMvbMovingCount}{43.5\xspace}
\newcommand{\sbTaskInternTwoMvbMovingCount}{44.0\xspace}
\newcommand{\sbTaskQwenFourMvbMovingCount}{42.0\xspace}
\newcommand{\sbTaskQwenSevenTwoFiveMvbMovingCount}{49.0\xspace}
\newcommand{\sbTaskLiveCcMvbMovingCount}{42.0\xspace}
\newcommand{\sbTaskStreamingVlmMvbMovingCount}{41.0\xspace}
\newcommand{\sbTaskTimeChatMvbMovingCount}{38.5\xspace}
\newcommand{\sbTaskQwenEightMvbMovingCount}{51.0\xspace}
\newcommand{\sbTaskTeacherMvbMovingCount}{49.0\xspace}
\newcommand{\sbTaskNMvbMovingAttribute}{200\xspace}
\newcommand{\sbTaskFloorMvbMovingAttribute}{39.0\xspace}
\newcommand{\sbTaskOursMvbMovingAttribute}{60.0\xspace}
\newcommand{\sbTaskLfmSmallMvbMovingAttribute}{50.5\xspace}
\newcommand{\sbTaskSmolSmallMvbMovingAttribute}{44.0\xspace}
\newcommand{\sbTaskFastVlmMvbMovingAttribute}{56.0\xspace}
\newcommand{\sbTaskMobileMvbMovingAttribute}{56.0\xspace}
\newcommand{\sbTaskInternOneMvbMovingAttribute}{60.0\xspace}
\newcommand{\sbTaskInternOneIntMvbMovingAttribute}{60.5\xspace}
\newcommand{\sbTaskLfmBigMvbMovingAttribute}{57.0\xspace}
\newcommand{\sbTaskQwenTwoMvbMovingAttribute}{61.0\xspace}
\newcommand{\sbTaskQwenTwoIntMvbMovingAttribute}{60.0\xspace}
\newcommand{\sbTaskSmolBigMvbMovingAttribute}{54.5\xspace}
\newcommand{\sbTaskInternTwoMvbMovingAttribute}{66.5\xspace}
\newcommand{\sbTaskQwenFourMvbMovingAttribute}{61.5\xspace}
\newcommand{\sbTaskQwenSevenTwoFiveMvbMovingAttribute}{58.0\xspace}
\newcommand{\sbTaskLiveCcMvbMovingAttribute}{48.0\xspace}
\newcommand{\sbTaskStreamingVlmMvbMovingAttribute}{64.0\xspace}
\newcommand{\sbTaskTimeChatMvbMovingAttribute}{60.5\xspace}
\newcommand{\sbTaskQwenEightMvbMovingAttribute}{60.0\xspace}
\newcommand{\sbTaskTeacherMvbMovingAttribute}{65.5\xspace}
\newcommand{\sbTaskNMvbCounterfactualInference}{200\xspace}
\newcommand{\sbTaskFloorMvbCounterfactualInference}{36.5\xspace}
\newcommand{\sbTaskOursMvbCounterfactualInference}{27.0\xspace}
\newcommand{\sbTaskLfmSmallMvbCounterfactualInference}{37.5\xspace}
\newcommand{\sbTaskSmolSmallMvbCounterfactualInference}{37.5\xspace}
\newcommand{\sbTaskFastVlmMvbCounterfactualInference}{31.5\xspace}
\newcommand{\sbTaskMobileMvbCounterfactualInference}{33.5\xspace}
\newcommand{\sbTaskInternOneMvbCounterfactualInference}{42.0\xspace}
\newcommand{\sbTaskInternOneIntMvbCounterfactualInference}{42.5\xspace}
\newcommand{\sbTaskLfmBigMvbCounterfactualInference}{39.0\xspace}
\newcommand{\sbTaskQwenTwoMvbCounterfactualInference}{38.5\xspace}
\newcommand{\sbTaskQwenTwoIntMvbCounterfactualInference}{36.5\xspace}
\newcommand{\sbTaskSmolBigMvbCounterfactualInference}{39.0\xspace}
\newcommand{\sbTaskInternTwoMvbCounterfactualInference}{37.5\xspace}
\newcommand{\sbTaskQwenFourMvbCounterfactualInference}{35.5\xspace}
\newcommand{\sbTaskQwenSevenTwoFiveMvbCounterfactualInference}{37.5\xspace}
\newcommand{\sbTaskLiveCcMvbCounterfactualInference}{33.0\xspace}
\newcommand{\sbTaskStreamingVlmMvbCounterfactualInference}{39.5\xspace}
\newcommand{\sbTaskTimeChatMvbCounterfactualInference}{43.0\xspace}
\newcommand{\sbTaskQwenEightMvbCounterfactualInference}{35.0\xspace}
\newcommand{\sbTaskTeacherMvbCounterfactualInference}{51.0\xspace}
\newcommand{\sbTaskNMvbObjectShuffle}{200\xspace}
\newcommand{\sbTaskFloorMvbObjectShuffle}{36.0\xspace}
\newcommand{\sbTaskOursMvbObjectShuffle}{31.5\xspace}
\newcommand{\sbTaskLfmSmallMvbObjectShuffle}{27.0\xspace}
\newcommand{\sbTaskSmolSmallMvbObjectShuffle}{30.5\xspace}
\newcommand{\sbTaskFastVlmMvbObjectShuffle}{29.5\xspace}
\newcommand{\sbTaskMobileMvbObjectShuffle}{32.0\xspace}
\newcommand{\sbTaskInternOneMvbObjectShuffle}{36.5\xspace}
\newcommand{\sbTaskInternOneIntMvbObjectShuffle}{31.5\xspace}
\newcommand{\sbTaskLfmBigMvbObjectShuffle}{34.0\xspace}
\newcommand{\sbTaskQwenTwoMvbObjectShuffle}{39.5\xspace}
\newcommand{\sbTaskQwenTwoIntMvbObjectShuffle}{33.5\xspace}
\newcommand{\sbTaskSmolBigMvbObjectShuffle}{27.5\xspace}
\newcommand{\sbTaskInternTwoMvbObjectShuffle}{29.5\xspace}
\newcommand{\sbTaskQwenFourMvbObjectShuffle}{37.5\xspace}
\newcommand{\sbTaskQwenSevenTwoFiveMvbObjectShuffle}{32.5\xspace}
\newcommand{\sbTaskLiveCcMvbObjectShuffle}{32.5\xspace}
\newcommand{\sbTaskStreamingVlmMvbObjectShuffle}{35.0\xspace}
\newcommand{\sbTaskTimeChatMvbObjectShuffle}{33.5\xspace}
\newcommand{\sbTaskQwenEightMvbObjectShuffle}{32.0\xspace}
\newcommand{\sbTaskTeacherMvbObjectShuffle}{34.0\xspace}
\newcommand{\sbTaskNMvbActionCount}{200\xspace}
\newcommand{\sbTaskFloorMvbActionCount}{35.0\xspace}
\newcommand{\sbTaskOursMvbActionCount}{35.5\xspace}
\newcommand{\sbTaskLfmSmallMvbActionCount}{31.5\xspace}
\newcommand{\sbTaskSmolSmallMvbActionCount}{34.0\xspace}
\newcommand{\sbTaskFastVlmMvbActionCount}{28.5\xspace}
\newcommand{\sbTaskMobileMvbActionCount}{34.0\xspace}
\newcommand{\sbTaskInternOneMvbActionCount}{32.5\xspace}
\newcommand{\sbTaskInternOneIntMvbActionCount}{34.0\xspace}
\newcommand{\sbTaskLfmBigMvbActionCount}{27.0\xspace}
\newcommand{\sbTaskQwenTwoMvbActionCount}{37.5\xspace}
\newcommand{\sbTaskQwenTwoIntMvbActionCount}{31.5\xspace}
\newcommand{\sbTaskSmolBigMvbActionCount}{27.5\xspace}
\newcommand{\sbTaskInternTwoMvbActionCount}{33.5\xspace}
\newcommand{\sbTaskQwenFourMvbActionCount}{33.5\xspace}
\newcommand{\sbTaskQwenSevenTwoFiveMvbActionCount}{32.5\xspace}
\newcommand{\sbTaskLiveCcMvbActionCount}{32.5\xspace}
\newcommand{\sbTaskStreamingVlmMvbActionCount}{33.0\xspace}
\newcommand{\sbTaskTimeChatMvbActionCount}{31.0\xspace}
\newcommand{\sbTaskQwenEightMvbActionCount}{30.0\xspace}
\newcommand{\sbTaskTeacherMvbActionCount}{34.0\xspace}
\newcommand{\sbTaskNMvbStateChange}{200\xspace}
\newcommand{\sbTaskFloorMvbStateChange}{9.5\xspace}
\newcommand{\sbTaskOursMvbStateChange}{25.5\xspace}
\newcommand{\sbTaskLfmSmallMvbStateChange}{23.5\xspace}
\newcommand{\sbTaskSmolSmallMvbStateChange}{23.0\xspace}
\newcommand{\sbTaskFastVlmMvbStateChange}{26.5\xspace}
\newcommand{\sbTaskMobileMvbStateChange}{18.5\xspace}
\newcommand{\sbTaskInternOneMvbStateChange}{40.0\xspace}
\newcommand{\sbTaskInternOneIntMvbStateChange}{37.5\xspace}
\newcommand{\sbTaskLfmBigMvbStateChange}{32.0\xspace}
\newcommand{\sbTaskQwenTwoMvbStateChange}{37.5\xspace}
\newcommand{\sbTaskQwenTwoIntMvbStateChange}{38.5\xspace}
\newcommand{\sbTaskSmolBigMvbStateChange}{26.5\xspace}
\newcommand{\sbTaskInternTwoMvbStateChange}{35.5\xspace}
\newcommand{\sbTaskQwenFourMvbStateChange}{39.0\xspace}
\newcommand{\sbTaskQwenSevenTwoFiveMvbStateChange}{27.0\xspace}
\newcommand{\sbTaskLiveCcMvbStateChange}{24.0\xspace}
\newcommand{\sbTaskStreamingVlmMvbStateChange}{47.5\xspace}
\newcommand{\sbTaskTimeChatMvbStateChange}{27.0\xspace}
\newcommand{\sbTaskQwenEightMvbStateChange}{43.0\xspace}
\newcommand{\sbTaskTeacherMvbStateChange}{38.0\xspace}
\newcommand{\sbTaskNMvbCharacterOrder}{200\xspace}
\newcommand{\sbTaskFloorMvbCharacterOrder}{48.0\xspace}
\newcommand{\sbTaskOursMvbCharacterOrder}{47.5\xspace}
\newcommand{\sbTaskLfmSmallMvbCharacterOrder}{47.0\xspace}
\newcommand{\sbTaskSmolSmallMvbCharacterOrder}{47.5\xspace}
\newcommand{\sbTaskFastVlmMvbCharacterOrder}{48.0\xspace}
\newcommand{\sbTaskMobileMvbCharacterOrder}{40.5\xspace}
\newcommand{\sbTaskInternOneMvbCharacterOrder}{51.0\xspace}
\newcommand{\sbTaskInternOneIntMvbCharacterOrder}{48.0\xspace}
\newcommand{\sbTaskLfmBigMvbCharacterOrder}{45.0\xspace}
\newcommand{\sbTaskQwenTwoMvbCharacterOrder}{48.5\xspace}
\newcommand{\sbTaskQwenTwoIntMvbCharacterOrder}{52.0\xspace}
\newcommand{\sbTaskSmolBigMvbCharacterOrder}{46.0\xspace}
\newcommand{\sbTaskInternTwoMvbCharacterOrder}{53.5\xspace}
\newcommand{\sbTaskQwenFourMvbCharacterOrder}{56.0\xspace}
\newcommand{\sbTaskQwenSevenTwoFiveMvbCharacterOrder}{50.5\xspace}
\newcommand{\sbTaskLiveCcMvbCharacterOrder}{39.0\xspace}
\newcommand{\sbTaskStreamingVlmMvbCharacterOrder}{50.5\xspace}
\newcommand{\sbTaskTimeChatMvbCharacterOrder}{53.0\xspace}
\newcommand{\sbTaskQwenEightMvbCharacterOrder}{55.0\xspace}
\newcommand{\sbTaskTeacherMvbCharacterOrder}{53.5\xspace}
\newcommand{\sbTaskNMvbActionLocalization}{200\xspace}
\newcommand{\sbTaskFloorMvbActionLocalization}{26.5\xspace}
\newcommand{\sbTaskOursMvbActionLocalization}{30.5\xspace}
\newcommand{\sbTaskLfmSmallMvbActionLocalization}{26.5\xspace}
\newcommand{\sbTaskSmolSmallMvbActionLocalization}{23.5\xspace}
\newcommand{\sbTaskFastVlmMvbActionLocalization}{27.0\xspace}
\newcommand{\sbTaskMobileMvbActionLocalization}{27.0\xspace}
\newcommand{\sbTaskInternOneMvbActionLocalization}{31.0\xspace}
\newcommand{\sbTaskInternOneIntMvbActionLocalization}{31.5\xspace}
\newcommand{\sbTaskLfmBigMvbActionLocalization}{26.5\xspace}
\newcommand{\sbTaskQwenTwoMvbActionLocalization}{31.0\xspace}
\newcommand{\sbTaskQwenTwoIntMvbActionLocalization}{26.0\xspace}
\newcommand{\sbTaskSmolBigMvbActionLocalization}{24.0\xspace}
\newcommand{\sbTaskInternTwoMvbActionLocalization}{29.0\xspace}
\newcommand{\sbTaskQwenFourMvbActionLocalization}{31.0\xspace}
\newcommand{\sbTaskQwenSevenTwoFiveMvbActionLocalization}{28.0\xspace}
\newcommand{\sbTaskLiveCcMvbActionLocalization}{28.5\xspace}
\newcommand{\sbTaskStreamingVlmMvbActionLocalization}{35.0\xspace}
\newcommand{\sbTaskTimeChatMvbActionLocalization}{28.5\xspace}
\newcommand{\sbTaskQwenEightMvbActionLocalization}{28.0\xspace}
\newcommand{\sbTaskTeacherMvbActionLocalization}{32.0\xspace}
\newcommand{\sbTaskNMvbSceneTransition}{200\xspace}
\newcommand{\sbTaskFloorMvbSceneTransition}{26.5\xspace}
\newcommand{\sbTaskOursMvbSceneTransition}{67.0\xspace}
\newcommand{\sbTaskLfmSmallMvbSceneTransition}{55.5\xspace}
\newcommand{\sbTaskSmolSmallMvbSceneTransition}{55.5\xspace}
\newcommand{\sbTaskFastVlmMvbSceneTransition}{63.5\xspace}
\newcommand{\sbTaskMobileMvbSceneTransition}{46.0\xspace}
\newcommand{\sbTaskInternOneMvbSceneTransition}{66.5\xspace}
\newcommand{\sbTaskInternOneIntMvbSceneTransition}{71.0\xspace}
\newcommand{\sbTaskLfmBigMvbSceneTransition}{61.5\xspace}
\newcommand{\sbTaskQwenTwoMvbSceneTransition}{74.0\xspace}
\newcommand{\sbTaskQwenTwoIntMvbSceneTransition}{73.5\xspace}
\newcommand{\sbTaskSmolBigMvbSceneTransition}{62.0\xspace}
\newcommand{\sbTaskInternTwoMvbSceneTransition}{73.5\xspace}
\newcommand{\sbTaskQwenFourMvbSceneTransition}{72.0\xspace}
\newcommand{\sbTaskQwenSevenTwoFiveMvbSceneTransition}{50.5\xspace}
\newcommand{\sbTaskLiveCcMvbSceneTransition}{52.5\xspace}
\newcommand{\sbTaskStreamingVlmMvbSceneTransition}{70.5\xspace}
\newcommand{\sbTaskTimeChatMvbSceneTransition}{73.5\xspace}
\newcommand{\sbTaskQwenEightMvbSceneTransition}{74.0\xspace}
\newcommand{\sbTaskTeacherMvbSceneTransition}{73.5\xspace}
\newcommand{\sbTaskNMvbEgocentricNavigation}{200\xspace}
\newcommand{\sbTaskFloorMvbEgocentricNavigation}{24.5\xspace}
\newcommand{\sbTaskOursMvbEgocentricNavigation}{25.5\xspace}
\newcommand{\sbTaskLfmSmallMvbEgocentricNavigation}{23.0\xspace}
\newcommand{\sbTaskSmolSmallMvbEgocentricNavigation}{24.5\xspace}
\newcommand{\sbTaskFastVlmMvbEgocentricNavigation}{25.0\xspace}
\newcommand{\sbTaskMobileMvbEgocentricNavigation}{25.0\xspace}
\newcommand{\sbTaskInternOneMvbEgocentricNavigation}{26.0\xspace}
\newcommand{\sbTaskInternOneIntMvbEgocentricNavigation}{25.0\xspace}
\newcommand{\sbTaskLfmBigMvbEgocentricNavigation}{22.5\xspace}
\newcommand{\sbTaskQwenTwoMvbEgocentricNavigation}{26.0\xspace}
\newcommand{\sbTaskQwenTwoIntMvbEgocentricNavigation}{24.0\xspace}
\newcommand{\sbTaskSmolBigMvbEgocentricNavigation}{24.5\xspace}
\newcommand{\sbTaskInternTwoMvbEgocentricNavigation}{22.0\xspace}
\newcommand{\sbTaskQwenFourMvbEgocentricNavigation}{27.5\xspace}
\newcommand{\sbTaskQwenSevenTwoFiveMvbEgocentricNavigation}{27.5\xspace}
\newcommand{\sbTaskLiveCcMvbEgocentricNavigation}{28.0\xspace}
\newcommand{\sbTaskStreamingVlmMvbEgocentricNavigation}{28.5\xspace}
\newcommand{\sbTaskTimeChatMvbEgocentricNavigation}{29.5\xspace}
\newcommand{\sbTaskQwenEightMvbEgocentricNavigation}{31.0\xspace}
\newcommand{\sbTaskTeacherMvbEgocentricNavigation}{30.0\xspace}
\newcommand{\sbTaskNMvbEpisodicReasoning}{200\xspace}
\newcommand{\sbTaskFloorMvbEpisodicReasoning}{34.5\xspace}
\newcommand{\sbTaskOursMvbEpisodicReasoning}{42.0\xspace}
\newcommand{\sbTaskLfmSmallMvbEpisodicReasoning}{38.0\xspace}
\newcommand{\sbTaskSmolSmallMvbEpisodicReasoning}{38.5\xspace}
\newcommand{\sbTaskFastVlmMvbEpisodicReasoning}{35.5\xspace}
\newcommand{\sbTaskMobileMvbEpisodicReasoning}{37.0\xspace}
\newcommand{\sbTaskInternOneMvbEpisodicReasoning}{39.5\xspace}
\newcommand{\sbTaskInternOneIntMvbEpisodicReasoning}{40.0\xspace}
\newcommand{\sbTaskLfmBigMvbEpisodicReasoning}{37.0\xspace}
\newcommand{\sbTaskQwenTwoMvbEpisodicReasoning}{39.5\xspace}
\newcommand{\sbTaskQwenTwoIntMvbEpisodicReasoning}{41.0\xspace}
\newcommand{\sbTaskSmolBigMvbEpisodicReasoning}{39.5\xspace}
\newcommand{\sbTaskInternTwoMvbEpisodicReasoning}{42.0\xspace}
\newcommand{\sbTaskQwenFourMvbEpisodicReasoning}{43.5\xspace}
\newcommand{\sbTaskQwenSevenTwoFiveMvbEpisodicReasoning}{38.5\xspace}
\newcommand{\sbTaskLiveCcMvbEpisodicReasoning}{35.5\xspace}
\newcommand{\sbTaskStreamingVlmMvbEpisodicReasoning}{38.0\xspace}
\newcommand{\sbTaskTimeChatMvbEpisodicReasoning}{39.5\xspace}
\newcommand{\sbTaskQwenEightMvbEpisodicReasoning}{43.5\xspace}
\newcommand{\sbTaskTeacherMvbEpisodicReasoning}{41.0\xspace}

\newcommand{\sbTaskNSbObjectPerception}{369\xspace}
\newcommand{\sbTaskFloorSbObjectPerception}{33.9\xspace}
\newcommand{\sbTaskOursSbObjectPerception}{51.5\xspace}
\newcommand{\sbTaskLfmSmallSbObjectPerception}{43.6\xspace}
\newcommand{\sbTaskSmolSmallSbObjectPerception}{45.3\xspace}
\newcommand{\sbTaskFastVlmSbObjectPerception}{47.2\xspace}
\newcommand{\sbTaskMobileSbObjectPerception}{43.4\xspace}
\newcommand{\sbTaskInternOneSbObjectPerception}{52.0\xspace}
\newcommand{\sbTaskInternOneIntSbObjectPerception}{53.1\xspace}
\newcommand{\sbTaskLfmBigSbObjectPerception}{46.6\xspace}
\newcommand{\sbTaskQwenTwoSbObjectPerception}{54.2\xspace}
\newcommand{\sbTaskQwenTwoIntSbObjectPerception}{54.2\xspace}
\newcommand{\sbTaskSmolBigSbObjectPerception}{50.9\xspace}
\newcommand{\sbTaskInternTwoSbObjectPerception}{50.9\xspace}
\newcommand{\sbTaskQwenFourSbObjectPerception}{55.3\xspace}
\newcommand{\sbTaskQwenSevenTwoFiveSbObjectPerception}{49.3\xspace}
\newcommand{\sbTaskLiveCcSbObjectPerception}{43.1\xspace}
\newcommand{\sbTaskStreamingVlmSbObjectPerception}{50.4\xspace}
\newcommand{\sbTaskTimeChatSbObjectPerception}{54.5\xspace}
\newcommand{\sbTaskQwenEightSbObjectPerception}{55.0\xspace}
\newcommand{\sbTaskTeacherSbObjectPerception}{53.1\xspace}
\newcommand{\sbTaskNSbActionPerception}{352\xspace}
\newcommand{\sbTaskFloorSbActionPerception}{28.1\xspace}
\newcommand{\sbTaskOursSbActionPerception}{46.9\xspace}
\newcommand{\sbTaskLfmSmallSbActionPerception}{38.9\xspace}
\newcommand{\sbTaskSmolSmallSbActionPerception}{43.8\xspace}
\newcommand{\sbTaskFastVlmSbActionPerception}{41.8\xspace}
\newcommand{\sbTaskMobileSbActionPerception}{38.6\xspace}
\newcommand{\sbTaskInternOneSbActionPerception}{48.6\xspace}
\newcommand{\sbTaskInternOneIntSbActionPerception}{48.9\xspace}
\newcommand{\sbTaskLfmBigSbActionPerception}{41.8\xspace}
\newcommand{\sbTaskQwenTwoSbActionPerception}{54.3\xspace}
\newcommand{\sbTaskQwenTwoIntSbActionPerception}{54.3\xspace}
\newcommand{\sbTaskSmolBigSbActionPerception}{42.6\xspace}
\newcommand{\sbTaskInternTwoSbActionPerception}{51.4\xspace}
\newcommand{\sbTaskQwenFourSbActionPerception}{56.5\xspace}
\newcommand{\sbTaskQwenSevenTwoFiveSbActionPerception}{49.7\xspace}
\newcommand{\sbTaskLiveCcSbActionPerception}{43.8\xspace}
\newcommand{\sbTaskStreamingVlmSbActionPerception}{48.3\xspace}
\newcommand{\sbTaskTimeChatSbActionPerception}{47.7\xspace}
\newcommand{\sbTaskQwenEightSbActionPerception}{56.5\xspace}
\newcommand{\sbTaskTeacherSbActionPerception}{57.1\xspace}
\newcommand{\sbTaskNSbTextRichUnderstanding}{321\xspace}
\newcommand{\sbTaskFloorSbTextRichUnderstanding}{30.2\xspace}
\newcommand{\sbTaskOursSbTextRichUnderstanding}{36.1\xspace}
\newcommand{\sbTaskLfmSmallSbTextRichUnderstanding}{37.1\xspace}
\newcommand{\sbTaskSmolSmallSbTextRichUnderstanding}{36.1\xspace}
\newcommand{\sbTaskFastVlmSbTextRichUnderstanding}{39.9\xspace}
\newcommand{\sbTaskMobileSbTextRichUnderstanding}{30.5\xspace}
\newcommand{\sbTaskInternOneSbTextRichUnderstanding}{44.2\xspace}
\newcommand{\sbTaskInternOneIntSbTextRichUnderstanding}{47.7\xspace}
\newcommand{\sbTaskLfmBigSbTextRichUnderstanding}{38.3\xspace}
\newcommand{\sbTaskQwenTwoSbTextRichUnderstanding}{42.1\xspace}
\newcommand{\sbTaskQwenTwoIntSbTextRichUnderstanding}{43.3\xspace}
\newcommand{\sbTaskSmolBigSbTextRichUnderstanding}{35.8\xspace}
\newcommand{\sbTaskInternTwoSbTextRichUnderstanding}{47.4\xspace}
\newcommand{\sbTaskQwenFourSbTextRichUnderstanding}{44.2\xspace}
\newcommand{\sbTaskQwenSevenTwoFiveSbTextRichUnderstanding}{33.6\xspace}
\newcommand{\sbTaskLiveCcSbTextRichUnderstanding}{32.7\xspace}
\newcommand{\sbTaskStreamingVlmSbTextRichUnderstanding}{43.6\xspace}
\newcommand{\sbTaskTimeChatSbTextRichUnderstanding}{46.4\xspace}
\newcommand{\sbTaskQwenEightSbTextRichUnderstanding}{48.0\xspace}
\newcommand{\sbTaskTeacherSbTextRichUnderstanding}{44.2\xspace}
\newcommand{\sbTaskNSbClipsSummarize}{317\xspace}
\newcommand{\sbTaskFloorSbClipsSummarize}{36.9\xspace}
\newcommand{\sbTaskOursSbClipsSummarize}{65.6\xspace}
\newcommand{\sbTaskLfmSmallSbClipsSummarize}{61.2\xspace}
\newcommand{\sbTaskSmolSmallSbClipsSummarize}{61.5\xspace}
\newcommand{\sbTaskFastVlmSbClipsSummarize}{62.5\xspace}
\newcommand{\sbTaskMobileSbClipsSummarize}{56.2\xspace}
\newcommand{\sbTaskInternOneSbClipsSummarize}{71.0\xspace}
\newcommand{\sbTaskInternOneIntSbClipsSummarize}{71.0\xspace}
\newcommand{\sbTaskLfmBigSbClipsSummarize}{65.9\xspace}
\newcommand{\sbTaskQwenTwoSbClipsSummarize}{69.7\xspace}
\newcommand{\sbTaskQwenTwoIntSbClipsSummarize}{69.7\xspace}
\newcommand{\sbTaskSmolBigSbClipsSummarize}{65.6\xspace}
\newcommand{\sbTaskInternTwoSbClipsSummarize}{70.7\xspace}
\newcommand{\sbTaskQwenFourSbClipsSummarize}{75.4\xspace}
\newcommand{\sbTaskQwenSevenTwoFiveSbClipsSummarize}{70.7\xspace}
\newcommand{\sbTaskLiveCcSbClipsSummarize}{71.0\xspace}
\newcommand{\sbTaskStreamingVlmSbClipsSummarize}{72.6\xspace}
\newcommand{\sbTaskTimeChatSbClipsSummarize}{65.9\xspace}
\newcommand{\sbTaskQwenEightSbClipsSummarize}{71.0\xspace}
\newcommand{\sbTaskTeacherSbClipsSummarize}{67.8\xspace}
\newcommand{\sbTaskNSbAttributePerception}{311\xspace}
\newcommand{\sbTaskFloorSbAttributePerception}{34.4\xspace}
\newcommand{\sbTaskOursSbAttributePerception}{55.9\xspace}
\newcommand{\sbTaskLfmSmallSbAttributePerception}{45.7\xspace}
\newcommand{\sbTaskSmolSmallSbAttributePerception}{43.4\xspace}
\newcommand{\sbTaskFastVlmSbAttributePerception}{47.9\xspace}
\newcommand{\sbTaskMobileSbAttributePerception}{43.1\xspace}
\newcommand{\sbTaskInternOneSbAttributePerception}{51.8\xspace}
\newcommand{\sbTaskInternOneIntSbAttributePerception}{53.1\xspace}
\newcommand{\sbTaskLfmBigSbAttributePerception}{42.4\xspace}
\newcommand{\sbTaskQwenTwoSbAttributePerception}{52.4\xspace}
\newcommand{\sbTaskQwenTwoIntSbAttributePerception}{51.8\xspace}
\newcommand{\sbTaskSmolBigSbAttributePerception}{51.8\xspace}
\newcommand{\sbTaskInternTwoSbAttributePerception}{53.4\xspace}
\newcommand{\sbTaskQwenFourSbAttributePerception}{53.1\xspace}
\newcommand{\sbTaskQwenSevenTwoFiveSbAttributePerception}{41.8\xspace}
\newcommand{\sbTaskLiveCcSbAttributePerception}{39.5\xspace}
\newcommand{\sbTaskStreamingVlmSbAttributePerception}{43.7\xspace}
\newcommand{\sbTaskTimeChatSbAttributePerception}{52.7\xspace}
\newcommand{\sbTaskQwenEightSbAttributePerception}{54.0\xspace}
\newcommand{\sbTaskTeacherSbAttributePerception}{56.9\xspace}
\newcommand{\sbTaskNSbSpatialUnderstanding}{246\xspace}
\newcommand{\sbTaskFloorSbSpatialUnderstanding}{30.5\xspace}
\newcommand{\sbTaskOursSbSpatialUnderstanding}{50.4\xspace}
\newcommand{\sbTaskLfmSmallSbSpatialUnderstanding}{45.1\xspace}
\newcommand{\sbTaskSmolSmallSbSpatialUnderstanding}{43.1\xspace}
\newcommand{\sbTaskFastVlmSbSpatialUnderstanding}{43.9\xspace}
\newcommand{\sbTaskMobileSbSpatialUnderstanding}{41.9\xspace}
\newcommand{\sbTaskInternOneSbSpatialUnderstanding}{43.5\xspace}
\newcommand{\sbTaskInternOneIntSbSpatialUnderstanding}{41.1\xspace}
\newcommand{\sbTaskLfmBigSbSpatialUnderstanding}{45.9\xspace}
\newcommand{\sbTaskQwenTwoSbSpatialUnderstanding}{44.3\xspace}
\newcommand{\sbTaskQwenTwoIntSbSpatialUnderstanding}{43.5\xspace}
\newcommand{\sbTaskSmolBigSbSpatialUnderstanding}{45.5\xspace}
\newcommand{\sbTaskInternTwoSbSpatialUnderstanding}{44.3\xspace}
\newcommand{\sbTaskQwenFourSbSpatialUnderstanding}{49.2\xspace}
\newcommand{\sbTaskQwenSevenTwoFiveSbSpatialUnderstanding}{38.2\xspace}
\newcommand{\sbTaskLiveCcSbSpatialUnderstanding}{41.1\xspace}
\newcommand{\sbTaskStreamingVlmSbSpatialUnderstanding}{46.3\xspace}
\newcommand{\sbTaskTimeChatSbSpatialUnderstanding}{45.5\xspace}
\newcommand{\sbTaskQwenEightSbSpatialUnderstanding}{47.6\xspace}
\newcommand{\sbTaskTeacherSbSpatialUnderstanding}{50.4\xspace}
\newcommand{\sbTaskNSbCounting}{188\xspace}
\newcommand{\sbTaskFloorSbCounting}{38.3\xspace}
\newcommand{\sbTaskOursSbCounting}{39.4\xspace}
\newcommand{\sbTaskLfmSmallSbCounting}{38.8\xspace}
\newcommand{\sbTaskSmolSmallSbCounting}{40.4\xspace}
\newcommand{\sbTaskFastVlmSbCounting}{37.2\xspace}
\newcommand{\sbTaskMobileSbCounting}{41.0\xspace}
\newcommand{\sbTaskInternOneSbCounting}{38.3\xspace}
\newcommand{\sbTaskInternOneIntSbCounting}{37.2\xspace}
\newcommand{\sbTaskLfmBigSbCounting}{37.8\xspace}
\newcommand{\sbTaskQwenTwoSbCounting}{37.2\xspace}
\newcommand{\sbTaskQwenTwoIntSbCounting}{36.7\xspace}
\newcommand{\sbTaskSmolBigSbCounting}{38.3\xspace}
\newcommand{\sbTaskInternTwoSbCounting}{36.2\xspace}
\newcommand{\sbTaskQwenFourSbCounting}{36.7\xspace}
\newcommand{\sbTaskQwenSevenTwoFiveSbCounting}{37.2\xspace}
\newcommand{\sbTaskLiveCcSbCounting}{37.8\xspace}
\newcommand{\sbTaskStreamingVlmSbCounting}{43.1\xspace}
\newcommand{\sbTaskTimeChatSbCounting}{37.2\xspace}
\newcommand{\sbTaskQwenEightSbCounting}{41.0\xspace}
\newcommand{\sbTaskTeacherSbCounting}{35.6\xspace}
\newcommand{\sbTaskNSbEventUnderstanding}{158\xspace}
\newcommand{\sbTaskFloorSbEventUnderstanding}{36.1\xspace}
\newcommand{\sbTaskOursSbEventUnderstanding}{50.6\xspace}
\newcommand{\sbTaskLfmSmallSbEventUnderstanding}{53.2\xspace}
\newcommand{\sbTaskSmolSmallSbEventUnderstanding}{52.5\xspace}
\newcommand{\sbTaskFastVlmSbEventUnderstanding}{55.7\xspace}
\newcommand{\sbTaskMobileSbEventUnderstanding}{50.6\xspace}
\newcommand{\sbTaskInternOneSbEventUnderstanding}{60.1\xspace}
\newcommand{\sbTaskInternOneIntSbEventUnderstanding}{62.0\xspace}
\newcommand{\sbTaskLfmBigSbEventUnderstanding}{56.3\xspace}
\newcommand{\sbTaskQwenTwoSbEventUnderstanding}{55.7\xspace}
\newcommand{\sbTaskQwenTwoIntSbEventUnderstanding}{58.2\xspace}
\newcommand{\sbTaskSmolBigSbEventUnderstanding}{53.2\xspace}
\newcommand{\sbTaskInternTwoSbEventUnderstanding}{60.1\xspace}
\newcommand{\sbTaskQwenFourSbEventUnderstanding}{57.6\xspace}
\newcommand{\sbTaskQwenSevenTwoFiveSbEventUnderstanding}{56.3\xspace}
\newcommand{\sbTaskLiveCcSbEventUnderstanding}{51.3\xspace}
\newcommand{\sbTaskStreamingVlmSbEventUnderstanding}{48.7\xspace}
\newcommand{\sbTaskTimeChatSbEventUnderstanding}{52.5\xspace}
\newcommand{\sbTaskQwenEightSbEventUnderstanding}{61.4\xspace}
\newcommand{\sbTaskTeacherSbEventUnderstanding}{58.9\xspace}
\newcommand{\sbTaskNSbCausalReasoning}{128\xspace}
\newcommand{\sbTaskFloorSbCausalReasoning}{33.6\xspace}
\newcommand{\sbTaskOursSbCausalReasoning}{50.8\xspace}
\newcommand{\sbTaskLfmSmallSbCausalReasoning}{50.8\xspace}
\newcommand{\sbTaskSmolSmallSbCausalReasoning}{53.9\xspace}
\newcommand{\sbTaskFastVlmSbCausalReasoning}{54.7\xspace}
\newcommand{\sbTaskMobileSbCausalReasoning}{55.5\xspace}
\newcommand{\sbTaskInternOneSbCausalReasoning}{63.3\xspace}
\newcommand{\sbTaskInternOneIntSbCausalReasoning}{60.9\xspace}
\newcommand{\sbTaskLfmBigSbCausalReasoning}{52.3\xspace}
\newcommand{\sbTaskQwenTwoSbCausalReasoning}{57.0\xspace}
\newcommand{\sbTaskQwenTwoIntSbCausalReasoning}{53.9\xspace}
\newcommand{\sbTaskSmolBigSbCausalReasoning}{58.6\xspace}
\newcommand{\sbTaskInternTwoSbCausalReasoning}{57.0\xspace}
\newcommand{\sbTaskQwenFourSbCausalReasoning}{60.9\xspace}
\newcommand{\sbTaskQwenSevenTwoFiveSbCausalReasoning}{54.7\xspace}
\newcommand{\sbTaskLiveCcSbCausalReasoning}{57.0\xspace}
\newcommand{\sbTaskStreamingVlmSbCausalReasoning}{51.6\xspace}
\newcommand{\sbTaskTimeChatSbCausalReasoning}{55.5\xspace}
\newcommand{\sbTaskQwenEightSbCausalReasoning}{60.2\xspace}
\newcommand{\sbTaskTeacherSbCausalReasoning}{57.8\xspace}
\newcommand{\sbTaskNSbProspectiveReasoning}{108\xspace}
\newcommand{\sbTaskFloorSbProspectiveReasoning}{42.6\xspace}
\newcommand{\sbTaskOursSbProspectiveReasoning}{53.7\xspace}
\newcommand{\sbTaskLfmSmallSbProspectiveReasoning}{49.1\xspace}
\newcommand{\sbTaskSmolSmallSbProspectiveReasoning}{45.4\xspace}
\newcommand{\sbTaskFastVlmSbProspectiveReasoning}{49.1\xspace}
\newcommand{\sbTaskMobileSbProspectiveReasoning}{43.5\xspace}
\newcommand{\sbTaskInternOneSbProspectiveReasoning}{58.3\xspace}
\newcommand{\sbTaskInternOneIntSbProspectiveReasoning}{58.3\xspace}
\newcommand{\sbTaskLfmBigSbProspectiveReasoning}{52.8\xspace}
\newcommand{\sbTaskQwenTwoSbProspectiveReasoning}{66.7\xspace}
\newcommand{\sbTaskQwenTwoIntSbProspectiveReasoning}{64.8\xspace}
\newcommand{\sbTaskSmolBigSbProspectiveReasoning}{50.0\xspace}
\newcommand{\sbTaskInternTwoSbProspectiveReasoning}{68.5\xspace}
\newcommand{\sbTaskQwenFourSbProspectiveReasoning}{75.9\xspace}
\newcommand{\sbTaskQwenSevenTwoFiveSbProspectiveReasoning}{52.8\xspace}
\newcommand{\sbTaskLiveCcSbProspectiveReasoning}{52.8\xspace}
\newcommand{\sbTaskStreamingVlmSbProspectiveReasoning}{64.8\xspace}
\newcommand{\sbTaskTimeChatSbProspectiveReasoning}{61.1\xspace}
\newcommand{\sbTaskQwenEightSbProspectiveReasoning}{74.1\xspace}
\newcommand{\sbTaskTeacherSbProspectiveReasoning}{75.0\xspace}

\newcommand{\sbTaskNVmObjectRecognition}{168\xspace}
\newcommand{\sbTaskFloorVmObjectRecognition}{29.2\xspace}
\newcommand{\sbTaskOursVmObjectRecognition}{47.0\xspace}
\newcommand{\sbTaskLfmSmallVmObjectRecognition}{40.5\xspace}
\newcommand{\sbTaskSmolSmallVmObjectRecognition}{44.6\xspace}
\newcommand{\sbTaskFastVlmVmObjectRecognition}{41.7\xspace}
\newcommand{\sbTaskMobileVmObjectRecognition}{38.1\xspace}
\newcommand{\sbTaskInternOneVmObjectRecognition}{50.6\xspace}
\newcommand{\sbTaskInternOneIntVmObjectRecognition}{50.0\xspace}
\newcommand{\sbTaskLfmBigVmObjectRecognition}{45.8\xspace}
\newcommand{\sbTaskQwenTwoVmObjectRecognition}{55.4\xspace}
\newcommand{\sbTaskQwenTwoIntVmObjectRecognition}{57.7\xspace}
\newcommand{\sbTaskSmolBigVmObjectRecognition}{44.6\xspace}
\newcommand{\sbTaskInternTwoVmObjectRecognition}{60.1\xspace}
\newcommand{\sbTaskQwenFourVmObjectRecognition}{57.1\xspace}
\newcommand{\sbTaskQwenSevenTwoFiveVmObjectRecognition}{47.6\xspace}
\newcommand{\sbTaskLiveCcVmObjectRecognition}{39.3\xspace}
\newcommand{\sbTaskStreamingVlmVmObjectRecognition}{45.8\xspace}
\newcommand{\sbTaskTimeChatVmObjectRecognition}{55.4\xspace}
\newcommand{\sbTaskQwenEightVmObjectRecognition}{59.5\xspace}
\newcommand{\sbTaskTeacherVmObjectRecognition}{51.2\xspace}
\newcommand{\sbTaskNVmActionRecognition}{131\xspace}
\newcommand{\sbTaskFloorVmActionRecognition}{23.7\xspace}
\newcommand{\sbTaskOursVmActionRecognition}{35.9\xspace}
\newcommand{\sbTaskLfmSmallVmActionRecognition}{42.0\xspace}
\newcommand{\sbTaskSmolSmallVmActionRecognition}{36.6\xspace}
\newcommand{\sbTaskFastVlmVmActionRecognition}{32.8\xspace}
\newcommand{\sbTaskMobileVmActionRecognition}{25.2\xspace}
\newcommand{\sbTaskInternOneVmActionRecognition}{48.9\xspace}
\newcommand{\sbTaskInternOneIntVmActionRecognition}{48.9\xspace}
\newcommand{\sbTaskLfmBigVmActionRecognition}{42.7\xspace}
\newcommand{\sbTaskQwenTwoVmActionRecognition}{50.4\xspace}
\newcommand{\sbTaskQwenTwoIntVmActionRecognition}{55.0\xspace}
\newcommand{\sbTaskSmolBigVmActionRecognition}{37.4\xspace}
\newcommand{\sbTaskInternTwoVmActionRecognition}{54.2\xspace}
\newcommand{\sbTaskQwenFourVmActionRecognition}{52.7\xspace}
\newcommand{\sbTaskQwenSevenTwoFiveVmActionRecognition}{41.2\xspace}
\newcommand{\sbTaskLiveCcVmActionRecognition}{38.2\xspace}
\newcommand{\sbTaskStreamingVlmVmActionRecognition}{52.7\xspace}
\newcommand{\sbTaskTimeChatVmActionRecognition}{54.2\xspace}
\newcommand{\sbTaskQwenEightVmActionRecognition}{50.4\xspace}
\newcommand{\sbTaskTeacherVmActionRecognition}{51.9\xspace}
\newcommand{\sbTaskNVmCountingProblem}{125\xspace}
\newcommand{\sbTaskFloorVmCountingProblem}{27.2\xspace}
\newcommand{\sbTaskOursVmCountingProblem}{28.8\xspace}
\newcommand{\sbTaskLfmSmallVmCountingProblem}{19.2\xspace}
\newcommand{\sbTaskSmolSmallVmCountingProblem}{23.2\xspace}
\newcommand{\sbTaskFastVlmVmCountingProblem}{20.8\xspace}
\newcommand{\sbTaskMobileVmCountingProblem}{20.8\xspace}
\newcommand{\sbTaskInternOneVmCountingProblem}{25.6\xspace}
\newcommand{\sbTaskInternOneIntVmCountingProblem}{23.2\xspace}
\newcommand{\sbTaskLfmBigVmCountingProblem}{24.0\xspace}
\newcommand{\sbTaskQwenTwoVmCountingProblem}{32.0\xspace}
\newcommand{\sbTaskQwenTwoIntVmCountingProblem}{26.4\xspace}
\newcommand{\sbTaskSmolBigVmCountingProblem}{24.0\xspace}
\newcommand{\sbTaskInternTwoVmCountingProblem}{31.2\xspace}
\newcommand{\sbTaskQwenFourVmCountingProblem}{30.4\xspace}
\newcommand{\sbTaskQwenSevenTwoFiveVmCountingProblem}{21.6\xspace}
\newcommand{\sbTaskLiveCcVmCountingProblem}{24.0\xspace}
\newcommand{\sbTaskStreamingVlmVmCountingProblem}{27.2\xspace}
\newcommand{\sbTaskTimeChatVmCountingProblem}{29.6\xspace}
\newcommand{\sbTaskQwenEightVmCountingProblem}{40.0\xspace}
\newcommand{\sbTaskTeacherVmCountingProblem}{31.2\xspace}
\newcommand{\sbTaskNVmAttributePerception}{122\xspace}
\newcommand{\sbTaskFloorVmAttributePerception}{25.4\xspace}
\newcommand{\sbTaskOursVmAttributePerception}{53.3\xspace}
\newcommand{\sbTaskLfmSmallVmAttributePerception}{45.9\xspace}
\newcommand{\sbTaskSmolSmallVmAttributePerception}{34.4\xspace}
\newcommand{\sbTaskFastVlmVmAttributePerception}{41.8\xspace}
\newcommand{\sbTaskMobileVmAttributePerception}{33.6\xspace}
\newcommand{\sbTaskInternOneVmAttributePerception}{52.5\xspace}
\newcommand{\sbTaskInternOneIntVmAttributePerception}{53.3\xspace}
\newcommand{\sbTaskLfmBigVmAttributePerception}{45.1\xspace}
\newcommand{\sbTaskQwenTwoVmAttributePerception}{54.9\xspace}
\newcommand{\sbTaskQwenTwoIntVmAttributePerception}{57.4\xspace}
\newcommand{\sbTaskSmolBigVmAttributePerception}{49.2\xspace}
\newcommand{\sbTaskInternTwoVmAttributePerception}{54.9\xspace}
\newcommand{\sbTaskQwenFourVmAttributePerception}{59.8\xspace}
\newcommand{\sbTaskQwenSevenTwoFiveVmAttributePerception}{39.3\xspace}
\newcommand{\sbTaskLiveCcVmAttributePerception}{36.9\xspace}
\newcommand{\sbTaskStreamingVlmVmAttributePerception}{41.0\xspace}
\newcommand{\sbTaskTimeChatVmAttributePerception}{55.7\xspace}
\newcommand{\sbTaskQwenEightVmAttributePerception}{56.6\xspace}
\newcommand{\sbTaskTeacherVmAttributePerception}{61.5\xspace}
\newcommand{\sbTaskNVmInformationSynopsis}{82\xspace}
\newcommand{\sbTaskFloorVmInformationSynopsis}{32.9\xspace}
\newcommand{\sbTaskOursVmInformationSynopsis}{67.1\xspace}
\newcommand{\sbTaskLfmSmallVmInformationSynopsis}{59.8\xspace}
\newcommand{\sbTaskSmolSmallVmInformationSynopsis}{59.8\xspace}
\newcommand{\sbTaskFastVlmVmInformationSynopsis}{57.3\xspace}
\newcommand{\sbTaskMobileVmInformationSynopsis}{54.9\xspace}
\newcommand{\sbTaskInternOneVmInformationSynopsis}{79.3\xspace}
\newcommand{\sbTaskInternOneIntVmInformationSynopsis}{78.0\xspace}
\newcommand{\sbTaskLfmBigVmInformationSynopsis}{67.1\xspace}
\newcommand{\sbTaskQwenTwoVmInformationSynopsis}{76.8\xspace}
\newcommand{\sbTaskQwenTwoIntVmInformationSynopsis}{75.6\xspace}
\newcommand{\sbTaskSmolBigVmInformationSynopsis}{64.6\xspace}
\newcommand{\sbTaskInternTwoVmInformationSynopsis}{86.6\xspace}
\newcommand{\sbTaskQwenFourVmInformationSynopsis}{80.5\xspace}
\newcommand{\sbTaskQwenSevenTwoFiveVmInformationSynopsis}{65.9\xspace}
\newcommand{\sbTaskLiveCcVmInformationSynopsis}{64.6\xspace}
\newcommand{\sbTaskStreamingVlmVmInformationSynopsis}{80.5\xspace}
\newcommand{\sbTaskTimeChatVmInformationSynopsis}{81.7\xspace}
\newcommand{\sbTaskQwenEightVmInformationSynopsis}{84.1\xspace}
\newcommand{\sbTaskTeacherVmInformationSynopsis}{80.5\xspace}
\newcommand{\sbTaskNVmObjectReasoning}{80\xspace}
\newcommand{\sbTaskFloorVmObjectReasoning}{28.8\xspace}
\newcommand{\sbTaskOursVmObjectReasoning}{52.5\xspace}
\newcommand{\sbTaskLfmSmallVmObjectReasoning}{50.0\xspace}
\newcommand{\sbTaskSmolSmallVmObjectReasoning}{51.2\xspace}
\newcommand{\sbTaskFastVlmVmObjectReasoning}{51.2\xspace}
\newcommand{\sbTaskMobileVmObjectReasoning}{46.2\xspace}
\newcommand{\sbTaskInternOneVmObjectReasoning}{62.5\xspace}
\newcommand{\sbTaskInternOneIntVmObjectReasoning}{62.5\xspace}
\newcommand{\sbTaskLfmBigVmObjectReasoning}{46.2\xspace}
\newcommand{\sbTaskQwenTwoVmObjectReasoning}{61.2\xspace}
\newcommand{\sbTaskQwenTwoIntVmObjectReasoning}{61.2\xspace}
\newcommand{\sbTaskSmolBigVmObjectReasoning}{41.2\xspace}
\newcommand{\sbTaskInternTwoVmObjectReasoning}{57.5\xspace}
\newcommand{\sbTaskQwenFourVmObjectReasoning}{66.2\xspace}
\newcommand{\sbTaskQwenSevenTwoFiveVmObjectReasoning}{45.0\xspace}
\newcommand{\sbTaskLiveCcVmObjectReasoning}{53.8\xspace}
\newcommand{\sbTaskStreamingVlmVmObjectReasoning}{62.5\xspace}
\newcommand{\sbTaskTimeChatVmObjectReasoning}{63.8\xspace}
\newcommand{\sbTaskQwenEightVmObjectReasoning}{57.5\xspace}
\newcommand{\sbTaskTeacherVmObjectReasoning}{61.2\xspace}
\newcommand{\sbTaskNVmOcrProblems}{57\xspace}
\newcommand{\sbTaskFloorVmOcrProblems}{36.8\xspace}
\newcommand{\sbTaskOursVmOcrProblems}{40.4\xspace}
\newcommand{\sbTaskLfmSmallVmOcrProblems}{36.8\xspace}
\newcommand{\sbTaskSmolSmallVmOcrProblems}{36.8\xspace}
\newcommand{\sbTaskFastVlmVmOcrProblems}{42.1\xspace}
\newcommand{\sbTaskMobileVmOcrProblems}{24.6\xspace}
\newcommand{\sbTaskInternOneVmOcrProblems}{52.6\xspace}
\newcommand{\sbTaskInternOneIntVmOcrProblems}{57.9\xspace}
\newcommand{\sbTaskLfmBigVmOcrProblems}{35.1\xspace}
\newcommand{\sbTaskQwenTwoVmOcrProblems}{42.1\xspace}
\newcommand{\sbTaskQwenTwoIntVmOcrProblems}{50.9\xspace}
\newcommand{\sbTaskSmolBigVmOcrProblems}{42.1\xspace}
\newcommand{\sbTaskInternTwoVmOcrProblems}{61.4\xspace}
\newcommand{\sbTaskQwenFourVmOcrProblems}{59.6\xspace}
\newcommand{\sbTaskQwenSevenTwoFiveVmOcrProblems}{29.8\xspace}
\newcommand{\sbTaskLiveCcVmOcrProblems}{31.6\xspace}
\newcommand{\sbTaskStreamingVlmVmOcrProblems}{45.6\xspace}
\newcommand{\sbTaskTimeChatVmOcrProblems}{56.1\xspace}
\newcommand{\sbTaskQwenEightVmOcrProblems}{61.4\xspace}
\newcommand{\sbTaskTeacherVmOcrProblems}{49.1\xspace}
\newcommand{\sbTaskNVmActionReasoning}{47\xspace}
\newcommand{\sbTaskFloorVmActionReasoning}{21.3\xspace}
\newcommand{\sbTaskOursVmActionReasoning}{38.3\xspace}
\newcommand{\sbTaskLfmSmallVmActionReasoning}{44.7\xspace}
\newcommand{\sbTaskSmolSmallVmActionReasoning}{29.8\xspace}
\newcommand{\sbTaskFastVlmVmActionReasoning}{40.4\xspace}
\newcommand{\sbTaskMobileVmActionReasoning}{46.8\xspace}
\newcommand{\sbTaskInternOneVmActionReasoning}{51.1\xspace}
\newcommand{\sbTaskInternOneIntVmActionReasoning}{51.1\xspace}
\newcommand{\sbTaskLfmBigVmActionReasoning}{46.8\xspace}
\newcommand{\sbTaskQwenTwoVmActionReasoning}{48.9\xspace}
\newcommand{\sbTaskQwenTwoIntVmActionReasoning}{48.9\xspace}
\newcommand{\sbTaskSmolBigVmActionReasoning}{42.6\xspace}
\newcommand{\sbTaskInternTwoVmActionReasoning}{61.7\xspace}
\newcommand{\sbTaskQwenFourVmActionReasoning}{51.1\xspace}
\newcommand{\sbTaskQwenSevenTwoFiveVmActionReasoning}{42.6\xspace}
\newcommand{\sbTaskLiveCcVmActionReasoning}{44.7\xspace}
\newcommand{\sbTaskStreamingVlmVmActionReasoning}{57.4\xspace}
\newcommand{\sbTaskTimeChatVmActionReasoning}{51.1\xspace}
\newcommand{\sbTaskQwenEightVmActionReasoning}{48.9\xspace}
\newcommand{\sbTaskTeacherVmActionReasoning}{44.7\xspace}
\newcommand{\sbTaskNVmSpatialPerception}{30\xspace}
\newcommand{\sbTaskFloorVmSpatialPerception}{26.7\xspace}
\newcommand{\sbTaskOursVmSpatialPerception}{46.7\xspace}
\newcommand{\sbTaskLfmSmallVmSpatialPerception}{43.3\xspace}
\newcommand{\sbTaskSmolSmallVmSpatialPerception}{36.7\xspace}
\newcommand{\sbTaskFastVlmVmSpatialPerception}{36.7\xspace}
\newcommand{\sbTaskMobileVmSpatialPerception}{36.7\xspace}
\newcommand{\sbTaskInternOneVmSpatialPerception}{40.0\xspace}
\newcommand{\sbTaskInternOneIntVmSpatialPerception}{36.7\xspace}
\newcommand{\sbTaskLfmBigVmSpatialPerception}{33.3\xspace}
\newcommand{\sbTaskQwenTwoVmSpatialPerception}{56.7\xspace}
\newcommand{\sbTaskQwenTwoIntVmSpatialPerception}{53.3\xspace}
\newcommand{\sbTaskSmolBigVmSpatialPerception}{43.3\xspace}
\newcommand{\sbTaskInternTwoVmSpatialPerception}{60.0\xspace}
\newcommand{\sbTaskQwenFourVmSpatialPerception}{66.7\xspace}
\newcommand{\sbTaskQwenSevenTwoFiveVmSpatialPerception}{46.7\xspace}
\newcommand{\sbTaskLiveCcVmSpatialPerception}{40.0\xspace}
\newcommand{\sbTaskStreamingVlmVmSpatialPerception}{40.0\xspace}
\newcommand{\sbTaskTimeChatVmSpatialPerception}{60.0\xspace}
\newcommand{\sbTaskQwenEightVmSpatialPerception}{56.7\xspace}
\newcommand{\sbTaskTeacherVmSpatialPerception}{56.7\xspace}
\newcommand{\sbTaskNVmSpatialReasoning}{27\xspace}
\newcommand{\sbTaskFloorVmSpatialReasoning}{37.0\xspace}
\newcommand{\sbTaskOursVmSpatialReasoning}{70.4\xspace}
\newcommand{\sbTaskLfmSmallVmSpatialReasoning}{59.3\xspace}
\newcommand{\sbTaskSmolSmallVmSpatialReasoning}{51.9\xspace}
\newcommand{\sbTaskFastVlmVmSpatialReasoning}{66.7\xspace}
\newcommand{\sbTaskMobileVmSpatialReasoning}{63.0\xspace}
\newcommand{\sbTaskInternOneVmSpatialReasoning}{77.8\xspace}
\newcommand{\sbTaskInternOneIntVmSpatialReasoning}{74.1\xspace}
\newcommand{\sbTaskLfmBigVmSpatialReasoning}{59.3\xspace}
\newcommand{\sbTaskQwenTwoVmSpatialReasoning}{74.1\xspace}
\newcommand{\sbTaskQwenTwoIntVmSpatialReasoning}{70.4\xspace}
\newcommand{\sbTaskSmolBigVmSpatialReasoning}{66.7\xspace}
\newcommand{\sbTaskInternTwoVmSpatialReasoning}{66.7\xspace}
\newcommand{\sbTaskQwenFourVmSpatialReasoning}{77.8\xspace}
\newcommand{\sbTaskQwenSevenTwoFiveVmSpatialReasoning}{63.0\xspace}
\newcommand{\sbTaskLiveCcVmSpatialReasoning}{44.4\xspace}
\newcommand{\sbTaskStreamingVlmVmSpatialReasoning}{70.4\xspace}
\newcommand{\sbTaskTimeChatVmSpatialReasoning}{81.5\xspace}
\newcommand{\sbTaskQwenEightVmSpatialReasoning}{74.1\xspace}
\newcommand{\sbTaskTeacherVmSpatialReasoning}{70.4\xspace}
\newcommand{\sbTaskNVmTemporalPerception}{18\xspace}
\newcommand{\sbTaskFloorVmTemporalPerception}{44.4\xspace}
\newcommand{\sbTaskOursVmTemporalPerception}{61.1\xspace}
\newcommand{\sbTaskLfmSmallVmTemporalPerception}{55.6\xspace}
\newcommand{\sbTaskSmolSmallVmTemporalPerception}{44.4\xspace}
\newcommand{\sbTaskFastVlmVmTemporalPerception}{55.6\xspace}
\newcommand{\sbTaskMobileVmTemporalPerception}{55.6\xspace}
\newcommand{\sbTaskInternOneVmTemporalPerception}{44.4\xspace}
\newcommand{\sbTaskInternOneIntVmTemporalPerception}{44.4\xspace}
\newcommand{\sbTaskLfmBigVmTemporalPerception}{44.4\xspace}
\newcommand{\sbTaskQwenTwoVmTemporalPerception}{55.6\xspace}
\newcommand{\sbTaskQwenTwoIntVmTemporalPerception}{55.6\xspace}
\newcommand{\sbTaskSmolBigVmTemporalPerception}{38.9\xspace}
\newcommand{\sbTaskInternTwoVmTemporalPerception}{38.9\xspace}
\newcommand{\sbTaskQwenFourVmTemporalPerception}{66.7\xspace}
\newcommand{\sbTaskQwenSevenTwoFiveVmTemporalPerception}{50.0\xspace}
\newcommand{\sbTaskLiveCcVmTemporalPerception}{66.7\xspace}
\newcommand{\sbTaskStreamingVlmVmTemporalPerception}{50.0\xspace}
\newcommand{\sbTaskTimeChatVmTemporalPerception}{72.2\xspace}
\newcommand{\sbTaskQwenEightVmTemporalPerception}{61.1\xspace}
\newcommand{\sbTaskTeacherVmTemporalPerception}{55.6\xspace}
\newcommand{\sbTaskNVmTemporalReasoning}{13\xspace}
\newcommand{\sbTaskFloorVmTemporalReasoning}{46.2\xspace}
\newcommand{\sbTaskOursVmTemporalReasoning}{53.8\xspace}
\newcommand{\sbTaskLfmSmallVmTemporalReasoning}{38.5\xspace}
\newcommand{\sbTaskSmolSmallVmTemporalReasoning}{46.2\xspace}
\newcommand{\sbTaskFastVlmVmTemporalReasoning}{30.8\xspace}
\newcommand{\sbTaskMobileVmTemporalReasoning}{53.8\xspace}
\newcommand{\sbTaskInternOneVmTemporalReasoning}{53.8\xspace}
\newcommand{\sbTaskInternOneIntVmTemporalReasoning}{46.2\xspace}
\newcommand{\sbTaskLfmBigVmTemporalReasoning}{53.8\xspace}
\newcommand{\sbTaskQwenTwoVmTemporalReasoning}{46.2\xspace}
\newcommand{\sbTaskQwenTwoIntVmTemporalReasoning}{38.5\xspace}
\newcommand{\sbTaskSmolBigVmTemporalReasoning}{61.5\xspace}
\newcommand{\sbTaskInternTwoVmTemporalReasoning}{69.2\xspace}
\newcommand{\sbTaskQwenFourVmTemporalReasoning}{53.8\xspace}
\newcommand{\sbTaskQwenSevenTwoFiveVmTemporalReasoning}{53.8\xspace}
\newcommand{\sbTaskLiveCcVmTemporalReasoning}{61.5\xspace}
\newcommand{\sbTaskStreamingVlmVmTemporalReasoning}{61.5\xspace}
\newcommand{\sbTaskTimeChatVmTemporalReasoning}{53.8\xspace}
\newcommand{\sbTaskQwenEightVmTemporalReasoning}{61.5\xspace}
\newcommand{\sbTaskTeacherVmTemporalReasoning}{61.5\xspace}

\newcommand{\sbTaskCount}{47\xspace}

\newcommand{\sbZeroFloorOvo}{28.4\xspace}
\newcommand{\sbZeroFloorMvb}{27.2\xspace}
\newcommand{\sbZeroFloorSb}{33.4\xspace}
\newcommand{\sbZeroFloorVm}{28.7\xspace}
\newcommand{\sbZeroFloorMean}{29.4\xspace}

\newcommand{\sbProtoFloorMean}{29.4\xspace}
\newcommand{\sbGainOursMean}{+2.5\xspace}
\newcommand{\sbGainOursMeanLo}{+1.2\xspace}
\newcommand{\sbGainOursMeanHi}{+3.7\xspace}

\newcommand{\sbZeroOursOvo}{47.2\xspace}
\newcommand{\sbZeroOursMvb}{36.4\xspace}
\newcommand{\sbZeroOursSb}{48.6\xspace}
\newcommand{\sbZeroOursVm}{40.1\xspace}
\newcommand{\sbZeroOursMean}{43.1\xspace}

\newcommand{\sbProtoOursMean}{45.6\xspace}
\newcommand{\sbGainLfmSmallMean}{$-$1.7\xspace}
\newcommand{\sbGainLfmSmallMeanLo}{$-$3.0\xspace}
\newcommand{\sbGainLfmSmallMeanHi}{$-$0.4\xspace}

\newcommand{\sbZeroLfmSmallOvo}{47.7\xspace}
\newcommand{\sbZeroLfmSmallMvb}{35.4\xspace}
\newcommand{\sbZeroLfmSmallSb}{48.0\xspace}
\newcommand{\sbZeroLfmSmallVm}{38.8\xspace}
\newcommand{\sbZeroLfmSmallMean}{42.5\xspace}

\newcommand{\sbProtoLfmSmallMean}{40.8\xspace}
\newcommand{\sbGainSmolSmallMean}{+1.5\xspace}
\newcommand{\sbGainSmolSmallMeanLo}{+0.3\xspace}
\newcommand{\sbGainSmolSmallMeanHi}{+2.8\xspace}

\newcommand{\sbZeroSmolSmallOvo}{41.7\xspace}
\newcommand{\sbZeroSmolSmallMvb}{34.0\xspace}
\newcommand{\sbZeroSmolSmallSb}{44.2\xspace}
\newcommand{\sbZeroSmolSmallVm}{36.0\xspace}
\newcommand{\sbZeroSmolSmallMean}{39.0\xspace}

\newcommand{\sbProtoSmolSmallMean}{40.5\xspace}
\newcommand{\sbGainFastVlmMean}{$-$0.5\xspace}
\newcommand{\sbGainFastVlmMeanLo}{$-$1.7\xspace}
\newcommand{\sbGainFastVlmMeanHi}{+0.8\xspace}

\newcommand{\sbZeroFastVlmOvo}{47.4\xspace}
\newcommand{\sbZeroFastVlmMvb}{37.2\xspace}
\newcommand{\sbZeroFastVlmSb}{48.6\xspace}
\newcommand{\sbZeroFastVlmVm}{37.8\xspace}
\newcommand{\sbZeroFastVlmMean}{42.8\xspace}

\newcommand{\sbProtoFastVlmMean}{42.3\xspace}
\newcommand{\sbGainMobileMean}{+1.4\xspace}
\newcommand{\sbGainMobileMeanLo}{+0.1\xspace}
\newcommand{\sbGainMobileMeanHi}{+2.7\xspace}

\newcommand{\sbZeroMobileOvo}{37.0\xspace}
\newcommand{\sbZeroMobileMvb}{32.4\xspace}
\newcommand{\sbZeroMobileSb}{41.9\xspace}
\newcommand{\sbZeroMobileVm}{31.2\xspace}
\newcommand{\sbZeroMobileMean}{35.6\xspace}

\newcommand{\sbProtoMobileMean}{37.0\xspace}
\newcommand{\sbGainInternOneMean}{+2.7\xspace}
\newcommand{\sbGainInternOneMeanLo}{+1.3\xspace}
\newcommand{\sbGainInternOneMeanHi}{+4.1\xspace}

\newcommand{\sbZeroInternOneOvo}{51.0\xspace}
\newcommand{\sbZeroInternOneMvb}{38.4\xspace}
\newcommand{\sbZeroInternOneSb}{52.4\xspace}
\newcommand{\sbZeroInternOneVm}{41.7\xspace}
\newcommand{\sbZeroInternOneMean}{45.9\xspace}

\newcommand{\sbProtoInternOneMean}{48.6\xspace}
\newcommand{\sbGainLfmBigMean}{$-$3.8\xspace}
\newcommand{\sbGainLfmBigMeanLo}{$-$5.2\xspace}
\newcommand{\sbGainLfmBigMeanHi}{$-$2.5\xspace}

\newcommand{\sbZeroLfmBigOvo}{52.7\xspace}
\newcommand{\sbZeroLfmBigMvb}{38.0\xspace}
\newcommand{\sbZeroLfmBigSb}{51.1\xspace}
\newcommand{\sbZeroLfmBigVm}{42.0\xspace}
\newcommand{\sbZeroLfmBigMean}{45.9\xspace}

\newcommand{\sbProtoLfmBigMean}{42.1\xspace}
\newcommand{\sbGainQwenTwoMean}{+3.2\xspace}
\newcommand{\sbGainQwenTwoMeanLo}{+1.9\xspace}
\newcommand{\sbGainQwenTwoMeanHi}{+4.5\xspace}

\newcommand{\sbZeroQwenTwoOvo}{53.0\xspace}
\newcommand{\sbZeroQwenTwoMvb}{39.8\xspace}
\newcommand{\sbZeroQwenTwoSb}{53.8\xspace}
\newcommand{\sbZeroQwenTwoVm}{41.9\xspace}
\newcommand{\sbZeroQwenTwoMean}{47.1\xspace}

\newcommand{\sbProtoQwenTwoMean}{50.3\xspace}
\newcommand{\sbGainSmolBigMean}{+1.5\xspace}
\newcommand{\sbGainSmolBigMeanLo}{+0.2\xspace}
\newcommand{\sbGainSmolBigMeanHi}{+2.8\xspace}
\newcommand{\sbTableOneSmolBigMean}{43.8\xspace}
\newcommand{\sbZeroSmolBigOvo}{46.0\xspace}
\newcommand{\sbZeroSmolBigMvb}{36.6\xspace}
\newcommand{\sbZeroSmolBigSb}{47.8\xspace}
\newcommand{\sbZeroSmolBigVm}{39.7\xspace}
\newcommand{\sbZeroSmolBigMean}{42.5\xspace}

\newcommand{\sbProtoSmolBigMean}{44.0\xspace}
\newcommand{\sbGainInternTwoMean}{+4.0\xspace}
\newcommand{\sbGainInternTwoMeanLo}{+2.6\xspace}
\newcommand{\sbGainInternTwoMeanHi}{+5.4\xspace}

\newcommand{\sbZeroInternTwoOvo}{53.8\xspace}
\newcommand{\sbZeroInternTwoMvb}{38.9\xspace}
\newcommand{\sbZeroInternTwoSb}{53.5\xspace}
\newcommand{\sbZeroInternTwoVm}{42.0\xspace}
\newcommand{\sbZeroInternTwoMean}{47.0\xspace}

\newcommand{\sbProtoInternTwoMean}{51.1\xspace}
\newcommand{\sbGainQwenFourMean}{+3.4\xspace}
\newcommand{\sbGainQwenFourMeanLo}{+2.1\xspace}
\newcommand{\sbGainQwenFourMeanHi}{+4.8\xspace}

\newcommand{\sbZeroQwenFourOvo}{54.5\xspace}
\newcommand{\sbZeroQwenFourMvb}{39.1\xspace}
\newcommand{\sbZeroQwenFourSb}{55.1\xspace}
\newcommand{\sbZeroQwenFourVm}{42.8\xspace}
\newcommand{\sbZeroQwenFourMean}{47.8\xspace}

\newcommand{\sbProtoQwenFourMean}{51.2\xspace}
\newcommand{\sbGainQwenSevenTwoFiveMean}{$-$2.7\xspace}
\newcommand{\sbGainQwenSevenTwoFiveMeanLo}{$-$4.1\xspace}
\newcommand{\sbGainQwenSevenTwoFiveMeanHi}{$-$1.3\xspace}

\newcommand{\sbZeroQwenSevenTwoFiveOvo}{51.9\xspace}
\newcommand{\sbZeroQwenSevenTwoFiveMvb}{38.3\xspace}
\newcommand{\sbZeroQwenSevenTwoFiveSb}{52.0\xspace}
\newcommand{\sbZeroQwenSevenTwoFiveVm}{42.6\xspace}
\newcommand{\sbZeroQwenSevenTwoFiveMean}{46.2\xspace}

\newcommand{\sbProtoQwenSevenTwoFiveMean}{43.5\xspace}
\newcommand{\sbGainLiveCcMean}{$-$0.8\xspace}
\newcommand{\sbGainLiveCcMeanLo}{$-$2.2\xspace}
\newcommand{\sbGainLiveCcMeanHi}{+0.4\xspace}

\newcommand{\sbZeroLiveCcOvo}{45.9\xspace}
\newcommand{\sbZeroLiveCcMvb}{36.2\xspace}
\newcommand{\sbZeroLiveCcSb}{48.0\xspace}
\newcommand{\sbZeroLiveCcVm}{36.7\xspace}
\newcommand{\sbZeroLiveCcMean}{41.7\xspace}

\newcommand{\sbProtoLiveCcMean}{40.8\xspace}
\newcommand{\sbGainStreamingVlmMean}{+6.0\xspace}
\newcommand{\sbGainStreamingVlmMeanLo}{+4.6\xspace}
\newcommand{\sbGainStreamingVlmMeanHi}{+7.3\xspace}

\newcommand{\sbZeroStreamingVlmOvo}{43.2\xspace}
\newcommand{\sbZeroStreamingVlmMvb}{36.3\xspace}
\newcommand{\sbZeroStreamingVlmSb}{46.9\xspace}
\newcommand{\sbZeroStreamingVlmVm}{38.0\xspace}
\newcommand{\sbZeroStreamingVlmMean}{41.1\xspace}

\newcommand{\sbProtoStreamingVlmMean}{47.1\xspace}
\newcommand{\sbGainTimeChatMean}{+3.3\xspace}
\newcommand{\sbGainTimeChatMeanLo}{+1.9\xspace}
\newcommand{\sbGainTimeChatMeanHi}{+4.6\xspace}

\newcommand{\sbZeroTimeChatOvo}{50.7\xspace}
\newcommand{\sbZeroTimeChatMvb}{38.0\xspace}
\newcommand{\sbZeroTimeChatSb}{53.0\xspace}
\newcommand{\sbZeroTimeChatVm}{44.1\xspace}
\newcommand{\sbZeroTimeChatMean}{46.5\xspace}

\newcommand{\sbProtoTimeChatMean}{49.7\xspace}
\newcommand{\sbGainQwenEightMean}{+4.1\xspace}
\newcommand{\sbGainQwenEightMeanLo}{+2.8\xspace}
\newcommand{\sbGainQwenEightMeanHi}{+5.4\xspace}

\newcommand{\sbZeroQwenEightOvo}{55.0\xspace}
\newcommand{\sbZeroQwenEightMvb}{38.7\xspace}
\newcommand{\sbZeroQwenEightSb}{54.1\xspace}
\newcommand{\sbZeroQwenEightVm}{43.2\xspace}
\newcommand{\sbZeroQwenEightMean}{47.7\xspace}

\newcommand{\sbProtoQwenEightMean}{51.9\xspace}
\newcommand{\sbGainTeacherMean}{+6.8\xspace}
\newcommand{\sbGainTeacherMeanLo}{+5.4\xspace}
\newcommand{\sbGainTeacherMeanHi}{+8.1\xspace}

\newcommand{\sbZeroTeacherOvo}{51.3\xspace}
\newcommand{\sbZeroTeacherMvb}{37.3\xspace}
\newcommand{\sbZeroTeacherSb}{53.0\xspace}
\newcommand{\sbZeroTeacherVm}{41.7\xspace}
\newcommand{\sbZeroTeacherMean}{45.8\xspace}

\newcommand{\sbProtoTeacherMean}{52.6\xspace}
\newcommand{\sbShiftSameMax}{0.34\xspace}
\newcommand{\sbShiftRecaptionedMax}{0.64\xspace}
\newcommand{\sbShiftRecaptionedCount}{9\xspace}

\newcommand{\sbGainModels}{17\xspace}

\newcommand{\sbSelSmolBigOffShare}{59\xspace}
\newcommand{\sbSelLfmSmallWzeroKzero}{0.2221\xspace}
\newcommand{\sbSelLfmSmallWzeroKsixteen}{0.2927\xspace}
\newcommand{\sbSelLfmSmallWzeroMean}{0.2574\xspace}
\newcommand{\sbSelLfmSmallWoneKzero}{0.3393\xspace}
\newcommand{\sbSelLfmSmallWoneKsixteen}{0.3307\xspace}
\newcommand{\sbSelLfmSmallWoneMean}{0.3350\xspace}
\newcommand{\sbSelLfmSmallWtwoKzero}{0.3245\xspace}
\newcommand{\sbSelLfmSmallWtwoKsixteen}{0.2042\xspace}
\newcommand{\sbSelLfmSmallWtwoMean}{0.2643\xspace}
\newcommand{\sbSelLfmSmallChosen}{w1\xspace}
\newcommand{\sbSelSmolSmallWzeroKzero}{0.2562\xspace}
\newcommand{\sbSelSmolSmallWzeroKsixteen}{0.3577\xspace}
\newcommand{\sbSelSmolSmallWzeroMean}{0.3070\xspace}
\newcommand{\sbSelSmolSmallWoneKzero}{0.2360\xspace}
\newcommand{\sbSelSmolSmallWoneKsixteen}{0.3456\xspace}
\newcommand{\sbSelSmolSmallWoneMean}{0.2908\xspace}
\newcommand{\sbSelSmolSmallWtwoKzero}{0.2605\xspace}
\newcommand{\sbSelSmolSmallWtwoKsixteen}{0.2661\xspace}
\newcommand{\sbSelSmolSmallWtwoMean}{0.2633\xspace}
\newcommand{\sbSelSmolSmallChosen}{w0\xspace}
\newcommand{\sbSelFastVlmWzeroKzero}{0.1548\xspace}
\newcommand{\sbSelFastVlmWzeroKsixteen}{0.1590\xspace}
\newcommand{\sbSelFastVlmWzeroMean}{0.1569\xspace}
\newcommand{\sbSelFastVlmWoneKzero}{0.2360\xspace}
\newcommand{\sbSelFastVlmWoneKsixteen}{0.2938\xspace}
\newcommand{\sbSelFastVlmWoneMean}{0.2649\xspace}
\newcommand{\sbSelFastVlmWtwoKzero}{0.0671\xspace}
\newcommand{\sbSelFastVlmWtwoKsixteen}{0.0000\xspace}
\newcommand{\sbSelFastVlmWtwoMean}{0.0335\xspace}
\newcommand{\sbSelFastVlmChosen}{w1\xspace}
\newcommand{\sbSelMobileWzeroKzero}{0.2149\xspace}
\newcommand{\sbSelMobileWzeroKsixteen}{0.2938\xspace}
\newcommand{\sbSelMobileWzeroMean}{0.2544\xspace}
\newcommand{\sbSelMobileWoneKzero}{0.1925\xspace}
\newcommand{\sbSelMobileWoneKsixteen}{0.2417\xspace}
\newcommand{\sbSelMobileWoneMean}{0.2171\xspace}
\newcommand{\sbSelMobileWtwoKzero}{0.2224\xspace}
\newcommand{\sbSelMobileWtwoKsixteen}{0.2868\xspace}
\newcommand{\sbSelMobileWtwoMean}{0.2546\xspace}
\newcommand{\sbSelMobileChosen}{w2\xspace}
\newcommand{\sbSelInternOneWzeroKzero}{0.3647\xspace}
\newcommand{\sbSelInternOneWzeroKsixteen}{0.4927\xspace}
\newcommand{\sbSelInternOneWzeroMean}{0.4287\xspace}
\newcommand{\sbSelInternOneWoneKzero}{0.3601\xspace}
\newcommand{\sbSelInternOneWoneKsixteen}{0.4758\xspace}
\newcommand{\sbSelInternOneWoneMean}{0.4179\xspace}
\newcommand{\sbSelInternOneWtwoKzero}{0.3568\xspace}
\newcommand{\sbSelInternOneWtwoKsixteen}{0.2107\xspace}
\newcommand{\sbSelInternOneWtwoMean}{0.2838\xspace}
\newcommand{\sbSelInternOneChosen}{w0\xspace}
\newcommand{\sbSelLfmBigWzeroKzero}{0.3371\xspace}
\newcommand{\sbSelLfmBigWzeroKsixteen}{0.3209\xspace}
\newcommand{\sbSelLfmBigWzeroMean}{0.3290\xspace}
\newcommand{\sbSelLfmBigWoneKzero}{0.3383\xspace}
\newcommand{\sbSelLfmBigWoneKsixteen}{0.3212\xspace}
\newcommand{\sbSelLfmBigWoneMean}{0.3298\xspace}
\newcommand{\sbSelLfmBigWtwoKzero}{0.3445\xspace}
\newcommand{\sbSelLfmBigWtwoKsixteen}{0.1720\xspace}
\newcommand{\sbSelLfmBigWtwoMean}{0.2582\xspace}
\newcommand{\sbSelLfmBigChosen}{w1\xspace}
\newcommand{\sbSelQwenTwoWzeroKzero}{0.3949\xspace}
\newcommand{\sbSelQwenTwoWzeroKsixteen}{0.5622\xspace}
\newcommand{\sbSelQwenTwoWzeroMean}{0.4785\xspace}
\newcommand{\sbSelQwenTwoWoneKzero}{0.4207\xspace}
\newcommand{\sbSelQwenTwoWoneKsixteen}{0.5082\xspace}
\newcommand{\sbSelQwenTwoWoneMean}{0.4645\xspace}
\newcommand{\sbSelQwenTwoWtwoKzero}{0.3964\xspace}
\newcommand{\sbSelQwenTwoWtwoKsixteen}{0.1961\xspace}
\newcommand{\sbSelQwenTwoWtwoMean}{0.2963\xspace}
\newcommand{\sbSelQwenTwoChosen}{w0\xspace}
\newcommand{\sbSelSmolBigWzeroKzero}{0.3187\xspace}
\newcommand{\sbSelSmolBigWzeroKsixteen}{0.4034\xspace}
\newcommand{\sbSelSmolBigWzeroMean}{0.3611\xspace}
\newcommand{\sbSelSmolBigWoneKzero}{0.3087\xspace}
\newcommand{\sbSelSmolBigWoneKsixteen}{0.3454\xspace}
\newcommand{\sbSelSmolBigWoneMean}{0.3270\xspace}
\newcommand{\sbSelSmolBigWtwoKzero}{0.3437\xspace}
\newcommand{\sbSelSmolBigWtwoKsixteen}{0.4034\xspace}
\newcommand{\sbSelSmolBigWtwoMean}{0.3736\xspace}
\newcommand{\sbSelSmolBigChosen}{w2\xspace}
\newcommand{\sbSelInternTwoWzeroKzero}{0.3505\xspace}
\newcommand{\sbSelInternTwoWzeroKsixteen}{0.4206\xspace}
\newcommand{\sbSelInternTwoWzeroMean}{0.3856\xspace}
\newcommand{\sbSelInternTwoWoneKzero}{0.3738\xspace}
\newcommand{\sbSelInternTwoWoneKsixteen}{0.2616\xspace}
\newcommand{\sbSelInternTwoWoneMean}{0.3177\xspace}
\newcommand{\sbSelInternTwoWtwoKzero}{0.3419\xspace}
\newcommand{\sbSelInternTwoWtwoKsixteen}{0.0522\xspace}
\newcommand{\sbSelInternTwoWtwoMean}{0.1970\xspace}
\newcommand{\sbSelInternTwoChosen}{w0\xspace}
\newcommand{\sbSelQwenFourWzeroKzero}{0.4317\xspace}
\newcommand{\sbSelQwenFourWzeroKsixteen}{0.5844\xspace}
\newcommand{\sbSelQwenFourWzeroMean}{0.5080\xspace}
\newcommand{\sbSelQwenFourWoneKzero}{0.4073\xspace}
\newcommand{\sbSelQwenFourWoneKsixteen}{0.5493\xspace}
\newcommand{\sbSelQwenFourWoneMean}{0.4783\xspace}
\newcommand{\sbSelQwenFourWtwoKzero}{0.3860\xspace}
\newcommand{\sbSelQwenFourWtwoKsixteen}{0.5910\xspace}
\newcommand{\sbSelQwenFourWtwoMean}{0.4885\xspace}
\newcommand{\sbSelQwenFourChosen}{w0\xspace}
\newcommand{\sbSelQwenSevenTwoFiveWzeroKzero}{0.3787\xspace}
\newcommand{\sbSelQwenSevenTwoFiveWzeroKsixteen}{0.3391\xspace}
\newcommand{\sbSelQwenSevenTwoFiveWzeroMean}{0.3589\xspace}
\newcommand{\sbSelQwenSevenTwoFiveWoneKzero}{0.3621\xspace}
\newcommand{\sbSelQwenSevenTwoFiveWoneKsixteen}{0.2577\xspace}
\newcommand{\sbSelQwenSevenTwoFiveWoneMean}{0.3099\xspace}
\newcommand{\sbSelQwenSevenTwoFiveWtwoKzero}{0.3811\xspace}
\newcommand{\sbSelQwenSevenTwoFiveWtwoKsixteen}{0.1653\xspace}
\newcommand{\sbSelQwenSevenTwoFiveWtwoMean}{0.2732\xspace}
\newcommand{\sbSelQwenSevenTwoFiveChosen}{w0\xspace}
\newcommand{\sbSelLiveCcWzeroKzero}{0.2843\xspace}
\newcommand{\sbSelLiveCcWzeroKsixteen}{0.2949\xspace}
\newcommand{\sbSelLiveCcWzeroMean}{0.2896\xspace}
\newcommand{\sbSelLiveCcWoneKzero}{0.2948\xspace}
\newcommand{\sbSelLiveCcWoneKsixteen}{0.2889\xspace}
\newcommand{\sbSelLiveCcWoneMean}{0.2919\xspace}
\newcommand{\sbSelLiveCcWtwoKzero}{0.2484\xspace}
\newcommand{\sbSelLiveCcWtwoKsixteen}{0.0689\xspace}
\newcommand{\sbSelLiveCcWtwoMean}{0.1587\xspace}
\newcommand{\sbSelLiveCcChosen}{w1\xspace}
\newcommand{\sbSelStreamingVlmWzeroKzero}{0.0323\xspace}
\newcommand{\sbSelStreamingVlmWzeroKsixteen}{0.0356\xspace}
\newcommand{\sbSelStreamingVlmWzeroMean}{0.0339\xspace}
\newcommand{\sbSelStreamingVlmWoneKzero}{0.0117\xspace}
\newcommand{\sbSelStreamingVlmWoneKsixteen}{0.0212\xspace}
\newcommand{\sbSelStreamingVlmWoneMean}{0.0164\xspace}
\newcommand{\sbSelStreamingVlmWtwoKzero}{0.0075\xspace}
\newcommand{\sbSelStreamingVlmWtwoKsixteen}{0.0126\xspace}
\newcommand{\sbSelStreamingVlmWtwoMean}{0.0100\xspace}
\newcommand{\sbSelStreamingVlmChosen}{w0\xspace}
\newcommand{\sbSelTimeChatWzeroKzero}{0.3515\xspace}
\newcommand{\sbSelTimeChatWzeroKsixteen}{0.4086\xspace}
\newcommand{\sbSelTimeChatWzeroMean}{0.3800\xspace}
\newcommand{\sbSelTimeChatWoneKzero}{0.3552\xspace}
\newcommand{\sbSelTimeChatWoneKsixteen}{0.2740\xspace}
\newcommand{\sbSelTimeChatWoneMean}{0.3146\xspace}
\newcommand{\sbSelTimeChatWtwoKzero}{0.3250\xspace}
\newcommand{\sbSelTimeChatWtwoKsixteen}{0.1803\xspace}
\newcommand{\sbSelTimeChatWtwoMean}{0.2527\xspace}
\newcommand{\sbSelTimeChatChosen}{w0\xspace}
\newcommand{\sbSelQwenEightWzeroKzero}{0.3976\xspace}
\newcommand{\sbSelQwenEightWzeroKsixteen}{0.5921\xspace}
\newcommand{\sbSelQwenEightWzeroMean}{0.4949\xspace}
\newcommand{\sbSelQwenEightWoneKzero}{0.3587\xspace}
\newcommand{\sbSelQwenEightWoneKsixteen}{0.5309\xspace}
\newcommand{\sbSelQwenEightWoneMean}{0.4448\xspace}
\newcommand{\sbSelQwenEightWtwoKzero}{0.3727\xspace}
\newcommand{\sbSelQwenEightWtwoKsixteen}{0.6179\xspace}
\newcommand{\sbSelQwenEightWtwoMean}{0.4953\xspace}
\newcommand{\sbSelQwenEightChosen}{w2\xspace}
\newcommand{\sbSelMarginMin}{0.0002\xspace}

%% file: numbers/supp_protocol.tex
\newcommand{\spOpenvidClips}{599,503\xspace}

\newcommand{\spStride}{8\xspace}
\newcommand{\spWindowsPerVideo}{6\xspace}
\newcommand{\spSftVal}{249\xspace}

\newcommand{\spRlTemperature}{0.7\xspace}

\newcommand{\spMinKept}{6\xspace}
\newcommand{\spQuestionsWritten}{8\xspace}
\newcommand{\spQuestionsKept}{6\xspace}
\newcommand{\spLeakRotations}{4\xspace}
\newcommand{\spLeakMaxRight}{1\xspace}
\newcommand{\spLabelMaxWords}{60\xspace}
\newcommand{\spLabelMaxTokens}{80\xspace}

\newcommand{\spPilotNoCeiling}{354\xspace}
\newcommand{\spPilotPerVideo}{27\xspace}

\newcommand{\spTestNoCeiling}{530\xspace}

\newcommand{\spOpenvidNoCeiling}{244\xspace}

\newcommand{\spPilotAudit}{581\xspace}
\newcommand{\spOpenvidSourceVideos}{405\xspace}
\newcommand{\spRlValInDev}{522\xspace}
\newcommand{\spRlValOutDev}{4\xspace}

\newcommand{\spCeilingOthers}{7\xspace}
\newcommand{\spCeilingNeed}{5\xspace}
\newcommand{\spPreWindows}{335,822\xspace}
\newcommand{\spPreSteps}{2,622\xspace}

\newcommand{\spSftEvalEvery}{250\xspace}

\newcommand{\spOneRlExit}{550\xspace}
\newcommand{\spOneRlJobHours}{19.2\xspace}
\newcommand{\spOneRlGpus}{3\xspace}
\newcommand{\spOneSftJobHours}{7.4\xspace}

\newcommand{\spSftGpu}{RTX 5090\xspace}
\newcommand{\spRlTrainerGpu}{RTX 4090\xspace}
\newcommand{\spRlJudgeGpu}{RTX 5090\xspace}
\newcommand{\spOneSftGpu}{A40\xspace}
\newcommand{\spOneRlGpu}{A100-40GB\xspace}

%% file: numbers/supp_training.tex
\newcommand{\stTuneLoraSftKzero}{43.8\xspace}
\newcommand{\stTuneLoraSftKtwo}{47.9\xspace}
\newcommand{\stTuneLoraSftKfour}{50.2\xspace}
\newcommand{\stTuneLoraSftKsixteen}{58.0\xspace}

\newcommand{\stTuneLoraGrpoKzero}{54.0\xspace}
\newcommand{\stTuneLoraGrpoKtwo}{57.7\xspace}
\newcommand{\stTuneLoraGrpoKfour}{59.1\xspace}
\newcommand{\stTuneLoraGrpoKsixteen}{66.9\xspace}
\newcommand{\stTuneBigKzero}{54.1\xspace}
\newcommand{\stTuneBigKtwo}{58.1\xspace}
\newcommand{\stTuneBigKfour}{63.5\xspace}
\newcommand{\stTuneBigKsixteen}{74.0\xspace}
\newcommand{\stTuneKdLoraKzero}{43.6\xspace}
\newcommand{\stTuneKdLoraKtwo}{47.8\xspace}
\newcommand{\stTuneKdLoraKfour}{50.6\xspace}
\newcommand{\stTuneKdLoraKsixteen}{57.2\xspace}
\newcommand{\stTuneOpLoraKzero}{43.1\xspace}
\newcommand{\stTuneOpLoraKtwo}{48.3\xspace}
\newcommand{\stTuneOpLoraKfour}{50.7\xspace}
\newcommand{\stTuneOpLoraKsixteen}{57.6\xspace}
\newcommand{\stTuneKdFullKzero}{47.9\xspace}
\newcommand{\stTuneKdFullKtwo}{52.1\xspace}
\newcommand{\stTuneKdFullKfour}{54.5\xspace}
\newcommand{\stTuneKdFullKsixteen}{65.7\xspace}
\newcommand{\stTuneFullSftOverLoraSftKzero}{+3.4\xspace}
\newcommand{\stTuneFullSftOverLoraSftKzeroLo}{+2.0\xspace}
\newcommand{\stTuneFullSftOverLoraSftKzeroHi}{+4.7\xspace}
\newcommand{\stTuneFullSftOverLoraSftKsixteen}{+7.3\xspace}
\newcommand{\stTuneFullSftOverLoraSftKsixteenLo}{+5.8\xspace}
\newcommand{\stTuneFullSftOverLoraSftKsixteenHi}{+8.9\xspace}
\newcommand{\stTuneLoraGrpoOverLoraSftKzero}{+10.2\xspace}
\newcommand{\stTuneLoraGrpoOverLoraSftKzeroLo}{+8.7\xspace}
\newcommand{\stTuneLoraGrpoOverLoraSftKzeroHi}{+11.8\xspace}
\newcommand{\stTuneLoraGrpoOverLoraSftKsixteen}{+8.9\xspace}
\newcommand{\stTuneLoraGrpoOverLoraSftKsixteenLo}{+7.1\xspace}
\newcommand{\stTuneLoraGrpoOverLoraSftKsixteenHi}{+10.5\xspace}
\newcommand{\stTuneBigOverLoraSftKzero}{+10.3\xspace}
\newcommand{\stTuneBigOverLoraSftKzeroLo}{+8.7\xspace}
\newcommand{\stTuneBigOverLoraSftKzeroHi}{+11.9\xspace}
\newcommand{\stTuneBigOverLoraSftKsixteen}{+16.0\xspace}
\newcommand{\stTuneBigOverLoraSftKsixteenLo}{+14.3\xspace}
\newcommand{\stTuneBigOverLoraSftKsixteenHi}{+17.7\xspace}
\newcommand{\stTuneKdLoraOverLoraSftKzero}{$-$0.2\xspace}
\newcommand{\stTuneKdLoraOverLoraSftKzeroLo}{$-$1.2\xspace}
\newcommand{\stTuneKdLoraOverLoraSftKzeroHi}{+0.7\xspace}
\newcommand{\stTuneKdLoraOverLoraSftKsixteen}{$-$0.8\xspace}
\newcommand{\stTuneKdLoraOverLoraSftKsixteenLo}{$-$2.0\xspace}
\newcommand{\stTuneKdLoraOverLoraSftKsixteenHi}{+0.4\xspace}
\newcommand{\stTuneOpLoraOverLoraSftKzero}{$-$0.7\xspace}
\newcommand{\stTuneOpLoraOverLoraSftKzeroLo}{$-$1.8\xspace}
\newcommand{\stTuneOpLoraOverLoraSftKzeroHi}{+0.4\xspace}
\newcommand{\stTuneOpLoraOverLoraSftKsixteen}{$-$0.4\xspace}
\newcommand{\stTuneOpLoraOverLoraSftKsixteenLo}{$-$1.7\xspace}
\newcommand{\stTuneOpLoraOverLoraSftKsixteenHi}{+0.9\xspace}
\newcommand{\stTuneKdFullOverFullSftKzero}{+0.7\xspace}
\newcommand{\stTuneKdFullOverFullSftKzeroLo}{$-$0.3\xspace}
\newcommand{\stTuneKdFullOverFullSftKzeroHi}{+1.7\xspace}
\newcommand{\stTuneKdFullOverFullSftKsixteen}{+0.4\xspace}
\newcommand{\stTuneKdFullOverFullSftKsixteenLo}{$-$0.8\xspace}
\newcommand{\stTuneKdFullOverFullSftKsixteenHi}{+1.5\xspace}
\newcommand{\stPoolSixtyFourKzero}{47.1\xspace}
\newcommand{\stPoolSixtyFourKone}{51.2\xspace}
\newcommand{\stPoolSixtyFourKtwo}{52.7\xspace}
\newcommand{\stPoolSixtyFourKfour}{54.8\xspace}
\newcommand{\stPoolSixtyFourPast}{256\xspace}
\newcommand{\stPoolSixtyFourFrames}{4\xspace}
\newcommand{\stPoolSixtyFourTokens}{64\xspace}
\newcommand{\stPoolSixteenKzero}{47.1\xspace}
\newcommand{\stPoolSixteenKone}{50.9\xspace}
\newcommand{\stPoolSixteenKtwo}{52.6\xspace}
\newcommand{\stPoolSixteenKfour}{55.2\xspace}
\newcommand{\stPoolSixteenKeight}{59.8\xspace}
\newcommand{\stPoolSixteenKsixteen}{65.0\xspace}
\newcommand{\stPoolSixteenPast}{256\xspace}
\newcommand{\stPoolSixteenFrames}{16\xspace}
\newcommand{\stPoolSixteenTokens}{16\xspace}
\newcommand{\stPoolFourKzero}{47.1\xspace}
\newcommand{\stPoolFourKone}{49.1\xspace}
\newcommand{\stPoolFourKtwo}{51.1\xspace}
\newcommand{\stPoolFourKfour}{53.2\xspace}
\newcommand{\stPoolFourKeight}{57.1\xspace}
\newcommand{\stPoolFourKsixteen}{61.0\xspace}
\newcommand{\stPoolFourPast}{64\xspace}
\newcommand{\stPoolFourFrames}{16\xspace}
\newcommand{\stPoolFourTokens}{4\xspace}

\newcommand{\stKlRecallSft}{0.487\xspace}
\newcommand{\stKlPrecisionSft}{0.502\xspace}
\newcommand{\stKlWordsSft}{44.1\xspace}

\newcommand{\stKlCloserSft}{0.4\xspace}
\newcommand{\stKlHedgeSft}{0.9\xspace}
\newcommand{\stKlTestWordsSft}{44.5\xspace}

\newcommand{\stKlRecallKl}{0.502\xspace}
\newcommand{\stKlPrecisionKl}{0.580\xspace}
\newcommand{\stKlWordsKl}{40.4\xspace}

\newcommand{\stKlCloserKl}{2.7\xspace}
\newcommand{\stKlHedgeKl}{1.7\xspace}
\newcommand{\stKlTestWordsKl}{40.3\xspace}

\newcommand{\stKlRecallNoKl}{0.453\xspace}
\newcommand{\stKlPrecisionNoKl}{0.741\xspace}
\newcommand{\stKlWordsNoKl}{46.9\xspace}

\newcommand{\stKlCloserNoKl}{86.2\xspace}
\newcommand{\stKlHedgeNoKl}{20.4\xspace}
\newcommand{\stKlTestWordsNoKl}{47.0\xspace}

\newcommand{\stKlOverNoKlCloser}{$-$83.6\xspace}
\newcommand{\stKlOverNoKlCloserLo}{$-$85.3\xspace}
\newcommand{\stKlOverNoKlCloserHi}{$-$81.8\xspace}
\newcommand{\stKlOverNoKlHedge}{$-$18.8\xspace}
\newcommand{\stKlOverNoKlHedgeLo}{$-$20.7\xspace}
\newcommand{\stKlOverNoKlHedgeHi}{$-$16.9\xspace}
\newcommand{\stKlOverNoKlTestWords}{$-$6.6\xspace}
\newcommand{\stKlOverNoKlTestWordsLo}{$-$6.9\xspace}
\newcommand{\stKlOverNoKlTestWordsHi}{$-$6.4\xspace}
\newcommand{\stBenchQuestionsOvo}{837\xspace}
\newcommand{\stBenchQuestionsMvb}{3,800\xspace}
\newcommand{\stBenchQuestionsSb}{2,498\xspace}
\newcommand{\stBenchQuestionsVm}{900\xspace}
\newcommand{\stBenchOneSftOvo}{43.8\xspace}
\newcommand{\stBenchOneSftMvb}{39.8\xspace}
\newcommand{\stBenchOneSftSb}{47.4\xspace}
\newcommand{\stBenchOneSftVm}{44.8\xspace}
\newcommand{\stBenchOneSftMean}{44.0\xspace}
\newcommand{\stBenchOneOvEqOvo}{43.5\xspace}
\newcommand{\stBenchOneOvEqMvb}{39.5\xspace}
\newcommand{\stBenchOneOvEqSb}{47.0\xspace}
\newcommand{\stBenchOneOvEqVm}{44.4\xspace}
\newcommand{\stBenchOneOvEqMean}{43.6\xspace}
\newcommand{\stBenchOneOvSftOvo}{46.7\xspace}
\newcommand{\stBenchOneOvSftMvb}{40.5\xspace}
\newcommand{\stBenchOneOvSftSb}{49.0\xspace}
\newcommand{\stBenchOneOvSftVm}{45.9\xspace}
\newcommand{\stBenchOneOvSftMean}{45.5\xspace}
\newcommand{\stBenchOneOvSftOverSftOvo}{+2.9\xspace}
\newcommand{\stBenchOneOvSftOverSftOvoLo}{$-$0.2\xspace}
\newcommand{\stBenchOneOvSftOverSftOvoHi}{+5.9\xspace}
\newcommand{\stBenchOneOvSftOverSftMvb}{+0.7\xspace}
\newcommand{\stBenchOneOvSftOverSftMvbLo}{$-$0.7\xspace}
\newcommand{\stBenchOneOvSftOverSftMvbHi}{+2.1\xspace}
\newcommand{\stBenchOneOvSftOverSftSb}{+1.6\xspace}
\newcommand{\stBenchOneOvSftOverSftSbLo}{$-$0.2\xspace}
\newcommand{\stBenchOneOvSftOverSftSbHi}{+3.4\xspace}
\newcommand{\stBenchOneOvSftOverSftVm}{+1.1\xspace}
\newcommand{\stBenchOneOvSftOverSftVmLo}{$-$1.6\xspace}
\newcommand{\stBenchOneOvSftOverSftVmHi}{+3.8\xspace}
\newcommand{\stBenchOneOvSftOverSftMean}{+1.6\xspace}
\newcommand{\stBenchOneOvSftOverSftMeanLo}{+0.4\xspace}
\newcommand{\stBenchOneOvSftOverSftMeanHi}{+2.7\xspace}
\newcommand{\stBenchOneOvEqOverSftOvo}{$-$0.4\xspace}
\newcommand{\stBenchOneOvEqOverSftOvoLo}{$-$3.5\xspace}
\newcommand{\stBenchOneOvEqOverSftOvoHi}{+2.7\xspace}
\newcommand{\stBenchOneOvEqOverSftMvb}{$-$0.3\xspace}
\newcommand{\stBenchOneOvEqOverSftMvbLo}{$-$1.7\xspace}
\newcommand{\stBenchOneOvEqOverSftMvbHi}{+1.0\xspace}
\newcommand{\stBenchOneOvEqOverSftSb}{$-$0.4\xspace}
\newcommand{\stBenchOneOvEqOverSftSbLo}{$-$2.2\xspace}
\newcommand{\stBenchOneOvEqOverSftSbHi}{+1.4\xspace}
\newcommand{\stBenchOneOvEqOverSftVm}{$-$0.3\xspace}
\newcommand{\stBenchOneOvEqOverSftVmLo}{$-$3.0\xspace}
\newcommand{\stBenchOneOvEqOverSftVmHi}{+2.3\xspace}
\newcommand{\stBenchOneOvEqOverSftMean}{$-$0.4\xspace}
\newcommand{\stBenchOneOvEqOverSftMeanLo}{$-$1.5\xspace}
\newcommand{\stBenchOneOvEqOverSftMeanHi}{+0.8\xspace}
\newcommand{\stBenchOneOvSftOverOvEqOvo}{+3.2\xspace}
\newcommand{\stBenchOneOvSftOverOvEqOvoLo}{0.0\xspace}
\newcommand{\stBenchOneOvSftOverOvEqOvoHi}{+6.5\xspace}
\newcommand{\stBenchOneOvSftOverOvEqMvb}{+1.0\xspace}
\newcommand{\stBenchOneOvSftOverOvEqMvbLo}{$-$0.4\xspace}
\newcommand{\stBenchOneOvSftOverOvEqMvbHi}{+2.4\xspace}
\newcommand{\stBenchOneOvSftOverOvEqSb}{+2.0\xspace}
\newcommand{\stBenchOneOvSftOverOvEqSbLo}{+0.2\xspace}
\newcommand{\stBenchOneOvSftOverOvEqSbHi}{+3.8\xspace}
\newcommand{\stBenchOneOvSftOverOvEqVm}{+1.4\xspace}
\newcommand{\stBenchOneOvSftOverOvEqVmLo}{$-$1.3\xspace}
\newcommand{\stBenchOneOvSftOverOvEqVmHi}{+4.2\xspace}
\newcommand{\stBenchOneOvSftOverOvEqMean}{+1.9\xspace}
\newcommand{\stBenchOneOvSftOverOvEqMeanLo}{+0.7\xspace}
\newcommand{\stBenchOneOvSftOverOvEqMeanHi}{+3.1\xspace}
\newcommand{\stBenchTwoSftOvo}{43.8\xspace}
\newcommand{\stBenchTwoSftMvb}{39.7\xspace}
\newcommand{\stBenchTwoSftSb}{47.4\xspace}
\newcommand{\stBenchTwoSftVm}{45.0\xspace}
\newcommand{\stBenchTwoSftMean}{44.0\xspace}
\newcommand{\stBenchTwoOvSftOvo}{46.6\xspace}
\newcommand{\stBenchTwoOvSftMvb}{40.3\xspace}
\newcommand{\stBenchTwoOvSftSb}{49.1\xspace}
\newcommand{\stBenchTwoOvSftVm}{46.0\xspace}
\newcommand{\stBenchTwoOvSftMean}{45.5\xspace}
\newcommand{\stBenchTwoKlOvo}{45.8\xspace}
\newcommand{\stBenchTwoKlMvb}{39.1\xspace}
\newcommand{\stBenchTwoKlSb}{48.3\xspace}
\newcommand{\stBenchTwoKlVm}{47.1\xspace}
\newcommand{\stBenchTwoKlMean}{45.1\xspace}
\newcommand{\stBenchTwoYoloOvo}{46.5\xspace}
\newcommand{\stBenchTwoYoloMvb}{39.4\xspace}
\newcommand{\stBenchTwoYoloSb}{50.6\xspace}
\newcommand{\stBenchTwoYoloVm}{46.8\xspace}
\newcommand{\stBenchTwoYoloMean}{45.8\xspace}
\newcommand{\stBenchTwoKlOverSftOvo}{+1.9\xspace}
\newcommand{\stBenchTwoKlOverSftOvoLo}{$-$1.1\xspace}
\newcommand{\stBenchTwoKlOverSftOvoHi}{+4.8\xspace}
\newcommand{\stBenchTwoKlOverSftMvb}{$-$0.6\xspace}
\newcommand{\stBenchTwoKlOverSftMvbLo}{$-$1.9\xspace}
\newcommand{\stBenchTwoKlOverSftMvbHi}{+0.7\xspace}
\newcommand{\stBenchTwoKlOverSftSb}{+0.9\xspace}
\newcommand{\stBenchTwoKlOverSftSbLo}{$-$0.8\xspace}
\newcommand{\stBenchTwoKlOverSftSbHi}{+2.6\xspace}
\newcommand{\stBenchTwoKlOverSftVm}{+2.1\xspace}
\newcommand{\stBenchTwoKlOverSftVmLo}{$-$0.6\xspace}
\newcommand{\stBenchTwoKlOverSftVmHi}{+4.8\xspace}
\newcommand{\stBenchTwoKlOverSftMean}{+1.1\xspace}
\newcommand{\stBenchTwoKlOverSftMeanLo}{0.0\xspace}
\newcommand{\stBenchTwoKlOverSftMeanHi}{+2.2\xspace}
\newcommand{\stBenchTwoYoloOverKlOvo}{+0.7\xspace}
\newcommand{\stBenchTwoYoloOverKlOvoLo}{$-$2.0\xspace}
\newcommand{\stBenchTwoYoloOverKlOvoHi}{+3.5\xspace}
\newcommand{\stBenchTwoYoloOverKlMvb}{+0.3\xspace}
\newcommand{\stBenchTwoYoloOverKlMvbLo}{$-$1.1\xspace}
\newcommand{\stBenchTwoYoloOverKlMvbHi}{+1.6\xspace}
\newcommand{\stBenchTwoYoloOverKlSb}{+2.3\xspace}
\newcommand{\stBenchTwoYoloOverKlSbLo}{+0.6\xspace}
\newcommand{\stBenchTwoYoloOverKlSbHi}{+4.1\xspace}
\newcommand{\stBenchTwoYoloOverKlVm}{$-$0.3\xspace}
\newcommand{\stBenchTwoYoloOverKlVmLo}{$-$2.9\xspace}
\newcommand{\stBenchTwoYoloOverKlVmHi}{+2.2\xspace}
\newcommand{\stBenchTwoYoloOverKlMean}{+0.7\xspace}
\newcommand{\stBenchTwoYoloOverKlMeanLo}{$-$0.3\xspace}
\newcommand{\stBenchTwoYoloOverKlMeanHi}{+1.8\xspace}
\newcommand{\stBenchTwoYoloOverOvSftOvo}{$-$0.1\xspace}
\newcommand{\stBenchTwoYoloOverOvSftOvoLo}{$-$3.1\xspace}
\newcommand{\stBenchTwoYoloOverOvSftOvoHi}{+2.9\xspace}
\newcommand{\stBenchTwoYoloOverOvSftMvb}{$-$0.9\xspace}
\newcommand{\stBenchTwoYoloOverOvSftMvbLo}{$-$2.3\xspace}
\newcommand{\stBenchTwoYoloOverOvSftMvbHi}{+0.3\xspace}
\newcommand{\stBenchTwoYoloOverOvSftSb}{+1.6\xspace}
\newcommand{\stBenchTwoYoloOverOvSftSbLo}{$-$0.1\xspace}
\newcommand{\stBenchTwoYoloOverOvSftSbHi}{+3.3\xspace}
\newcommand{\stBenchTwoYoloOverOvSftVm}{+0.8\xspace}
\newcommand{\stBenchTwoYoloOverOvSftVmLo}{$-$1.8\xspace}
\newcommand{\stBenchTwoYoloOverOvSftVmHi}{+3.3\xspace}
\newcommand{\stBenchTwoYoloOverOvSftMean}{+0.3\xspace}
\newcommand{\stBenchTwoYoloOverOvSftMeanLo}{$-$0.8\xspace}
\newcommand{\stBenchTwoYoloOverOvSftMeanHi}{+1.5\xspace}
\newcommand{\stBenchTwoYoloOverSftOvo}{+2.6\xspace}
\newcommand{\stBenchTwoYoloOverSftOvoLo}{$-$0.4\xspace}
\newcommand{\stBenchTwoYoloOverSftOvoHi}{+5.6\xspace}
\newcommand{\stBenchTwoYoloOverSftMvb}{$-$0.3\xspace}
\newcommand{\stBenchTwoYoloOverSftMvbLo}{$-$1.7\xspace}
\newcommand{\stBenchTwoYoloOverSftMvbHi}{+1.2\xspace}
\newcommand{\stBenchTwoYoloOverSftSb}{+3.2\xspace}
\newcommand{\stBenchTwoYoloOverSftSbLo}{+1.4\xspace}
\newcommand{\stBenchTwoYoloOverSftSbHi}{+5.1\xspace}
\newcommand{\stBenchTwoYoloOverSftVm}{+1.8\xspace}
\newcommand{\stBenchTwoYoloOverSftVmLo}{$-$0.9\xspace}
\newcommand{\stBenchTwoYoloOverSftVmHi}{+4.4\xspace}
\newcommand{\stBenchTwoYoloOverSftMean}{+1.8\xspace}
\newcommand{\stBenchTwoYoloOverSftMeanLo}{+0.7\xspace}
\newcommand{\stBenchTwoYoloOverSftMeanHi}{+3.0\xspace}
\newcommand{\stHeldOvSftOKzero}{79.5\xspace}
\newcommand{\stHeldOvSftOKsixteen}{88.0\xspace}
\newcommand{\stHeldOvSftTKzero}{48.6\xspace}
\newcommand{\stHeldOvSftTKsixteen}{66.5\xspace}
\newcommand{\stHeldKlOKzero}{77.5\xspace}
\newcommand{\stHeldKlOKsixteen}{83.3\xspace}
\newcommand{\stHeldKlTKzero}{53.7\xspace}
\newcommand{\stHeldKlTKsixteen}{72.0\xspace}
\newcommand{\stHeldYoloOKzero}{83.8\xspace}
\newcommand{\stHeldYoloOKsixteen}{92.6\xspace}
\newcommand{\stHeldYoloTKzero}{53.3\xspace}
\newcommand{\stHeldYoloTKsixteen}{71.8\xspace}
\newcommand{\stHeldYoloOverKlOKzero}{+6.3\xspace}
\newcommand{\stHeldYoloOverKlOKzeroLo}{+5.1\xspace}
\newcommand{\stHeldYoloOverKlOKzeroHi}{+7.6\xspace}
\newcommand{\stHeldYoloOverKlOKsixteen}{+9.4\xspace}
\newcommand{\stHeldYoloOverKlOKsixteenLo}{+8.2\xspace}
\newcommand{\stHeldYoloOverKlOKsixteenHi}{+10.6\xspace}
\newcommand{\stHeldYoloOverKlTKzero}{$-$0.4\xspace}
\newcommand{\stHeldYoloOverKlTKzeroLo}{$-$1.5\xspace}
\newcommand{\stHeldYoloOverKlTKzeroHi}{+0.8\xspace}
\newcommand{\stHeldYoloOverKlTKsixteen}{$-$0.2\xspace}
\newcommand{\stHeldYoloOverKlTKsixteenLo}{$-$1.4\xspace}
\newcommand{\stHeldYoloOverKlTKsixteenHi}{+1.0\xspace}

\newcommand{\stVarSftOKzero}{69.7\xspace}
\newcommand{\stVarSftOKsixteen}{75.8\xspace}
\newcommand{\stVarSftTKzero}{47.9\xspace}
\newcommand{\stVarSftTKsixteen}{64.9\xspace}
\newcommand{\stVarKlOKzero}{77.4\xspace}
\newcommand{\stVarKlOKsixteen}{83.3\xspace}
\newcommand{\stVarKlTKzero}{54.8\xspace}
\newcommand{\stVarKlTKsixteen}{72.1\xspace}
\newcommand{\stVarNoKlOKzero}{79.0\xspace}
\newcommand{\stVarNoKlOKsixteen}{83.9\xspace}
\newcommand{\stVarNoKlTKzero}{57.8\xspace}
\newcommand{\stVarNoKlTKsixteen}{74.3\xspace}
\newcommand{\stVarKlOverSftOKzero}{+7.6\xspace}
\newcommand{\stVarKlOverSftOKzeroLo}{+6.4\xspace}
\newcommand{\stVarKlOverSftOKzeroHi}{+8.8\xspace}
\newcommand{\stVarKlOverSftOKsixteen}{+7.5\xspace}
\newcommand{\stVarKlOverSftOKsixteenLo}{+6.4\xspace}
\newcommand{\stVarKlOverSftOKsixteenHi}{+8.7\xspace}
\newcommand{\stVarKlOverSftTKzero}{+6.9\xspace}
\newcommand{\stVarKlOverSftTKzeroLo}{+5.8\xspace}
\newcommand{\stVarKlOverSftTKzeroHi}{+8.1\xspace}
\newcommand{\stVarKlOverSftTKsixteen}{+7.2\xspace}
\newcommand{\stVarKlOverSftTKsixteenLo}{+5.9\xspace}
\newcommand{\stVarKlOverSftTKsixteenHi}{+8.5\xspace}
\newcommand{\stVarKlOverNoKlOKzero}{$-$1.6\xspace}
\newcommand{\stVarKlOverNoKlOKzeroLo}{$-$2.8\xspace}
\newcommand{\stVarKlOverNoKlOKzeroHi}{$-$0.4\xspace}
\newcommand{\stVarKlOverNoKlOKsixteen}{$-$0.6\xspace}
\newcommand{\stVarKlOverNoKlOKsixteenLo}{$-$1.6\xspace}
\newcommand{\stVarKlOverNoKlOKsixteenHi}{+0.4\xspace}
\newcommand{\stVarKlOverNoKlTKzero}{$-$3.0\xspace}
\newcommand{\stVarKlOverNoKlTKzeroLo}{$-$4.2\xspace}
\newcommand{\stVarKlOverNoKlTKzeroHi}{$-$1.7\xspace}
\newcommand{\stVarKlOverNoKlTKsixteen}{$-$2.3\xspace}
\newcommand{\stVarKlOverNoKlTKsixteenLo}{$-$3.6\xspace}
\newcommand{\stVarKlOverNoKlTKsixteenHi}{$-$0.9\xspace}

\newcommand{\stDevYoloRecall}{0.505\xspace}
\newcommand{\stDevYoloPrecision}{0.580\xspace}
\newcommand{\stDevYoloWords}{40.5\xspace}

\newcommand{\stDevTeacherRecall}{0.841\xspace}
\newcommand{\stDevTeacherPrecision}{0.887\xspace}
\newcommand{\stDevTeacherWords}{44.1\xspace}

\newcommand{\stDevScored}{1,246\xspace}

\newcommand{\stDrawsDev}{4,000\xspace}
\newcommand{\stDrawsReport}{2,000\xspace}
\newcommand{\stDrawsQuestions}{10,000\xspace}
\newcommand{\stStyleWindows}{1,752\xspace}
\newcommand{\stSeedSpreadEnds}{1.1\xspace}

%% file: numbers/supp_interp.tex
\newcommand{\ycInterpPlainSixteen}{72.1\xspace}

\newcommand{\ycInterpSmoothSixteen}{70.7\xspace}

\newcommand{\ycInterpDenseSixteen}{71.0\xspace}

\newcommand{\ycInterpDupSixteen}{72.1\xspace}

\newcommand{\ycInterpPlainEight}{65.1\xspace}

\newcommand{\ycInterpDenseEight}{65.3\xspace}

\newcommand{\ycInterpSmoothGap}{$-$1.5\xspace}
\newcommand{\ycInterpSmoothGapLo}{$-$2.6\xspace}
\newcommand{\ycInterpSmoothGapHi}{$-$0.3\xspace}

\newcommand{\ycInterpDenseGap}{$-$1.1\xspace}
\newcommand{\ycInterpDenseGapLo}{$-$2.3\xspace}
\newcommand{\ycInterpDenseGapHi}{+0.1\xspace}
\newcommand{\ycInterpDenseEightGap}{+0.2\xspace}
\newcommand{\ycInterpDenseEightGapLo}{$-$1.0\xspace}
\newcommand{\ycInterpDenseEightGapHi}{+1.3\xspace}

\newcommand{\ycInterpDupGap}{+0.0\xspace}
\newcommand{\ycInterpDupGapLo}{$-$1.2\xspace}
\newcommand{\ycInterpDupGapHi}{+1.2\xspace}

%% file: floats/fig_lookback.tex
\begin{center}
  \includegraphics[width=\textwidth]{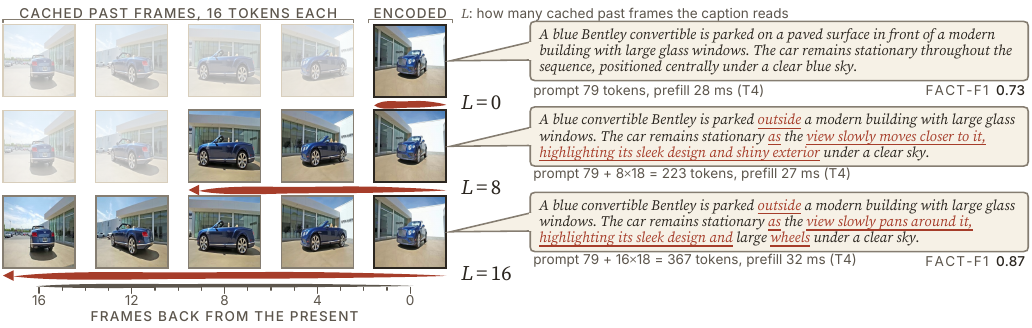}
  \captionof{figure}{\textbf{\model scans the video, reading more of the cached past.} 
  One frame per second, every fourth drawn frame $t$ is encoded; each past frame was 
  encoded once and cached as 16 tokens, read without a new vision pass. The lookback $L$ 
  (bracket: bright frames are read) is set per caption at run time. 
  Captions of a held-out OpenVid window; underlined words are absent from the $L{=}0$ caption;
  fact-F1 only where scored. Latency was recorded on a T4.}
  \label{fig:lookback}
\end{center}
\vspace{0.5em}

%% file: sec/0_abstract.tex
\begin{abstract}
  We introduce a new model for real-time streaming captioning on a single GPU. \model is a \totalParams-parameter
  streaming captioner that provides Pareto-frontier
  performance among generative vision models in its class.
  Distilled from a 27B teacher by SFT and GRPO, it leads every model under 1B parameters on
  OVO-Bench, MVBench, StreamingBench and Video-MME, answered from its captions, and beats LFM2.5-VL-1.6B and SmolVLM2-2.2B.
  \model is designed to read more frames cheaply and make them informative, encoding each frame once and caching it as
  16 tokens. This allows the model to incorporate and infer from past frames without additional vision passes,
  making it efficient and effective. We also provide a detailed recipe for training \model, with its stages and
  data, as well as ablations that demonstrate the contribution of each component.

\end{abstract}

%% file: sec/1_intro.tex
\section{Introduction}
\label{sec:intro}
\input{floats/fig_arch}

While deep learning has enabled significant advancements in computer vision, offering new capabilities~\cite{resnet2016,yolo2016,vit2021,clip2021},
beyond traditional tasks like image classification and object detection, such as image captioning,
image generation and their video counterparts~\cite{showtell2015,ldm2022,llavavideo2024,videodiffusion2022}, the deployment of these models in real-time settings with high-volume data
remains a challenge~\cite{edgevideo2022,powerhungry2024,effmllm2024}. In particular, the need for generative outputs from live video streams, such as descriptions,
commentary, answers, and alerts, requires models that can operate efficiently on small GPUs at the edge~\cite{videollmonline2024,livecc2025,hazardnet2025,vlmedge2025}.
In order to operate effectively in these scenarios, a few constraints must be met, including the ability to process and generate outputs
in real time while maintaining high-quality performance without incurring significant computational costs~\cite{speedacc2017,streamingvlm2025,effvlm2025}. 
This has led to the development of smaller vision-language models (VLMs) that are designed for question answering on images or clips
under a budget constraint~\cite{mobilevlm2023,smolvlm2025,fastvlm2025,lfm2,mobilevideogpt2025}.
In this work, we are particularly interested in making a sub-500M-parameter model that can scale with the number of frames it can read
efficiently and effectively, adjustable at run time for different scenarios and comparable to larger models on four
benchmarks curated for captioning and answered from the captions alone: OVO-Bench Real-Time~\cite{ovobench2025}, MVBench~\cite{mvbench2024}, StreamingBench Real-Time~\cite{streamingbench2024}, and Video-MME~\cite{videomme2024}.

We formulate streaming captioning as one fresh description per second of a sliding recent window. Caption generation
is stateless: each caption reads the current frame and $L$ past frames sampled at one frame per second, without the previous caption
or language-model state. The visual cache persists, but generated text does not. A fixed-size cache and bounded
per-frame work keep memory bounded and latency predictable, while $L$ remains adjustable at run time. But overlapping
windows also produce overlapping descriptions: persistent facts recur even without copying the previous caption.
For a rolling state log, each caption can update the current state rather than append another report of the same
event. In video surveillance and anomaly detection, an alert monitor can read the descriptions continuously and
merge repeated alerts for the same event. These consumers must treat overlap as repeated coverage, not independent
evidence. We evaluate the captions, not a complete monitoring system.

\model is designed to read more frames cheaply and infer from them without additional vision passes. It also sets the quality--cost
frontier among generative vision models in its class, and we provide a detailed recipe for training it that we think could be useful for other
tasks as well, along with ablations that demonstrate the contribution of each component.
Our model is distilled from a larger teacher model and trained using a combination of supervised fine-tuning and reinforcement
learning. We observe that the model learns to understand the temporal context of the video stream and to generate high-quality captions.
Our contributions can be summarized as follows:
\begin{enumerate}
\item We introduce a new model for real-time streaming captioning on a single GPU,
\model, which is a \totalParams-parameter streaming captioner that provides Pareto-frontier
performance among generative vision models in its class (\cref{fig:performance}).
\item We introduce an architecture that reads more frames cheaply and makes them informative, 
encoding each frame once and caching it as 16 tokens (\cref{fig:lookback}).
This allows the model to incorporate and infer from past frames without additional vision passes, 
making it efficient and effective.
\item We provide a detailed recipe for training \model, including its stages and data,
as well as ablations that demonstrate the contribution of each component.
\end{enumerate}

%% file: floats/fig_arch.tex
\begin{figure*}[t]
  \centering
  \includegraphics[width=\textwidth]{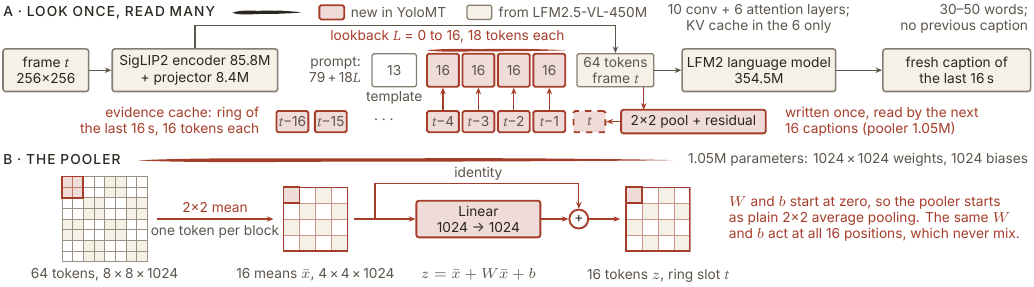}
  \caption{\textbf{Look once, read many.} (a) Each sampled frame (one per second) passes the vision encoder once. Its 64 tokens enter the prompt as the present and, pooled to 16, a ring of the last 16 frames that later captions read; a caption reads $L$ of these blocks, its lookback, $0\le L\le16$, chosen at run time with one set of weights. (b) The pooler. Each $2{\times}2$ block of the projected $8{\times}8$ grid becomes one mean token (highlighted: one block followed through the pooler). A linear residual shared by the 16 positions transforms each mean without mixing positions; with zero-initialized weights and bias, the pooler starts as average pooling.}
  \label{fig:arch}
\end{figure*}

%% file: sec/2_related.tex
\section{Related work}
\label{sec:related}
\input{floats/fig_recipe}

Traditional computer vision tasks, such as image classification and object detection, have been significantly advanced by deep learning
techniques. The generative AI paradigm has expanded the scope of computer vision to include new tasks such as image captioning and
image generation, but this expansion came at the cost of an explosion in model size and computational requirements~\cite{kaplan2020,greenai2019,powerhungry2024}, creating a split
between large generalist models, including proprietary ones, and smaller models that are more efficient but less capable~\cite{gpt42023,gemini2023,llavakd2025}.
This split has shaped the field of small vision-language models (VLMs)~\cite{effmllm2024,effvlm2025}, which are designed for image and video
understanding~\cite{lfm2,smolvlm2025,tinyllavavideo2025} but are often limited in their ability to generate high-quality outputs in real-time scenarios~\cite{streamingbench2024,ovobench2025,rtvbench2025}.
The main challenge has thus become how to design and train small VLMs that can operate efficiently in real-time settings
while maintaining the high-quality performance of larger models~\cite{mobilevideogpt2025,mobilevlmv2,vicostream2026}.
Mobile-VideoGPT addresses this through lightweight dual visual encoders, frame selection and visual-token pruning~\cite{mobilevideogpt2025}.
Other approaches optimize the architecture and training of small VLMs~\cite{fastvlm2025,llavamini2025}, or change how much
visual information the language model reads. Matryoshka Multimodal Models and the Matryoshka Query Transformer train
one model to represent an image with different numbers of visual tokens at inference~\cite{matryoshka2024,mqt2024}.
This allows a trade-off between spatial detail and computation. In our setting, the adjustable budget instead controls
how many past moments the model reads, with a fixed number of tokens for each cached frame.

The streaming setting adds another requirement: the model must generate outputs as new frames arrive, rather than
after receiving a complete clip. VideoLLM-online introduces a streaming training objective and inference pipeline for
conversation within a continuous video stream~\cite{videollmonline2024}. LiveCC trains on speech transcripts interleaved
with video frames and learns real-time commentary~\cite{livecc2025}. We study the same need for continuous generative
outputs, but focus on a fresh caption of the recent past every second, within the memory and per-frame time budgets
of a T4.

Keeping the past available has led to several forms of streaming memory. Flash-VStream combines a compact temporal
context memory with a larger memory for retrieving spatial detail~\cite{flashvstream2025}. TimeChat-Online drops visual
tokens that do not change between frames~\cite{timechatonline2025}, while ReKV stores video key-value caches and retrieves
those relevant to a question~\cite{rekv2025}. Our model instead caches pooled visual tokens before the language model,
and does not carry the language model's state from one caption to the next. Each sampled frame is encoded once and kept as 16 tokens in a ring
of recent frames. A caption reads the current frame and a chosen number of cached past frames, without another vision
pass. This is a bounded recent history for continuous captioning, rather than a memory for retrieving events from an
entire long video.

The distinction is where state lives, rather than whether memory is bounded. StreamingVLM also bounds its KV cache,
but carries attention sinks, recent visual tokens and a longer text history~\cite{streamingvlm2025}. Our captions
instead restart from the selected visual window. Each caption is thus independent of earlier outputs, and $L$ can
change from one caption to the next, but nothing enforces novelty across overlapping captions. Text is also an interface for anomaly detection: LAVAD captions frames, then
uses an LLM to aggregate descriptions over time and estimate anomaly scores~\cite{lavad2024}. Such aggregation belongs
to the downstream consumer.
Socratic Models and LLoVi answer video questions from captions~\cite{socratic2023,llovi2024}; VDCscore~\cite{auroracap2025}
and CaptionQA~\cite{captionqa2026} score captions by the answers they support. We bring this to streaming, where
captions precede the questions.

More history is not always more useful. SimpleStream shows that a recent-frame window can match more elaborate
streaming models, and that adding history can weaken current-scene perception~\cite{simplestream2026}. SelectStream
addresses this by exposing history through a compact, query-conditioned evidence budget while keeping the current
frame visible~\cite{whatshouldremember2026}. These findings motivate both parts of our approach: making past frames
cheap to read, and training the model to infer from them. Our controls separate the information in those frames from
the tokens and computation used to read them (\cref{sec:ablations}).

Previous work has also explored distillation techniques that transfer knowledge from larger models to smaller
ones~\cite{kd2015,seqkd2016,goodteacher2022,gkd2024,llavakd2025}. We use this to specialize a small model for captioning,
first through the teacher's captions and then through reinforcement learning. To serve different lookbacks with one
set of weights, we apply the sandwich rule of slimmable networks~\cite{slimmable2019,usnet2019} to the number of past
frames rather than network width. Our work brings these architectural and training choices together in a streaming
model that runs within the budget of a T4, and measures what the main choices contribute. While we focus on streaming captioning, we
believe the recipe could also be useful for other generative vision tasks~\cite{videollmonline2024,livecc2025,dircast2026,strad2026}.

%% file: floats/fig_recipe.tex
\begin{figure*}[t]
  \centering
  \includegraphics[width=\textwidth]{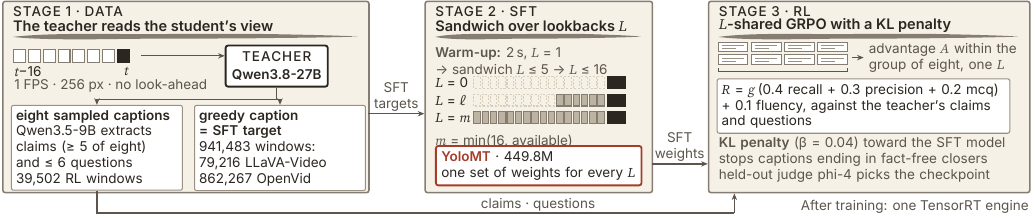}
  \caption{\textbf{Distilling a larger teacher into \model in three stages.} The teacher labels the student's own view: a greedy caption for SFT, and sampled captions that yield claims and questions for RL. SFT starts with shorter captions on 2\,s windows at $L{=}1$, then uses the sandwich rule up to $L{=}5$ and finally $L{=}16$ (shown). The last SFT phase trains all weights; GRPO shares one $L$ in each group, and a KL penalty toward the SFT model keeps captions from ending in fact-free closers. On held-out OpenVid the result reaches \tenOKzero\% ($L{=}0$) to \tenOKsixteen\% ($L{=}16$) of the teacher ceiling in fact-F1, from one engine with p99 \eyPNinetyNineMin--\eyPNinetyNineMax~ms per frame on a T4.}
  \label{fig:recipe}
\end{figure*}

%% file: sec/3_method.tex
\section{Method}
\label{sec:method}
\input{floats/fig_performance}
In order to build a small, performant model for streaming captioning, we propose a method that
combines architectural optimizations and a multi-stage training recipe based on distillation and
reinforcement learning.

\subsection{Architecture: look once, read many}
\label{sec:arch}
\model employs the architecture of LFM2.5-VL-450M~\cite{lfm2}, with a vision encoder and a language model, and adds a
pooler (\cref{fig:arch}a). The vision encoder, SigLIP2~\cite{siglip2}, processes each frame once, and the pooler caches
its representation as 16 tokens, allowing the model to read multiple past frames without re-encoding them. The language
model, LFM2.5-350M~\cite{lfm2}, generates captions based on the current frame and the cached representations of past
frames. The idea is to allow the model to look at past frames without incurring the computational cost of re-encoding 
them, enabling efficient streaming captioning in real time. 
The number of cached frames that the model reads, 16 visual tokens each, is referred to as the lookback $L$,
which can be adjusted at run time to the amount of past context the model should consider when generating captions. The
input prompt to the language model consists of a template, the cached blocks of past frames, and the tokens of the
current frame. The model is trained to generate captions of 30--50 words that describe the last 16 seconds of video; the
previous caption is not part of the prompt.

The pooler (\poolerParams{} parameters; \cref{fig:arch}b) arranges the 64 projected
tokens as an $8\times8$ grid $x_{i,j}\in\mathbb{R}^{d}$, $d=1024$, and maps each
$2\times2$ block, indexed by $u,v\in\{0,\dots,3\}$, to a single token:
\begin{equation}\label{eq:pooler}
\begin{aligned}
  \bar x_{u,v} &= \frac{1}{4}\sum_{a,b\in\{0,1\}} x_{2u+a,\,2v+b},\\
  z_{u,v}      &= \bar x_{u,v} + W\bar x_{u,v} + b.
\end{aligned}
\end{equation}
$W\in\mathbb{R}^{d\times d}$ and $b\in\mathbb{R}^{d}$ are shared across all 16 positions
and initialized to zero, so the pooler starts as exact average pooling and learns a
residual correction on top of it. The 16 pooled tokens $z_{u,v}$ enter the ring cache,
while the current frame keeps all 64 tokens in the prompt.

At deployment, we spread the work of a caption across the incoming frames of the 30\,FPS stream (\cref{fig:schedule}).
The first frame of each second runs the vision encoder and the pooler. Each following frame processes one chunk of at
most \maxChunk\ prompt tokens~\cite{sarathi2023,sarathiserve2024}; once the prompt is processed, each frame runs
\decodeSteps\ decoding steps until the caption ends. A longer lookback adds prompt work and can delay the caption, but does not require another vision
pass or a longer chunk of work per frame. The deployment budget is 16\,ms of GPU work per incoming frame, under 2\,GB
of memory, and a completed caption before the next second. We provide the architecture and engine details in the
supplementary material (\cref{sec:supp:arch,sec:supp:runtime}).

\input{floats/fig_schedule}

\subsection{Training recipe: distillation in three stages}
\label{sec:training}
The training of \model is based on classical distillation techniques, where a larger teacher model is
used to provide supervision for the smaller student model. The training process consists of three stages.

\paragraph{Stage 1: data collection.} We run the teacher, Qwen3.8-27B~\cite{qwen38_27b}\footnote{Hugging Face ids:
Qwen/Qwen3.8-27B-FP8, Qwen/Qwen3.5-9B, microsoft/phi-4, LiquidAI/LFM2.5-VL-450M.}, on two datasets,
LLaVA-Video-178K~\cite{llavavideo2024} and OpenVid-1M~\cite{openvid2025}, to generate captions for the student model.
More specifically, we sample one frame per second from a 30\,FPS stream and prompt the teacher to generate captions of
30--50 words that describe the last 16 seconds of the video. We collect a total of \sftTrainTen\ windows of video data,
which we use for the supervised fine-tuning (SFT) stage. For the reinforcement learning (RL) stage, we collect \rlTen\
windows of video data through the same process, but we generate eight additional captions for each window at a
temperature of 0.7, and use a smaller model, Qwen3.5-9B~\cite{qwen35_9b}, to extract atomic claims from the
captions and generate multiple-choice questions based on the captions. The union contains claims from the greedy
caption and all retained samples; consensus keeps those shared by at least five of eight samples (the same fraction
when length filtering leaves fewer). These claims and questions are prepared offline for the RL stage.

\paragraph{Stage 2: supervised fine-tuning (SFT).} In this stage, we train the student model to mimic the teacher's
captions for different lookbacks $L$. We use the sandwich rule of slimmable networks~\cite{slimmable2019,usnet2019},
where each training window is processed with three different lookbacks: $L=0$, a random $\ell$ drawn uniformly from $1$
to $m-1$, and $L=m$, where $m=\min(16,L_{\mathrm{avail}})$ and $L_{\mathrm{avail}}$ is the number of past frames
available. The loss is computed over the tokens of the teacher's caption under each of the three lookbacks, and the
model is trained for \sftEpochs\ epochs with a learning rate of \sftLr\ for the language model and \sftVisionLr\ for the
vision encoder. The loss is defined as:
\begin{equation}
  \mathcal{L}(\theta)=\frac{1}{N}\sum_{w\in B}\,\sum_{L\in\{0,\ell_w,m_w\}}\,\sum_{i}-\log p_\theta\!\left(y^w_i\mid y^w_{<i},\,c^w_L\right),
\end{equation}
where $B$ is the set of windows of a training step, $N$ its number of target tokens, $y^w$ the teacher's caption of
window $w$, and $c^w_L$ the prompt of $w$ at lookback $L$.
The student does not start from the base model. Following short-to-long schedules for language
models~\cite{seqlenwarmup2021}, we first train it on shorter captions of 2\,s windows reading one past frame, then with
the sandwich rule up to $L{=}5$, before the training above extends it to $L{=}16$. Starting from the base model instead
costs 1.6 points at $L{=}16$ (\cref{sec:supp:ablations}).

\paragraph{Stage 3: reinforcement learning (RL).} In the final stage, we use reinforcement learning to further improve
the quality of the captions generated by the student model. We use the GRPO~\cite{grpo2024} algorithm, where we sample
\rlWindowsStep\ windows per step and generate \rlGroup\ captions for each window at one shared lookback $L$.
We choose the largest available $L$ with probability $\tfrac12$, otherwise uniformly from zero through that maximum.
For each distinct valid caption among the \rlRollouts\ rollouts, Qwen3.5-9B is queried online. It extracts atomic facts
and checks each against the union of the teacher's claims for precision. In the other direction, it checks whether the
caption states or implies each consensus claim for recall. The same judge answers the window's multiple-choice
questions from the caption alone and rates its fluency; a frozen small language model provides a separate fluency score.
We batch these text-only requests, with duplicate captions of a window scored once per step (\cref{sec:supp:reward}).
The advantage is computed as $A_i=(R_i-\bar{R})/\sigma_R$ within the group, and groups with equal rewards are dropped
as in~\cite{dapo2025}. The update is the PPO-clipped surrogate~\cite{ppo2017}, averaged over the tokens of the step
with clip \rlClip, and a per-token KL penalty with coefficient $\beta{=}\rlBeta$ pulls towards the frozen SFT model
(\cref{sec:supp:grpo}). Without it the model drifts towards repeating certain phrases and producing fact-free closers
(\cref{sec:ablations}). The training stops after
three evaluation checks without gain (phi-4~\cite{phi4} scores greedy dev captions every \rlEvalEvery\ steps), and a
tripwire is used to tighten the reward when the training reward rises but the evaluation score does not. The reward is
defined as:
\begin{equation}
  R=g\,(\lambda_1\,\mathrm{recall}+\lambda_2\,\mathrm{precision}+\lambda_3\,\mathrm{mcq})+\lambda_4\,\mathrm{fluency},
  \label{eq:reward}
\end{equation}
where $\lambda_1=0.4$, $\lambda_2=0.3$, $\lambda_3=0.2$ and $\lambda_4=0.1$ weight the four terms.
The gate $g\in[0,1]$ scales the factual terms using the judge's fluency rating and a repetition penalty.
The additive fluency term instead rewards likelihood under the frozen language model; both are normalized to $[0,1]$
(\cref{sec:supp:reward}). A caption over 60 words receives $R=0$ for the whole sum, including the additive fluency term.
The final model is selected based on the best evaluation score achieved during training. \Cref{fig:recipe} provides a
visual overview of the training process, showing the three stages of data collection, supervised fine-tuning, and
reinforcement learning, along with the flow of data and the interactions between the teacher and student models.

%% file: floats/fig_performance.tex
\begin{figure*}[t]
  \centering
  \includegraphics[width=\textwidth]{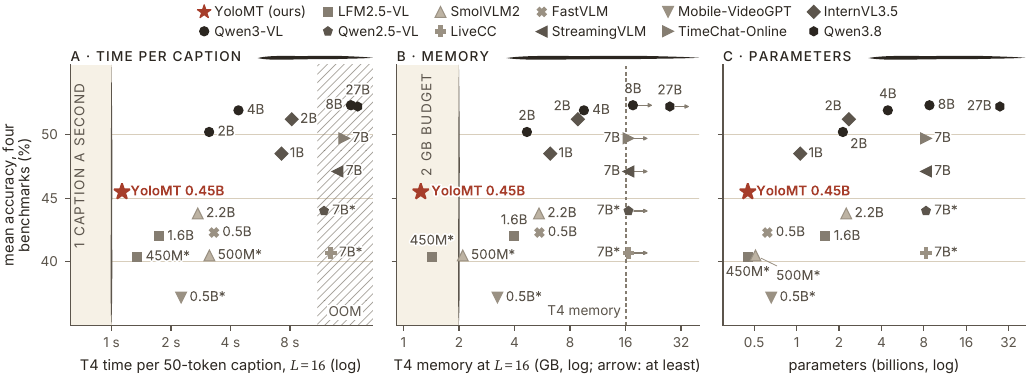}
  \caption{\textbf{Accuracy against time, memory and size: \cref{tab:main} in three plots.} 
  The mean accuracy of the four benchmarks (caption then answer, 
  $L{=}\min(t,16)$ past frames) against (a) the
  median T4 time of a 50-token caption at $L{=}16$, eager PyTorch FP16 
  (cream: one caption a second or faster; hatched: models whose weights 
  alone exceed the T4's memory), (b) the T4 process memory while captioning at 
  $L{=}16$ (cream: the 2\,GB budget; arrow: a lower bound, the weights alone), 
  and (c) the parameters. A marker is a model family and its tag the model's size;
  $^*$: under 30 words per caption; \cref{tab:main} also reports the lead of \model after deliberately truncating its captions to the three small baselines' word counts.
  Qwen3.8-27B is the teacher. }
  \label{fig:performance}
\end{figure*}

%% file: floats/fig_schedule.tex
\begin{figure}[h]
  \centering
  \includegraphics[width=\linewidth]{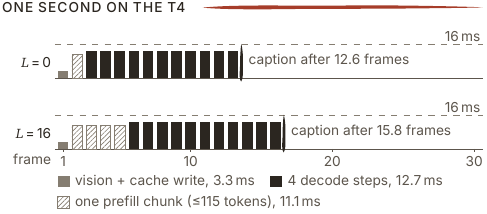}
  \caption{\textbf{One second on the T4.} Each frame carries one kind of work, so p99 per frame is \eyPNinetyNineMin--\eyPNinetyNineMax~ms at every measured $L$; a larger $L$ adds prefill frames and caption lag, not frame time. Bars: the p99 GPU time of each kind of frame at $L{=}16$, among the 30 frames of 33\,ms in a second; dashed: the 16\,ms budget per frame. Frame counts follow the engine's prefill schedule and the mean caption lag. The schedule is new in \model.}
  \label{fig:schedule}
\end{figure}

%% file: sec/4_results.tex
\input{floats/tab_main}
\section{Results}
\label{sec:results}
We evaluate \model on four public benchmarks with questions built from human annotations:
OVO-Bench~\cite{ovobench2025}, StreamingBench~\cite{streamingbench2024}, Video-MME~\cite{videomme2024} and
MVBench~\cite{mvbench2024}.
We curate these benchmarks for streaming captioning by keeping only the parts that captions of the last 16
frames can answer: we leave out questions that need audio, memory far beyond 16\,s, dialogue state or a
decision of when to speak (\cref{sec:supp:bench}).
Each model captions the videos as a live stream, from a fixed prompt and blind to the question, and a text-only
answerer\footnote{Mistral-Small-24B-Instruct-2501, temperature 0, one option letter.} answers from the captions
alone. This setup, new for streaming captioners, separates what a model sees from how it reasons: one answerer
serves every model, so accuracy measures what the captions carry, and every answer can be inspected against the
captions behind it.
The captioning model is run at 1\,FPS and 256$\times$256 resolution, reading $L{=}\min(t,16)$ past frames at
the question's last second $t$.
For each baseline, we select among the same three fixed prompts: plain description, live narration and concrete facts.
The wording with the best mean phi-4 fact-F1 at $L{=}0$ and 16 on \promptPilotWindows\ separate pilot windows wins.
These windows enter no reported result. We freeze the wording before any scored captioning, benchmarks included;
\model keeps its training prompt (\cref{sec:supp:prompts}). This gives each baseline a choice rather than imposing
one wording on every model.
We also score the captions themselves by fact-F1 on held-out captions from LLaVA-Video test and OpenVid, unseen in
training but from the same distribution.
The fact-F1 score measures the recall of the teacher's consensus claims and the precision of the caption's atomic facts,
as judged by phi-4~\cite{phi4}, and is reported as a percentage of the teacher's agreement with itself. On these sets,
\cref{tab:main} reports fact-F1 without a word-limit penalty, with the original 60-word-rule scores in parentheses
where they differ. That cutoff was a training constraint for \model, but the baselines were only prompted for
30--50 words.
\Cref{tab:main} provides the main results of our evaluation, showing that \model outperforms the other models
in its class on all four benchmarks, and is comparable to models several times its size.
On the mean of the four benchmarks, \model leads LFM2.5-VL-450M by \oneOverLfmSmallMean\
[\oneOverLfmSmallMeanLo, \oneOverLfmSmallMeanHi] points, SmolVLM2-500M by \oneOverSmolSmallMean\
[\oneOverSmolSmallMeanLo, \oneOverSmolSmallMeanHi], Mobile-VideoGPT-0.5B by \oneOverMobileMean\
[\oneOverMobileMeanLo, \oneOverMobileMeanHi] and FastVLM-0.5B by \nbOverFastVlmMean\
[\nbOverFastVlmMeanLo, \nbOverFastVlmMeanHi]; on held-out LLaVA-Video test captions at $L{=}16$, it reaches
\freeTOursKsixteen\% of the teacher ceiling, against \freeTLfmSmallKsixteen\% and \freeTSmolSmallKsixteen\% for the first two.
Without the length penalty, InternVL3.5-2B leads \model on test (\freeTInternTwoKsixteen\% against
\freeTOursKsixteen\%), as the Qwen3-VL models and TimeChat-Online-7B (\freeTTimeChatKsixteen\%) do; under the rule,
InternVL3.5-2B scores \tsInternTwoKsixteen\%.
Among larger models, \model leads LFM2.5-VL-1.6B by \oneOverLfmBigMean\ [\oneOverLfmBigMeanLo, \oneOverLfmBigMeanHi]
and SmolVLM2-2.2B by \oneOverSmolBigMean\ [\oneOverSmolBigMeanLo, \oneOverSmolBigMeanHi], as well as Qwen2.5-VL-7B and
LiveCC-7B, which write shorter captions; InternVL3.5-1B and InternVL3.5-2B lead it by \oneLeadInternOneMean\ and
\oneLeadInternTwoMean\ points, and the Qwen3-VL models by \oneLeadQwenTwoMean\ to \oneLeadQwenEightMean. The streaming
models StreamingVLM-7B and TimeChat-Online-7B lead it by \nbLeadStreamingVlmMean\ [\nbLeadStreamingVlmMeanLo,
\nbLeadStreamingVlmMeanHi] and \nbLeadTimeChatMean\ [\nbLeadTimeChatMeanLo, \nbLeadTimeChatMeanHi] points.
\model also runs in real time on a single T4: with frames run back to back, its TensorRT engine keeps the p99 time
per frame at \eyPNinetyNineMin--\eyPNinetyNineMax\,ms at every $L$, within the 16\,ms we allow per frame at 30\,FPS,
with \eyGB\,GB of memory (\cref{fig:schedule}). With arrivals paced at 30\,FPS, the p99 is
\stOneMin--\stOneMax\,ms, still within the same budget. At an equal 45 tokens per caption in the same TensorRT engine,
\model shows its caption within \efYoloLagKsixteenSecMax\,s at $L{=}16$, against \efLfmLagKsixteenSecMax\,s for
LFM2.5-VL-450M; SmolVLM2-500M finishes no 45-token caption within its second at $L\ge8$ (\cref{sec:supp:runtime}).
Even in eager PyTorch, \model is the fastest model we timed: it writes a 50-token caption in \latSecOursKsixteen\,s at $L{=}16$, against \latSecLfmSmallKsixteen\,s for LFM2.5-VL-450M and up to
\latSecInternTwoKsixteen\,s for the others, and the 7B and 8B models do not fit on a T4 at all. Quantized
to INT4 and served with vLLM, Qwen3-VL-2B writes a caption in \latSecQwenTwoIntKsixteen\,s, but with
\runGBQwenTwoInt\,GB of memory, over the 2\,GB budget.
We also show the performance of every model against its time per caption, memory usage and size
(\cref{fig:performance}).
We also evaluate the effect of reading more past frames on different models on the four benchmarks,
and show that not only can \model read more past frames without significant computational cost,
but reading more past frames also improves its accuracy (\cref{sec:supp:frames}): from $L{=}0$ to 16, its mean
accuracy rises by \gaGainOursMean\ [\gaGainOursMeanLo, \gaGainOursMeanHi] points, while its 50-token caption takes
no longer (\latSecOursKzero\ and \latSecOursKsixteen\,s). This is not a property that its base model, LFM2.5-VL-450M, has by default: it loses
accuracy with more past frames (\gaGainLfmSmallMean\ [\gaGainLfmSmallMeanLo, \gaGainLfmSmallMeanHi], \cref{sec:supp:frames}). We attribute
this to the temporal context that \model learns to use through its training.
On held-out videos, we can directly observe how the captions generated by \model evolve
with additional temporal context (\cref{fig:lookback}). The examples in \cref{sec:supp:examples} show how the past changes
what the captions describe: with the past, \model describes a water lily opening and the camera moving around a parked car;
without it, the captions describe the flowers and the car largely as still scenes. The past supplies the change in
the scene or the viewpoint, rather than just more details of the current frame.
\subsection{Ablations}
\label{sec:ablations}
\begin{table}[t]
  \centering
  \caption{\textbf{What the main ablations change.} Differences in dev fact-F1 (\% of the teacher ceiling) with paired 95\% intervals, positive in favour of the first setting of each row. The first row uses \model; the others are SFT or GRPO runs on the 0.1$\times$ data. Each row is a separate comparison, not a cumulative gain.}
  \label{tab:ablations}
  \scriptsize
  \setlength{\tabcolsep}{3pt}
  \begin{tabular}{@{}llr@{}}
    \toprule
    Comparison & measured at & change [95\% interval] \\
    \midrule
    Real past vs current frame repeated & $L{=}16$ & +\cyRepeatDrop\ [\cyRepeatDropLo, \cyRepeatDropHi] \\
    16 frames of 16 vs 4 of 64 tokens & 256 past tokens & \eFourMoreSecondsBig\ [\eFourMoreSecondsBigLo, \eFourMoreSecondsBigHi] \\
    Many lookbacks vs $L{=}16$ only & $L{=}0$ & \ablBaseOverOnlyKKzero\ [\ablBaseOverOnlyKKzeroLo, \ablBaseOverOnlyKKzeroHi] \\
     & $L{=}16$ & \ablBaseOverOnlyKKsixteen\ [\ablBaseOverOnlyKKsixteenLo, \ablBaseOverOnlyKKsixteenHi] \\
    GRPO with KL vs SFT & $L{=}16$ & \mainOverSFTKsixteen\ [\mainOverSFTKsixteenLo, \mainOverSFTKsixteenHi] \\
    \bottomrule
  \end{tabular}
\end{table}
To explain the performance of \model, we check the information in the cached frames, the allocation of past tokens,
and the stages of the training recipe (\cref{tab:ablations}). We ablate training choices on the 0.1$\times$ data,
\sftTrain\ LLaVA-Video windows against the \sftTrainTen\ windows of \model, then check data scaling separately.
The evidence controls run on \model itself. The 0.1$\times$ model already recovers a good portion of the full model's
performance; its results and the full training comparisons are in the supplementary material (\cref{sec:supp:ablations}).
Scores here are fact-F1 with the original length rule on the LLaVA-Video dev split unless stated otherwise.

The first claim we check is the central claim of the paper: for a strong captioning model,
reading more past frames is important, which is why making it cheap is the key to a strong model (\cref{sec:supp:controls}).
We first check whether compute explains the gain in performance: we provide the current frame 16 times
to the model, which keeps the prompt and its length, and check whether it gains performance. It gains nothing over
$L{=}0$ (\cyRepeatOverZero\ [\cyRepeatOverZeroLo, \cyRepeatOverZeroHi]) and stays \cyRepeatDrop\ points below the
real past.
We then check whether the model is reading the past frames: we provide the past frames of a different video.
The model then scores \cyForeign\%, far below no past frames at all (\cyPlainKzero\%), which confirms that the model
is indeed reading the past frames and inferring from them. The gain therefore comes from the evidence in those
frames, not just from a longer prompt.
We end these controls by checking the temporal order, keeping the same past frames. Shuffling them costs
\cyShuffledCost\ [\cyShuffledCostLo, \cyShuffledCostHi] points, an interval that includes zero. Fact-F1 matches the
facts a caption states, not the order of the events, so this control cannot show whether the model reads frame order.

The second claim we check justifies why we give each past frame a small latent representation rather than richer
representations of fewer frames: at an equal number of past tokens, 16 frames of 16 tokens beat 4 frames of 64
tokens by \eFourMoreSecondsBig\ [\eFourMoreSecondsBigLo, \eFourMoreSecondsBigHi] points. The comparison also holds
at a smaller token budget: 16 frames of 4 tokens beat 4 frames of 16 by \eFourMoreSecondsSmall\
[\eFourMoreSecondsSmallLo, \eFourMoreSecondsSmallHi] points (SFT models). At both budgets, covering more moments
helps more than representing fewer moments in greater detail.

We then check the training recipe, one stage at a time (\cref{sec:supp:ablations}).
The third claim we check is that a single set of weights can serve every lookback. Training with the sandwich rule
over three lookbacks gains \ablBaseOverOnlyKKzero\ [\ablBaseOverOnlyKKzeroLo, \ablBaseOverOnlyKKzeroHi] points at
$L{=}0$ over training at $L{=}16$ only, while the difference at $L{=}16$ is \ablBaseOverOnlyKKsixteen\
[\ablBaseOverOnlyKKsixteenLo, \ablBaseOverOnlyKKsixteenHi] points: a small cost at the full lookback, in exchange for a model that works
at every lookback. Drawing one random lookback per window instead of the sandwich gives similar results
(\ablBaseOverRandKzero\ at $L{=}0$, \ablBaseOverRandKsixteen\ at $L{=}16$). Serving different lookbacks therefore does
not depend on the sandwich rule alone. We also check how much of the student to train: updating all weights rather
than LoRA adapters gains \fullOverLoraKzero\ points at $L{=}0$ and \fullOverLoraKsixteen\ at $L{=}16$.

\label{sec:abl:rl}
The fourth claim we check is that reinforcement learning improves the captions beyond supervised imitation.
GRPO with the KL penalty adds \mainOverSFTKzero\ [\mainOverSFTKzeroLo, \mainOverSFTKzeroHi] points at $L{=}0$ and
\mainOverSFTKsixteen\ [\mainOverSFTKsixteenLo, \mainOverSFTKsixteenHi] at $L{=}16$ over its SFT model, and
\esixOverSftTenMean\ [\esixOverSftTenMeanLo, \esixOverSftTenMeanHi] on the mean of the four benchmarks. It raises
precision from \precisionSFTKsixteen\ to \precisionMainKsixteen\ at $L{=}16$, while keeping recall
(\recallSFTKsixteen\ to \recallMainKsixteen).

We then isolate what the KL penalty prevents. Without it, precision climbs to \precisionRLKsixteen\ but recall
falls to \recallRLKsixteen, and \genCloserRL\% of the test captions end with a fact-free closer, against
\genCloserKl\% with it. Against the run without it, the penalty moves fact-F1 on LLaVA-Video test by
\tKlMinusRLKsixteen\ [\tKlMinusRLKsixteenLo, \tKlMinusRLKsixteenHi] points. We accept this loss in exchange for
preserving recall and reducing fact-free endings: the penalty controls a failure mode that fact-F1 alone does not
fully capture.

\label{sec:abl:data}
We end by checking what the full data brings. SFT on the full data gains \ovSftOverSftMean\
[\ovSftOverSftMeanLo, \ovSftOverSftMeanHi] points on the mean of the four benchmarks over SFT on the 0.1$\times$ data.
Trained on the full data for the same \sftSteps\ steps as on the 0.1$\times$ data, it gains nothing
(\ovEqOverSftMean\ [\ovEqOverSftMeanLo, \ovEqOverSftMeanHi]). The improvement comes with longer training; a larger
pool alone does not recover it at the same number of steps. After RL, \model ends \tenOverEsixMean\
[\tenOverEsixMeanLo, \tenOverEsixMeanHi] points above the 0.1$\times$ model on the benchmarks. On held-out captions,
the full data helps on OpenVid, one of its training sources (\tenOverEsixOKsixteen\ [\tenOverEsixOKsixteenLo,
\tenOverEsixOKsixteenHi]), but not on LLaVA-Video test (\tenOverEsixTKsixteen\ [\tenOverEsixTKsixteenLo,
\tenOverEsixTKsixteenHi]). The clearest gain is on the added training source, not on LLaVA-Video test.

Finally, we check whether the student or the labels limit caption quality. After the same LoRA SFT on the same
labels, LFM2.5-VL-1.6B reaches \shareBigKsixteen\% of the ceiling at $L{=}16$, against \shareLoraKsixteen\% for the
450M model. Distilling the 1.6B model into the 450M one moves it by at most \distMaxAbs\ points, with every interval
spanning zero. The labels support higher scores in a larger student, but that gain does not transfer back to the
450M model through this distillation. This points to student capacity as a limit of the current recipe.

%% file: floats/tab_main.tex
\begin{table*}[t]
  \centering
  \caption{\textbf{Every model on the same questions and the same T4, by size.} $L$: lookback, the number of past frames a caption reads. T4: eager PyTorch FP16, batch 1. Memory in GB ($10^9$ bytes): the process memory while captioning at $L{=}16$. Time: the median time of a caption forced to 50 new tokens, so that every model decodes the same work; \model reads its cached past frames (p99 at $L{=}16$: \latPNinetyNineOursKsixteen\,s). Every timed model with the same script on a T4. $^\ddagger$: vLLM with weight-only INT8 or INT4, peak device memory. $^\dagger$: the weights alone (FP16; the teacher FP8). OOM: the weights alone exceed the T4's \tFourGB\,GB. Fact-F1 as a percentage of the teacher ceiling (judge: phi-4) on held-out OpenVid (a training source of \model, held out by video) and LLaVA-Video test (test), without a test-time word-limit penalty. Parentheses give the original score with the 60-word rule where the rounded values differ. This rule was part of \model's RL reward; the baselines were only prompted for 30--50 words. Accuracy (\%) by caption-then-answer, every row with the same answerer and settings over the same questions. TimeChat-Online-7B keeps its token drop; StreamingVLM-7B, silent when asked for one caption, decodes with its silent tokens banned (\cref{sec:supp:prompts}). Each describer reads $L{=}\min(t,16)$ past frames at the question's second $t$ (OVO-RT: OVO-Bench Real-Time; StrBench: StreamingBench Real-Time; VMME: Video-MME short; mean: of the four). Words: mean words per caption on the four benchmarks; the task asks for 30--50. $^*$: under 30 words per caption. As a deliberate handicap to \model, we cut its captions request by request to the word count of LFM2.5-VL-450M, SmolVLM2-500M or Mobile-VideoGPT-0.5B, even mid-sentence. Despite losing part of their content, they still lead by \oneCutOverLfmSmallMean\ [\oneCutOverLfmSmallMeanLo, \oneCutOverLfmSmallMeanHi], \oneCutOverSmolSmallMean\ [\oneCutOverSmolSmallMeanLo, \oneCutOverSmolSmallMeanHi] and \oneCutOverMobileMean\ [\oneCutOverMobileMeanLo, \oneCutOverMobileMeanHi] points on the mean. These are truncated outputs, not captions generated for a shorter word budget. No truncation comparison was run for Qwen2.5-VL-7B or LiveCC-7B. --: not scored. \Cref{fig:performance} plots the mean against time, memory and parameters.}
  \label{tab:main}
  \scriptsize
  \setlength{\tabcolsep}{3pt}
  \input{tables/main}
\end{table*}

%% file: tables/main.tex
\begin{tabular}{@{}lrrrrrrrrrrrrr@{}}
\toprule
 & & \multicolumn{3}{c}{T4, batch 1} & \multicolumn{3}{c}{fact-F1, \% of ceiling} & & \multicolumn{5}{c}{accuracy (\%), $L{=}\min(t,16)$} \\
\cmidrule(lr){3-5}\cmidrule(lr){6-8}\cmidrule(l){10-14}
 & & memory & \multicolumn{2}{c}{s per caption} & \multicolumn{2}{c}{OpenVid} & test & & & & & & \\
\cmidrule(lr){4-5}\cmidrule(lr){6-7}
Model & params (B) & (GB) & $L{=}0$ & $L{=}16$ & $L{=}0$ & $L{=}16$ & $L{=}16$ & words & OVO-RT & MVBench & StrBench & VMME & mean \\
\midrule
no captions (floor) & -- & -- & -- & -- & -- & -- & -- & -- & \oneAccFloorOvo & \oneAccFloorMvb & \oneAccFloorSb & \oneAccFloorVm & \oneAccFloorMean \\
\midrule
\multicolumn{14}{@{}l}{\emph{under 1B parameters}} \\
\textbf{\model} & \parOurs & \runGBOurs & \latSecOursKzero & \latSecOursKsixteen & \textbf{\lengthCellOOursKzero} & \textbf{\lengthCellOOursKsixteen} & \textbf{\lengthCellTOursKsixteen} & \oneWordsOurs & \textbf{\oneAccOursOvo} & \textbf{\oneAccOursMvb} & \textbf{\oneAccOursSb} & \textbf{\oneAccOursVm} & \textbf{\oneAccOursMean} \\
LFM2.5-VL-450M$^*$ & \parLfmSmall & \runGBLfmSmall & \latSecLfmSmallKzero & \latSecLfmSmallKsixteen & \lengthCellOLfmSmallKzero & \lengthCellOLfmSmallKsixteen & \lengthCellTLfmSmallKsixteen & \oneWordsLfmSmall & \oneAccLfmSmallOvo & \oneAccLfmSmallMvb & \oneAccLfmSmallSb & \oneAccLfmSmallVm & \oneAccLfmSmallMean \\
SmolVLM2-500M$^*$ & \parSmolSmall & \runGBSmolSmall & \latSecSmolSmallKzero & \latSecSmolSmallKsixteen & \lengthCellOSmolSmallKzero & \lengthCellOSmolSmallKsixteen & \lengthCellTSmolSmallKsixteen & \oneWordsSmolSmall & \oneAccSmolSmallOvo & \oneAccSmolSmallMvb & \oneAccSmolSmallSb & \oneAccSmolSmallVm & \oneAccSmolSmallMean \\
FastVLM-0.5B & \parFastVlm & \runGBFastVlm & \latSecFastVlmKzero & \latSecFastVlmKsixteen & \lengthCellOFastVlmKzero & \lengthCellOFastVlmKsixteen & \lengthCellTFastVlmKsixteen & \nbWordsFastVlm & \nbAccFastVlmOvo & \nbAccFastVlmMvb & \nbAccFastVlmSb & \nbAccFastVlmVm & \nbAccFastVlmMean \\
Mobile-VideoGPT-0.5B$^*$ & \parMobile & \runGBMobile & \latSecMobileKzero & \latSecMobileKsixteen & \lengthCellOMobileKzero & \lengthCellOMobileKsixteen & \lengthCellTMobileKsixteen & \oneWordsMobile & \oneAccMobileOvo & \oneAccMobileMvb & \oneAccMobileSb & \oneAccMobileVm & \oneAccMobileMean \\
\midrule
\multicolumn{14}{@{}l}{\emph{1B to 3B parameters}} \\
InternVL3.5-1B~\cite{internvl35} & \parInternOne & \runGBInternOne & \latSecInternOneKzero & \latSecInternOneKsixteen & \lengthCellOInternOneKzero & \lengthCellOInternOneKsixteen & \lengthCellTInternOneKsixteen & \oneWordsInternOne & \oneAccInternOneOvo & \oneAccInternOneMvb & \oneAccInternOneSb & \oneAccInternOneVm & \oneAccInternOneMean \\
\quad INT8$^\ddagger$ & \parInternOne & \runGBInternOneInt & \latSecInternOneIntKzero & \latSecInternOneIntKsixteen & \lengthCellOInternOneIntKzero & \lengthCellOInternOneIntKsixteen & \lengthCellTInternOneIntKsixteen & \oneWordsInternOneInt & \oneAccInternOneIntOvo & \oneAccInternOneIntMvb & \oneAccInternOneIntSb & \oneAccInternOneIntVm & \oneAccInternOneIntMean \\
LFM2.5-VL-1.6B & \parLfmBig & \runGBLfmBig & \latSecLfmBigKzero & \latSecLfmBigKsixteen & \lengthCellOLfmBigKzero & \lengthCellOLfmBigKsixteen & \lengthCellTLfmBigKsixteen & \oneWordsLfmBig & \oneAccLfmBigOvo & \oneAccLfmBigMvb & \oneAccLfmBigSb & \oneAccLfmBigVm & \oneAccLfmBigMean \\
Qwen3-VL-2B~\cite{qwen3vl2025} & \parQwenTwo & \runGBQwenTwo & \latSecQwenTwoKzero & \latSecQwenTwoKsixteen & \lengthCellOQwenTwoKzero & \lengthCellOQwenTwoKsixteen & \lengthCellTQwenTwoKsixteen & \oneWordsQwenTwo & \oneAccQwenTwoOvo & \oneAccQwenTwoMvb & \oneAccQwenTwoSb & \oneAccQwenTwoVm & \oneAccQwenTwoMean \\
\quad INT4$^\ddagger$ & \parQwenTwo & \runGBQwenTwoInt & \latSecQwenTwoIntKzero & \latSecQwenTwoIntKsixteen & \lengthCellOQwenTwoIntKzero & \lengthCellOQwenTwoIntKsixteen & \lengthCellTQwenTwoIntKsixteen & \oneWordsQwenTwoInt & \oneAccQwenTwoIntOvo & \oneAccQwenTwoIntMvb & \oneAccQwenTwoIntSb & \oneAccQwenTwoIntVm & \oneAccQwenTwoIntMean \\
SmolVLM2-2.2B & \parSmolBig & \runGBSmolBig & \latSecSmolBigKzero & \latSecSmolBigKsixteen & \lengthCellOSmolBigKzero & \lengthCellOSmolBigKsixteen & \lengthCellTSmolBigKsixteen & \oneWordsSmolBig & \oneAccSmolBigOvo & \oneAccSmolBigMvb & \oneAccSmolBigSb & \oneAccSmolBigVm & \oneAccSmolBigMean \\
InternVL3.5-2B & \parInternTwo & \runGBInternTwo & \latSecInternTwoKzero & \latSecInternTwoKsixteen & \lengthCellOInternTwoKzero & \lengthCellOInternTwoKsixteen & \lengthCellTInternTwoKsixteen & \oneWordsInternTwo & \oneAccInternTwoOvo & \oneAccInternTwoMvb & \oneAccInternTwoSb & \oneAccInternTwoVm & \oneAccInternTwoMean \\
\midrule
\multicolumn{14}{@{}l}{\emph{above 3B parameters}} \\
Qwen3-VL-4B & \parQwenFour & \runGBQwenFour & \latSecQwenFourKzero & \latSecQwenFourKsixteen & \lengthCellOQwenFourKzero & \lengthCellOQwenFourKsixteen & \lengthCellTQwenFourKsixteen & \oneWordsQwenFour & \oneAccQwenFourOvo & \oneAccQwenFourMvb & \oneAccQwenFourSb & \oneAccQwenFourVm & \oneAccQwenFourMean \\
Qwen2.5-VL-7B$^*$~\cite{qwen25vl2025} & \parQwenSevenTwoFive & $\ge$\wgtGBQwenSevenTwoFive$^\dagger$ & OOM & OOM & \lengthCellOQwenSevenTwoFiveKzero & \lengthCellOQwenSevenTwoFiveKsixteen & \lengthCellTQwenSevenTwoFiveKsixteen & \oneWordsQwenSevenTwoFive & \oneAccQwenSevenTwoFiveOvo & \oneAccQwenSevenTwoFiveMvb & \oneAccQwenSevenTwoFiveSb & \oneAccQwenSevenTwoFiveVm & \oneAccQwenSevenTwoFiveMean \\
LiveCC-7B$^*$ & \parLiveCc & $\ge$\wgtGBLiveCc$^\dagger$ & OOM & OOM & \lengthCellOLiveCcKzero & \lengthCellOLiveCcKsixteen & \lengthCellTLiveCcKsixteen & \oneWordsLiveCc & \oneAccLiveCcOvo & \oneAccLiveCcMvb & \oneAccLiveCcSb & \oneAccLiveCcVm & \oneAccLiveCcMean \\
StreamingVLM-7B & \parStreamingVlm & $\ge$\wgtGBStreamingVlm$^\dagger$ & OOM & OOM & \lengthCellOStreamingVlmKzero & \lengthCellOStreamingVlmKsixteen & \lengthCellTStreamingVlmKsixteen & \nbWordsStreamingVlm & \nbAccStreamingVlmOvo & \nbAccStreamingVlmMvb & \nbAccStreamingVlmSb & \nbAccStreamingVlmVm & \nbAccStreamingVlmMean \\
TimeChat-Online-7B & \parTimeChat & $\ge$\wgtGBTimeChat$^\dagger$ & OOM & OOM & \lengthCellOTimeChatKzero & \lengthCellOTimeChatKsixteen & \lengthCellTTimeChatKsixteen & \nbWordsTimeChat & \nbAccTimeChatOvo & \nbAccTimeChatMvb & \nbAccTimeChatSb & \nbAccTimeChatVm & \nbAccTimeChatMean \\
Qwen3-VL-8B & \parQwenEight & $\ge$\wgtGBQwenEight$^\dagger$ & OOM & OOM & \lengthCellOQwenEightKzero & \lengthCellOQwenEightKsixteen & \lengthCellTQwenEightKsixteen & \oneWordsQwenEight & \oneAccQwenEightOvo & \oneAccQwenEightMvb & \oneAccQwenEightSb & \oneAccQwenEightVm & \oneAccQwenEightMean \\
\midrule
Qwen3.8-27B (teacher) & \parTeacher & $\ge$\wgtGBTeacher$^\dagger$ & OOM & OOM & -- & -- & -- & \oneWordsTeacher & \oneAccTeacherOvo & \oneAccTeacherMvb & \oneAccTeacherSb & \oneAccTeacherVm & \oneAccTeacherMean \\
\bottomrule
\end{tabular}

%% file: sec/5_discussion.tex
\section{Discussion and limitations}
\label{sec:discussion}
\paragraph{Discussion.}
Nothing in the architecture or in the recipe is specific to captioning: the cached
frames, the lookback chosen at run time, the per-frame schedule, SFT over many lookbacks and GRPO with a shared lookback
and a KL penalty would apply as they are to other generative tasks on video streams, such as answering questions or
raising alerts. A new task enters the recipe only through the teacher's labels and the reward. We have, however, run the
recipe once: on streaming captioning, with one student family and a horizon of 16\,s.
On its own view, the teacher scores \shareTeacherKsixteen\% of the ceiling at
$L{=}16$ on dev, and \model \devTenKsixteen\%. Our ablations point to the student's capacity as a limit on its performance
(\cref{sec:ablations}): a larger student trained on the same labels scores far higher, and \model misses recall and
precision about equally (\devTenRecallKsixteen\ against \teacherRecallKsixteen, \devTenPrecisionKsixteen\ against
\teacherPrecisionKsixteen). More data helped only with more training, so we expect the gap to close with the capacity of
the student rather than with more labels, which is the trade-off we make by staying under 500M parameters.

\paragraph{Limitations.}
While the recipe is general, we have run it on only one task, with one model architecture as the student.
The gap to the teacher's performance is hypothesized to be due to the student's capacity; whether the size
of the student or its architecture is the limiting factor is unknown.
The teacher model, Qwen3.8-27B, while performant in our tests, is still far from the state of the art
in vision capabilities, and whether a student model can be further improved by a teacher with higher-quality
outputs is left to future work.
We have also read the benchmarks through captions, keeping the parts that captions can answer. This setup shows
what each model's captions carry, but it scores general-purpose models on the task ours was
trained for, a limitation that labels covering more general vision tasks could address.

\model shows that a small streaming model can use more of the past without encoding it again, and turn that past into
better captions within a T4 budget.

%% file: sec/6_llm.tex
\section*{LLM Usage Statement}
We used large language models in three ways. First, as components of the method: a 27B vision-language model wrote
the training labels, a 9B model extracted facts and scored the rewards, a 14B model judged the held-out captions, and a
24B model answered the benchmark questions from the captions (\cref{sec:training,sec:results}). Second, as coding
agents: Claude Opus 5.5 and Claude Sonnet 5.5 (Anthropic) wrote code for training, evaluation, deployment and the
figure and table scripts, and launched and monitored experiments; the authors audited the code. Third, for this paper:
the same agents searched the literature through the arXiv API, built the reference list from arXiv records, checked
the numbers in the text against the result files, drafted parts of the text and corrected grammar and typos. The
authors wrote most of the text, revised every passage, and verified every cited paper.

%% file: tables/supp_prompts.tex
\lstdefinestyle{spprompt}{basicstyle=\ttfamily\scriptsize, breaklines=true, breakatwhitespace=true, breakindent=0pt, columns=fullflexible,
  keepspaces=true, showstringspaces=false, upquote=false, frame=single, framesep=2pt, xleftmargin=2.4pt, xrightmargin=2.4pt,
  aboveskip=2pt, belowskip=6pt, postbreak=\mbox{$\hookrightarrow$}\space}
\newcommand{\spPromptStudent}{\par\nolinenumbers\lstinputlisting[style=spprompt]{tables/prompts/Student.txt}\linenumbers}
\newcommand{\spPromptTitleStudent}{The prompt of YoloMT at lookback L (training and inference; K in the code is L)}
\newcommand{\spPromptTeacher}{\par\nolinenumbers\lstinputlisting[style=spprompt]{tables/prompts/Teacher.txt}\linenumbers}
\newcommand{\spPromptTitleTeacher}{The teacher's labelling prompt, sent with the video: the SFT caption (greedy) and the RL samples (temperature 0.7)}
\newcommand{\spPromptFacts}{\par\nolinenumbers\lstinputlisting[style=spprompt]{tables/prompts/Facts.txt}\linenumbers}
\newcommand{\spPromptTitleFacts}{Fact extraction from one caption (labels, reward, held-out judge)}
\newcommand{\spPromptCluster}{\par\nolinenumbers\lstinputlisting[style=spprompt]{tables/prompts/Cluster.txt}\linenumbers}
\newcommand{\spPromptTitleCluster}{Grouping the facts of the nine teacher captions into claims (labels)}
\newcommand{\spPromptQuestion}{\par\nolinenumbers\lstinputlisting[style=spprompt]{tables/prompts/Question.txt}\linenumbers}
\newcommand{\spPromptTitleQuestion}{Writing the multiple-choice questions from the consensus claims (labels)}
\newcommand{\spPromptAnswer}{\par\nolinenumbers\lstinputlisting[style=spprompt]{tables/prompts/Answer.txt}\linenumbers}
\newcommand{\spPromptTitleAnswer}{Answering the questions from a caption (reward); with no caption, \{caption\} is (none given), the label filter}
\newcommand{\spPromptEntail}{\par\nolinenumbers\lstinputlisting[style=spprompt]{tables/prompts/Entail.txt}\linenumbers}
\newcommand{\spPromptTitleEntail}{Verifying the consensus claims against a caption (recall; reward and held-out judge)}
\newcommand{\spPromptSupport}{\par\nolinenumbers\lstinputlisting[style=spprompt]{tables/prompts/Support.txt}\linenumbers}
\newcommand{\spPromptTitleSupport}{Verifying a caption's facts against the union of the teacher's claims (precision; reward and held-out judge)}
\newcommand{\spPromptRate}{\par\nolinenumbers\lstinputlisting[style=spprompt]{tables/prompts/Rate.txt}\linenumbers}
\newcommand{\spPromptTitleRate}{Rating the fluency of a caption from 1 to 5 (the gate; reward and held-out judge)}
\newcommand{\spPromptAnswerer}{\par\nolinenumbers\lstinputlisting[style=spprompt]{tables/prompts/Answerer.txt}\linenumbers}
\newcommand{\spPromptTitleAnswerer}{The benchmark answerer's messages when captions are given (Mistral-Small-24B-Instruct-2501, temperature 0, at most 8 new tokens)}
\newcommand{\spPromptAnswererFloor}{\par\nolinenumbers\lstinputlisting[style=spprompt]{tables/prompts/AnswererFloor.txt}\linenumbers}
\newcommand{\spPromptTitleAnswererFloor}{The answerer's user message without captions (the floor)}

%% file: sec/X_suppl.tex
\clearpage
\setcounter{page}{1}
\maketitlesupplementary
\appendix
\raggedbottom
\newcommand{\keeptogether}[1]{\par\vbox{\parindent=0pt #1}\par}
\newcommand{\promptbox}[2]{\medskip\keeptogether{{\raggedright\textbf{#1.}\par}#2}}

The input stream runs at 30\,FPS, but \model encodes one frame per second: the first frame of each second runs the
vision encoder, and the following frame arrivals carry the prompt processing and decoding of the caption
(\cref{fig:schedule}). \Cref{tab:runtime} therefore reports two different timings: the GPU time per incoming frame,
which stays within 16\,ms at every $L$, and the lag until the caption is complete, which grows with $L$.

{\small\setcounter{tocdepth}{2}\tableofcontents}

\paragraph{Terms and numbers.} $L$, the lookback, is the number of past frames a caption reads, one per second. On the
benchmarks, a caption reads the protocol lookback $L{=}\min(t,16)$ at the question's last second $t$
(\cref{sec:supp:bench}). Dev is the LLaVA-Video~\cite{llavavideo2024} dev split, and the held-out sets are LLaVA-Video test and
OpenVid~\cite{openvid2025} (\cref{sec:supp:sets}). The 0.1$\times$ data is the LLaVA-Video training windows alone; the full data adds OpenVid
(\cref{sec:supp:data}). Unless a table says otherwise, fact-F1 is in \% of the teacher ceiling and judged by
phi-4~\cite{phi4}; the ceiling is the teacher's agreement with itself, not an upper bound, so a score can exceed
100\% (\cref{sec:supp:sets} defines fact-F1 and its variants). A judging run scores a set of models and the ceiling with one judge; an answer call answers the
benchmark questions from a set of models' captions. Each run and each call scores anew, so the same model can differ by
a few tenths of a point between tables; compare rows within a table. Intervals are paired 95\% bootstrap intervals~\cite{bootstrap1993}, over
windows for fact-F1 (\stDrawsDev\ draws on dev, \stDrawsReport\ on the held-out sets) and over questions for accuracy
(\stDrawsQuestions\ draws).

\section{Model and engine}
\label{sec:supp:model}

\subsection{Architecture}
\label{sec:supp:arch}
The base is LFM2.5-VL-450M~\cite{lfm2}, and all \totalParams\ parameters are trained (\cref{fig:arch}a). The vision
encoder, SigLIP2~\cite{siglip2} with 12 layers of width 768, cuts a 256$\times$256 frame into $16{\times}16$ patches. A
projector ($2{\times}2$ pixel unshuffle~\cite{pixelshuffle2016}, then an MLP) maps them to the 64 tokens of the
present. The pooler of \cref{eq:pooler} has $1024^2+1024$ parameters, counted as \poolerParams; no spatial attention
or nonlinear layer follows its averaging.

The prompt holds $\promptBase{}+\blockTokens{}L$ tokens: the template (\templateTokens\ tokens), the $L$ blocks oldest
first, and the 64 tokens of the present, each block between two delimiters and without timestamps. The language model is LFM2.5-350M,
of the LFM2 architecture~\cite{lfm2}, with \lmParams\ parameters. It has 10 gated short-convolution layers and 6
grouped-query attention~\cite{gqa2023} layers, so only 6 layers keep a KV cache. Decoding is greedy, with at most 80
new tokens. We leave out the previous caption because our earlier captioners of 2\,s windows copied it in
\feedbackRepeatMin--\feedbackRepeatMax\% of seconds when it was fed back, but in at most \freshRepeatMax\% when each
caption was written fresh.

\subsection{Model size on a T4}
\label{sec:supp:sizing}
Real time on a T4 admits only small models. At batch 1, every generated token reads all language-model weights, and a
T4 reads 320\,GB/s. Qwen3-VL-2B~\cite{qwen3vl2025} has \qwenTwoLmParams\ language-model parameters, \qwenTwoLmGB\,GB at FP16, so a token
takes at least \qwenTwoTokenMs\,ms and a caption of \captionTokens\ tokens, the mean of \model at $L{=}16$, at least
\qwenTwoCaptionMs\,ms. Thirty frames of at
most 16\,ms each give \gpuMsPerSecond\,ms of GPU time per second, less than this caption needs. \model reads \eyLmGB\,GB per token with its cut output head (\cref{sec:supp:runtime}), at least
\eyTokenMs\,ms. These are floors at peak bandwidth, computed from the model
configurations.

\subsection{The real-time engine}
\label{sec:supp:runtime}
\input{floats/tab_runtime}
\model runs as one TensorRT engine (NVIDIA TensorRT \trtVersion, FP16, built strongly typed), on the schedule of
\cref{fig:schedule}. Its output head is cut to \eyHeadRows\ token ids, which cover \eyHeadCover\% of the tokens of the
teacher's SFT labels, plus byte-level and special tokens. A caption still open at the next second would be abandoned
(\cref{tab:runtime}). Where FP16 captions differ from the float32 reference, they differ in wording at near-ties.

Paced by a 30\,FPS clock, one stream's p99 (\cref{sec:results}) is above the back-to-back times of \cref{tab:runtime}
because the GPU slows down between frames; its host wall time stays at most \stOneHostMax\,ms. Separate processes
share one T4 when their frame arrivals are staggered over the second.
Three streams keep every p99 at most \stThreeMax\,ms at $L{=}0$, 8 and 16 (host \stThreeHostMax\,ms), and four do not
(\stFourMin\,ms and more). Arrivals that coincide collide: two aligned streams reach \stAlignedTwoMin\,ms.

\section{Training}
\label{sec:supp:recipe}
Every model and setting of the recipe is a slot (\cref{fig:recipe}): the student, the teacher, the claim extractor and reward judge, the
held-out judge and the device. This paper fills them with LFM2.5-VL-450M, Qwen3.8-27B~\cite{qwen38_27b}, Qwen3.5-9B~\cite{qwen35_9b}, phi-4
and a T4.
Another GPU admits another student, and another teacher, judge or device changes a slot, not the stages.

\subsection{Reward}
\label{sec:supp:reward}

\paragraph{Online claim scoring.}
Qwen3.5-9B, an LLM judge~\cite{llmjudge2023} in text-only mode, scores the captions that pass the hard filters below
against the claims and questions of \cref{sec:supp:data}. The first round batches three separate requests per
caption: extract up to 16 atomic facts~\cite{factscore2023}, verify every consensus claim against the caption, and
answer the window's questions from the caption alone. A second round verifies the extracted student facts against the
teacher's full claim union, which gives the recall and precision of fact-F1 (\cref{sec:supp:sets}); an unmentioned
claim receives a ``no''.
Extraction, verdicts and question answers are constrained by JSON schemas. The judge is served with
vLLM~\cite{vllm2023}; neither the frames nor the 27B teacher are queried during these calls.

\paragraph{The multiplicative gate.}
Without a gate, an earlier RL run learned fact-dense word salad, which the factual terms alone still reward. The gate
$g$ multiplies these terms in \cref{eq:reward}, so a caption earns them in full only if it reads as fluent English and
does not repeat itself:
\begin{equation}
\begin{aligned}
  g(y)&=\frac{\operatorname{clip}_{[0,1]}((s(y)-1)/4)}{1+e(y)},\\
  e(y)&=[r_{\mathrm{c}}(y)-\tau_{\mathrm{c}}]_+ + [r_{\mathrm{b}}(y)-\tau_{\mathrm{b}}]_+,
\end{aligned}
\end{equation}
where $[a]_+{=}\max(0,a)$. The numerator maps a fluency rating $s$ from 1--5 to $[0,1]$. In a request of its own
(\cref{sec:supp:allprompts}), Qwen3.5-9B rates the English of the caption from 1 (a list of words or broken
fragments) to 5 (fluent, natural sentences). It decodes greedily, so the rating it writes is an integer, and the same 5
can come with near certainty or with doubt. As in G-Eval~\cite{geval2023}, we use instead its expected rating
$s(y){=}\sum_{k=1}^5 k\,p(k\mid y)$, which tells such captions apart; $p(k\mid y)$ is the probability of answer $k$,
renormalized over the five answers so that $s$ is a mean between 1 and 5. The denominator penalizes repeats. Each
count sums the occurrences after the first, $r_{\mathrm{c}}{=}\sum_v[n_v-1]_+$ over content words and
$r_{\mathrm{b}}{=}\sum_{u,v}[n_{uv}-1]_+$ over adjacent word bigrams. Words are lowercased and split at characters
other than letters, digits and apostrophes; content words exclude a fixed English stopword list, and bigrams include
stopwords. The thresholds $\tau_{\mathrm{c}}{=}\fluencyContentThreshold$ and
$\tau_{\mathrm{b}}{=}\fluencyBigramThreshold$ are the 95th percentiles of these counts on \fluencyCaptions\ training
teacher captions, so at most 5\% of them exceed each threshold; a single repeat beyond the thresholds halves the gate.

\paragraph{The additive fluency term.}
The gate only scales the factual terms, so fluency pays only in proportion to the facts a caption gets right, and the
gate's rating comes from the judge that RL trains against. The additive term rewards natural text in its own right,
measured by a second model independent of the judge: a frozen, text-only LFM2.5-350M that reads the caption alone,
without frames or an instruction. It rests on language-model likelihood as a measure of acceptability and
fluency~\cite{acceptability2017,fluencyeval2018}: a sentence that a language model finds predictable reads as natural,
and the same words shuffled do not. With the weight $\lambda_4{=}0.1$, it adds at most 0.1 to a reward of at most 1.
For caption $y$ with tokens $y_i$ and $C(y)$ characters, the model's surprise in bits per character and the fluency
term are
\begin{equation}
\begin{aligned}
  b(y)&=-\frac{1}{C(y)}\sum_i\log_2 p_{\mathrm{frozen}}(y_i\mid y_{<i}),\\
  \mathrm{fluency}(y)&=\operatorname{clip}_{[0,1]}\!\left(\frac{b_{\mathrm{shuf}}-b(y)}{b_{\mathrm{shuf}}-b_{05}}\right).
\end{aligned}
\end{equation}
Per character, $b$ is an average, so a longer caption is not penalized for its length; the token history starts with
the model's beginning-of-sequence token. The scale is set once on the same teacher captions. Fluency is 1 at
$b_{05}{=}\fluencyBGood$, the fifth percentile of their $b$, as predictable as the most predictable 5\% of teacher
captions, and 0 at $b_{\mathrm{shuf}}{=}\fluencyBShuffled$, the median $b$ after shuffling each caption's words, the
level of word salad. It is linear in between, and a caption at or above $b_{\mathrm{shuf}}$ gets $R{=}0$ (below).

\paragraph{Hard filters and checkpoint selection.}
Besides a caption over 60 whitespace-separated words, an empty caption, a decoding replacement character, a
special-token marker, three identical words in a row and $b(y)\ge b_{\mathrm{shuf}}$ also give $R{=}0$. The gated fact-F1 of greedy
captions on the \rlVal\ selection windows (\cref{sec:supp:sets}) at $L{=}16$ selects the checkpoint. The tripwire
counts a strike when the mean training reward rises by more than two standard errors and this held-out score does
not; after two strikes in a row the precision weight becomes 0.4 and the word limit 50.

\subsection{GRPO update}
\label{sec:supp:grpo}
All captions of a group read the same lookback. For window $w$, let $m_w{=}\min(16,L_{\mathrm{avail}}^w)$; its
group's lookback is drawn as
\begin{equation}
  \Pr(L_w{=}\ell)=\alpha\,\mathbf{1}[\ell{=}m_w]+\frac{1-\alpha}{m_w+1},
  \quad \ell\in\{0,\ldots,m_w\},
\end{equation}
with $\alpha{=}\rlMaxProb$. The uniform branch includes $m_w$, so its total probability is slightly above one half.
Each of the \rlWindowsStep\ windows generates $G{=}\rlGroupSize$ captions at temperature \rlTemperature, with at most
\rlMaxNew\ new tokens. Within a group, $A_{wi}{=}(R_{wi}-\bar R_w)/\sigma_w$, using the population standard
deviation. Groups with $\sigma_w\le10^{-6}$ or an incomplete judge response are dropped.

Let $h_{wit}$ be the prompt and the prefix of caption $i$ before token $t$. The sampled-token ratios are
\begin{equation}
\begin{aligned}
  \rho_{wit}&=\frac{p_\theta(y^{wi}_t\mid h_{wit})}{p_{\mathrm{old}}(y^{wi}_t\mid h_{wit})},\\
  q_{wit}&=\frac{p_{\mathrm{SFT}}(y^{wi}_t\mid h_{wit})}{p_\theta(y^{wi}_t\mid h_{wit})}.
\end{aligned}
\end{equation}
Here $p_{\mathrm{old}}$ generated the rollouts and $p_{\mathrm{SFT}}$ is the frozen initialization of RL.
With $\tilde\rho{=}\operatorname{clip}(\rho,1-\epsilon,1+\epsilon)$, as in PPO~\cite{ppo2017}, we maximize
\begin{equation}
\begin{aligned}
  u_{wit}&=\min(\rho_{wit}A_{wi},\tilde\rho_{wit}A_{wi}),\\
  \mathcal J(\theta)&=\frac{1}{N}\sum_{w,i,t}\left[u_{wit}-\beta(q_{wit}-\log q_{wit}-1)\right],
\end{aligned}
\end{equation}
where $\epsilon{=}\rlClip$ and $\beta{=}\rlBeta$. The second term is the $k_3$ KL estimator used in
GRPO~\cite{grpo2024}. The sum runs over action tokens of kept groups, including the end-of-caption token; $N$ is
their total count, not a mean over caption lengths. We take one AdamW~\cite{adamw2019} update per batch of rollouts, accumulating
gradients across replay micro-batches and clipping their norm to 1. The learning rate is \rlLr, and \rlVisionLr\ for
the vision encoder. All student weights train.

\subsection{Training ablations}
\label{sec:supp:ablations}
\input{floats/tab_recipe}
\input{floats/tab_supp_pool}
\input{floats/tab_supp_kl}
\input{floats/tab_supp_bench}
GRPO adds fact-F1 over every SFT model, the warm-up helps at $L{=}16$ and all weights beat LoRA adapters, while
distilling a 1.6B student into the 450M model does not help (\cref{tab:recipe}). For the same past tokens, more frames
with fewer tokens each beat fewer, richer frames (\cref{tab:supp:pool}). The KL penalty trades some fact-F1 for recall
and removes the closing formula (\cref{tab:supp:kl}). More data helps on the benchmarks only with longer training,
and on held-out captions only on OpenVid, one of its sources (\cref{tab:supp:bench}). The sandwich~\cite{slimmable2019,usnet2019} and
random-$L$ comparisons of \cref{sec:ablations} are between the three rows of \cref{tab:recipe} without the 2\,s
pre-stage; the random-$L$ ones have the intervals [\ablBaseOverRandKzeroLo, \ablBaseOverRandKzeroHi] at $L{=}0$ and [\ablBaseOverRandKsixteenLo,
\ablBaseOverRandKsixteenHi] at $L{=}16$. Three seeds of the 0.1$\times$ recipe without KL spread by up to
\stSeedSpreadEnds\ points at $L{=}0$ and 16. On dev at $L{=}16$, \model trails its teacher in recall and precision
alike, \stDevYoloRecall\ and \stDevYoloPrecision\ against \stDevTeacherRecall\ and \stDevTeacherPrecision, with
captions of \stDevYoloWords\ and \stDevTeacherWords\ words.

\subsection{Data}
\label{sec:supp:data}
The SFT videos come from \sftSources\ subsets of LLaVA-Video-178K~\cite{llavavideo2024}, whose captions we do not use, and
from OpenVid-1M~\cite{openvid2025}. Window ends are at least \spStride\,s apart, with at most \spWindowsPerVideo\ windows per
video. The split hashes the name of each source video, so the training windows and the three evaluation sets
(\cref{sec:supp:sets}) share no video. The 0.1$\times$ data is the \sftTrain\ LLaVA-Video training windows. The full data adds
\sftOpenvidTen\ OpenVid windows from \spOpenvidClips\ clips, \sftTrainTen\ windows in all. The SFT target of a window is the
teacher's greedy caption of the student's own frames, kept if it has at most \spLabelMaxWords\ words and
\spLabelMaxTokens\ student tokens.

For the consensus and the union of the RL windows (\cref{sec:training}), Qwen3.5-9B splits every teacher caption into
atomic claims and groups the equal ones, so that both count claims, not wordings. The consensus fraction is rounded up
when length filtering keeps fewer samples, and a window with fewer than \spMinKept\ kept samples has no consensus.
From the consensus, Qwen3.5-9B writes \spQuestionsWritten\ multiple-choice questions. A question is kept only if Qwen3.5-9B, given no caption, answers it correctly in at most
\spLeakMaxRight\ of the \spLeakRotations\ rotations of its options, and a window keeps at most \spQuestionsKept. A window
enters RL only with a consensus and a kept question.

\subsection{Schedule and compute}
\label{sec:supp:schedule}
\begin{table}[t]
  \centering
  \caption{\textbf{The training phases of \model, in order.} Target: the windows the teacher's captions describe;
  lookbacks: the past frames the student reads with the current one. The warm-up starts from our 2\,s captioner,
  which reads one past frame (\cref{sec:training}). Sandwich: $L{=}0$, a random $L$ and the largest available, up to
  the bound; GRPO draws one $L$ per group (\cref{sec:supp:grpo}). 0.1$\times$ runs in parentheses.}
  \label{tab:supp:phases}
  \scriptsize
  \setlength{\tabcolsep}{3pt}
  \begin{tabular}{@{}lllll@{}}
    \toprule
    Phase & Target & Lookbacks & Weights & Steps \\
    \midrule
    warm-up SFT & 2\,s & sandwich, $L\le\preKmax$ & LoRA, merged & \spPreSteps \\
    SFT & 16\,s & sandwich, $L\le16$ & all & \sftStepsTen\ (\sftSteps) \\
    GRPO & claims of 16\,s & one $L$ per group & all & \rlExitTen\ (\spOneRlExit) \\
    \bottomrule
  \end{tabular}
\end{table}
The teacher's captions describe a fixed window, while the student reads $L$ past frames
(\cref{tab:supp:phases}): the warm-up learns captions of 2\,s windows from up to \preKmax\ past frames, over two
epochs of \spPreWindows\ windows, and the SFT then learns captions of 16\,s windows from 0 to 16. The SFT of \model
takes \sftTenHours\,h on one \spSftGpu, \sftEpochs\ epochs in batches of \sftBatch\ windows; its kept step,
\sftBestTen, has the lowest mean validation loss over the lookbacks on \spSftVal\ windows, checked every
\spSftEvalEvery\ steps. RL trains on one \spRlTrainerGpu, while Qwen3.5-9B and phi-4 take turns on one \spRlJudgeGpu.
Its tripwire fired at step \rlTripTen\ and the exit rule stopped it after \rlTenHours\,h; step \rlBestTen, the best on
the selection windows, is kept, and the 0.1$\times$ run fired and kept at the same steps. The 0.1$\times$ runs used
one \spOneSftGpu\ for SFT (\spOneSftJobHours\,h with frame staging) and \spOneRlGpus\ \spOneRlGpu\ GPUs for RL
(\spOneRlJobHours\,h).

\section{Evaluation protocol}
\label{sec:supp:eval}

\subsection{Benchmark parts}
\label{sec:supp:bench}
\input{floats/tab_benchsubsets}
\Cref{tab:benchsubsets} gives the parts of OVO-Bench~\cite{ovobench2025}, StreamingBench~\cite{streamingbench2024},
Video-MME~\cite{videomme2024} and MVBench~\cite{mvbench2024} that we keep, and why. For each question, the answerer
reads one caption: the last one, written at the question's last second $t$, in the answerer prompt of
\cref{sec:supp:allprompts}. The accuracy per task is in \cref{sec:supp:tasks}.

\subsection{Prompts for baseline captioners}
\label{sec:supp:prompts}
Every off-the-shelf baseline (SmolVLM2~\cite{smolvlm2025}, InternVL3.5~\cite{internvl35}, Qwen3-VL~\cite{qwen3vl2025},
Qwen2.5-VL~\cite{qwen25vl2025}, LFM2.5-VL~\cite{lfm2}, LiveCC~\cite{livecc2025}, FastVLM~\cite{fastvlm2025},
Mobile-VideoGPT~\cite{mobilevideogpt2025}, StreamingVLM~\cite{streamingvlm2025} and
TimeChat-Online~\cite{timechatonline2025}) gets the same three candidate wordings, w0 to w2, each with a single-frame
and a multiple-frame version. All request 30--50 words; the multiple-frame versions state the 1\,FPS sampling and
oldest-first order.

\paragraph{Selection and freeze.}
The \promptPilotWindows\ pilot windows are LLaVA-Video windows without a teacher ceiling (\cref{sec:supp:sets}).
Phi-4 scores their raw fact-F1 under the 60-word rule, and a tie goes to the lowest index
(\cref{tab:supp:promptselect}). Quantized variants inherit the full-precision model's wording, and the teacher uses
its labelling prompt (\cref{sec:supp:allprompts}). For SmolVLM2-2.2B, \cref{tab:main} uses the plain-description
wording (w0) on OVO-Bench and MVBench, \sbSelSmolBigOffShare\% of its benchmark captions, rather than its selected
concrete-facts wording (w2). With w2 throughout, its four-benchmark mean is \sbProtoSmolBigMean\% in the separate
answer call of \cref{tab:supp:kzero}, against \sbTableOneSmolBigMean\% in \cref{tab:main}.

Each box below gives a wording for $L{=}0$, then, after a blank line, for $L{>}0$, where the placeholder \{N\} is the
number of input frames, $L{+}1$.

\par\medskip\vbox{\parindent=0pt
{\raggedright\textbf{w0: plain description.}\par}
\nolinenumbers
\begin{lstlisting}[style=spprompt]
This image is the current frame of a video. Describe in 30 to 50 words what is happening.

These {N} frames are from a video, sampled at 1 frame per second over the last {N} seconds, oldest first; the last frame is the current moment. Describe in 30 to 50 words what happens.
\end{lstlisting}
\linenumbers}\par

\par\medskip\vbox{\parindent=0pt
{\raggedright\textbf{w1: live narration.}\par}
\nolinenumbers
\begin{lstlisting}[style=spprompt]
You are describing a live video once per second. This image is its current frame. Write one caption of 30 to 50 words saying what is happening now. Output only the caption.

You are describing a live video once per second. These {N} frames are the video at 1 frame per second over the last {N} seconds, oldest first; the final frame is the present. Write one caption of 30 to 50 words saying what is happening now and what led to it. Output only the caption.
\end{lstlisting}
\linenumbers}\par

\par\medskip\vbox{\parindent=0pt
{\raggedright\textbf{w2: concrete facts.}\par}
\nolinenumbers
\begin{lstlisting}[style=spprompt]
This image is one frame of a video. In 30 to 50 words, state only concrete facts: who or what is in view, the setting and what they are doing. No speculation, no preamble.

These {N} frames are consecutive frames at 1 frame per second covering the last {N} seconds of a video, oldest first. In 30 to 50 words, state only concrete facts: who or what is in view, the setting, what they do and how the scene changes from the first frame to the last. No speculation, no preamble.
\end{lstlisting}
\linenumbers}\par

\input{floats/tab_supp_promptselect}
\paragraph{The prompt of \model.}
\model reads one fixed prompt in training and at inference. Its text never changes: the slot of the previous caption always
reads ``none'' (\cref{sec:supp:arch}). The $L$ pooled blocks follow, oldest first, then the 64 tokens of the present; the
listing writes $L$ as K.\par\medskip
\keeptogether{\spPromptStudent}

\paragraph{The three newest baselines.} The protocol is the same for FastVLM-0.5B, StreamingVLM-7B and TimeChat-Online-7B:
the $L{+}1$ frames in one request, the selected wording, greedy decoding and at most 80 new tokens. The two streaming
models therefore do not run in their own interaction format. TimeChat-Online-7B keeps its token drop, which removes every
visual token whose cosine similarity to the token at the same position one frame pair earlier exceeds 0.5; it keeps
\nbTimeChatKept\% of its visual tokens at the protocol $L$, and all of them at $L{=}0$, where no earlier pair exists.
StreamingVLM-7B, trained to comment on a stream turn by turn, answers such a request with its silent token. We therefore
ban its silent tokens, require at least one token and apply its own repetition penalty. Its captions then start as a
description and repeat themselves until the token limit (\nbWordsStreamingVlm\ words on average on the benchmarks); on
the held-out sets, the 60-word rule scores most of them 0 (\cref{tab:main}). Its numbers measure this adaptation, not its own
streaming commentary.

\subsection{Held-out captions and a second judge}
\label{sec:supp:judges}
\input{floats/tab_judges}
Since OpenVid is a training source of \model, the benchmarks and LLaVA-Video test carry the comparison with the
baselines. \Cref{tab:main} drops the 60-word rule but keeps the same captions and teacher ceiling.
Mobile-VideoGPT-0.5B answers in one short sentence whatever the wording. Under Qwen3.5-9B, \model also gains
\devQTenGain\ [\devQTenGainLo, \devQTenGainHi] points on dev from $L{=}0$ to 16 and stays level with the 0.1$\times$
model at $L{=}16$ on dev and test.

\subsection{Evaluation sets and fact-F1}
\label{sec:supp:sets}
\paragraph{Evaluation sets.}
Fact-F1 uses three sets of windows, disjoint by source video from each other and from the training data. The LLaVA-Video
dev split has \pilotWindows\ windows of \pilotVideos\ videos, LLaVA-Video test \tWindows\ windows of \tVideos\ videos, and
held-out OpenVid \ovWindows\ windows of \ovVideos\ clips from \spOpenvidSourceVideos\ videos. These are the windows with a
ceiling. A window has none when the derivation of its claims did not complete, when its consensus is empty or when none of
its questions survived; this leaves out \spPilotNoCeiling, \spTestNoCeiling\ and \spOpenvidNoCeiling\ windows with 16 stored
past seconds. Dev and test windows are consecutive seconds of the same episodes, a median of \spPilotPerVideo\ windows per
video. \spRlValInDev\ of the \rlVal\ windows that select the RL checkpoint (\cref{sec:supp:reward}) are dev windows, the
other \spRlValOutDev\ have no ceiling, and the remaining \spPilotAudit\ dev windows select nothing. Dev therefore serves curves, ablations and controls, while the comparisons
with the baselines use the held-out sets and the benchmarks. The \promptPilotWindows\ prompt-pilot windows of
\cref{sec:supp:prompts} come from dev videos but have no ceiling, so they are in none of these sets.

\paragraph{Fact-F1.}
The score of one caption on one window comes first. The judge splits the caption into atomic facts and checks them
against the teacher's claims of the window, with the prompts of \cref{sec:supp:allprompts}: recall is the share of the
consensus claims that the caption states or clearly implies, precision the share of the caption's facts that the
union states or clearly implies, and raw fact-F1 their harmonic mean, from 0 to 1. The target is the consensus of all
kept teacher samples and the union of every teacher caption of the window. The original scoring protocol then gives 0
to a caption over 60 words, the 60-word rule: \cref{tab:main} reports held-out scores without it and gives the rule's
scores in parentheses where they differ, while the dev results, as in \cref{sec:ablations}, and
\cref{tab:supp:promptselect,tab:judges} keep it. Only checkpoint selection multiplies raw fact-F1 by the gate $g$ of
\cref{sec:supp:reward}, as gated fact-F1, and only \cref{tab:supp:promptselect} reports raw fact-F1, since the pilot
windows have no ceiling. Everywhere else, fact-F1 is a share of the teacher ceiling: a model's raw fact-F1 summed over
the windows of a set, divided by the ceiling's sum over the same windows. The ceiling of a window is the raw fact-F1 of
one teacher caption, the first of the window's samples at temperature \spRlTemperature, against a target built without
it: the claims of at least \spCeilingNeed\ of the other \spCeilingOthers\ kept samples (the same fraction, rounded up,
when fewer are kept) and the union of every other caption, the greedy one included. Every judging run scores the
ceiling again. The ceiling is therefore the teacher's agreement with itself, not an upper bound: 100\% means agreeing
with the target as well as one teacher sample agrees with the others, not recovering every visible fact. The teacher's
greedy caption meets a union that holds the claims of its own stored label, so it exceeds the ceiling, as on dev at
$L{=}16$ (\cref{fig:dial}a).

\input{floats/fig_dial}
\input{floats/tab_supp_kzero}
\input{floats/tab_lookback}
\input{floats/fig_examples}
\input{floats/tab_supp_tasksall}
\section{Further results}
\label{sec:supp:results}

\subsection{More frames}
\label{sec:supp:frames}
Each past frame adds fact-F1 for \model at almost no prefill time, since \model reads cached blocks while the baselines
re-encode every frame (\cref{fig:dial}). On the benchmarks, \model gains from past frames where its base model loses
(\cref{tab:supp:kzero}).

\subsection{More evidence or more compute}
\label{sec:supp:controls}
Repeating the current frame keeps the prompt and its length but removes the gain from history, and another video's
past falls far below no past at all, so the gain comes from what the past frames show (\cref{tab:lookback}).
Shuffling the real past costs much less; fact-F1 ignores the order of events, so this control does not test it. A
consensus of 16 samples~\cite{selfconsistency2023}, at 16 times the decoding, recovers at most \cyMbrOfGain\% of the
real past's gain. Blending cached seconds into extra blocks at test time gains nothing: the cache is not read at a
finer grain than it was trained on.

\paragraph{Earlier tests at 2\,s.}
\label{sec:supp:loops}
More compute on the same 2\,s did not help our earlier 2\,s captioners, but more past did (\twoSecWindows\ validation
windows). Looping~\cite{universaltransformer2019,recurrentdepth2025} layers 4--11 of LFM2 up to eight times scores
\depthMin--\depthMax\% against \depthControl\% for one pass, and carrying a latent state of 32 slots across seconds
reaches at best \latentBest\% against \latentBase\%, while reading more past rises from \twoSecRLOne\% at $L{=}1$ to
\twoSecRLFive\% at $L{=}5$.

\subsection{Accuracy per task}
\label{sec:supp:tasks}
\Cref{tab:supp:tasksall} gives the accuracy per task of every model of \cref{tab:main}.

\subsection{Caption examples}
\label{sec:supp:examples}
\Cref{fig:examples} shows what the past adds to the captions of \model, its SFT model and two baselines, on two held-out
OpenVid windows where the view changes.

\section{Prompts of the teacher, the judges and the answerer}
\label{sec:supp:allprompts}
The prompts below are those of the code that produced every number of the paper, placeholders in braces. The teacher
receives the frames of the window with its prompt. Qwen3.5-9B runs without thinking. The judges and the answerer decode
greedily, constrained to a JSON schema or to a fixed set of choices.
\promptbox{\spPromptTitleTeacher}{\spPromptTeacher}
\promptbox{\spPromptTitleFacts}{\spPromptFacts}
\promptbox{\spPromptTitleCluster}{\spPromptCluster}
\promptbox{\spPromptTitleQuestion}{\spPromptQuestion}
\promptbox{\spPromptTitleAnswer}{\spPromptAnswer}
\promptbox{\spPromptTitleEntail}{\spPromptEntail}
\promptbox{\spPromptTitleSupport}{\spPromptSupport}
\promptbox{\spPromptTitleRate}{\spPromptRate}
\promptbox{\spPromptTitleAnswerer}{\spPromptAnswerer}
\promptbox{\spPromptTitleAnswererFloor}{\spPromptAnswererFloor}
\FloatBarrier

%% file: floats/tab_runtime.tex
\begin{table}[t]
  \centering
  \caption{\textbf{The TensorRT engine holds p99 time per frame at every $L$ and pays for $L$ in caption lag.} FP16 on a T4, batch 1, worst clip, frames back to back; peak memory \eyGB\,GB. Prompt: tokens and prefill frames; lag: mean frames from the second's first frame to the finished caption, at 30\,FPS; stage sum: GPU busy time per caption; exact: share of captions equal to the float32 reference on 64 windows. Bottom: every model writes 45-token captions in the same engine, end token ignored; steps: decode steps per frame; p99: range over sessions; lag (seconds into the second) and GPU time per caption: the largest; --: no caption finishes within its second.}
  \label{tab:runtime}
  \scriptsize
  \setlength{\tabcolsep}{4pt}
  \input{tables/runtime}
  \par\medskip
  \begin{tabular}{@{}lrrrrrr@{}}
  \toprule
   & & & \multicolumn{2}{c}{$L{=}8$} & \multicolumn{2}{c}{$L{=}16$} \\
  \cmidrule(lr){4-5}\cmidrule(l){6-7}
  Model & steps & p99 (ms) & lag (s) & GPU (s) & lag (s) & GPU (s) \\
  \midrule
  \model & \efYoloSteps & \efYoloPNinetyNineMin--\efYoloPNinetyNineMax & \efYoloLagKeightSecMax & \efYoloGpuKeightMax & \efYoloLagKsixteenSecMax & \efYoloGpuKsixteenMax \\
  LFM2.5-VL-450M & \efLfmSteps & \efLfmPNinetyNineMin--\efLfmPNinetyNineMax & \efLfmLagKeightSecMax & \efLfmGpuKeightMax & \efLfmLagKsixteenSecMax & \efLfmGpuKsixteenMax \\
  LFM2.5-VL-450M & \efLfmTwoSteps & \efLfmTwoPNinetyNineMin--\efLfmTwoPNinetyNineMax & \efLfmTwoLagKeightSecMax & \efLfmTwoGpuKeightMax & -- & -- \\
  SmolVLM2-500M & \efSmolSteps & \efSmolPNinetyNineMin--\efSmolPNinetyNineMax & -- & -- & -- & -- \\
  \bottomrule
  \end{tabular}
\end{table}

%% file: tables/runtime.tex
\begin{tabular}{@{}rrrrrrr@{}}
\toprule
$L$ & prompt & chunks & p99 (ms) & lag (frames) & stage sum (s) & exact (\%) \\
\midrule
0 & 79 & 1 & 12.66 & 12.6 & 0.157 & 84.4 \\
1 & 97 & 1 & 12.66 & 12.1 & 0.153 & 85.9 \\
2 & 115 & 1 & 12.69 & 12.3 & 0.154 & 85.9 \\
4 & 151 & 2 & 12.67 & 13.2 & 0.161 & 89.1 \\
8 & 223 & 2 & 12.69 & 13.7 & 0.170 & 84.4 \\
16 & 367 & 4 & 12.67 & 15.8 & 0.197 & 93.8 \\
\bottomrule
\end{tabular}

%% file: floats/tab_recipe.tex
\begin{table*}[t]
  \centering
  \caption{\textbf{Training ablations on the LLaVA-Video dev split.} Fact-F1 at four lookbacks and the change against the reference row: GRPO rows against the SFT model of the same data, full SFT against 0.1$\times$ SFT, SFT ablations against the full SFT recipe (one seed each), student variants against the LoRA SFT. Each block is one judging run with its own reference row, except the student variants. \model is level with the 0.1$\times$ GRPO model judged in the same run: \devTenOverEsixKzero\ [\devTenOverEsixKzeroLo, \devTenOverEsixKzeroHi] at $L{=}0$ and \devTenOverEsixKsixteen\ [\devTenOverEsixKsixteenLo, \devTenOverEsixKsixteenHi] at $L{=}16$. All weights beat LoRA adapters~\cite{lora2022}, which sit on the language model and the projector with the vision encoder frozen. The 1.6B student, LFM2.5-VL-1.6B with the LoRA recipe, scores far higher, but distilling its token distribution~\cite{kd2015} into the 450M model, on the teacher's caption (off-policy) or on the student's own samples (on-policy~\cite{gkd2024}), changes nothing measurable.}
  \label{tab:recipe}
  \scriptsize
  \setlength{\tabcolsep}{3pt}
  \input{tables/recipe}
\end{table*}

%% file: tables/recipe.tex
\begin{tabular}{@{}lrrrrcc@{}}
\toprule
 & $L{=}0$ & $L{=}2$ & $L{=}4$ & $L{=}16$ & change at $L{=}0$ & change at $L{=}16$ \\
\midrule
0.1$\times$ data, SFT & 47.2 & 52.7 & 55.2 & 65.3 & -- & -- \\
0.1$\times$ data, SFT + GRPO + KL & 52.3 & 57.7 & 60.6 & 71.9 & +5.2 [+3.8, +6.5] & +6.6 [+5.1, +8.0] \\
0.1$\times$ data, SFT + GRPO, no KL & 57.5 & 63.9 & 66.3 & 75.6 & +10.3 [+8.8, +11.8] & +10.3 [+8.6, +12.0] \\
\midrule
\multicolumn{7}{@{}l}{\emph{Full data (one judging run, the 0.1$\times$ rows judged again)}} \\
0.1$\times$ data, SFT & \devSftKzero & \devSftKtwo & \devSftKfour & \devSftKsixteen & -- & -- \\
full data, SFT & \devOvSftKzero & \devOvSftKtwo & \devOvSftKfour & \devOvSftKsixteen & \devOvSftOverSftKzero\ [\devOvSftOverSftKzeroLo, \devOvSftOverSftKzeroHi] & \devOvSftOverSftKsixteen\ [\devOvSftOverSftKsixteenLo, \devOvSftOverSftKsixteenHi] \\
0.1$\times$ data, SFT + GRPO + KL & \devEsixKzero & \devEsixKtwo & \devEsixKfour & \devEsixKsixteen & \devEsixOverSftKzero\ [\devEsixOverSftKzeroLo, \devEsixOverSftKzeroHi] & \devEsixOverSftKsixteen\ [\devEsixOverSftKsixteenLo, \devEsixOverSftKsixteenHi] \\
\textbf{\model}: full data, SFT + GRPO + KL & \devTenKzero & \devTenKtwo & \devTenKfour & \devTenKsixteen & \devTenOverOvSftKzero\ [\devTenOverOvSftKzeroLo, \devTenOverOvSftKzeroHi] & \devTenOverOvSftKsixteen\ [\devTenOverOvSftKsixteenLo, \devTenOverOvSftKsixteenHi] \\
\midrule
\multicolumn{7}{@{}l}{\emph{SFT ablations (one seed each; one judging run, changes against the full recipe judged again)}} \\
full recipe (2\,s pre-stage, sandwich over $L$) & 47.3 & 52.1 & 54.1 & 65.2 & -- & -- \\
no 2\,s pre-stage & 47.1 & 52.4 & 54.7 & 63.7 & $-$0.2 [$-$1.3, +1.0] & $-$1.6 [$-$2.9, $-$0.2] \\
no pre-stage, $L{=}16$ only & 44.4 & 51.3 & 55.0 & 66.0 & $-$2.9 [$-$4.3, $-$1.5] & +0.8 [$-$0.4, +2.1] \\
no pre-stage, one random $L$ & 46.9 & 52.1 & 54.4 & 63.9 & $-$0.4 [$-$1.4, +0.7] & $-$1.3 [$-$2.6, 0.0] \\
\midrule
\multicolumn{7}{@{}l}{\emph{Student variants on the 0.1$\times$ data (each row judged in its own run; changes against the LoRA SFT, the last row against the first row)}} \\
LoRA adapters, SFT & \stTuneLoraSftKzero & \stTuneLoraSftKtwo & \stTuneLoraSftKfour & \stTuneLoraSftKsixteen & -- & -- \\
all weights, SFT & \multicolumn{4}{c}{\emph{the first row}} & \stTuneFullSftOverLoraSftKzero\ [\stTuneFullSftOverLoraSftKzeroLo, \stTuneFullSftOverLoraSftKzeroHi] & \stTuneFullSftOverLoraSftKsixteen\ [\stTuneFullSftOverLoraSftKsixteenLo, \stTuneFullSftOverLoraSftKsixteenHi] \\
LoRA adapters, SFT + GRPO & \stTuneLoraGrpoKzero & \stTuneLoraGrpoKtwo & \stTuneLoraGrpoKfour & \stTuneLoraGrpoKsixteen & \stTuneLoraGrpoOverLoraSftKzero\ [\stTuneLoraGrpoOverLoraSftKzeroLo, \stTuneLoraGrpoOverLoraSftKzeroHi] & \stTuneLoraGrpoOverLoraSftKsixteen\ [\stTuneLoraGrpoOverLoraSftKsixteenLo, \stTuneLoraGrpoOverLoraSftKsixteenHi] \\
1.6B student, LoRA SFT & \stTuneBigKzero & \stTuneBigKtwo & \stTuneBigKfour & \stTuneBigKsixteen & \stTuneBigOverLoraSftKzero\ [\stTuneBigOverLoraSftKzeroLo, \stTuneBigOverLoraSftKzeroHi] & \stTuneBigOverLoraSftKsixteen\ [\stTuneBigOverLoraSftKsixteenLo, \stTuneBigOverLoraSftKsixteenHi] \\
450M, distilled from the 1.6B (off-policy), LoRA & \stTuneKdLoraKzero & \stTuneKdLoraKtwo & \stTuneKdLoraKfour & \stTuneKdLoraKsixteen & \stTuneKdLoraOverLoraSftKzero\ [\stTuneKdLoraOverLoraSftKzeroLo, \stTuneKdLoraOverLoraSftKzeroHi] & \stTuneKdLoraOverLoraSftKsixteen\ [\stTuneKdLoraOverLoraSftKsixteenLo, \stTuneKdLoraOverLoraSftKsixteenHi] \\
450M, distilled from the 1.6B (on-policy), LoRA & \stTuneOpLoraKzero & \stTuneOpLoraKtwo & \stTuneOpLoraKfour & \stTuneOpLoraKsixteen & \stTuneOpLoraOverLoraSftKzero\ [\stTuneOpLoraOverLoraSftKzeroLo, \stTuneOpLoraOverLoraSftKzeroHi] & \stTuneOpLoraOverLoraSftKsixteen\ [\stTuneOpLoraOverLoraSftKsixteenLo, \stTuneOpLoraOverLoraSftKsixteenHi] \\
450M, distilled from the 1.6B (off-policy), all weights & \stTuneKdFullKzero & \stTuneKdFullKtwo & \stTuneKdFullKfour & \stTuneKdFullKsixteen & \stTuneKdFullOverFullSftKzero\ [\stTuneKdFullOverFullSftKzeroLo, \stTuneKdFullOverFullSftKzeroHi] & \stTuneKdFullOverFullSftKsixteen\ [\stTuneKdFullOverFullSftKsixteenLo, \stTuneKdFullOverFullSftKsixteenHi] \\
\bottomrule
\end{tabular}

%% file: floats/tab_supp_pool.tex
\begin{table}[t]
  \centering
  \caption{\textbf{At equal past tokens, more seconds beat more tokens per second.} SFT on the 0.1$\times$ data with every past frame pooled to the tokens of its row, dev split, one judging run, at the lookbacks each arm was measured at. Frames $\times$ tokens: the largest $L$ an arm was trained for and its tokens per past frame; past tokens: their product. The $16\times16$ arm is the 0.1$\times$ SFT model of \cref{tab:recipe}, judged again. The paired comparisons of \cref{sec:ablations} take, at 256 past tokens, the $16\times16$ arm at $L{=}16$ against the $4\times64$ arm at $L{=}4$, and at 64, the $16\times4$ arm at $L{=}16$ against the $16\times16$ arm at $L{=}4$.}
  \label{tab:supp:pool}
  \scriptsize
  \setlength{\tabcolsep}{3pt}
  \input{tables/supp_training_pool}
\end{table}

%% file: tables/supp_training_pool.tex
\begin{tabular}{@{}lrrrrrrr@{}}
\toprule
frames $\times$ tokens & past tokens & $L{=}0$ & $L{=}1$ & $L{=}2$ & $L{=}4$ & $L{=}8$ & $L{=}16$ \\
\midrule
$\stPoolSixtyFourFrames\times\stPoolSixtyFourTokens$ & \stPoolSixtyFourPast & \stPoolSixtyFourKzero & \stPoolSixtyFourKone & \stPoolSixtyFourKtwo & \stPoolSixtyFourKfour & -- & -- \\
$\stPoolSixteenFrames\times\stPoolSixteenTokens$ & \stPoolSixteenPast & \stPoolSixteenKzero & \stPoolSixteenKone & \stPoolSixteenKtwo & \stPoolSixteenKfour & \stPoolSixteenKeight & \stPoolSixteenKsixteen \\
$\stPoolFourFrames\times\stPoolFourTokens$ & \stPoolFourPast & \stPoolFourKzero & \stPoolFourKone & \stPoolFourKtwo & \stPoolFourKfour & \stPoolFourKeight & \stPoolFourKsixteen \\
\bottomrule
\end{tabular}

%% file: floats/tab_supp_kl.tex
\begin{table*}[t]
  \centering
  \caption{\textbf{The KL penalty gives up some fact-F1 to keep recall and to remove the closing formula.} GRPO on the 0.1$\times$ data with and without the KL penalty, against its SFT model; their dev fact-F1 is in \cref{tab:recipe}. Held-out: OpenVid and LLaVA-Video test. Dev: recall, precision and words over the \stDevScored\ dev windows scored. Captions: shares of the \stStyleWindows\ test captions at $L{=}16$ that end with the closer ``The scene captures \ldots'', counted by its verb ``captures'', or that hedge (``appears to'', ``possibly'', ``something'').}
  \label{tab:supp:kl}
  \scriptsize
  \setlength{\tabcolsep}{3pt}
  \input{tables/supp_training_kl}
\end{table*}

%% file: tables/supp_training_kl.tex
\begin{tabular}{@{}lrrrcc@{}}
\toprule
 & SFT & SFT + GRPO + KL & SFT + GRPO, no KL & GRPO + KL minus SFT & GRPO + KL minus no KL \\
\midrule
\multicolumn{6}{@{}l}{\emph{Held-out fact-F1 (\% of the teacher ceiling), one judging run}} \\
OpenVid, $L{=}0$ & \stVarSftOKzero & \stVarKlOKzero & \stVarNoKlOKzero & \stVarKlOverSftOKzero\ [\stVarKlOverSftOKzeroLo, \stVarKlOverSftOKzeroHi] & \stVarKlOverNoKlOKzero\ [\stVarKlOverNoKlOKzeroLo, \stVarKlOverNoKlOKzeroHi] \\
OpenVid, $L{=}16$ & \stVarSftOKsixteen & \stVarKlOKsixteen & \stVarNoKlOKsixteen & \stVarKlOverSftOKsixteen\ [\stVarKlOverSftOKsixteenLo, \stVarKlOverSftOKsixteenHi] & \stVarKlOverNoKlOKsixteen\ [\stVarKlOverNoKlOKsixteenLo, \stVarKlOverNoKlOKsixteenHi] \\
LLaVA-Video test, $L{=}0$ & \stVarSftTKzero & \stVarKlTKzero & \stVarNoKlTKzero & \stVarKlOverSftTKzero\ [\stVarKlOverSftTKzeroLo, \stVarKlOverSftTKzeroHi] & \stVarKlOverNoKlTKzero\ [\stVarKlOverNoKlTKzeroLo, \stVarKlOverNoKlTKzeroHi] \\
LLaVA-Video test, $L{=}16$ & \stVarSftTKsixteen & \stVarKlTKsixteen & \stVarNoKlTKsixteen & \stVarKlOverSftTKsixteen\ [\stVarKlOverSftTKsixteenLo, \stVarKlOverSftTKsixteenHi] & \stVarKlOverNoKlTKsixteen\ [\stVarKlOverNoKlTKsixteenLo, \stVarKlOverNoKlTKsixteenHi] \\
\midrule
\multicolumn{6}{@{}l}{\emph{Dev split at $L{=}16$, means over the dev windows scored}} \\
recall & \stKlRecallSft & \stKlRecallKl & \stKlRecallNoKl & -- & -- \\
precision & \stKlPrecisionSft & \stKlPrecisionKl & \stKlPrecisionNoKl & -- & -- \\
words & \stKlWordsSft & \stKlWordsKl & \stKlWordsNoKl & -- & -- \\
\midrule
\multicolumn{6}{@{}l}{\emph{LLaVA-Video test captions at $L{=}16$ (share in \%, points for the difference)}} \\
with a fact-free closer & \stKlCloserSft & \stKlCloserKl & \stKlCloserNoKl & -- & \stKlOverNoKlCloser\ [\stKlOverNoKlCloserLo, \stKlOverNoKlCloserHi] \\
with a hedge & \stKlHedgeSft & \stKlHedgeKl & \stKlHedgeNoKl & -- & \stKlOverNoKlHedge\ [\stKlOverNoKlHedgeLo, \stKlOverNoKlHedgeHi] \\
words & \stKlTestWordsSft & \stKlTestWordsKl & \stKlTestWordsNoKl & -- & \stKlOverNoKlTestWords\ [\stKlOverNoKlTestWordsLo, \stKlOverNoKlTestWordsHi] \\
\bottomrule
\end{tabular}

%% file: floats/tab_supp_bench.tex
\begin{table*}[t]
  \centering
  \caption{\textbf{The full data helps on the benchmarks only with longer training, and on held-out captions only on OpenVid, one of its sources.} Top: caption-then-answer accuracy (\%) at $L{=}\min(t,16)$, answerer and questions of \cref{tab:main}; each block is one answer call, and the 0.1$\times$ SFT model is in both. Full data: SFT run to its end or stopped at the step count of the 0.1$\times$ run (\cref{sec:supp:schedule}). Mean: unweighted, over the four benchmarks. Bottom: held-out OpenVid and LLaVA-Video test (\cref{sec:supp:sets}); the 0.1$\times$ SFT + GRPO + KL model is judged again in \cref{tab:supp:kl}, and \cref{tab:judges} gives the gain of \model over its SFT model.}
  \label{tab:supp:bench}
  \scriptsize
  \setlength{\tabcolsep}{3pt}
  \input{tables/supp_training_bench}
  \par\medskip
  \setlength{\tabcolsep}{4pt}
  \input{tables/supp_training_heldout}
\end{table*}

%% file: tables/supp_training_bench.tex
\begin{tabular}{@{}lrrrrr@{}}
\toprule
 & OVO-RT & MVBench & StrBench & VMME & mean \\
\midrule
questions & \stBenchQuestionsOvo & \stBenchQuestionsMvb & \stBenchQuestionsSb & \stBenchQuestionsVm & -- \\
\midrule
\multicolumn{6}{@{}l}{\emph{Answer call 1 (the SFT models): accuracy (\%)}} \\
0.1$\times$ data, SFT & \stBenchOneSftOvo & \stBenchOneSftMvb & \stBenchOneSftSb & \stBenchOneSftVm & \stBenchOneSftMean \\
full data, SFT, 618 steps & \stBenchOneOvEqOvo & \stBenchOneOvEqMvb & \stBenchOneOvEqSb & \stBenchOneOvEqVm & \stBenchOneOvEqMean \\
full data, SFT, 7,354 steps & \stBenchOneOvSftOvo & \stBenchOneOvSftMvb & \stBenchOneOvSftSb & \stBenchOneOvSftVm & \stBenchOneOvSftMean \\
\multicolumn{6}{@{}l}{\emph{Answer call 1: paired difference [95\% interval] over questions}} \\
full SFT minus 0.1$\times$ SFT & \stBenchOneOvSftOverSftOvo\ [\stBenchOneOvSftOverSftOvoLo, \stBenchOneOvSftOverSftOvoHi] & \stBenchOneOvSftOverSftMvb\ [\stBenchOneOvSftOverSftMvbLo, \stBenchOneOvSftOverSftMvbHi] & \stBenchOneOvSftOverSftSb\ [\stBenchOneOvSftOverSftSbLo, \stBenchOneOvSftOverSftSbHi] & \stBenchOneOvSftOverSftVm\ [\stBenchOneOvSftOverSftVmLo, \stBenchOneOvSftOverSftVmHi] & \stBenchOneOvSftOverSftMean\ [\stBenchOneOvSftOverSftMeanLo, \stBenchOneOvSftOverSftMeanHi] \\
full SFT at 618 steps minus 0.1$\times$ SFT & \stBenchOneOvEqOverSftOvo\ [\stBenchOneOvEqOverSftOvoLo, \stBenchOneOvEqOverSftOvoHi] & \stBenchOneOvEqOverSftMvb\ [\stBenchOneOvEqOverSftMvbLo, \stBenchOneOvEqOverSftMvbHi] & \stBenchOneOvEqOverSftSb\ [\stBenchOneOvEqOverSftSbLo, \stBenchOneOvEqOverSftSbHi] & \stBenchOneOvEqOverSftVm\ [\stBenchOneOvEqOverSftVmLo, \stBenchOneOvEqOverSftVmHi] & \stBenchOneOvEqOverSftMean\ [\stBenchOneOvEqOverSftMeanLo, \stBenchOneOvEqOverSftMeanHi] \\
full SFT minus full SFT at 618 steps & \stBenchOneOvSftOverOvEqOvo\ [\stBenchOneOvSftOverOvEqOvoLo, \stBenchOneOvSftOverOvEqOvoHi] & \stBenchOneOvSftOverOvEqMvb\ [\stBenchOneOvSftOverOvEqMvbLo, \stBenchOneOvSftOverOvEqMvbHi] & \stBenchOneOvSftOverOvEqSb\ [\stBenchOneOvSftOverOvEqSbLo, \stBenchOneOvSftOverOvEqSbHi] & \stBenchOneOvSftOverOvEqVm\ [\stBenchOneOvSftOverOvEqVmLo, \stBenchOneOvSftOverOvEqVmHi] & \stBenchOneOvSftOverOvEqMean\ [\stBenchOneOvSftOverOvEqMeanLo, \stBenchOneOvSftOverOvEqMeanHi] \\
\midrule
\multicolumn{6}{@{}l}{\emph{Answer call 2 (the headline models): accuracy (\%)}} \\
0.1$\times$ data, SFT & \stBenchTwoSftOvo & \stBenchTwoSftMvb & \stBenchTwoSftSb & \stBenchTwoSftVm & \stBenchTwoSftMean \\
full data, SFT & \stBenchTwoOvSftOvo & \stBenchTwoOvSftMvb & \stBenchTwoOvSftSb & \stBenchTwoOvSftVm & \stBenchTwoOvSftMean \\
0.1$\times$ data, SFT + GRPO + KL & \stBenchTwoKlOvo & \stBenchTwoKlMvb & \stBenchTwoKlSb & \stBenchTwoKlVm & \stBenchTwoKlMean \\
\textbf{\model}: full data, SFT + GRPO + KL & \stBenchTwoYoloOvo & \stBenchTwoYoloMvb & \stBenchTwoYoloSb & \stBenchTwoYoloVm & \stBenchTwoYoloMean \\
\multicolumn{6}{@{}l}{\emph{Answer call 2: paired difference [95\% interval] over questions}} \\
0.1$\times$ GRPO + KL minus 0.1$\times$ SFT & \stBenchTwoKlOverSftOvo\ [\stBenchTwoKlOverSftOvoLo, \stBenchTwoKlOverSftOvoHi] & \stBenchTwoKlOverSftMvb\ [\stBenchTwoKlOverSftMvbLo, \stBenchTwoKlOverSftMvbHi] & \stBenchTwoKlOverSftSb\ [\stBenchTwoKlOverSftSbLo, \stBenchTwoKlOverSftSbHi] & \stBenchTwoKlOverSftVm\ [\stBenchTwoKlOverSftVmLo, \stBenchTwoKlOverSftVmHi] & \stBenchTwoKlOverSftMean\ [\stBenchTwoKlOverSftMeanLo, \stBenchTwoKlOverSftMeanHi] \\
\model minus 0.1$\times$ GRPO + KL & \stBenchTwoYoloOverKlOvo\ [\stBenchTwoYoloOverKlOvoLo, \stBenchTwoYoloOverKlOvoHi] & \stBenchTwoYoloOverKlMvb\ [\stBenchTwoYoloOverKlMvbLo, \stBenchTwoYoloOverKlMvbHi] & \stBenchTwoYoloOverKlSb\ [\stBenchTwoYoloOverKlSbLo, \stBenchTwoYoloOverKlSbHi] & \stBenchTwoYoloOverKlVm\ [\stBenchTwoYoloOverKlVmLo, \stBenchTwoYoloOverKlVmHi] & \stBenchTwoYoloOverKlMean\ [\stBenchTwoYoloOverKlMeanLo, \stBenchTwoYoloOverKlMeanHi] \\
\model minus full SFT & \stBenchTwoYoloOverOvSftOvo\ [\stBenchTwoYoloOverOvSftOvoLo, \stBenchTwoYoloOverOvSftOvoHi] & \stBenchTwoYoloOverOvSftMvb\ [\stBenchTwoYoloOverOvSftMvbLo, \stBenchTwoYoloOverOvSftMvbHi] & \stBenchTwoYoloOverOvSftSb\ [\stBenchTwoYoloOverOvSftSbLo, \stBenchTwoYoloOverOvSftSbHi] & \stBenchTwoYoloOverOvSftVm\ [\stBenchTwoYoloOverOvSftVmLo, \stBenchTwoYoloOverOvSftVmHi] & \stBenchTwoYoloOverOvSftMean\ [\stBenchTwoYoloOverOvSftMeanLo, \stBenchTwoYoloOverOvSftMeanHi] \\
\model minus 0.1$\times$ SFT & \stBenchTwoYoloOverSftOvo\ [\stBenchTwoYoloOverSftOvoLo, \stBenchTwoYoloOverSftOvoHi] & \stBenchTwoYoloOverSftMvb\ [\stBenchTwoYoloOverSftMvbLo, \stBenchTwoYoloOverSftMvbHi] & \stBenchTwoYoloOverSftSb\ [\stBenchTwoYoloOverSftSbLo, \stBenchTwoYoloOverSftSbHi] & \stBenchTwoYoloOverSftVm\ [\stBenchTwoYoloOverSftVmLo, \stBenchTwoYoloOverSftVmHi] & \stBenchTwoYoloOverSftMean\ [\stBenchTwoYoloOverSftMeanLo, \stBenchTwoYoloOverSftMeanHi] \\
\bottomrule
\end{tabular}

%% file: tables/supp_training_heldout.tex
\begin{tabular}{@{}lrrrr@{}}
\toprule
 & \multicolumn{2}{c}{held-out OpenVid} & \multicolumn{2}{c}{LLaVA-Video test} \\
\cmidrule(lr){2-3}\cmidrule(l){4-5}
 & $L{=}0$ & $L{=}16$ & $L{=}0$ & $L{=}16$ \\
\midrule
\multicolumn{5}{@{}l}{\emph{Held-out fact-F1 (\% of the teacher ceiling), one judging run}} \\
full data, SFT & \stHeldOvSftOKzero & \stHeldOvSftOKsixteen & \stHeldOvSftTKzero & \stHeldOvSftTKsixteen \\
0.1$\times$ data, SFT + GRPO + KL & \stHeldKlOKzero & \stHeldKlOKsixteen & \stHeldKlTKzero & \stHeldKlTKsixteen \\
\textbf{\model}: full data, SFT + GRPO + KL & \stHeldYoloOKzero & \stHeldYoloOKsixteen & \stHeldYoloTKzero & \stHeldYoloTKsixteen \\
\multicolumn{5}{@{}l}{\emph{Paired difference [95\% interval]}} \\
\model minus 0.1$\times$ GRPO + KL & \stHeldYoloOverKlOKzero\ [\stHeldYoloOverKlOKzeroLo, \stHeldYoloOverKlOKzeroHi] & \stHeldYoloOverKlOKsixteen\ [\stHeldYoloOverKlOKsixteenLo, \stHeldYoloOverKlOKsixteenHi] & \stHeldYoloOverKlTKzero\ [\stHeldYoloOverKlTKzeroLo, \stHeldYoloOverKlTKzeroHi] & \stHeldYoloOverKlTKsixteen\ [\stHeldYoloOverKlTKsixteenLo, \stHeldYoloOverKlTKsixteenHi] \\
\bottomrule
\end{tabular}

%% file: floats/tab_benchsubsets.tex
\begin{table}[t]
  \centering
  \caption{\textbf{The benchmark parts that fit streaming captioning.} The parts kept and left out under the rules of \cref{sec:results}; we also keep only data that is openly downloadable.}
  \label{tab:benchsubsets}
  \scriptsize
  \setlength{\tabcolsep}{3pt}
  \input{tables/bench_subsets}
\end{table}

%% file: tables/bench_subsets.tex
\begin{tabular}{@{}p{0.21\linewidth}p{0.29\linewidth}p{0.42\linewidth}@{}}
\toprule
Benchmark & Used & Left out, and why \\
\midrule
OVO-Bench & Real-Time Visual Perception: 6 tasks, \ovoQuestions\ questions & Backward Tracing (631) needs memory far beyond 16\,s; Forward Active Responding (172) is about when to answer \\
StreamingBench & Real-Time Visual Understanding: 10 tasks, \nbQuestionsSb\ of 2,500 questions & Omni-Source and Contextual Understanding (8 tasks) need audio, dialogue history or deciding when to speak \\
Video-MME & Short videos ($<$2\,min), no subtitles: \nbQuestionsVm\ questions & Medium and long videos (1,800): their questions need memory far beyond 16\,s \\
MVBench & 19 of 20 tasks: \mvbQuestions\ questions & Fine-grained Pose (200): its videos are released through a request form, outside our open-data rule \\
\bottomrule
\end{tabular}

%% file: floats/tab_supp_promptselect.tex
\begin{table*}[t]
  \centering
  \caption{\textbf{Every baseline's wording is chosen on pilot windows, before any reported score.} Raw fact-F1 (0 to 1; phi-4, with the 60-word rule) of each baseline's captions of the pilot windows, in each of the three wordings of \cref{sec:supp:prompts}, at the lookbacks $L{=}0$ and 16, and its mean over the two; the chosen wording is in bold. Four decimals, because the mean decides: Mobile-VideoGPT-0.5B's two best wordings are \sbSelMarginMin\ apart.}
  \label{tab:supp:promptselect}
  \scriptsize
  \setlength{\tabcolsep}{4pt}
  \input{tables/supp_prompt_select}
\end{table*}

%% file: tables/supp_prompt_select.tex
\begin{tabular}{@{}lrrrrrrrrrr@{}}
\toprule
 & \multicolumn{3}{c}{w0: plain description} & \multicolumn{3}{c}{w1: live narration} & \multicolumn{3}{c}{w2: concrete facts} & \\
\cmidrule(lr){2-4}\cmidrule(lr){5-7}\cmidrule(lr){8-10}
Model & $L{=}0$ & $L{=}16$ & mean & $L{=}0$ & $L{=}16$ & mean & $L{=}0$ & $L{=}16$ & mean & chosen \\
\midrule
LFM2.5-VL-450M & \sbSelLfmSmallWzeroKzero & \sbSelLfmSmallWzeroKsixteen & \sbSelLfmSmallWzeroMean & \sbSelLfmSmallWoneKzero & \sbSelLfmSmallWoneKsixteen & \textbf{\sbSelLfmSmallWoneMean} & \sbSelLfmSmallWtwoKzero & \sbSelLfmSmallWtwoKsixteen & \sbSelLfmSmallWtwoMean & \textbf{\sbSelLfmSmallChosen} \\
SmolVLM2-500M & \sbSelSmolSmallWzeroKzero & \sbSelSmolSmallWzeroKsixteen & \textbf{\sbSelSmolSmallWzeroMean} & \sbSelSmolSmallWoneKzero & \sbSelSmolSmallWoneKsixteen & \sbSelSmolSmallWoneMean & \sbSelSmolSmallWtwoKzero & \sbSelSmolSmallWtwoKsixteen & \sbSelSmolSmallWtwoMean & \textbf{\sbSelSmolSmallChosen} \\
FastVLM-0.5B & \sbSelFastVlmWzeroKzero & \sbSelFastVlmWzeroKsixteen & \sbSelFastVlmWzeroMean & \sbSelFastVlmWoneKzero & \sbSelFastVlmWoneKsixteen & \textbf{\sbSelFastVlmWoneMean} & \sbSelFastVlmWtwoKzero & \sbSelFastVlmWtwoKsixteen & \sbSelFastVlmWtwoMean & \textbf{\sbSelFastVlmChosen} \\
Mobile-VideoGPT-0.5B & \sbSelMobileWzeroKzero & \sbSelMobileWzeroKsixteen & \sbSelMobileWzeroMean & \sbSelMobileWoneKzero & \sbSelMobileWoneKsixteen & \sbSelMobileWoneMean & \sbSelMobileWtwoKzero & \sbSelMobileWtwoKsixteen & \textbf{\sbSelMobileWtwoMean} & \textbf{\sbSelMobileChosen} \\
\midrule
InternVL3.5-1B & \sbSelInternOneWzeroKzero & \sbSelInternOneWzeroKsixteen & \textbf{\sbSelInternOneWzeroMean} & \sbSelInternOneWoneKzero & \sbSelInternOneWoneKsixteen & \sbSelInternOneWoneMean & \sbSelInternOneWtwoKzero & \sbSelInternOneWtwoKsixteen & \sbSelInternOneWtwoMean & \textbf{\sbSelInternOneChosen} \\
LFM2.5-VL-1.6B & \sbSelLfmBigWzeroKzero & \sbSelLfmBigWzeroKsixteen & \sbSelLfmBigWzeroMean & \sbSelLfmBigWoneKzero & \sbSelLfmBigWoneKsixteen & \textbf{\sbSelLfmBigWoneMean} & \sbSelLfmBigWtwoKzero & \sbSelLfmBigWtwoKsixteen & \sbSelLfmBigWtwoMean & \textbf{\sbSelLfmBigChosen} \\
Qwen3-VL-2B & \sbSelQwenTwoWzeroKzero & \sbSelQwenTwoWzeroKsixteen & \textbf{\sbSelQwenTwoWzeroMean} & \sbSelQwenTwoWoneKzero & \sbSelQwenTwoWoneKsixteen & \sbSelQwenTwoWoneMean & \sbSelQwenTwoWtwoKzero & \sbSelQwenTwoWtwoKsixteen & \sbSelQwenTwoWtwoMean & \textbf{\sbSelQwenTwoChosen} \\
SmolVLM2-2.2B & \sbSelSmolBigWzeroKzero & \sbSelSmolBigWzeroKsixteen & \sbSelSmolBigWzeroMean & \sbSelSmolBigWoneKzero & \sbSelSmolBigWoneKsixteen & \sbSelSmolBigWoneMean & \sbSelSmolBigWtwoKzero & \sbSelSmolBigWtwoKsixteen & \textbf{\sbSelSmolBigWtwoMean} & \textbf{\sbSelSmolBigChosen} \\
InternVL3.5-2B & \sbSelInternTwoWzeroKzero & \sbSelInternTwoWzeroKsixteen & \textbf{\sbSelInternTwoWzeroMean} & \sbSelInternTwoWoneKzero & \sbSelInternTwoWoneKsixteen & \sbSelInternTwoWoneMean & \sbSelInternTwoWtwoKzero & \sbSelInternTwoWtwoKsixteen & \sbSelInternTwoWtwoMean & \textbf{\sbSelInternTwoChosen} \\
\midrule
Qwen3-VL-4B & \sbSelQwenFourWzeroKzero & \sbSelQwenFourWzeroKsixteen & \textbf{\sbSelQwenFourWzeroMean} & \sbSelQwenFourWoneKzero & \sbSelQwenFourWoneKsixteen & \sbSelQwenFourWoneMean & \sbSelQwenFourWtwoKzero & \sbSelQwenFourWtwoKsixteen & \sbSelQwenFourWtwoMean & \textbf{\sbSelQwenFourChosen} \\
Qwen2.5-VL-7B & \sbSelQwenSevenTwoFiveWzeroKzero & \sbSelQwenSevenTwoFiveWzeroKsixteen & \textbf{\sbSelQwenSevenTwoFiveWzeroMean} & \sbSelQwenSevenTwoFiveWoneKzero & \sbSelQwenSevenTwoFiveWoneKsixteen & \sbSelQwenSevenTwoFiveWoneMean & \sbSelQwenSevenTwoFiveWtwoKzero & \sbSelQwenSevenTwoFiveWtwoKsixteen & \sbSelQwenSevenTwoFiveWtwoMean & \textbf{\sbSelQwenSevenTwoFiveChosen} \\
LiveCC-7B & \sbSelLiveCcWzeroKzero & \sbSelLiveCcWzeroKsixteen & \sbSelLiveCcWzeroMean & \sbSelLiveCcWoneKzero & \sbSelLiveCcWoneKsixteen & \textbf{\sbSelLiveCcWoneMean} & \sbSelLiveCcWtwoKzero & \sbSelLiveCcWtwoKsixteen & \sbSelLiveCcWtwoMean & \textbf{\sbSelLiveCcChosen} \\
StreamingVLM-7B & \sbSelStreamingVlmWzeroKzero & \sbSelStreamingVlmWzeroKsixteen & \textbf{\sbSelStreamingVlmWzeroMean} & \sbSelStreamingVlmWoneKzero & \sbSelStreamingVlmWoneKsixteen & \sbSelStreamingVlmWoneMean & \sbSelStreamingVlmWtwoKzero & \sbSelStreamingVlmWtwoKsixteen & \sbSelStreamingVlmWtwoMean & \textbf{\sbSelStreamingVlmChosen} \\
TimeChat-Online-7B & \sbSelTimeChatWzeroKzero & \sbSelTimeChatWzeroKsixteen & \textbf{\sbSelTimeChatWzeroMean} & \sbSelTimeChatWoneKzero & \sbSelTimeChatWoneKsixteen & \sbSelTimeChatWoneMean & \sbSelTimeChatWtwoKzero & \sbSelTimeChatWtwoKsixteen & \sbSelTimeChatWtwoMean & \textbf{\sbSelTimeChatChosen} \\
Qwen3-VL-8B & \sbSelQwenEightWzeroKzero & \sbSelQwenEightWzeroKsixteen & \sbSelQwenEightWzeroMean & \sbSelQwenEightWoneKzero & \sbSelQwenEightWoneKsixteen & \sbSelQwenEightWoneMean & \sbSelQwenEightWtwoKzero & \sbSelQwenEightWtwoKsixteen & \textbf{\sbSelQwenEightWtwoMean} & \textbf{\sbSelQwenEightChosen} \\
\bottomrule
\end{tabular}

%% file: floats/tab_judges.tex
\begin{table}[t]
  \centering
  \caption{\textbf{A second judge keeps every comparison with \model.} Fact-F1 (\% of each judge's own teacher ceiling) of the held-out captions of \cref{tab:main} (test: LLaVA-Video test), judged by phi-4, the judge of the paper, and by Qwen3.5-9B, which also scores the RL reward of \model. These scores keep the 60-word rule; \cref{tab:main} also gives them without it. Bottom: the gain of \model over its SFT model before GRPO. A policy that games its reward would gain more under the reward's judge; the gain is about the same under both.}
  \label{tab:judges}
  \scriptsize
  \setlength{\tabcolsep}{3pt}
  \input{tables/judges}
  \par\medskip
  \begin{tabular}{@{}lrll@{}}
  \toprule
  Gain over SFT & $L$ & phi-4 & Qwen \\
  \midrule
  dev & 0 & \devTenOverOvSftKzero\ [\devTenOverOvSftKzeroLo, \devTenOverOvSftKzeroHi] & \devQTenOverOvSftKzero\ [\devQTenOverOvSftKzeroLo, \devQTenOverOvSftKzeroHi] \\
  dev & 16 & \devTenOverOvSftKsixteen\ [\devTenOverOvSftKsixteenLo, \devTenOverOvSftKsixteenHi] & \devQTenOverOvSftKsixteen\ [\devQTenOverOvSftKsixteenLo, \devQTenOverOvSftKsixteenHi] \\
  OpenVid & 0 & \tenOverOvSftOKzero\ [\tenOverOvSftOKzeroLo, \tenOverOvSftOKzeroHi] & \jqTenOverOvSftOKzero\ [\jqTenOverOvSftOKzeroLo, \jqTenOverOvSftOKzeroHi] \\
  OpenVid & 16 & \tenOverOvSftOKsixteen\ [\tenOverOvSftOKsixteenLo, \tenOverOvSftOKsixteenHi] & \jqTenOverOvSftOKsixteen\ [\jqTenOverOvSftOKsixteenLo, \jqTenOverOvSftOKsixteenHi] \\
  test & 0 & \tenOverOvSftTKzero\ [\tenOverOvSftTKzeroLo, \tenOverOvSftTKzeroHi] & \jqTenOverOvSftTKzero\ [\jqTenOverOvSftTKzeroLo, \jqTenOverOvSftTKzeroHi] \\
  test & 16 & \tenOverOvSftTKsixteen\ [\tenOverOvSftTKsixteenLo, \tenOverOvSftTKsixteenHi] & \jqTenOverOvSftTKsixteen\ [\jqTenOverOvSftTKsixteenLo, \jqTenOverOvSftTKsixteenHi] \\
  \bottomrule
  \end{tabular}
\end{table}

%% file: tables/judges.tex
\begin{tabular}{@{}lrrrrrr@{}}
\toprule
 & \multicolumn{2}{c}{OpenVid, $L{=}0$} & \multicolumn{2}{c}{OpenVid, $L{=}16$} & \multicolumn{2}{c}{test, $L{=}16$} \\
\cmidrule(lr){2-3}\cmidrule(lr){4-5}\cmidrule(l){6-7}
Model & phi-4 & Qwen & phi-4 & Qwen & phi-4 & Qwen \\
\midrule
\textbf{\model} & \textbf{\tenOKzero} & \textbf{\jqTenOKzero} & \textbf{\tenOKsixteen} & \textbf{\jqTenOKsixteen} & \textbf{\tenTKsixteen} & \textbf{\jqTenTKsixteen} \\
LFM2.5-VL-450M & \oLfmSmallKzero & \jqOLfmSmallKzero & \oLfmSmallKsixteen & \jqOLfmSmallKsixteen & \tbLfmKsixteen & \tbQjLfmKsixteen \\
SmolVLM2-500M & \oSmolSmallKzero & \jqOSmolSmallKzero & \oSmolSmallKsixteen & \jqOSmolSmallKsixteen & \tbSmolKsixteen & \tbQjSmolKsixteen \\
FastVLM-0.5B & \nfOFastVlmKzero & \nfQOFastVlmKzero & \nfOFastVlmKsixteen & \nfQOFastVlmKsixteen & \nfTFastVlmKsixteen & \nfQTFastVlmKsixteen \\
Mobile-VideoGPT-0.5B & \fcOMobileKzero & \fcQOMobileKzero & \fcOMobileKsixteen & \fcQOMobileKsixteen & \fcTMobileKsixteen & \fcQTMobileKsixteen \\
\midrule
InternVL3.5-1B & \oInternOneKzero & \jqOInternOneKzero & \oInternOneKsixteen & \jqOInternOneKsixteen & \tsInternOneKsixteen & \jqTsInternOneKsixteen \\
Qwen3-VL-2B & \oQwenTwoKzero & \jqOQwenTwoKzero & \oQwenTwoKsixteen & \jqOQwenTwoKsixteen & \tQwenTwoKsixteen & \jqTQwenTwoKsixteen \\
InternVL3.5-1B, INT8 & \qOQIntKzero & \qOQIntKzeroReward & \qOQIntKsixteen & \qOQIntKsixteenReward & \fcTInternOneIntKsixteen & \fcQTInternOneIntKsixteen \\
Qwen3-VL-2B, INT4 & \qOQQwenKzero & \qOQQwenKzeroReward & \qOQQwenKsixteen & \qOQQwenKsixteenReward & \fcTQwenTwoIntKsixteen & \fcQTQwenTwoIntKsixteen \\
LFM2.5-VL-1.6B & \oLfmBigKzero & \jqOLfmBigKzero & \oLfmBigKsixteen & \jqOLfmBigKsixteen & \tsLfmBigKsixteen & \jqTsLfmBigKsixteen \\
SmolVLM2-2.2B & \oSmolBigKzero & \jqOSmolBigKzero & \oSmolBigKsixteen & \jqOSmolBigKsixteen & \tsSmolBigKsixteen & \jqTsSmolBigKsixteen \\
InternVL3.5-2B & \oInternTwoKzero & \jqOInternTwoKzero & \oInternTwoKsixteen & \jqOInternTwoKsixteen & \tsInternTwoKsixteen & \jqTsInternTwoKsixteen \\
Qwen3-VL-4B & \oQwenFourKzero & \jqOQwenFourKzero & \oQwenFourKsixteen & \jqOQwenFourKsixteen & \tQwenFourKsixteen & \jqTQwenFourKsixteen \\
Qwen2.5-VL-7B & \oQwenSevenTwoFiveKzero & \jqOQwenSevenTwoFiveKzero & \oQwenSevenTwoFiveKsixteen & \jqOQwenSevenTwoFiveKsixteen & \tsQwenSevenTwoFiveKsixteen & \jqTsQwenSevenTwoFiveKsixteen \\
LiveCC-7B & \oLiveCcKzero & \jqOLiveCcKzero & \oLiveCcKsixteen & \jqOLiveCcKsixteen & \tsLiveCcKsixteen & \jqTsLiveCcKsixteen \\
StreamingVLM-7B & \nfOStreamingVlmKzero & \nfQOStreamingVlmKzero & \nfOStreamingVlmKsixteen & \nfQOStreamingVlmKsixteen & \nfTStreamingVlmKsixteen & \nfQTStreamingVlmKsixteen \\
TimeChat-Online-7B & \nfOTimeChatKzero & \nfQOTimeChatKzero & \nfOTimeChatKsixteen & \nfQOTimeChatKsixteen & \nfTTimeChatKsixteen & \nfQTTimeChatKsixteen \\
Qwen3-VL-8B & \oQwenEightKzero & \jqOQwenEightKzero & \oQwenEightKsixteen & \jqOQwenEightKsixteen & \tQwenEightKsixteen & \jqTQwenEightKsixteen \\
\bottomrule
\end{tabular}

%% file: floats/fig_dial.tex
\begin{figure*}[tp]
  \centering
  \includegraphics[width=\linewidth]{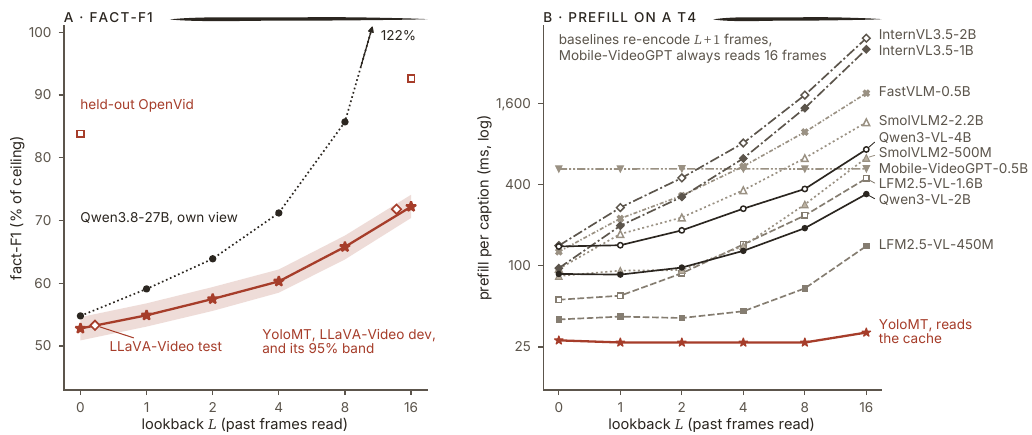}
  \caption{\textbf{More frames for \model: informative and nearly free.} (a) Fact-F1 against $L$ on the dev split with a 95\% band, and on held-out OpenVid and test at $L{=}0$ and 16; the teacher, Qwen3.8-27B, reads its own view. (b) Prefill time per caption on a T4 (eager FP16, median of 100): \model reads cached blocks; the baselines re-encode every frame, and Mobile-VideoGPT-0.5B always reads 16 resampled frames. Benchmark gains are in \cref{tab:supp:kzero}.}
  \label{fig:dial}
\end{figure*}

%% file: floats/tab_supp_kzero.tex
\begin{table*}[tp]
  \centering
  \caption{\textbf{What more frames buy, for every model.} Caption-then-answer accuracy (\%) on the questions of \cref{tab:benchsubsets} with no past frame, and the four-benchmark mean with none and with the protocol lookback $L{=}\min(t,16)$: \sbGainModels\ models in one answer call; the INT8 and INT4 rows of \cref{tab:main} were not run at $L{=}0$. Gain: the second mean minus the first, then the gain of \model minus the model's, with paired 95\% intervals. The means at $L{=}\min(t,16)$ are within \sbShiftSameMax\ points of \cref{tab:main}, or \sbShiftRecaptionedMax\ for the \sbShiftRecaptionedCount\ models captioned again.}
  \label{tab:supp:kzero}
  \scriptsize
  \setlength{\tabcolsep}{3pt}
  \input{tables/supp_kzero}
\end{table*}

%% file: tables/supp_kzero.tex
\begin{tabular}{@{}lrrrrrrcc@{}}
\toprule
 & \multicolumn{5}{c}{accuracy (\%) at $L{=}0$} & mean at & \multicolumn{2}{c}{gain of the mean [95\% interval]} \\
\cmidrule(lr){2-6}\cmidrule(l){8-9}
Model & OVO-RT & MVBench & StrBench & VMME & mean & $L{=}\min(t,16)$ & the model's & \model's minus it \\
\midrule
no captions & \sbZeroFloorOvo & \sbZeroFloorMvb & \sbZeroFloorSb & \sbZeroFloorVm & \sbZeroFloorMean & \sbProtoFloorMean & -- & -- \\
\midrule
\multicolumn{9}{@{}l}{\emph{under 1B}} \\
\model & \sbZeroOursOvo & \sbZeroOursMvb & \sbZeroOursSb & \sbZeroOursVm & \sbZeroOursMean & \sbProtoOursMean & \sbGainOursMean\ [\sbGainOursMeanLo, \sbGainOursMeanHi] & -- \\
LFM2.5-VL-450M & \sbZeroLfmSmallOvo & \sbZeroLfmSmallMvb & \sbZeroLfmSmallSb & \sbZeroLfmSmallVm & \sbZeroLfmSmallMean & \sbProtoLfmSmallMean & \sbGainLfmSmallMean\ [\sbGainLfmSmallMeanLo, \sbGainLfmSmallMeanHi] & \gaOverLfmSmallMean\ [\gaOverLfmSmallMeanLo, \gaOverLfmSmallMeanHi] \\
SmolVLM2-500M & \sbZeroSmolSmallOvo & \sbZeroSmolSmallMvb & \sbZeroSmolSmallSb & \sbZeroSmolSmallVm & \sbZeroSmolSmallMean & \sbProtoSmolSmallMean & \sbGainSmolSmallMean\ [\sbGainSmolSmallMeanLo, \sbGainSmolSmallMeanHi] & \gaOverSmolSmallMean\ [\gaOverSmolSmallMeanLo, \gaOverSmolSmallMeanHi] \\
FastVLM-0.5B & \sbZeroFastVlmOvo & \sbZeroFastVlmMvb & \sbZeroFastVlmSb & \sbZeroFastVlmVm & \sbZeroFastVlmMean & \sbProtoFastVlmMean & \sbGainFastVlmMean\ [\sbGainFastVlmMeanLo, \sbGainFastVlmMeanHi] & \gaOverFastVlmMean\ [\gaOverFastVlmMeanLo, \gaOverFastVlmMeanHi] \\
Mobile-VideoGPT-0.5B & \sbZeroMobileOvo & \sbZeroMobileMvb & \sbZeroMobileSb & \sbZeroMobileVm & \sbZeroMobileMean & \sbProtoMobileMean & \sbGainMobileMean\ [\sbGainMobileMeanLo, \sbGainMobileMeanHi] & \gaOverMobileMean\ [\gaOverMobileMeanLo, \gaOverMobileMeanHi] \\
\midrule
\multicolumn{9}{@{}l}{\emph{1B to 3B}} \\
InternVL3.5-1B & \sbZeroInternOneOvo & \sbZeroInternOneMvb & \sbZeroInternOneSb & \sbZeroInternOneVm & \sbZeroInternOneMean & \sbProtoInternOneMean & \sbGainInternOneMean\ [\sbGainInternOneMeanLo, \sbGainInternOneMeanHi] & \gaOverInternOneMean\ [\gaOverInternOneMeanLo, \gaOverInternOneMeanHi] \\
LFM2.5-VL-1.6B & \sbZeroLfmBigOvo & \sbZeroLfmBigMvb & \sbZeroLfmBigSb & \sbZeroLfmBigVm & \sbZeroLfmBigMean & \sbProtoLfmBigMean & \sbGainLfmBigMean\ [\sbGainLfmBigMeanLo, \sbGainLfmBigMeanHi] & \gaOverLfmBigMean\ [\gaOverLfmBigMeanLo, \gaOverLfmBigMeanHi] \\
Qwen3-VL-2B & \sbZeroQwenTwoOvo & \sbZeroQwenTwoMvb & \sbZeroQwenTwoSb & \sbZeroQwenTwoVm & \sbZeroQwenTwoMean & \sbProtoQwenTwoMean & \sbGainQwenTwoMean\ [\sbGainQwenTwoMeanLo, \sbGainQwenTwoMeanHi] & \gaOverQwenTwoMean\ [\gaOverQwenTwoMeanLo, \gaOverQwenTwoMeanHi] \\
SmolVLM2-2.2B & \sbZeroSmolBigOvo & \sbZeroSmolBigMvb & \sbZeroSmolBigSb & \sbZeroSmolBigVm & \sbZeroSmolBigMean & \sbProtoSmolBigMean & \sbGainSmolBigMean\ [\sbGainSmolBigMeanLo, \sbGainSmolBigMeanHi] & \gaOverSmolBigMean\ [\gaOverSmolBigMeanLo, \gaOverSmolBigMeanHi] \\
InternVL3.5-2B & \sbZeroInternTwoOvo & \sbZeroInternTwoMvb & \sbZeroInternTwoSb & \sbZeroInternTwoVm & \sbZeroInternTwoMean & \sbProtoInternTwoMean & \sbGainInternTwoMean\ [\sbGainInternTwoMeanLo, \sbGainInternTwoMeanHi] & \gaOverInternTwoMean\ [\gaOverInternTwoMeanLo, \gaOverInternTwoMeanHi] \\
\midrule
\multicolumn{9}{@{}l}{\emph{above 3B}} \\
Qwen3-VL-4B & \sbZeroQwenFourOvo & \sbZeroQwenFourMvb & \sbZeroQwenFourSb & \sbZeroQwenFourVm & \sbZeroQwenFourMean & \sbProtoQwenFourMean & \sbGainQwenFourMean\ [\sbGainQwenFourMeanLo, \sbGainQwenFourMeanHi] & \gaOverQwenFourMean\ [\gaOverQwenFourMeanLo, \gaOverQwenFourMeanHi] \\
Qwen2.5-VL-7B & \sbZeroQwenSevenTwoFiveOvo & \sbZeroQwenSevenTwoFiveMvb & \sbZeroQwenSevenTwoFiveSb & \sbZeroQwenSevenTwoFiveVm & \sbZeroQwenSevenTwoFiveMean & \sbProtoQwenSevenTwoFiveMean & \sbGainQwenSevenTwoFiveMean\ [\sbGainQwenSevenTwoFiveMeanLo, \sbGainQwenSevenTwoFiveMeanHi] & \gaOverQwenSevenTwoFiveMean\ [\gaOverQwenSevenTwoFiveMeanLo, \gaOverQwenSevenTwoFiveMeanHi] \\
LiveCC-7B & \sbZeroLiveCcOvo & \sbZeroLiveCcMvb & \sbZeroLiveCcSb & \sbZeroLiveCcVm & \sbZeroLiveCcMean & \sbProtoLiveCcMean & \sbGainLiveCcMean\ [\sbGainLiveCcMeanLo, \sbGainLiveCcMeanHi] & \gaOverLiveCcMean\ [\gaOverLiveCcMeanLo, \gaOverLiveCcMeanHi] \\
StreamingVLM-7B & \sbZeroStreamingVlmOvo & \sbZeroStreamingVlmMvb & \sbZeroStreamingVlmSb & \sbZeroStreamingVlmVm & \sbZeroStreamingVlmMean & \sbProtoStreamingVlmMean & \sbGainStreamingVlmMean\ [\sbGainStreamingVlmMeanLo, \sbGainStreamingVlmMeanHi] & \gaOverStreamingVlmMean\ [\gaOverStreamingVlmMeanLo, \gaOverStreamingVlmMeanHi] \\
TimeChat-Online-7B & \sbZeroTimeChatOvo & \sbZeroTimeChatMvb & \sbZeroTimeChatSb & \sbZeroTimeChatVm & \sbZeroTimeChatMean & \sbProtoTimeChatMean & \sbGainTimeChatMean\ [\sbGainTimeChatMeanLo, \sbGainTimeChatMeanHi] & \gaOverTimeChatMean\ [\gaOverTimeChatMeanLo, \gaOverTimeChatMeanHi] \\
Qwen3-VL-8B & \sbZeroQwenEightOvo & \sbZeroQwenEightMvb & \sbZeroQwenEightSb & \sbZeroQwenEightVm & \sbZeroQwenEightMean & \sbProtoQwenEightMean & \sbGainQwenEightMean\ [\sbGainQwenEightMeanLo, \sbGainQwenEightMeanHi] & \gaOverQwenEightMean\ [\gaOverQwenEightMeanLo, \gaOverQwenEightMeanHi] \\
\midrule
Qwen3.8-27B (teacher) & \sbZeroTeacherOvo & \sbZeroTeacherMvb & \sbZeroTeacherSb & \sbZeroTeacherVm & \sbZeroTeacherMean & \sbProtoTeacherMean & \sbGainTeacherMean\ [\sbGainTeacherMeanLo, \sbGainTeacherMeanHi] & \gaOverTeacherMean\ [\gaOverTeacherMeanLo, \gaOverTeacherMeanHi] \\
\bottomrule
\end{tabular}

%% file: floats/tab_lookback.tex
\begin{table*}[tp]
  \centering
  \caption{\textbf{Real past frames pay; the same frame repeated, another video's past or blended frames do not.} Dev split, same weights, inference only; changes against the reference row of each block at the same $L$. Upper block: the prompt, its length and its $L$ blocks stay, and only the pixels they are pooled from change; the consensus decodes 16 samples per caption. Lower block: cached seconds blended at test time; $2L{-}1$ blocks put a midpoint between every two seconds, blended from both or copied from the older one.}
  \label{tab:lookback}
  \scriptsize
  \setlength{\tabcolsep}{4pt}
  \input{tables/lookback}
\end{table*}

%% file: tables/lookback.tex
\begin{tabular}{@{}lrrr@{}}
\toprule
 & $L$ & fact-F1 & change [95\% interval] \\
\midrule
\multicolumn{4}{@{}l}{\emph{\model, one judging run}} \\
no past frames & 0 & \cyPlainKzero & reference \\
consensus of 16 samples & 0 & \cyConsensusKzero & \cyConsensusGapKzero\ [\cyConsensusGapKzeroLo, \cyConsensusGapKzeroHi] \\
the real past & 16 & \cyPlainKsixteen & reference \\
frame $t$, repeated & 16 & \cyRepeat & \cyRepeatVsReal\ [\cyRepeatVsRealLo, \cyRepeatVsRealHi] \\
another video's past & 16 & \cyForeign & \cyForeignVsReal\ [\cyForeignVsRealLo, \cyForeignVsRealHi] \\
the real past, shuffled & 16 & \cyShuffled & \cyShuffledVsReal\ [\cyShuffledVsRealLo, \cyShuffledVsRealHi] \\
consensus of 16 samples & 16 & \cyConsensusKsixteen & \cyConsensusGapKsixteen\ [\cyConsensusGapKsixteenLo, \cyConsensusGapKsixteenHi] \\
\midrule
\multicolumn{4}{@{}l}{\emph{\model, another judging run}} \\
as trained & 8 & \ycInterpPlainEight & reference \\
midpoints blended, $2L{-}1$ blocks & 8 & \ycInterpDenseEight & \ycInterpDenseEightGap\ [\ycInterpDenseEightGapLo, \ycInterpDenseEightGapHi] \\
as trained & 16 & \ycInterpPlainSixteen & reference \\
neighbours blended & 16 & \ycInterpSmoothSixteen & \ycInterpSmoothGap\ [\ycInterpSmoothGapLo, \ycInterpSmoothGapHi] \\
midpoints blended, $2L{-}1$ blocks & 16 & \ycInterpDenseSixteen & \ycInterpDenseGap\ [\ycInterpDenseGapLo, \ycInterpDenseGapHi] \\
midpoints copied, $2L{-}1$ blocks & 16 & \ycInterpDupSixteen & \ycInterpDupGap\ [\ycInterpDupGapLo, \ycInterpDupGapHi] \\
\bottomrule
\end{tabular}

%% file: floats/fig_examples.tex
\begin{figure*}[tp]
  \centering
  \includegraphics[width=\linewidth]{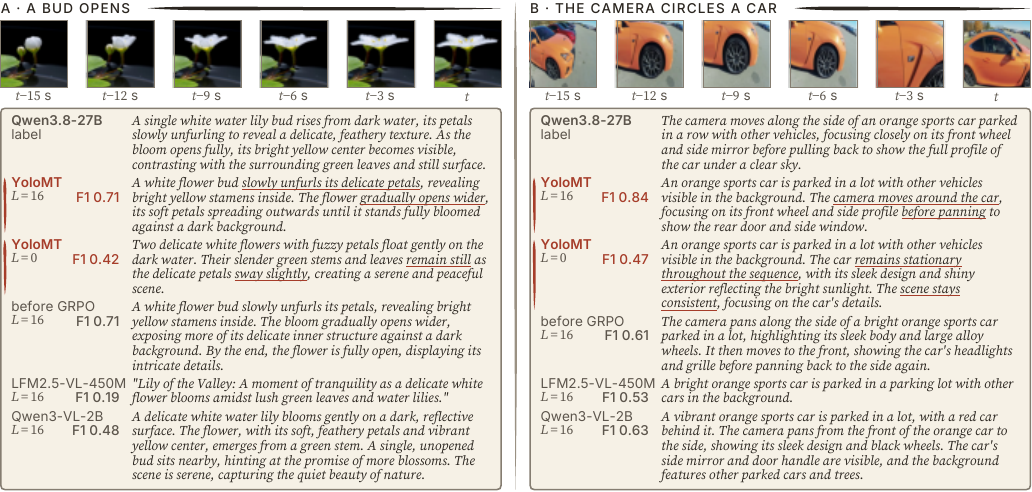}
  \caption{\textbf{What the past adds: two held-out OpenVid windows where the view changes.} (a) The scene changes; (b) the camera moves. Picked by eye among the \examplesPool\ held-out windows with no person in the teacher's label or in \model's captions, from another video than \cref{fig:lookback}. They show what the past is for, not a typical window: \model's caption at $L{=}16$ beats its caption at $L{=}0$ on \examplesGainPos\ of the \examplesPool, and over all held-out windows the median fact-F1 at $L{=}16$ is \examplesMedianF\ and the median gain \examplesMedianGain. Six of the 17 frames read at $L{=}16$; captions as written, with their fact-F1. Underlines mark the temporal descriptions compared, not per-fact correctness.}
  \label{fig:examples}
\end{figure*}

%% file: floats/tab_supp_tasksall.tex
\begin{table*}[p]
  \centering
  \caption{\textbf{Accuracy per task of every model of \cref{tab:main}.} Caption-then-answer accuracy (\%) at $L{=}\min(t,16)$ on the \sbTaskCount\ tasks of \cref{tab:benchsubsets} (questions per task in the second column), from the answer calls of \cref{tab:main}, which gives each benchmark's accuracy over all its tasks. $^\dagger$: the second call, for the three newest baselines. $^\ddagger$: OVO-Bench and MVBench captions in w0, not the selected w2 (\cref{sec:supp:prompts}). Per-task samples are small: an accuracy near 50\% has a 95\% interval of about $\pm$10 points at 100 questions and $\pm$7 at 200, and no interval is given per task.}
  \label{tab:supp:tasksall}
  \scriptsize
  \setlength{\tabcolsep}{2pt}
  \input{tables/supp_tasks_all}
\end{table*}

%% file: tables/supp_tasks_all.tex
\begin{tabular}{@{}lrrrrrrrrrrrrrrrrrrrrr@{}}
\toprule
 & &  & \multicolumn{5}{c}{under 1B} & \multicolumn{7}{c}{1B to 3B} & \multicolumn{6}{c}{above 3B} &  \\
\cmidrule(lr){4-8}\cmidrule(lr){9-15}\cmidrule(lr){16-21}
Task & questions & \rotatebox{90}{no captions} & \rotatebox{90}{\textbf{\model}} & \rotatebox{90}{LFM2.5-VL-450M} & \rotatebox{90}{SmolVLM2-500M} & \rotatebox{90}{FastVLM-0.5B$^\dagger$} & \rotatebox{90}{Mobile-VideoGPT-0.5B} & \rotatebox{90}{InternVL3.5-1B} & \rotatebox{90}{InternVL3.5-1B INT8} & \rotatebox{90}{LFM2.5-VL-1.6B} & \rotatebox{90}{Qwen3-VL-2B} & \rotatebox{90}{Qwen3-VL-2B INT4} & \rotatebox{90}{SmolVLM2-2.2B$^\ddagger$} & \rotatebox{90}{InternVL3.5-2B} & \rotatebox{90}{Qwen3-VL-4B} & \rotatebox{90}{Qwen2.5-VL-7B} & \rotatebox{90}{LiveCC-7B} & \rotatebox{90}{StreamingVLM-7B$^\dagger$} & \rotatebox{90}{TimeChat-Online-7B$^\dagger$} & \rotatebox{90}{Qwen3-VL-8B} & \rotatebox{90}{Qwen3.8-27B (teacher)} \\
\midrule
\multicolumn{22}{@{}l}{\emph{OVO-Bench Real-Time}} \\
Object Recognition & \sbTaskNOvoOjr & \sbTaskFloorOvoOjr & \sbTaskOursOvoOjr & \sbTaskLfmSmallOvoOjr & \sbTaskSmolSmallOvoOjr & \sbTaskFastVlmOvoOjr & \sbTaskMobileOvoOjr & \sbTaskInternOneOvoOjr & \sbTaskInternOneIntOvoOjr & \sbTaskLfmBigOvoOjr & \sbTaskQwenTwoOvoOjr & \sbTaskQwenTwoIntOvoOjr & \sbTaskSmolBigOvoOjr & \sbTaskInternTwoOvoOjr & \sbTaskQwenFourOvoOjr & \sbTaskQwenSevenTwoFiveOvoOjr & \sbTaskLiveCcOvoOjr & \sbTaskStreamingVlmOvoOjr & \sbTaskTimeChatOvoOjr & \sbTaskQwenEightOvoOjr & \sbTaskTeacherOvoOjr \\
Spatial Understanding & \sbTaskNOvoStu & \sbTaskFloorOvoStu & \sbTaskOursOvoStu & \sbTaskLfmSmallOvoStu & \sbTaskSmolSmallOvoStu & \sbTaskFastVlmOvoStu & \sbTaskMobileOvoStu & \sbTaskInternOneOvoStu & \sbTaskInternOneIntOvoStu & \sbTaskLfmBigOvoStu & \sbTaskQwenTwoOvoStu & \sbTaskQwenTwoIntOvoStu & \sbTaskSmolBigOvoStu & \sbTaskInternTwoOvoStu & \sbTaskQwenFourOvoStu & \sbTaskQwenSevenTwoFiveOvoStu & \sbTaskLiveCcOvoStu & \sbTaskStreamingVlmOvoStu & \sbTaskTimeChatOvoStu & \sbTaskQwenEightOvoStu & \sbTaskTeacherOvoStu \\
Optical Character Recognition & \sbTaskNOvoOcr & \sbTaskFloorOvoOcr & \sbTaskOursOvoOcr & \sbTaskLfmSmallOvoOcr & \sbTaskSmolSmallOvoOcr & \sbTaskFastVlmOvoOcr & \sbTaskMobileOvoOcr & \sbTaskInternOneOvoOcr & \sbTaskInternOneIntOvoOcr & \sbTaskLfmBigOvoOcr & \sbTaskQwenTwoOvoOcr & \sbTaskQwenTwoIntOvoOcr & \sbTaskSmolBigOvoOcr & \sbTaskInternTwoOvoOcr & \sbTaskQwenFourOvoOcr & \sbTaskQwenSevenTwoFiveOvoOcr & \sbTaskLiveCcOvoOcr & \sbTaskStreamingVlmOvoOcr & \sbTaskTimeChatOvoOcr & \sbTaskQwenEightOvoOcr & \sbTaskTeacherOvoOcr \\
Attribute Recognition & \sbTaskNOvoAtr & \sbTaskFloorOvoAtr & \sbTaskOursOvoAtr & \sbTaskLfmSmallOvoAtr & \sbTaskSmolSmallOvoAtr & \sbTaskFastVlmOvoAtr & \sbTaskMobileOvoAtr & \sbTaskInternOneOvoAtr & \sbTaskInternOneIntOvoAtr & \sbTaskLfmBigOvoAtr & \sbTaskQwenTwoOvoAtr & \sbTaskQwenTwoIntOvoAtr & \sbTaskSmolBigOvoAtr & \sbTaskInternTwoOvoAtr & \sbTaskQwenFourOvoAtr & \sbTaskQwenSevenTwoFiveOvoAtr & \sbTaskLiveCcOvoAtr & \sbTaskStreamingVlmOvoAtr & \sbTaskTimeChatOvoAtr & \sbTaskQwenEightOvoAtr & \sbTaskTeacherOvoAtr \\
Action Recognition & \sbTaskNOvoAcr & \sbTaskFloorOvoAcr & \sbTaskOursOvoAcr & \sbTaskLfmSmallOvoAcr & \sbTaskSmolSmallOvoAcr & \sbTaskFastVlmOvoAcr & \sbTaskMobileOvoAcr & \sbTaskInternOneOvoAcr & \sbTaskInternOneIntOvoAcr & \sbTaskLfmBigOvoAcr & \sbTaskQwenTwoOvoAcr & \sbTaskQwenTwoIntOvoAcr & \sbTaskSmolBigOvoAcr & \sbTaskInternTwoOvoAcr & \sbTaskQwenFourOvoAcr & \sbTaskQwenSevenTwoFiveOvoAcr & \sbTaskLiveCcOvoAcr & \sbTaskStreamingVlmOvoAcr & \sbTaskTimeChatOvoAcr & \sbTaskQwenEightOvoAcr & \sbTaskTeacherOvoAcr \\
Future Prediction & \sbTaskNOvoFpd & \sbTaskFloorOvoFpd & \sbTaskOursOvoFpd & \sbTaskLfmSmallOvoFpd & \sbTaskSmolSmallOvoFpd & \sbTaskFastVlmOvoFpd & \sbTaskMobileOvoFpd & \sbTaskInternOneOvoFpd & \sbTaskInternOneIntOvoFpd & \sbTaskLfmBigOvoFpd & \sbTaskQwenTwoOvoFpd & \sbTaskQwenTwoIntOvoFpd & \sbTaskSmolBigOvoFpd & \sbTaskInternTwoOvoFpd & \sbTaskQwenFourOvoFpd & \sbTaskQwenSevenTwoFiveOvoFpd & \sbTaskLiveCcOvoFpd & \sbTaskStreamingVlmOvoFpd & \sbTaskTimeChatOvoFpd & \sbTaskQwenEightOvoFpd & \sbTaskTeacherOvoFpd \\
\midrule
\multicolumn{22}{@{}l}{\emph{MVBench}} \\
Action sequence & \sbTaskNMvbActionSequence & \sbTaskFloorMvbActionSequence & \sbTaskOursMvbActionSequence & \sbTaskLfmSmallMvbActionSequence & \sbTaskSmolSmallMvbActionSequence & \sbTaskFastVlmMvbActionSequence & \sbTaskMobileMvbActionSequence & \sbTaskInternOneMvbActionSequence & \sbTaskInternOneIntMvbActionSequence & \sbTaskLfmBigMvbActionSequence & \sbTaskQwenTwoMvbActionSequence & \sbTaskQwenTwoIntMvbActionSequence & \sbTaskSmolBigMvbActionSequence & \sbTaskInternTwoMvbActionSequence & \sbTaskQwenFourMvbActionSequence & \sbTaskQwenSevenTwoFiveMvbActionSequence & \sbTaskLiveCcMvbActionSequence & \sbTaskStreamingVlmMvbActionSequence & \sbTaskTimeChatMvbActionSequence & \sbTaskQwenEightMvbActionSequence & \sbTaskTeacherMvbActionSequence \\
Action prediction & \sbTaskNMvbActionPrediction & \sbTaskFloorMvbActionPrediction & \sbTaskOursMvbActionPrediction & \sbTaskLfmSmallMvbActionPrediction & \sbTaskSmolSmallMvbActionPrediction & \sbTaskFastVlmMvbActionPrediction & \sbTaskMobileMvbActionPrediction & \sbTaskInternOneMvbActionPrediction & \sbTaskInternOneIntMvbActionPrediction & \sbTaskLfmBigMvbActionPrediction & \sbTaskQwenTwoMvbActionPrediction & \sbTaskQwenTwoIntMvbActionPrediction & \sbTaskSmolBigMvbActionPrediction & \sbTaskInternTwoMvbActionPrediction & \sbTaskQwenFourMvbActionPrediction & \sbTaskQwenSevenTwoFiveMvbActionPrediction & \sbTaskLiveCcMvbActionPrediction & \sbTaskStreamingVlmMvbActionPrediction & \sbTaskTimeChatMvbActionPrediction & \sbTaskQwenEightMvbActionPrediction & \sbTaskTeacherMvbActionPrediction \\
Object interaction & \sbTaskNMvbObjectInteraction & \sbTaskFloorMvbObjectInteraction & \sbTaskOursMvbObjectInteraction & \sbTaskLfmSmallMvbObjectInteraction & \sbTaskSmolSmallMvbObjectInteraction & \sbTaskFastVlmMvbObjectInteraction & \sbTaskMobileMvbObjectInteraction & \sbTaskInternOneMvbObjectInteraction & \sbTaskInternOneIntMvbObjectInteraction & \sbTaskLfmBigMvbObjectInteraction & \sbTaskQwenTwoMvbObjectInteraction & \sbTaskQwenTwoIntMvbObjectInteraction & \sbTaskSmolBigMvbObjectInteraction & \sbTaskInternTwoMvbObjectInteraction & \sbTaskQwenFourMvbObjectInteraction & \sbTaskQwenSevenTwoFiveMvbObjectInteraction & \sbTaskLiveCcMvbObjectInteraction & \sbTaskStreamingVlmMvbObjectInteraction & \sbTaskTimeChatMvbObjectInteraction & \sbTaskQwenEightMvbObjectInteraction & \sbTaskTeacherMvbObjectInteraction \\
Action antonym & \sbTaskNMvbActionAntonym & \sbTaskFloorMvbActionAntonym & \sbTaskOursMvbActionAntonym & \sbTaskLfmSmallMvbActionAntonym & \sbTaskSmolSmallMvbActionAntonym & \sbTaskFastVlmMvbActionAntonym & \sbTaskMobileMvbActionAntonym & \sbTaskInternOneMvbActionAntonym & \sbTaskInternOneIntMvbActionAntonym & \sbTaskLfmBigMvbActionAntonym & \sbTaskQwenTwoMvbActionAntonym & \sbTaskQwenTwoIntMvbActionAntonym & \sbTaskSmolBigMvbActionAntonym & \sbTaskInternTwoMvbActionAntonym & \sbTaskQwenFourMvbActionAntonym & \sbTaskQwenSevenTwoFiveMvbActionAntonym & \sbTaskLiveCcMvbActionAntonym & \sbTaskStreamingVlmMvbActionAntonym & \sbTaskTimeChatMvbActionAntonym & \sbTaskQwenEightMvbActionAntonym & \sbTaskTeacherMvbActionAntonym \\
Fine grained action & \sbTaskNMvbFineGrainedAction & \sbTaskFloorMvbFineGrainedAction & \sbTaskOursMvbFineGrainedAction & \sbTaskLfmSmallMvbFineGrainedAction & \sbTaskSmolSmallMvbFineGrainedAction & \sbTaskFastVlmMvbFineGrainedAction & \sbTaskMobileMvbFineGrainedAction & \sbTaskInternOneMvbFineGrainedAction & \sbTaskInternOneIntMvbFineGrainedAction & \sbTaskLfmBigMvbFineGrainedAction & \sbTaskQwenTwoMvbFineGrainedAction & \sbTaskQwenTwoIntMvbFineGrainedAction & \sbTaskSmolBigMvbFineGrainedAction & \sbTaskInternTwoMvbFineGrainedAction & \sbTaskQwenFourMvbFineGrainedAction & \sbTaskQwenSevenTwoFiveMvbFineGrainedAction & \sbTaskLiveCcMvbFineGrainedAction & \sbTaskStreamingVlmMvbFineGrainedAction & \sbTaskTimeChatMvbFineGrainedAction & \sbTaskQwenEightMvbFineGrainedAction & \sbTaskTeacherMvbFineGrainedAction \\
Unexpected action & \sbTaskNMvbUnexpectedAction & \sbTaskFloorMvbUnexpectedAction & \sbTaskOursMvbUnexpectedAction & \sbTaskLfmSmallMvbUnexpectedAction & \sbTaskSmolSmallMvbUnexpectedAction & \sbTaskFastVlmMvbUnexpectedAction & \sbTaskMobileMvbUnexpectedAction & \sbTaskInternOneMvbUnexpectedAction & \sbTaskInternOneIntMvbUnexpectedAction & \sbTaskLfmBigMvbUnexpectedAction & \sbTaskQwenTwoMvbUnexpectedAction & \sbTaskQwenTwoIntMvbUnexpectedAction & \sbTaskSmolBigMvbUnexpectedAction & \sbTaskInternTwoMvbUnexpectedAction & \sbTaskQwenFourMvbUnexpectedAction & \sbTaskQwenSevenTwoFiveMvbUnexpectedAction & \sbTaskLiveCcMvbUnexpectedAction & \sbTaskStreamingVlmMvbUnexpectedAction & \sbTaskTimeChatMvbUnexpectedAction & \sbTaskQwenEightMvbUnexpectedAction & \sbTaskTeacherMvbUnexpectedAction \\
Object existence & \sbTaskNMvbObjectExistence & \sbTaskFloorMvbObjectExistence & \sbTaskOursMvbObjectExistence & \sbTaskLfmSmallMvbObjectExistence & \sbTaskSmolSmallMvbObjectExistence & \sbTaskFastVlmMvbObjectExistence & \sbTaskMobileMvbObjectExistence & \sbTaskInternOneMvbObjectExistence & \sbTaskInternOneIntMvbObjectExistence & \sbTaskLfmBigMvbObjectExistence & \sbTaskQwenTwoMvbObjectExistence & \sbTaskQwenTwoIntMvbObjectExistence & \sbTaskSmolBigMvbObjectExistence & \sbTaskInternTwoMvbObjectExistence & \sbTaskQwenFourMvbObjectExistence & \sbTaskQwenSevenTwoFiveMvbObjectExistence & \sbTaskLiveCcMvbObjectExistence & \sbTaskStreamingVlmMvbObjectExistence & \sbTaskTimeChatMvbObjectExistence & \sbTaskQwenEightMvbObjectExistence & \sbTaskTeacherMvbObjectExistence \\
Moving direction & \sbTaskNMvbMovingDirection & \sbTaskFloorMvbMovingDirection & \sbTaskOursMvbMovingDirection & \sbTaskLfmSmallMvbMovingDirection & \sbTaskSmolSmallMvbMovingDirection & \sbTaskFastVlmMvbMovingDirection & \sbTaskMobileMvbMovingDirection & \sbTaskInternOneMvbMovingDirection & \sbTaskInternOneIntMvbMovingDirection & \sbTaskLfmBigMvbMovingDirection & \sbTaskQwenTwoMvbMovingDirection & \sbTaskQwenTwoIntMvbMovingDirection & \sbTaskSmolBigMvbMovingDirection & \sbTaskInternTwoMvbMovingDirection & \sbTaskQwenFourMvbMovingDirection & \sbTaskQwenSevenTwoFiveMvbMovingDirection & \sbTaskLiveCcMvbMovingDirection & \sbTaskStreamingVlmMvbMovingDirection & \sbTaskTimeChatMvbMovingDirection & \sbTaskQwenEightMvbMovingDirection & \sbTaskTeacherMvbMovingDirection \\
Moving count & \sbTaskNMvbMovingCount & \sbTaskFloorMvbMovingCount & \sbTaskOursMvbMovingCount & \sbTaskLfmSmallMvbMovingCount & \sbTaskSmolSmallMvbMovingCount & \sbTaskFastVlmMvbMovingCount & \sbTaskMobileMvbMovingCount & \sbTaskInternOneMvbMovingCount & \sbTaskInternOneIntMvbMovingCount & \sbTaskLfmBigMvbMovingCount & \sbTaskQwenTwoMvbMovingCount & \sbTaskQwenTwoIntMvbMovingCount & \sbTaskSmolBigMvbMovingCount & \sbTaskInternTwoMvbMovingCount & \sbTaskQwenFourMvbMovingCount & \sbTaskQwenSevenTwoFiveMvbMovingCount & \sbTaskLiveCcMvbMovingCount & \sbTaskStreamingVlmMvbMovingCount & \sbTaskTimeChatMvbMovingCount & \sbTaskQwenEightMvbMovingCount & \sbTaskTeacherMvbMovingCount \\
Moving attribute & \sbTaskNMvbMovingAttribute & \sbTaskFloorMvbMovingAttribute & \sbTaskOursMvbMovingAttribute & \sbTaskLfmSmallMvbMovingAttribute & \sbTaskSmolSmallMvbMovingAttribute & \sbTaskFastVlmMvbMovingAttribute & \sbTaskMobileMvbMovingAttribute & \sbTaskInternOneMvbMovingAttribute & \sbTaskInternOneIntMvbMovingAttribute & \sbTaskLfmBigMvbMovingAttribute & \sbTaskQwenTwoMvbMovingAttribute & \sbTaskQwenTwoIntMvbMovingAttribute & \sbTaskSmolBigMvbMovingAttribute & \sbTaskInternTwoMvbMovingAttribute & \sbTaskQwenFourMvbMovingAttribute & \sbTaskQwenSevenTwoFiveMvbMovingAttribute & \sbTaskLiveCcMvbMovingAttribute & \sbTaskStreamingVlmMvbMovingAttribute & \sbTaskTimeChatMvbMovingAttribute & \sbTaskQwenEightMvbMovingAttribute & \sbTaskTeacherMvbMovingAttribute \\
Counterfactual inference & \sbTaskNMvbCounterfactualInference & \sbTaskFloorMvbCounterfactualInference & \sbTaskOursMvbCounterfactualInference & \sbTaskLfmSmallMvbCounterfactualInference & \sbTaskSmolSmallMvbCounterfactualInference & \sbTaskFastVlmMvbCounterfactualInference & \sbTaskMobileMvbCounterfactualInference & \sbTaskInternOneMvbCounterfactualInference & \sbTaskInternOneIntMvbCounterfactualInference & \sbTaskLfmBigMvbCounterfactualInference & \sbTaskQwenTwoMvbCounterfactualInference & \sbTaskQwenTwoIntMvbCounterfactualInference & \sbTaskSmolBigMvbCounterfactualInference & \sbTaskInternTwoMvbCounterfactualInference & \sbTaskQwenFourMvbCounterfactualInference & \sbTaskQwenSevenTwoFiveMvbCounterfactualInference & \sbTaskLiveCcMvbCounterfactualInference & \sbTaskStreamingVlmMvbCounterfactualInference & \sbTaskTimeChatMvbCounterfactualInference & \sbTaskQwenEightMvbCounterfactualInference & \sbTaskTeacherMvbCounterfactualInference \\
Object shuffle & \sbTaskNMvbObjectShuffle & \sbTaskFloorMvbObjectShuffle & \sbTaskOursMvbObjectShuffle & \sbTaskLfmSmallMvbObjectShuffle & \sbTaskSmolSmallMvbObjectShuffle & \sbTaskFastVlmMvbObjectShuffle & \sbTaskMobileMvbObjectShuffle & \sbTaskInternOneMvbObjectShuffle & \sbTaskInternOneIntMvbObjectShuffle & \sbTaskLfmBigMvbObjectShuffle & \sbTaskQwenTwoMvbObjectShuffle & \sbTaskQwenTwoIntMvbObjectShuffle & \sbTaskSmolBigMvbObjectShuffle & \sbTaskInternTwoMvbObjectShuffle & \sbTaskQwenFourMvbObjectShuffle & \sbTaskQwenSevenTwoFiveMvbObjectShuffle & \sbTaskLiveCcMvbObjectShuffle & \sbTaskStreamingVlmMvbObjectShuffle & \sbTaskTimeChatMvbObjectShuffle & \sbTaskQwenEightMvbObjectShuffle & \sbTaskTeacherMvbObjectShuffle \\
Action count & \sbTaskNMvbActionCount & \sbTaskFloorMvbActionCount & \sbTaskOursMvbActionCount & \sbTaskLfmSmallMvbActionCount & \sbTaskSmolSmallMvbActionCount & \sbTaskFastVlmMvbActionCount & \sbTaskMobileMvbActionCount & \sbTaskInternOneMvbActionCount & \sbTaskInternOneIntMvbActionCount & \sbTaskLfmBigMvbActionCount & \sbTaskQwenTwoMvbActionCount & \sbTaskQwenTwoIntMvbActionCount & \sbTaskSmolBigMvbActionCount & \sbTaskInternTwoMvbActionCount & \sbTaskQwenFourMvbActionCount & \sbTaskQwenSevenTwoFiveMvbActionCount & \sbTaskLiveCcMvbActionCount & \sbTaskStreamingVlmMvbActionCount & \sbTaskTimeChatMvbActionCount & \sbTaskQwenEightMvbActionCount & \sbTaskTeacherMvbActionCount \\
State change & \sbTaskNMvbStateChange & \sbTaskFloorMvbStateChange & \sbTaskOursMvbStateChange & \sbTaskLfmSmallMvbStateChange & \sbTaskSmolSmallMvbStateChange & \sbTaskFastVlmMvbStateChange & \sbTaskMobileMvbStateChange & \sbTaskInternOneMvbStateChange & \sbTaskInternOneIntMvbStateChange & \sbTaskLfmBigMvbStateChange & \sbTaskQwenTwoMvbStateChange & \sbTaskQwenTwoIntMvbStateChange & \sbTaskSmolBigMvbStateChange & \sbTaskInternTwoMvbStateChange & \sbTaskQwenFourMvbStateChange & \sbTaskQwenSevenTwoFiveMvbStateChange & \sbTaskLiveCcMvbStateChange & \sbTaskStreamingVlmMvbStateChange & \sbTaskTimeChatMvbStateChange & \sbTaskQwenEightMvbStateChange & \sbTaskTeacherMvbStateChange \\
Character order & \sbTaskNMvbCharacterOrder & \sbTaskFloorMvbCharacterOrder & \sbTaskOursMvbCharacterOrder & \sbTaskLfmSmallMvbCharacterOrder & \sbTaskSmolSmallMvbCharacterOrder & \sbTaskFastVlmMvbCharacterOrder & \sbTaskMobileMvbCharacterOrder & \sbTaskInternOneMvbCharacterOrder & \sbTaskInternOneIntMvbCharacterOrder & \sbTaskLfmBigMvbCharacterOrder & \sbTaskQwenTwoMvbCharacterOrder & \sbTaskQwenTwoIntMvbCharacterOrder & \sbTaskSmolBigMvbCharacterOrder & \sbTaskInternTwoMvbCharacterOrder & \sbTaskQwenFourMvbCharacterOrder & \sbTaskQwenSevenTwoFiveMvbCharacterOrder & \sbTaskLiveCcMvbCharacterOrder & \sbTaskStreamingVlmMvbCharacterOrder & \sbTaskTimeChatMvbCharacterOrder & \sbTaskQwenEightMvbCharacterOrder & \sbTaskTeacherMvbCharacterOrder \\
Action localization & \sbTaskNMvbActionLocalization & \sbTaskFloorMvbActionLocalization & \sbTaskOursMvbActionLocalization & \sbTaskLfmSmallMvbActionLocalization & \sbTaskSmolSmallMvbActionLocalization & \sbTaskFastVlmMvbActionLocalization & \sbTaskMobileMvbActionLocalization & \sbTaskInternOneMvbActionLocalization & \sbTaskInternOneIntMvbActionLocalization & \sbTaskLfmBigMvbActionLocalization & \sbTaskQwenTwoMvbActionLocalization & \sbTaskQwenTwoIntMvbActionLocalization & \sbTaskSmolBigMvbActionLocalization & \sbTaskInternTwoMvbActionLocalization & \sbTaskQwenFourMvbActionLocalization & \sbTaskQwenSevenTwoFiveMvbActionLocalization & \sbTaskLiveCcMvbActionLocalization & \sbTaskStreamingVlmMvbActionLocalization & \sbTaskTimeChatMvbActionLocalization & \sbTaskQwenEightMvbActionLocalization & \sbTaskTeacherMvbActionLocalization \\
Scene transition & \sbTaskNMvbSceneTransition & \sbTaskFloorMvbSceneTransition & \sbTaskOursMvbSceneTransition & \sbTaskLfmSmallMvbSceneTransition & \sbTaskSmolSmallMvbSceneTransition & \sbTaskFastVlmMvbSceneTransition & \sbTaskMobileMvbSceneTransition & \sbTaskInternOneMvbSceneTransition & \sbTaskInternOneIntMvbSceneTransition & \sbTaskLfmBigMvbSceneTransition & \sbTaskQwenTwoMvbSceneTransition & \sbTaskQwenTwoIntMvbSceneTransition & \sbTaskSmolBigMvbSceneTransition & \sbTaskInternTwoMvbSceneTransition & \sbTaskQwenFourMvbSceneTransition & \sbTaskQwenSevenTwoFiveMvbSceneTransition & \sbTaskLiveCcMvbSceneTransition & \sbTaskStreamingVlmMvbSceneTransition & \sbTaskTimeChatMvbSceneTransition & \sbTaskQwenEightMvbSceneTransition & \sbTaskTeacherMvbSceneTransition \\
Egocentric navigation & \sbTaskNMvbEgocentricNavigation & \sbTaskFloorMvbEgocentricNavigation & \sbTaskOursMvbEgocentricNavigation & \sbTaskLfmSmallMvbEgocentricNavigation & \sbTaskSmolSmallMvbEgocentricNavigation & \sbTaskFastVlmMvbEgocentricNavigation & \sbTaskMobileMvbEgocentricNavigation & \sbTaskInternOneMvbEgocentricNavigation & \sbTaskInternOneIntMvbEgocentricNavigation & \sbTaskLfmBigMvbEgocentricNavigation & \sbTaskQwenTwoMvbEgocentricNavigation & \sbTaskQwenTwoIntMvbEgocentricNavigation & \sbTaskSmolBigMvbEgocentricNavigation & \sbTaskInternTwoMvbEgocentricNavigation & \sbTaskQwenFourMvbEgocentricNavigation & \sbTaskQwenSevenTwoFiveMvbEgocentricNavigation & \sbTaskLiveCcMvbEgocentricNavigation & \sbTaskStreamingVlmMvbEgocentricNavigation & \sbTaskTimeChatMvbEgocentricNavigation & \sbTaskQwenEightMvbEgocentricNavigation & \sbTaskTeacherMvbEgocentricNavigation \\
Episodic reasoning & \sbTaskNMvbEpisodicReasoning & \sbTaskFloorMvbEpisodicReasoning & \sbTaskOursMvbEpisodicReasoning & \sbTaskLfmSmallMvbEpisodicReasoning & \sbTaskSmolSmallMvbEpisodicReasoning & \sbTaskFastVlmMvbEpisodicReasoning & \sbTaskMobileMvbEpisodicReasoning & \sbTaskInternOneMvbEpisodicReasoning & \sbTaskInternOneIntMvbEpisodicReasoning & \sbTaskLfmBigMvbEpisodicReasoning & \sbTaskQwenTwoMvbEpisodicReasoning & \sbTaskQwenTwoIntMvbEpisodicReasoning & \sbTaskSmolBigMvbEpisodicReasoning & \sbTaskInternTwoMvbEpisodicReasoning & \sbTaskQwenFourMvbEpisodicReasoning & \sbTaskQwenSevenTwoFiveMvbEpisodicReasoning & \sbTaskLiveCcMvbEpisodicReasoning & \sbTaskStreamingVlmMvbEpisodicReasoning & \sbTaskTimeChatMvbEpisodicReasoning & \sbTaskQwenEightMvbEpisodicReasoning & \sbTaskTeacherMvbEpisodicReasoning \\
\midrule
\multicolumn{22}{@{}l}{\emph{StreamingBench Real-Time}} \\
Object Perception & \sbTaskNSbObjectPerception & \sbTaskFloorSbObjectPerception & \sbTaskOursSbObjectPerception & \sbTaskLfmSmallSbObjectPerception & \sbTaskSmolSmallSbObjectPerception & \sbTaskFastVlmSbObjectPerception & \sbTaskMobileSbObjectPerception & \sbTaskInternOneSbObjectPerception & \sbTaskInternOneIntSbObjectPerception & \sbTaskLfmBigSbObjectPerception & \sbTaskQwenTwoSbObjectPerception & \sbTaskQwenTwoIntSbObjectPerception & \sbTaskSmolBigSbObjectPerception & \sbTaskInternTwoSbObjectPerception & \sbTaskQwenFourSbObjectPerception & \sbTaskQwenSevenTwoFiveSbObjectPerception & \sbTaskLiveCcSbObjectPerception & \sbTaskStreamingVlmSbObjectPerception & \sbTaskTimeChatSbObjectPerception & \sbTaskQwenEightSbObjectPerception & \sbTaskTeacherSbObjectPerception \\
Action Perception & \sbTaskNSbActionPerception & \sbTaskFloorSbActionPerception & \sbTaskOursSbActionPerception & \sbTaskLfmSmallSbActionPerception & \sbTaskSmolSmallSbActionPerception & \sbTaskFastVlmSbActionPerception & \sbTaskMobileSbActionPerception & \sbTaskInternOneSbActionPerception & \sbTaskInternOneIntSbActionPerception & \sbTaskLfmBigSbActionPerception & \sbTaskQwenTwoSbActionPerception & \sbTaskQwenTwoIntSbActionPerception & \sbTaskSmolBigSbActionPerception & \sbTaskInternTwoSbActionPerception & \sbTaskQwenFourSbActionPerception & \sbTaskQwenSevenTwoFiveSbActionPerception & \sbTaskLiveCcSbActionPerception & \sbTaskStreamingVlmSbActionPerception & \sbTaskTimeChatSbActionPerception & \sbTaskQwenEightSbActionPerception & \sbTaskTeacherSbActionPerception \\
Text-Rich Understanding & \sbTaskNSbTextRichUnderstanding & \sbTaskFloorSbTextRichUnderstanding & \sbTaskOursSbTextRichUnderstanding & \sbTaskLfmSmallSbTextRichUnderstanding & \sbTaskSmolSmallSbTextRichUnderstanding & \sbTaskFastVlmSbTextRichUnderstanding & \sbTaskMobileSbTextRichUnderstanding & \sbTaskInternOneSbTextRichUnderstanding & \sbTaskInternOneIntSbTextRichUnderstanding & \sbTaskLfmBigSbTextRichUnderstanding & \sbTaskQwenTwoSbTextRichUnderstanding & \sbTaskQwenTwoIntSbTextRichUnderstanding & \sbTaskSmolBigSbTextRichUnderstanding & \sbTaskInternTwoSbTextRichUnderstanding & \sbTaskQwenFourSbTextRichUnderstanding & \sbTaskQwenSevenTwoFiveSbTextRichUnderstanding & \sbTaskLiveCcSbTextRichUnderstanding & \sbTaskStreamingVlmSbTextRichUnderstanding & \sbTaskTimeChatSbTextRichUnderstanding & \sbTaskQwenEightSbTextRichUnderstanding & \sbTaskTeacherSbTextRichUnderstanding \\
Clips Summarize & \sbTaskNSbClipsSummarize & \sbTaskFloorSbClipsSummarize & \sbTaskOursSbClipsSummarize & \sbTaskLfmSmallSbClipsSummarize & \sbTaskSmolSmallSbClipsSummarize & \sbTaskFastVlmSbClipsSummarize & \sbTaskMobileSbClipsSummarize & \sbTaskInternOneSbClipsSummarize & \sbTaskInternOneIntSbClipsSummarize & \sbTaskLfmBigSbClipsSummarize & \sbTaskQwenTwoSbClipsSummarize & \sbTaskQwenTwoIntSbClipsSummarize & \sbTaskSmolBigSbClipsSummarize & \sbTaskInternTwoSbClipsSummarize & \sbTaskQwenFourSbClipsSummarize & \sbTaskQwenSevenTwoFiveSbClipsSummarize & \sbTaskLiveCcSbClipsSummarize & \sbTaskStreamingVlmSbClipsSummarize & \sbTaskTimeChatSbClipsSummarize & \sbTaskQwenEightSbClipsSummarize & \sbTaskTeacherSbClipsSummarize \\
Attribute Perception & \sbTaskNSbAttributePerception & \sbTaskFloorSbAttributePerception & \sbTaskOursSbAttributePerception & \sbTaskLfmSmallSbAttributePerception & \sbTaskSmolSmallSbAttributePerception & \sbTaskFastVlmSbAttributePerception & \sbTaskMobileSbAttributePerception & \sbTaskInternOneSbAttributePerception & \sbTaskInternOneIntSbAttributePerception & \sbTaskLfmBigSbAttributePerception & \sbTaskQwenTwoSbAttributePerception & \sbTaskQwenTwoIntSbAttributePerception & \sbTaskSmolBigSbAttributePerception & \sbTaskInternTwoSbAttributePerception & \sbTaskQwenFourSbAttributePerception & \sbTaskQwenSevenTwoFiveSbAttributePerception & \sbTaskLiveCcSbAttributePerception & \sbTaskStreamingVlmSbAttributePerception & \sbTaskTimeChatSbAttributePerception & \sbTaskQwenEightSbAttributePerception & \sbTaskTeacherSbAttributePerception \\
Spatial Understanding & \sbTaskNSbSpatialUnderstanding & \sbTaskFloorSbSpatialUnderstanding & \sbTaskOursSbSpatialUnderstanding & \sbTaskLfmSmallSbSpatialUnderstanding & \sbTaskSmolSmallSbSpatialUnderstanding & \sbTaskFastVlmSbSpatialUnderstanding & \sbTaskMobileSbSpatialUnderstanding & \sbTaskInternOneSbSpatialUnderstanding & \sbTaskInternOneIntSbSpatialUnderstanding & \sbTaskLfmBigSbSpatialUnderstanding & \sbTaskQwenTwoSbSpatialUnderstanding & \sbTaskQwenTwoIntSbSpatialUnderstanding & \sbTaskSmolBigSbSpatialUnderstanding & \sbTaskInternTwoSbSpatialUnderstanding & \sbTaskQwenFourSbSpatialUnderstanding & \sbTaskQwenSevenTwoFiveSbSpatialUnderstanding & \sbTaskLiveCcSbSpatialUnderstanding & \sbTaskStreamingVlmSbSpatialUnderstanding & \sbTaskTimeChatSbSpatialUnderstanding & \sbTaskQwenEightSbSpatialUnderstanding & \sbTaskTeacherSbSpatialUnderstanding \\
Counting & \sbTaskNSbCounting & \sbTaskFloorSbCounting & \sbTaskOursSbCounting & \sbTaskLfmSmallSbCounting & \sbTaskSmolSmallSbCounting & \sbTaskFastVlmSbCounting & \sbTaskMobileSbCounting & \sbTaskInternOneSbCounting & \sbTaskInternOneIntSbCounting & \sbTaskLfmBigSbCounting & \sbTaskQwenTwoSbCounting & \sbTaskQwenTwoIntSbCounting & \sbTaskSmolBigSbCounting & \sbTaskInternTwoSbCounting & \sbTaskQwenFourSbCounting & \sbTaskQwenSevenTwoFiveSbCounting & \sbTaskLiveCcSbCounting & \sbTaskStreamingVlmSbCounting & \sbTaskTimeChatSbCounting & \sbTaskQwenEightSbCounting & \sbTaskTeacherSbCounting \\
Event Understanding & \sbTaskNSbEventUnderstanding & \sbTaskFloorSbEventUnderstanding & \sbTaskOursSbEventUnderstanding & \sbTaskLfmSmallSbEventUnderstanding & \sbTaskSmolSmallSbEventUnderstanding & \sbTaskFastVlmSbEventUnderstanding & \sbTaskMobileSbEventUnderstanding & \sbTaskInternOneSbEventUnderstanding & \sbTaskInternOneIntSbEventUnderstanding & \sbTaskLfmBigSbEventUnderstanding & \sbTaskQwenTwoSbEventUnderstanding & \sbTaskQwenTwoIntSbEventUnderstanding & \sbTaskSmolBigSbEventUnderstanding & \sbTaskInternTwoSbEventUnderstanding & \sbTaskQwenFourSbEventUnderstanding & \sbTaskQwenSevenTwoFiveSbEventUnderstanding & \sbTaskLiveCcSbEventUnderstanding & \sbTaskStreamingVlmSbEventUnderstanding & \sbTaskTimeChatSbEventUnderstanding & \sbTaskQwenEightSbEventUnderstanding & \sbTaskTeacherSbEventUnderstanding \\
Causal Reasoning & \sbTaskNSbCausalReasoning & \sbTaskFloorSbCausalReasoning & \sbTaskOursSbCausalReasoning & \sbTaskLfmSmallSbCausalReasoning & \sbTaskSmolSmallSbCausalReasoning & \sbTaskFastVlmSbCausalReasoning & \sbTaskMobileSbCausalReasoning & \sbTaskInternOneSbCausalReasoning & \sbTaskInternOneIntSbCausalReasoning & \sbTaskLfmBigSbCausalReasoning & \sbTaskQwenTwoSbCausalReasoning & \sbTaskQwenTwoIntSbCausalReasoning & \sbTaskSmolBigSbCausalReasoning & \sbTaskInternTwoSbCausalReasoning & \sbTaskQwenFourSbCausalReasoning & \sbTaskQwenSevenTwoFiveSbCausalReasoning & \sbTaskLiveCcSbCausalReasoning & \sbTaskStreamingVlmSbCausalReasoning & \sbTaskTimeChatSbCausalReasoning & \sbTaskQwenEightSbCausalReasoning & \sbTaskTeacherSbCausalReasoning \\
Prospective Reasoning & \sbTaskNSbProspectiveReasoning & \sbTaskFloorSbProspectiveReasoning & \sbTaskOursSbProspectiveReasoning & \sbTaskLfmSmallSbProspectiveReasoning & \sbTaskSmolSmallSbProspectiveReasoning & \sbTaskFastVlmSbProspectiveReasoning & \sbTaskMobileSbProspectiveReasoning & \sbTaskInternOneSbProspectiveReasoning & \sbTaskInternOneIntSbProspectiveReasoning & \sbTaskLfmBigSbProspectiveReasoning & \sbTaskQwenTwoSbProspectiveReasoning & \sbTaskQwenTwoIntSbProspectiveReasoning & \sbTaskSmolBigSbProspectiveReasoning & \sbTaskInternTwoSbProspectiveReasoning & \sbTaskQwenFourSbProspectiveReasoning & \sbTaskQwenSevenTwoFiveSbProspectiveReasoning & \sbTaskLiveCcSbProspectiveReasoning & \sbTaskStreamingVlmSbProspectiveReasoning & \sbTaskTimeChatSbProspectiveReasoning & \sbTaskQwenEightSbProspectiveReasoning & \sbTaskTeacherSbProspectiveReasoning \\
\midrule
\multicolumn{22}{@{}l}{\emph{Video-MME short}} \\
Object Recognition & \sbTaskNVmObjectRecognition & \sbTaskFloorVmObjectRecognition & \sbTaskOursVmObjectRecognition & \sbTaskLfmSmallVmObjectRecognition & \sbTaskSmolSmallVmObjectRecognition & \sbTaskFastVlmVmObjectRecognition & \sbTaskMobileVmObjectRecognition & \sbTaskInternOneVmObjectRecognition & \sbTaskInternOneIntVmObjectRecognition & \sbTaskLfmBigVmObjectRecognition & \sbTaskQwenTwoVmObjectRecognition & \sbTaskQwenTwoIntVmObjectRecognition & \sbTaskSmolBigVmObjectRecognition & \sbTaskInternTwoVmObjectRecognition & \sbTaskQwenFourVmObjectRecognition & \sbTaskQwenSevenTwoFiveVmObjectRecognition & \sbTaskLiveCcVmObjectRecognition & \sbTaskStreamingVlmVmObjectRecognition & \sbTaskTimeChatVmObjectRecognition & \sbTaskQwenEightVmObjectRecognition & \sbTaskTeacherVmObjectRecognition \\
Action Recognition & \sbTaskNVmActionRecognition & \sbTaskFloorVmActionRecognition & \sbTaskOursVmActionRecognition & \sbTaskLfmSmallVmActionRecognition & \sbTaskSmolSmallVmActionRecognition & \sbTaskFastVlmVmActionRecognition & \sbTaskMobileVmActionRecognition & \sbTaskInternOneVmActionRecognition & \sbTaskInternOneIntVmActionRecognition & \sbTaskLfmBigVmActionRecognition & \sbTaskQwenTwoVmActionRecognition & \sbTaskQwenTwoIntVmActionRecognition & \sbTaskSmolBigVmActionRecognition & \sbTaskInternTwoVmActionRecognition & \sbTaskQwenFourVmActionRecognition & \sbTaskQwenSevenTwoFiveVmActionRecognition & \sbTaskLiveCcVmActionRecognition & \sbTaskStreamingVlmVmActionRecognition & \sbTaskTimeChatVmActionRecognition & \sbTaskQwenEightVmActionRecognition & \sbTaskTeacherVmActionRecognition \\
Counting Problem & \sbTaskNVmCountingProblem & \sbTaskFloorVmCountingProblem & \sbTaskOursVmCountingProblem & \sbTaskLfmSmallVmCountingProblem & \sbTaskSmolSmallVmCountingProblem & \sbTaskFastVlmVmCountingProblem & \sbTaskMobileVmCountingProblem & \sbTaskInternOneVmCountingProblem & \sbTaskInternOneIntVmCountingProblem & \sbTaskLfmBigVmCountingProblem & \sbTaskQwenTwoVmCountingProblem & \sbTaskQwenTwoIntVmCountingProblem & \sbTaskSmolBigVmCountingProblem & \sbTaskInternTwoVmCountingProblem & \sbTaskQwenFourVmCountingProblem & \sbTaskQwenSevenTwoFiveVmCountingProblem & \sbTaskLiveCcVmCountingProblem & \sbTaskStreamingVlmVmCountingProblem & \sbTaskTimeChatVmCountingProblem & \sbTaskQwenEightVmCountingProblem & \sbTaskTeacherVmCountingProblem \\
Attribute Perception & \sbTaskNVmAttributePerception & \sbTaskFloorVmAttributePerception & \sbTaskOursVmAttributePerception & \sbTaskLfmSmallVmAttributePerception & \sbTaskSmolSmallVmAttributePerception & \sbTaskFastVlmVmAttributePerception & \sbTaskMobileVmAttributePerception & \sbTaskInternOneVmAttributePerception & \sbTaskInternOneIntVmAttributePerception & \sbTaskLfmBigVmAttributePerception & \sbTaskQwenTwoVmAttributePerception & \sbTaskQwenTwoIntVmAttributePerception & \sbTaskSmolBigVmAttributePerception & \sbTaskInternTwoVmAttributePerception & \sbTaskQwenFourVmAttributePerception & \sbTaskQwenSevenTwoFiveVmAttributePerception & \sbTaskLiveCcVmAttributePerception & \sbTaskStreamingVlmVmAttributePerception & \sbTaskTimeChatVmAttributePerception & \sbTaskQwenEightVmAttributePerception & \sbTaskTeacherVmAttributePerception \\
Information Synopsis & \sbTaskNVmInformationSynopsis & \sbTaskFloorVmInformationSynopsis & \sbTaskOursVmInformationSynopsis & \sbTaskLfmSmallVmInformationSynopsis & \sbTaskSmolSmallVmInformationSynopsis & \sbTaskFastVlmVmInformationSynopsis & \sbTaskMobileVmInformationSynopsis & \sbTaskInternOneVmInformationSynopsis & \sbTaskInternOneIntVmInformationSynopsis & \sbTaskLfmBigVmInformationSynopsis & \sbTaskQwenTwoVmInformationSynopsis & \sbTaskQwenTwoIntVmInformationSynopsis & \sbTaskSmolBigVmInformationSynopsis & \sbTaskInternTwoVmInformationSynopsis & \sbTaskQwenFourVmInformationSynopsis & \sbTaskQwenSevenTwoFiveVmInformationSynopsis & \sbTaskLiveCcVmInformationSynopsis & \sbTaskStreamingVlmVmInformationSynopsis & \sbTaskTimeChatVmInformationSynopsis & \sbTaskQwenEightVmInformationSynopsis & \sbTaskTeacherVmInformationSynopsis \\
Object Reasoning & \sbTaskNVmObjectReasoning & \sbTaskFloorVmObjectReasoning & \sbTaskOursVmObjectReasoning & \sbTaskLfmSmallVmObjectReasoning & \sbTaskSmolSmallVmObjectReasoning & \sbTaskFastVlmVmObjectReasoning & \sbTaskMobileVmObjectReasoning & \sbTaskInternOneVmObjectReasoning & \sbTaskInternOneIntVmObjectReasoning & \sbTaskLfmBigVmObjectReasoning & \sbTaskQwenTwoVmObjectReasoning & \sbTaskQwenTwoIntVmObjectReasoning & \sbTaskSmolBigVmObjectReasoning & \sbTaskInternTwoVmObjectReasoning & \sbTaskQwenFourVmObjectReasoning & \sbTaskQwenSevenTwoFiveVmObjectReasoning & \sbTaskLiveCcVmObjectReasoning & \sbTaskStreamingVlmVmObjectReasoning & \sbTaskTimeChatVmObjectReasoning & \sbTaskQwenEightVmObjectReasoning & \sbTaskTeacherVmObjectReasoning \\
OCR Problems & \sbTaskNVmOcrProblems & \sbTaskFloorVmOcrProblems & \sbTaskOursVmOcrProblems & \sbTaskLfmSmallVmOcrProblems & \sbTaskSmolSmallVmOcrProblems & \sbTaskFastVlmVmOcrProblems & \sbTaskMobileVmOcrProblems & \sbTaskInternOneVmOcrProblems & \sbTaskInternOneIntVmOcrProblems & \sbTaskLfmBigVmOcrProblems & \sbTaskQwenTwoVmOcrProblems & \sbTaskQwenTwoIntVmOcrProblems & \sbTaskSmolBigVmOcrProblems & \sbTaskInternTwoVmOcrProblems & \sbTaskQwenFourVmOcrProblems & \sbTaskQwenSevenTwoFiveVmOcrProblems & \sbTaskLiveCcVmOcrProblems & \sbTaskStreamingVlmVmOcrProblems & \sbTaskTimeChatVmOcrProblems & \sbTaskQwenEightVmOcrProblems & \sbTaskTeacherVmOcrProblems \\
Action Reasoning & \sbTaskNVmActionReasoning & \sbTaskFloorVmActionReasoning & \sbTaskOursVmActionReasoning & \sbTaskLfmSmallVmActionReasoning & \sbTaskSmolSmallVmActionReasoning & \sbTaskFastVlmVmActionReasoning & \sbTaskMobileVmActionReasoning & \sbTaskInternOneVmActionReasoning & \sbTaskInternOneIntVmActionReasoning & \sbTaskLfmBigVmActionReasoning & \sbTaskQwenTwoVmActionReasoning & \sbTaskQwenTwoIntVmActionReasoning & \sbTaskSmolBigVmActionReasoning & \sbTaskInternTwoVmActionReasoning & \sbTaskQwenFourVmActionReasoning & \sbTaskQwenSevenTwoFiveVmActionReasoning & \sbTaskLiveCcVmActionReasoning & \sbTaskStreamingVlmVmActionReasoning & \sbTaskTimeChatVmActionReasoning & \sbTaskQwenEightVmActionReasoning & \sbTaskTeacherVmActionReasoning \\
Spatial Perception & \sbTaskNVmSpatialPerception & \sbTaskFloorVmSpatialPerception & \sbTaskOursVmSpatialPerception & \sbTaskLfmSmallVmSpatialPerception & \sbTaskSmolSmallVmSpatialPerception & \sbTaskFastVlmVmSpatialPerception & \sbTaskMobileVmSpatialPerception & \sbTaskInternOneVmSpatialPerception & \sbTaskInternOneIntVmSpatialPerception & \sbTaskLfmBigVmSpatialPerception & \sbTaskQwenTwoVmSpatialPerception & \sbTaskQwenTwoIntVmSpatialPerception & \sbTaskSmolBigVmSpatialPerception & \sbTaskInternTwoVmSpatialPerception & \sbTaskQwenFourVmSpatialPerception & \sbTaskQwenSevenTwoFiveVmSpatialPerception & \sbTaskLiveCcVmSpatialPerception & \sbTaskStreamingVlmVmSpatialPerception & \sbTaskTimeChatVmSpatialPerception & \sbTaskQwenEightVmSpatialPerception & \sbTaskTeacherVmSpatialPerception \\
Spatial Reasoning & \sbTaskNVmSpatialReasoning & \sbTaskFloorVmSpatialReasoning & \sbTaskOursVmSpatialReasoning & \sbTaskLfmSmallVmSpatialReasoning & \sbTaskSmolSmallVmSpatialReasoning & \sbTaskFastVlmVmSpatialReasoning & \sbTaskMobileVmSpatialReasoning & \sbTaskInternOneVmSpatialReasoning & \sbTaskInternOneIntVmSpatialReasoning & \sbTaskLfmBigVmSpatialReasoning & \sbTaskQwenTwoVmSpatialReasoning & \sbTaskQwenTwoIntVmSpatialReasoning & \sbTaskSmolBigVmSpatialReasoning & \sbTaskInternTwoVmSpatialReasoning & \sbTaskQwenFourVmSpatialReasoning & \sbTaskQwenSevenTwoFiveVmSpatialReasoning & \sbTaskLiveCcVmSpatialReasoning & \sbTaskStreamingVlmVmSpatialReasoning & \sbTaskTimeChatVmSpatialReasoning & \sbTaskQwenEightVmSpatialReasoning & \sbTaskTeacherVmSpatialReasoning \\
Temporal Perception & \sbTaskNVmTemporalPerception & \sbTaskFloorVmTemporalPerception & \sbTaskOursVmTemporalPerception & \sbTaskLfmSmallVmTemporalPerception & \sbTaskSmolSmallVmTemporalPerception & \sbTaskFastVlmVmTemporalPerception & \sbTaskMobileVmTemporalPerception & \sbTaskInternOneVmTemporalPerception & \sbTaskInternOneIntVmTemporalPerception & \sbTaskLfmBigVmTemporalPerception & \sbTaskQwenTwoVmTemporalPerception & \sbTaskQwenTwoIntVmTemporalPerception & \sbTaskSmolBigVmTemporalPerception & \sbTaskInternTwoVmTemporalPerception & \sbTaskQwenFourVmTemporalPerception & \sbTaskQwenSevenTwoFiveVmTemporalPerception & \sbTaskLiveCcVmTemporalPerception & \sbTaskStreamingVlmVmTemporalPerception & \sbTaskTimeChatVmTemporalPerception & \sbTaskQwenEightVmTemporalPerception & \sbTaskTeacherVmTemporalPerception \\
Temporal Reasoning & \sbTaskNVmTemporalReasoning & \sbTaskFloorVmTemporalReasoning & \sbTaskOursVmTemporalReasoning & \sbTaskLfmSmallVmTemporalReasoning & \sbTaskSmolSmallVmTemporalReasoning & \sbTaskFastVlmVmTemporalReasoning & \sbTaskMobileVmTemporalReasoning & \sbTaskInternOneVmTemporalReasoning & \sbTaskInternOneIntVmTemporalReasoning & \sbTaskLfmBigVmTemporalReasoning & \sbTaskQwenTwoVmTemporalReasoning & \sbTaskQwenTwoIntVmTemporalReasoning & \sbTaskSmolBigVmTemporalReasoning & \sbTaskInternTwoVmTemporalReasoning & \sbTaskQwenFourVmTemporalReasoning & \sbTaskQwenSevenTwoFiveVmTemporalReasoning & \sbTaskLiveCcVmTemporalReasoning & \sbTaskStreamingVlmVmTemporalReasoning & \sbTaskTimeChatVmTemporalReasoning & \sbTaskQwenEightVmTemporalReasoning & \sbTaskTeacherVmTemporalReasoning \\
\bottomrule
\end{tabular}

%% file: arxiv.bbl
\begin{thebibliography}{85}
\providecommand{\natexlab}[1]{#1}
\providecommand{\url}[1]{\texttt{#1}}
\expandafter\ifx\csname urlstyle\endcsname\relax
  \providecommand{\doi}[1]{doi: #1}\else
  \providecommand{\doi}{doi: \begingroup \urlstyle{rm}\Url}\fi

\bibitem[Abdin et~al.(2024)Abdin, Aneja, Behl, Bubeck, Eldan, Gunasekar,
  Harrison, Hewett, Javaheripi, Kauffmann, et~al.]{phi4}
Marah Abdin, Jyoti Aneja, Harkirat Behl, Sébastien Bubeck, Ronen Eldan, Suriya
  Gunasekar, Michael Harrison, Russell~J. Hewett, Mojan Javaheripi, Piero
  Kauffmann, et~al.
\newblock Phi-4 technical report.
\newblock \emph{arXiv preprint arXiv:2412.08905}, 2024.

\bibitem[Agarwal et~al.(2024)Agarwal, Vieillard, Zhou, Stanczyk, Ramos, Geist,
  and Bachem]{gkd2024}
Rishabh Agarwal, Nino Vieillard, Yongchao Zhou, Piotr Stanczyk, Sabela Ramos,
  Matthieu Geist, and Olivier Bachem.
\newblock On-policy distillation of language models: Learning from
  self-generated mistakes.
\newblock In \emph{ICLR}, 2024.
\newblock arXiv:2306.13649.

\bibitem[Agrawal et~al.(2023)Agrawal, Panwar, Mohan, Kwatra, Gulavani, and
  Ramjee]{sarathi2023}
Amey Agrawal, Ashish Panwar, Jayashree Mohan, Nipun Kwatra, Bhargav~S.
  Gulavani, and Ramachandran Ramjee.
\newblock {SARATHI}: Efficient {LLM} inference by piggybacking decodes with
  chunked prefills.
\newblock \emph{arXiv preprint arXiv:2308.16369}, 2023.

\bibitem[Agrawal et~al.(2024)Agrawal, Kedia, Panwar, Mohan, Kwatra, Gulavani,
  Tumanov, and Ramjee]{sarathiserve2024}
Amey Agrawal, Nitin Kedia, Ashish Panwar, Jayashree Mohan, Nipun Kwatra,
  Bhargav~S. Gulavani, Alexey Tumanov, and Ramachandran Ramjee.
\newblock Taming {Throughput-Latency} tradeoff in {LLM} inference with
  {Sarathi-Serve}.
\newblock In \emph{OSDI}, 2024.
\newblock arXiv:2403.02310.

\bibitem[Ainslie et~al.(2023)Ainslie, Lee-Thorp, de~Jong, Zemlyanskiy,
  Lebr{\'o}n, and Sanghai]{gqa2023}
Joshua Ainslie, James Lee-Thorp, Michiel de Jong, Yury Zemlyanskiy, Federico
  Lebr{\'o}n, and Sumit Sanghai.
\newblock {GQA}: Training generalized multi-query transformer models from
  multi-head checkpoints.
\newblock In \emph{EMNLP}, 2023.
\newblock arXiv:2305.13245.

\bibitem[Amini et~al.(2025)Amini, Banaszak, Benoit, Böök, Dakhran, Duong,
  Eng, Fernandes, Härkönen, Harrington, et~al.]{lfm2}
Alexander Amini, Anna Banaszak, Harold Benoit, Arthur Böök, Tarek Dakhran,
  Song Duong, Alfred Eng, Fernando Fernandes, Marc Härkönen, Anne Harrington,
  et~al.
\newblock {LFM2} technical report.
\newblock \emph{arXiv preprint arXiv:2511.23404}, 2025.

\bibitem[Bai et~al.(2025{\natexlab{a}})Bai, Cai, Chen, Chen, Chen, Cheng, Deng,
  Ding, Gao, Ge, et~al.]{qwen3vl2025}
Shuai Bai, Yuxuan Cai, Ruizhe Chen, Keqin Chen, Xionghui Chen, Zesen Cheng,
  Lianghao Deng, Wei Ding, Chang Gao, Chunjiang Ge, et~al.
\newblock {Qwen3-VL} technical report.
\newblock \emph{arXiv preprint arXiv:2511.21631}, 2025{\natexlab{a}}.

\bibitem[Bai et~al.(2025{\natexlab{b}})Bai, Chen, Liu, Wang, Ge, Song, Dang,
  Wang, Wang, Tang, et~al.]{qwen25vl2025}
Shuai Bai, Keqin Chen, Xuejing Liu, Jialin Wang, Wenbin Ge, Sibo Song, Kai
  Dang, Peng Wang, Shijie Wang, Jun Tang, et~al.
\newblock {Qwen2.5-VL} technical report.
\newblock \emph{arXiv preprint arXiv:2502.13923}, 2025{\natexlab{b}}.

\bibitem[Beyer et~al.(2022)Beyer, Zhai, Royer, Markeeva, Anil, and
  Kolesnikov]{goodteacher2022}
Lucas Beyer, Xiaohua Zhai, Amélie Royer, Larisa Markeeva, Rohan Anil, and
  Alexander Kolesnikov.
\newblock Knowledge distillation: A good teacher is patient and consistent.
\newblock In \emph{CVPR}, 2022.
\newblock arXiv:2106.05237.

\bibitem[Cai et~al.(2025{\natexlab{a}})Cai, Yang, Gao, and Lee]{matryoshka2024}
Mu Cai, Jianwei Yang, Jianfeng Gao, and Yong~Jae Lee.
\newblock Matryoshka multimodal models.
\newblock In \emph{ICLR}, 2025{\natexlab{a}}.
\newblock arXiv:2405.17430.

\bibitem[Cai et~al.(2025{\natexlab{b}})Cai, Zhang, He, He, Tong, Gan, Wang,
  Xue, Liu, and Bai]{llavakd2025}
Yuxuan Cai, Jiangning Zhang, Haoyang He, Xinwei He, Ao Tong, Zhenye Gan,
  Chengjie Wang, Zhucun Xue, Yong Liu, and Xiang Bai.
\newblock {LLaVA-KD}: A framework of distilling multimodal large language
  models.
\newblock In \emph{ICCV}, 2025{\natexlab{b}}.
\newblock arXiv:2410.16236.

\bibitem[Chai et~al.(2025)Chai, Song, Du, Meng, Madhavan, Bar-Tal, Hwang, Xie,
  and Manning]{auroracap2025}
Wenhao Chai, Enxin Song, Yilun Du, Chenlin Meng, Vashisht Madhavan, Omer
  Bar-Tal, Jenq-Neng Hwang, Saining Xie, and Christopher~D. Manning.
\newblock {AuroraCap}: Efficient, performant video detailed captioning and a
  new benchmark.
\newblock In \emph{ICLR}, 2025.
\newblock arXiv:2410.03051.

\bibitem[Chen et~al.(2024)Chen, Lv, Wu, Lin, Song, Gao, Liu, Gao, Mao, and
  Shou]{videollmonline2024}
Joya Chen, Zhaoyang Lv, Shiwei Wu, Kevin~Qinghong Lin, Chenan Song, Difei Gao,
  Jia-Wei Liu, Ziteng Gao, Dongxing Mao, and Mike~Zheng Shou.
\newblock {VideoLLM-online}: Online video large language model for streaming
  video.
\newblock In \emph{CVPR}, 2024.
\newblock arXiv:2406.11816.

\bibitem[Chen et~al.(2025)Chen, Zeng, Lin, Li, Ma, and Shou]{livecc2025}
Joya Chen, Ziyun Zeng, Yiqi Lin, Wei Li, Zejun Ma, and Mike~Zheng Shou.
\newblock {LiveCC}: Learning video {LLM} with streaming speech transcription at
  scale.
\newblock In \emph{CVPR}, 2025.
\newblock arXiv:2504.16030.

\bibitem[Chu et~al.(2023)Chu, Qiao, Lin, Xu, Yang, Hu, Wei, Zhang, Zhang, Wei,
  et~al.]{mobilevlm2023}
Xiangxiang Chu, Limeng Qiao, Xinyang Lin, Shuang Xu, Yang Yang, Yiming Hu, Fei
  Wei, Xinyu Zhang, Bo Zhang, Xiaolin Wei, et~al.
\newblock {MobileVLM} : A fast, strong and open vision language assistant for
  mobile devices.
\newblock \emph{arXiv preprint arXiv:2312.16886}, 2023.

\bibitem[Chu et~al.(2024)Chu, Qiao, Zhang, Xu, Wei, Yang, Sun, Hu, Lin, Zhang,
  et~al.]{mobilevlmv2}
Xiangxiang Chu, Limeng Qiao, Xinyu Zhang, Shuang Xu, Fei Wei, Yang Yang,
  Xiaofei Sun, Yiming Hu, Xinyang Lin, Bo Zhang, et~al.
\newblock {MobileVLM} v2: Faster and stronger baseline for vision language
  model.
\newblock \emph{arXiv preprint arXiv:2402.03766}, 2024.

\bibitem[Dehghani et~al.(2019)Dehghani, Gouws, Vinyals, Uszkoreit, and
  Kaiser]{universaltransformer2019}
Mostafa Dehghani, Stephan Gouws, Oriol Vinyals, Jakob Uszkoreit, and Łukasz
  Kaiser.
\newblock Universal transformers.
\newblock In \emph{ICLR}, 2019.
\newblock arXiv:1807.03819.

\bibitem[Di et~al.(2025)Di, Yu, Zhang, Li, Zhong, Cheng, Li, He, Shu, and
  Jiang]{rekv2025}
Shangzhe Di, Zhelun Yu, Guanghao Zhang, Haoyuan Li, Tao Zhong, Hao Cheng, Bolin
  Li, Wanggui He, Fangxun Shu, and Hao Jiang.
\newblock Streaming video {Question-Answering} with in-context video {KV-Cache}
  retrieval.
\newblock In \emph{ICLR}, 2025.
\newblock arXiv:2503.00540.

\bibitem[Dosovitskiy et~al.(2021)Dosovitskiy, Beyer, Kolesnikov, Weissenborn,
  Zhai, Unterthiner, Dehghani, Minderer, Heigold, Gelly, et~al.]{vit2021}
Alexey Dosovitskiy, Lucas Beyer, Alexander Kolesnikov, Dirk Weissenborn,
  Xiaohua Zhai, Thomas Unterthiner, Mostafa Dehghani, Matthias Minderer, Georg
  Heigold, Sylvain Gelly, et~al.
\newblock An image is worth 16x16 words: Transformers for image recognition at
  scale.
\newblock In \emph{ICLR}, 2021.
\newblock arXiv:2010.11929.

\bibitem[Efron and Tibshirani(1993)]{bootstrap1993}
Bradley Efron and Robert~J. Tibshirani.
\newblock \emph{An Introduction to the Bootstrap}.
\newblock Chapman \& Hall/CRC, 1993.

\bibitem[Fu et~al.(2025)Fu, Dai, Luo, Li, Ren, Zhang, Wang, Zhou, Shen, Zhang,
  et~al.]{videomme2024}
Chaoyou Fu, Yuhan Dai, Yongdong Luo, Lei Li, Shuhuai Ren, Renrui Zhang, Zihan
  Wang, Chenyu Zhou, Yunhang Shen, Mengdan Zhang, et~al.
\newblock {Video-MME}: The first-ever comprehensive evaluation benchmark of
  multi-modal {LLMs} in video analysis.
\newblock In \emph{CVPR}, 2025.
\newblock arXiv:2405.21075.

\bibitem[Gao et~al.(2026)Gao, Xiu, Luo, Cao, Cai, Li, Wu, and
  Zhang]{dircast2026}
Zixian Gao, Lixin Xiu, Penghao Luo, Xiangcheng Cao, Ziqi Cai, Zhen Li, Erwin
  Wu, and Kaipeng Zhang.
\newblock {DirCast}: Live esports directing and commentary with a single
  streaming {Video-Language} model.
\newblock \emph{arXiv preprint arXiv:2610.31734}, 2026.

\bibitem[Ge et~al.(2026)Ge, Wang, Wu, and Cai]{whatshouldremember2026}
Haonan Ge, Yiwei Wang, Hang Wu, and Yujun Cai.
\newblock What should a streaming video model remember?
\newblock In \emph{NeurIPS}, 2026.
\newblock arXiv:2606.16353.

\bibitem[Geiping et~al.(2025)Geiping, McLeish, Jain, Kirchenbauer, Singh,
  Bartoldson, Kailkhura, Bhatele, and Goldstein]{recurrentdepth2025}
Jonas Geiping, Sean McLeish, Neel Jain, John Kirchenbauer, Siddharth Singh,
  Brian~R. Bartoldson, Bhavya Kailkhura, Abhinav Bhatele, and Tom Goldstein.
\newblock Scaling up test-time compute with latent reasoning: A recurrent depth
  approach.
\newblock \emph{arXiv preprint arXiv:2502.05171}, 2025.

\bibitem[He et~al.(2016)He, Zhang, Ren, and Sun]{resnet2016}
Kaiming He, Xiangyu Zhang, Shaoqing Ren, and Jian Sun.
\newblock Deep residual learning for image recognition.
\newblock In \emph{CVPR}, 2016.
\newblock arXiv:1512.03385.

\bibitem[Hinton et~al.(2014)Hinton, Vinyals, and Dean]{kd2015}
Geoffrey Hinton, Oriol Vinyals, and Jeff Dean.
\newblock Distilling the knowledge in a neural network.
\newblock In \emph{NIPS 2014 Deep Learning Workshop}, 2014.
\newblock arXiv:1503.02531.

\bibitem[Ho et~al.(2022)Ho, Salimans, Gritsenko, Chan, Norouzi, and
  Fleet]{videodiffusion2022}
Jonathan Ho, Tim Salimans, Alexey Gritsenko, William Chan, Mohammad Norouzi,
  and David~J. Fleet.
\newblock Video diffusion models.
\newblock In \emph{NeurIPS}, 2022.
\newblock arXiv:2204.03458.

\bibitem[Hu et~al.(2022)Hu, Shen, Wallis, Allen-Zhu, Li, Wang, Wang, and
  Chen]{lora2022}
Edward~J. Hu, Yelong Shen, Phillip Wallis, Zeyuan Allen-Zhu, Yuanzhi Li, Shean
  Wang, Lu Wang, and Weizhu Chen.
\newblock {LoRA}: Low-rank adaptation of large language models.
\newblock In \emph{ICLR}, 2022.
\newblock arXiv:2106.09685.

\bibitem[Hu et~al.(2024)Hu, Dou, Li, Kamath, Peng, and Chang]{mqt2024}
Wenbo Hu, Zi-Yi Dou, Liunian~Harold Li, Amita Kamath, Nanyun Peng, and Kai-Wei
  Chang.
\newblock Matryoshka query transformer for large {Vision-Language} models.
\newblock In \emph{NeurIPS}, 2024.
\newblock arXiv:2405.19315.

\bibitem[Huang et~al.(2017)Huang, Rathod, Sun, Zhu, Korattikara, Fathi,
  Fischer, Wojna, Song, Guadarrama, et~al.]{speedacc2017}
Jonathan Huang, Vivek Rathod, Chen Sun, Menglong Zhu, Anoop Korattikara,
  Alireza Fathi, Ian Fischer, Zbigniew Wojna, Yang Song, Sergio Guadarrama,
  et~al.
\newblock Speed/accuracy trade-offs for modern convolutional object detectors.
\newblock In \emph{CVPR}, 2017.
\newblock arXiv:1611.10012.

\bibitem[Jin et~al.(2024)Jin, Li, Liu, Gu, Wu, Jiang, He, Zhao, Tan, Gan,
  et~al.]{effmllm2024}
Yizhang Jin, Jian Li, Yexin Liu, Tianjun Gu, Kai Wu, Zhengkai Jiang, Muyang He,
  Bo Zhao, Xin Tan, Zhenye Gan, et~al.
\newblock Efficient multimodal large language models: A survey.
\newblock \emph{arXiv preprint arXiv:2405.10739}, 2024.

\bibitem[Kann et~al.(2018)Kann, Rothe, and Filippova]{fluencyeval2018}
Katharina Kann, Sascha Rothe, and Katja Filippova.
\newblock Sentence-level fluency evaluation: References help, but can be
  spared!
\newblock In \emph{CoNLL}, 2018.

\bibitem[Kaplan et~al.(2020)Kaplan, McCandlish, Henighan, Brown, Chess, Child,
  Gray, Radford, Wu, and Amodei]{kaplan2020}
Jared Kaplan, Sam McCandlish, Tom Henighan, Tom~B. Brown, Benjamin Chess, Rewon
  Child, Scott Gray, Alec Radford, Jeffrey Wu, and Dario Amodei.
\newblock Scaling laws for neural language models.
\newblock \emph{arXiv preprint arXiv:2001.08361}, 2020.

\bibitem[Kim and Rush(2016)]{seqkd2016}
Yoon Kim and Alexander~M. Rush.
\newblock {Sequence-Level} knowledge distillation.
\newblock In \emph{EMNLP}, 2016.
\newblock arXiv:1606.07947.

\bibitem[Kwon et~al.(2023)Kwon, Li, Zhuang, Sheng, Zheng, Yu, Gonzalez, Zhang,
  and Stoica]{vllm2023}
Woosuk Kwon, Zhuohan Li, Siyuan Zhuang, Ying Sheng, Lianmin Zheng, Cody~Hao Yu,
  Joseph~E. Gonzalez, Hao Zhang, and Ion Stoica.
\newblock Efficient memory management for large language model serving with
  {PagedAttention}.
\newblock In \emph{SOSP}, 2023.
\newblock arXiv:2309.06180.

\bibitem[Lau et~al.(2017)Lau, Clark, and Lappin]{acceptability2017}
Jey~Han Lau, Alexander Clark, and Shalom Lappin.
\newblock Grammaticality, acceptability, and probability: A probabilistic view
  of linguistic knowledge.
\newblock \emph{Cognitive Science}, 41\penalty0 (5):\penalty0 1202--1241, 2017.

\bibitem[Li et~al.(2022)Li, Zhang, and He]{seqlenwarmup2021}
Conglong Li, Minjia Zhang, and Yuxiong He.
\newblock The {Stability-Efficiency} dilemma: Investigating sequence length
  warmup for training {GPT} models.
\newblock In \emph{NeurIPS}, 2022.
\newblock arXiv:2108.06084.

\bibitem[Li et~al.(2024)Li, Wang, He, Li, Wang, Liu, Wang, Xu, Chen, Luo,
  et~al.]{mvbench2024}
Kunchang Li, Yali Wang, Yinan He, Yizhuo Li, Yi Wang, Yi Liu, Zun Wang, Jilan
  Xu, Guo Chen, Ping Luo, et~al.
\newblock {MVBench}: A comprehensive multi-modal video understanding benchmark.
\newblock In \emph{CVPR}, 2024.
\newblock arXiv:2311.17005.

\bibitem[Li et~al.(2025)Li, Niu, Miao, Ge, Zhou, He, Dong, Duan, Ding, Qian,
  et~al.]{ovobench2025}
Yifei Li, Junbo Niu, Ziyang Miao, Chunjiang Ge, Yuanhang Zhou, Qihao He, Xiaoyi
  Dong, Haodong Duan, Shuangrui Ding, Rui Qian, et~al.
\newblock {OVO-Bench}: How far is your {Video-LLMs} from real-world online
  video understanding?
\newblock In \emph{CVPR}, 2025.
\newblock arXiv:2501.05510.

\bibitem[Lin et~al.(2024)Lin, Fang, Chen, Wan, Luo, Li, Liu, and
  Sun]{streamingbench2024}
Junming Lin, Zheng Fang, Chi Chen, Zihao Wan, Fuwen Luo, Peng Li, Yang Liu, and
  Maosong Sun.
\newblock {StreamingBench}: Assessing the gap for {MLLMs} to achieve streaming
  video understanding.
\newblock \emph{arXiv preprint arXiv:2411.03628}, 2024.

\bibitem[Liu et~al.(2023)Liu, Iter, Xu, Wang, Xu, and Zhu]{geval2023}
Yang Liu, Dan Iter, Yichong Xu, Shuohang Wang, Ruochen Xu, and Chenguang Zhu.
\newblock {G-Eval}: {NLG} evaluation using {GPT-4} with better human alignment.
\newblock In \emph{EMNLP}, 2023.
\newblock arXiv:2303.16634.

\bibitem[Loshchilov and Hutter(2019)]{adamw2019}
Ilya Loshchilov and Frank Hutter.
\newblock Decoupled weight decay regularization.
\newblock In \emph{ICLR}, 2019.
\newblock arXiv:1711.05101.

\bibitem[Luccioni et~al.(2024)Luccioni, Jernite, and Strubell]{powerhungry2024}
Alexandra~Sasha Luccioni, Yacine Jernite, and Emma Strubell.
\newblock Power hungry processing: Watts driving the cost of {AI} deployment?
\newblock In \emph{FAccT}, 2024.
\newblock arXiv:2311.16863.

\bibitem[Marafioti et~al.(2025)Marafioti, Zohar, Farré, Noyan, Bakouch,
  Cuenca, Zakka, Allal, Lozhkov, Tazi, et~al.]{smolvlm2025}
Andrés Marafioti, Orr Zohar, Miquel Farré, Merve Noyan, Elie Bakouch, Pedro
  Cuenca, Cyril Zakka, Loubna~Ben Allal, Anton Lozhkov, Nouamane Tazi, et~al.
\newblock {SmolVLM}: Redefining small and efficient multimodal models.
\newblock \emph{arXiv preprint arXiv:2504.05299}, 2025.

\bibitem[Min et~al.(2023)Min, Krishna, Lyu, Lewis, Yih, Koh, Iyyer,
  Zettlemoyer, and Hajishirzi]{factscore2023}
Sewon Min, Kalpesh Krishna, Xinxi Lyu, Mike Lewis, Wen-tau Yih, Pang~Wei Koh,
  Mohit Iyyer, Luke Zettlemoyer, and Hannaneh Hajishirzi.
\newblock {FActScore}: Fine-grained atomic evaluation of factual precision in
  long form text generation.
\newblock In \emph{EMNLP}, 2023.
\newblock arXiv:2305.14251.

\bibitem[Nan et~al.(2025)Nan, Xie, Zhou, Fan, Yang, Chen, Li, Yang, and
  Tai]{openvid2025}
Kepan Nan, Rui Xie, Penghao Zhou, Tiehan Fan, Zhenheng Yang, Zhijie Chen, Xiang
  Li, Jian Yang, and Ying Tai.
\newblock {OpenVid-1M}: A large-scale high-quality dataset for text-to-video
  generation.
\newblock In \emph{ICLR}, 2025.
\newblock arXiv:2407.02371.

\bibitem[OpenAI et~al.(2023)OpenAI, Achiam, Adler, Agarwal, Ahmad, Akkaya,
  Aleman, Almeida, Altenschmidt, Altman, et~al.]{gpt42023}
OpenAI, Josh Achiam, Steven Adler, Sandhini Agarwal, Lama Ahmad, Ilge Akkaya,
  Florencia~Leoni Aleman, Diogo Almeida, Janko Altenschmidt, Sam Altman, et~al.
\newblock {GPT-4} technical report.
\newblock \emph{arXiv preprint arXiv:2303.08774}, 2023.

\bibitem[{Qwen Team}(2026{\natexlab{a}})]{qwen35_9b}
{Qwen Team}.
\newblock {Qwen3.5-9B}.
\newblock \url{https://huggingface.co/Qwen/Qwen3.5-9B}, 2026{\natexlab{a}}.

\bibitem[{Qwen Team}(2026{\natexlab{b}})]{qwen38_27b}
{Qwen Team}.
\newblock {Qwen3.8-27B}.
\newblock \url{https://huggingface.co/Qwen/Qwen3.8-27B}, 2026{\natexlab{b}}.

\bibitem[Radford et~al.(2021)Radford, Kim, Hallacy, Ramesh, Goh, Agarwal,
  Sastry, Askell, Mishkin, Clark, et~al.]{clip2021}
Alec Radford, Jong~Wook Kim, Chris Hallacy, Aditya Ramesh, Gabriel Goh,
  Sandhini Agarwal, Girish Sastry, Amanda Askell, Pamela Mishkin, Jack Clark,
  et~al.
\newblock Learning transferable visual models from natural language
  supervision.
\newblock In \emph{ICML}, 2021.
\newblock arXiv:2103.00020.

\bibitem[Redmon et~al.(2016)Redmon, Divvala, Girshick, and Farhadi]{yolo2016}
Joseph Redmon, Santosh Divvala, Ross Girshick, and Ali Farhadi.
\newblock You only look once: Unified, {Real-Time} object detection.
\newblock In \emph{CVPR}, 2016.
\newblock arXiv:1506.02640.

\bibitem[Rombach et~al.(2022)Rombach, Blattmann, Lorenz, Esser, and
  Ommer]{ldm2022}
Robin Rombach, Andreas Blattmann, Dominik Lorenz, Patrick Esser, and Björn
  Ommer.
\newblock {High-Resolution} image synthesis with latent diffusion models.
\newblock In \emph{CVPR}, 2022.
\newblock arXiv:2112.10752.

\bibitem[Schulman et~al.(2017)Schulman, Wolski, Dhariwal, Radford, and
  Klimov]{ppo2017}
John Schulman, Filip Wolski, Prafulla Dhariwal, Alec Radford, and Oleg Klimov.
\newblock Proximal policy optimization algorithms.
\newblock \emph{arXiv preprint arXiv:1707.06347}, 2017.

\bibitem[Schwartz et~al.(2020)Schwartz, Dodge, Smith, and Etzioni]{greenai2019}
Roy Schwartz, Jesse Dodge, Noah~A. Smith, and Oren Etzioni.
\newblock Green {AI}.
\newblock \emph{Communications of the ACM}, 63\penalty0 (12):\penalty0 54--63,
  2020.
\newblock arXiv:1907.10597.

\bibitem[Shaker et~al.(2025)Shaker, Maaz, Gou, Rezatofighi, Khan, and
  Khan]{mobilevideogpt2025}
Abdelrahman Shaker, Muhammad Maaz, Chenhui Gou, Hamid Rezatofighi, Salman Khan,
  and Fahad~Shahbaz Khan.
\newblock {Mobile-VideoGPT}: Fast and accurate model for mobile video
  understanding.
\newblock \emph{arXiv preprint arXiv:2503.21782}, 2025.

\bibitem[Shao et~al.(2024)Shao, Wang, Zhu, Xu, Song, Bi, Zhang, Zhang, Li, Wu,
  et~al.]{grpo2024}
Zhihong Shao, Peiyi Wang, Qihao Zhu, Runxin Xu, Junxiao Song, Xiao Bi, Haowei
  Zhang, Mingchuan Zhang, Y.~K. Li, Y. Wu, et~al.
\newblock {DeepSeekMath}: Pushing the limits of mathematical reasoning in open
  language models.
\newblock \emph{arXiv preprint arXiv:2402.03300}, 2024.

\bibitem[Sharshar et~al.(2025)Sharshar, Khan, Ullah, and Guizani]{vlmedge2025}
Ahmed Sharshar, Latif~U. Khan, Waseem Ullah, and Mohsen Guizani.
\newblock {Vision-Language} models for edge networks: A comprehensive survey.
\newblock \emph{arXiv preprint arXiv:2502.07855}, 2025.

\bibitem[Shen et~al.(2026)Shen, Tian, Yang, and Liu]{simplestream2026}
Yujiao Shen, Shulin Tian, Jingkang Yang, and Ziwei Liu.
\newblock A simple baseline for streaming video understanding.
\newblock \emph{arXiv preprint arXiv:2604.02317}, 2026.

\bibitem[Shi et~al.(2016)Shi, Caballero, Husz{\'a}r, Totz, Aitken, Bishop,
  Rueckert, and Wang]{pixelshuffle2016}
Wenzhe Shi, Jose Caballero, Ferenc Husz{\'a}r, Johannes Totz, Andrew~P. Aitken,
  Rob Bishop, Daniel Rueckert, and Zehan Wang.
\newblock Real-time single image and video super-resolution using an efficient
  sub-pixel convolutional neural network.
\newblock In \emph{CVPR}, 2016.
\newblock arXiv:1609.05158.

\bibitem[Shinde et~al.(2025)Shinde, Ravi, Dey, Sakib, Rampure, and
  Roy]{effvlm2025}
Gaurav Shinde, Anuradha Ravi, Emon Dey, Shadman Sakib, Milind Rampure, and
  Nirmalya Roy.
\newblock A survey on efficient {Vision-Language} models.
\newblock \emph{arXiv preprint arXiv:2504.09724}, 2025.

\bibitem[Spravil et~al.(2026)Spravil, Houben, and Behnke]{strad2026}
Julian Spravil, Sebastian Houben, and Sven Behnke.
\newblock {StrAD}: A streaming method and benchmark for audio description
  generation for long-form videos.
\newblock \emph{arXiv preprint arXiv:2608.12549}, 2026.

\bibitem[Tami et~al.(2025)Tami, Elhenawy, and Ashqar]{hazardnet2025}
Mohammad~Abu Tami, Mohammed Elhenawy, and Huthaifa~I. Ashqar.
\newblock {HazardNet}: A {Small-Scale} vision language model for {Real-Time}
  traffic safety detection at edge devices.
\newblock \emph{arXiv preprint arXiv:2502.20572}, 2025.

\bibitem[Tan et~al.(2026)Tan, Tong, Yue, Wu, Fang, and Shen]{vicostream2026}
Yang Tan, Junlong Tong, Linan Yue, Hao Wu, Pengfei Fang, and Xiaoyu Shen.
\newblock {ViCoStream}: Streaming {VideoLLMs} can run beyond 100 {FPS} with
  {Stage-Wise} coordinated inference.
\newblock \emph{arXiv preprint arXiv:2606.19849}, 2026.

\bibitem[Team et~al.(2023)Team, Anil, Borgeaud, Alayrac, Yu, Soricut,
  Schalkwyk, Dai, Hauth, Millican, et~al.]{gemini2023}
Gemini Team, Rohan Anil, Sebastian Borgeaud, Jean-Baptiste Alayrac, Jiahui Yu,
  Radu Soricut, Johan Schalkwyk, Andrew~M. Dai, Anja Hauth, Katie Millican,
  et~al.
\newblock Gemini: A family of highly capable multimodal models.
\newblock \emph{arXiv preprint arXiv:2312.11805}, 2023.

\bibitem[Tschannen et~al.(2025)Tschannen, Gritsenko, Wang, Naeem,
  Alabdulmohsin, Parthasarathy, Evans, Beyer, Xia, Mustafa, et~al.]{siglip2}
Michael Tschannen, Alexey Gritsenko, Xiao Wang, Muhammad~Ferjad Naeem, Ibrahim
  Alabdulmohsin, Nikhil Parthasarathy, Talfan Evans, Lucas Beyer, Ye Xia, Basil
  Mustafa, et~al.
\newblock {SigLIP} 2: Multilingual vision-language encoders with improved
  semantic understanding, localization, and dense features.
\newblock \emph{arXiv preprint arXiv:2502.14786}, 2025.

\bibitem[Vasu et~al.(2025)Vasu, Faghri, Li, Koc, True, Antony, Santhanam,
  Gabriel, Grasch, Tuzel, et~al.]{fastvlm2025}
Pavan Kumar~Anasosalu Vasu, Fartash Faghri, Chun-Liang Li, Cem Koc, Nate True,
  Albert Antony, Gokul Santhanam, James Gabriel, Peter Grasch, Oncel Tuzel,
  et~al.
\newblock {FastVLM}: Efficient vision encoding for vision language models.
\newblock In \emph{CVPR}, 2025.
\newblock arXiv:2412.13303.

\bibitem[Vinyals et~al.(2015)Vinyals, Toshev, Bengio, and Erhan]{showtell2015}
Oriol Vinyals, Alexander Toshev, Samy Bengio, and Dumitru Erhan.
\newblock Show and tell: A neural image caption generator.
\newblock In \emph{CVPR}, 2015.
\newblock arXiv:1411.4555.

\bibitem[Wang et~al.(2025)Wang, Gao, Gu, Pu, Cui, Wei, Liu, Jing, Ye, Shao,
  et~al.]{internvl35}
Weiyun Wang, Zhangwei Gao, Lixin Gu, Hengjun Pu, Long Cui, Xingguang Wei,
  Zhaoyang Liu, Linglin Jing, Shenglong Ye, Jie Shao, et~al.
\newblock {InternVL3.5}: Advancing open-source multimodal models in
  versatility, reasoning, and efficiency.
\newblock \emph{arXiv preprint arXiv:2508.18265}, 2025.

\bibitem[Wang et~al.(2023)Wang, Wei, Schuurmans, Le, Chi, Narang, Chowdhery,
  and Zhou]{selfconsistency2023}
Xuezhi Wang, Jason Wei, Dale Schuurmans, Quoc Le, Ed Chi, Sharan Narang,
  Aakanksha Chowdhery, and Denny Zhou.
\newblock Self-consistency improves chain of thought reasoning in language
  models.
\newblock In \emph{ICLR}, 2023.
\newblock arXiv:2203.11171.

\bibitem[Xu et~al.(2022)Xu, Razavi, and Zheng]{edgevideo2022}
Renjie Xu, Saiedeh Razavi, and Rong Zheng.
\newblock Edge video analytics: A survey on applications, systems and enabling
  techniques.
\newblock \emph{arXiv preprint arXiv:2211.15751}, 2022.

\bibitem[Xu et~al.(2026)Xu, Xiao, Chen, He, Peng, Lu, and
  Han]{streamingvlm2025}
Ruyi Xu, Guangxuan Xiao, Yukang Chen, Liuning He, Kelly Peng, Yao Lu, and Song
  Han.
\newblock {StreamingVLM}: Real-time understanding for infinite video streams.
\newblock In \emph{ICLR}, 2026.
\newblock arXiv:2510.09608.

\bibitem[Xun et~al.(2025)Xun, Tao, Li, Shi, Lin, Zhu, Yan, Li, Zhang, Wang,
  et~al.]{rtvbench2025}
Shuhang Xun, Sicheng Tao, Jungang Li, Yibo Shi, Zhixin Lin, Zhanhui Zhu, Yibo
  Yan, Hanqian Li, Linghao Zhang, Shikang Wang, et~al.
\newblock {RTV-Bench}: Benchmarking {MLLM} continuous perception, understanding
  and reasoning through {Real-Time} video.
\newblock In \emph{NeurIPS Datasets and Benchmarks Track}, 2025.
\newblock arXiv:2505.02064.

\bibitem[Yang et~al.(2026)Yang, Liu, Zhai, Sun, Liu, Barsoum, Li, and
  Xu]{captionqa2026}
Shijia Yang, Yunong Liu, Bohan Zhai, Ximeng Sun, Zicheng Liu, Emad Barsoum,
  Manling Li, and Chenfeng Xu.
\newblock {CaptionQA}: Is your caption as useful as the image itself?
\newblock In \emph{CVPR}, 2026.
\newblock arXiv:2511.21025.

\bibitem[Yao et~al.(2025)Yao, Li, Wei, Li, Ren, Liu, Ouyang, Wang, Li, Li,
  et~al.]{timechatonline2025}
Linli Yao, Yicheng Li, Yuancheng Wei, Lei Li, Shuhuai Ren, Yuanxin Liu, Kun
  Ouyang, Lean Wang, Shicheng Li, Sida Li, et~al.
\newblock {TimeChat-Online}: 80\% visual tokens are naturally redundant in
  streaming videos.
\newblock \emph{arXiv preprint arXiv:2504.17343}, 2025.

\bibitem[Yu and Huang(2019)]{usnet2019}
Jiahui Yu and Thomas Huang.
\newblock Universally slimmable networks and improved training techniques.
\newblock In \emph{ICCV}, 2019.
\newblock arXiv:1903.05134.

\bibitem[Yu et~al.(2019)Yu, Yang, Xu, Yang, and Huang]{slimmable2019}
Jiahui Yu, Linjie Yang, Ning Xu, Jianchao Yang, and Thomas Huang.
\newblock Slimmable neural networks.
\newblock In \emph{ICLR}, 2019.
\newblock arXiv:1812.08928.

\bibitem[Yu et~al.(2025)Yu, Zhang, Zhu, Yuan, Zuo, Yue, Dai, Fan, Liu, Liu,
  et~al.]{dapo2025}
Qiying Yu, Zheng Zhang, Ruofei Zhu, Yufeng Yuan, Xiaochen Zuo, Yu Yue, Weinan
  Dai, Tiantian Fan, Gaohong Liu, Lingjun Liu, et~al.
\newblock {DAPO}: An {Open-Source} {LLM} reinforcement learning system at
  scale.
\newblock \emph{arXiv preprint arXiv:2503.14476}, 2025.

\bibitem[Zanella et~al.(2024)Zanella, Menapace, Mancini, Wang, and
  Ricci]{lavad2024}
Luca Zanella, Willi Menapace, Massimiliano Mancini, Yiming Wang, and Elisa
  Ricci.
\newblock Harnessing large language models for training-free video anomaly
  detection.
\newblock In \emph{CVPR}, 2024.
\newblock arXiv:2404.01014.

\bibitem[Zeng et~al.(2023)Zeng, Attarian, Ichter, Choromanski, Wong, Welker,
  Tombari, Purohit, Ryoo, Sindhwani, Lee, Vanhoucke, and
  Florence]{socratic2023}
Andy Zeng, Maria Attarian, Brian Ichter, Krzysztof Choromanski, Adrian Wong,
  Stefan Welker, Federico Tombari, Aveek Purohit, Michael Ryoo, Vikas
  Sindhwani, Johnny Lee, Vincent Vanhoucke, and Pete Florence.
\newblock {Socratic Models}: Composing zero-shot multimodal reasoning with
  language.
\newblock In \emph{ICLR}, 2023.
\newblock arXiv:2204.00598.

\bibitem[Zhang et~al.(2024)Zhang, Lu, Islam, Wang, Yu, Bansal, and
  Bertasius]{llovi2024}
Ce Zhang, Taixi Lu, Md~Mohaiminul Islam, Ziyang Wang, Shoubin Yu, Mohit Bansal,
  and Gedas Bertasius.
\newblock A simple {LLM} framework for {Long-Range} video {Question-Answering}.
\newblock In \emph{EMNLP}, 2024.
\newblock arXiv:2312.17235.

\bibitem[Zhang et~al.(2025{\natexlab{a}})Zhang, Wang, Tang, Liu, Feng, and
  Jin]{flashvstream2025}
Haoji Zhang, Yiqin Wang, Yansong Tang, Yong Liu, Jiashi Feng, and Xiaojie Jin.
\newblock {Flash-VStream}: Efficient {Real-Time} understanding for long video
  streams.
\newblock In \emph{ICCV}, 2025{\natexlab{a}}.
\newblock arXiv:2506.23825.

\bibitem[Zhang et~al.(2025{\natexlab{b}})Zhang, Fang, Yang, and
  Feng]{llavamini2025}
Shaolei Zhang, Qingkai Fang, Zhe Yang, and Yang Feng.
\newblock {LLaVA-Mini}: Efficient image and video large multimodal models with
  one vision token.
\newblock In \emph{ICLR}, 2025{\natexlab{b}}.
\newblock arXiv:2501.03895.

\bibitem[Zhang et~al.(2025{\natexlab{c}})Zhang, Weng, Yue, Fan, Wu, and
  Huang]{tinyllavavideo2025}
Xingjian Zhang, Xi Weng, Yihao Yue, Zhaoxin Fan, Wenjun Wu, and Lei Huang.
\newblock {TinyLLaVA-Video}: Towards smaller {LMMs} for video understanding
  with group resampler.
\newblock \emph{arXiv preprint arXiv:2501.15513}, 2025{\natexlab{c}}.

\bibitem[Zhang et~al.(2025{\natexlab{d}})Zhang, Wu, Li, Li, Ma, Liu, and
  Li]{llavavideo2024}
Yuanhan Zhang, Jinming Wu, Wei Li, Bo Li, Zejun Ma, Ziwei Liu, and Chunyuan Li.
\newblock {LLaVA-Video}: Video instruction tuning with synthetic data.
\newblock \emph{TMLR}, 2025{\natexlab{d}}.
\newblock arXiv:2410.02713.

\bibitem[Zheng et~al.(2023)Zheng, Chiang, Sheng, Zhuang, Wu, Zhuang, Lin, Li,
  Li, Xing, et~al.]{llmjudge2023}
Lianmin Zheng, Wei-Lin Chiang, Ying Sheng, Siyuan Zhuang, Zhanghao Wu, Yonghao
  Zhuang, Zi Lin, Zhuohan Li, Dacheng Li, Eric~P. Xing, et~al.
\newblock Judging {LLM-as-a-Judge} with {MT-Bench} and {Chatbot Arena}.
\newblock In \emph{NeurIPS Datasets and Benchmarks Track}, 2023.
\newblock arXiv:2306.05685.

\end{thebibliography}
